\documentclass{article}

\newif\ifdraftfigures

\draftfiguresfalse    %

\ifdraftfigures
    \PassOptionsToPackage{draft}{graphicx}
\else
    \PassOptionsToPackage{final}{graphicx}
\fi

\usepackage[utf8]{inputenc}
\usepackage{authblk}
\usepackage{main}
\usepackage{microtype}
\usepackage{graphicx}

\usepackage{times}
\usepackage{latexsym}
\usepackage{amsmath,amsfonts,amssymb,amsthm}
\usepackage{float}
\usepackage{footnote}
\usepackage{enumitem}
\usepackage{bm}
\usepackage{arydshln}
\usepackage{booktabs}
\usepackage{multicol}
\usepackage{multirow}
\usepackage{color}
\usepackage{xcolor}     
\usepackage{colortbl}
\usepackage[table]{xcolor} %
\usepackage{algorithm}
\usepackage{algpseudocode}

\algrenewcommand{\algorithmicrequire}{\textbf{Input:}}
\algrenewcommand{\algorithmicensure}{\textbf{Output:}}
\definecolor{royalblue}{RGB}{65,105,225}
\definecolor{purple}{RGB}{128,0,128}
\definecolor{selfevolagent_dark}{HTML}{C2410C}
\definecolor{selfevolagent_light}{HTML}{FDBA74}
\definecolor{selfevolagent_lighter}{HTML}{FFF7ED}
\definecolor{muse_deep}{HTML}{9A3412}
\definecolor{muse_burnt}{HTML}{C84B09}
\definecolor{muse_mid}{HTML}{E56E1F}
\definecolor{muse_soft}{HTML}{FED7AA}
\definecolor{muse_pale}{HTML}{FEEAD5}
\definecolor{muse_warmgray}{HTML}{FFFDF3}
\definecolor{spectrum_stage_1}{HTML}{93C5FD}
\definecolor{spectrum_stage_2}{HTML}{86EFAC}
\definecolor{spectrum_stage_3}{HTML}{FDE68A}
\definecolor{spectrum_stage_4}{HTML}{FDBA74}
\definecolor{spectrum_stage_5}{HTML}{FCA5A5}
\definecolor{stage_1_dark}{RGB}{20, 80, 150}   %
\definecolor{stage_2_dark}{RGB}{20, 100, 40}   %
\definecolor{stage_3_dark}{RGB}{180, 120, 0}   %
\definecolor{stage_4_dark}{RGB}{180, 80, 10}   %
\definecolor{stage_5_dark}{RGB}{150, 20, 40}   %
\providecolor{notationblue}{RGB}{137,176,211}
\providecolor{notationorange}{RGB}{225,181,137}
\makeatletter
\newcommand{\appmodule}{\@startsection{paragraph}{4}{\z@}{-1.5ex plus -0.5ex minus -.2ex}{0.5ex plus .2ex}{\normalfont\normalsize\bfseries\raggedright}}
\makeatother

\usepackage{bbding}
\usepackage{makecell}
\usepackage{mathtools}
\usepackage{imakeidx}
\usepackage{longtable}
\usepackage{wrapfig}
\usepackage{subcaption}
\makeindex
\usepackage{arydshln}
\usepackage{lipsum}
\usepackage{natbib}
\usepackage[edges]{forest}
\usepackage[normalem]{ulem}
\usepackage{titletoc}
\definecolor{mydarkblue}{rgb}{0,0.08,0.45}
\usepackage[colorlinks=true,linkcolor=mydarkblue,citecolor=mydarkblue,filecolor=mydarkblue,urlcolor=mydarkblue]{hyperref}
\usepackage{caption}
\usepackage{CJKutf8}
\usepackage{awesomebox} %
\usepackage{bbding}
\usepackage[most]{tcolorbox}
\usepackage{caption}

\usepackage[tikz]{bclogo}
\usepackage[framemethod=tikz]{mdframed}
\definecolor{bgblue}{RGB}{245,243,253}
\definecolor{ttblue}{RGB}{91,194,224}
\usetikzlibrary{positioning, shapes.geometric, shapes, arrows.meta}
\usepackage{tabularx}
\usepackage{geometry}
\usepackage{pifont}
\definecolor{myblue}{RGB}{0,80,160}
\definecolor{myyellow}{RGB}{210,170,0}
\definecolor{mygray}{RGB}{100,100,100}
\definecolor{mypurple}{RGB}{120,80,160}

\mdfdefinestyle{mystyle}{%
  rightline=true,
  innerleftmargin=10,
  innerrightmargin=10,
  outerlinewidth=3pt,
  topline=false,
  rightline=true,
  bottomline=false,
  skipabove=\topsep,
  skipbelow=\topsep
}

\newtcolorbox{myboxi}[1][]{
  breakable,
  title=#1,
  colback=red!5,
  colbacktitle=red!5,
  coltitle=black,
  fonttitle=\bfseries,
  bottomrule=0pt,
  toprule=0pt,
  leftrule=2pt,
  rightrule=2pt,
  titlerule=0pt,
  arc=0pt,
  outer arc=0pt,
  colframe=red,
}

\newtcolorbox{myboxnote}[1][]{
  breakable,
  title=#1,
  colback=orange!0,
  colbacktitle=orange!0,
  coltitle=black,
  fonttitle=\bfseries,
  bottomrule=0pt,
  toprule=0pt,
  leftrule=2pt,
  rightrule=2pt,
  titlerule=0pt,
  arc=0pt,
  outer arc=0pt,
  colframe=orange,
}

\newtcolorbox{myboxii}[1][]{
  breakable,
  freelance,
  title=#1,
  colback=white,
  colbacktitle=white,
  coltitle=black,
  fonttitle=\bfseries,
  bottomrule=0pt,
  boxrule=0pt,
  colframe=white,
  overlay unbroken and first={
  \draw[red!75!black,line width=3pt]
    ([xshift=5pt]frame.north west) -- 
    (frame.north west) -- 
    (frame.south west);
  \draw[red!75!black,line width=3pt]
    ([xshift=-5pt]frame.north east) -- 
    (frame.north east) -- 
    (frame.south east);
  },
  overlay unbroken app={
  \draw[red!75!black,line width=3pt,line cap=rect]
    (frame.south west) -- 
    ([xshift=5pt]frame.south west);
  \draw[red!75!black,line width=3pt,line cap=rect]
    (frame.south east) -- 
    ([xshift=-5pt]frame.south east);
  },
  overlay middle and last={
  \draw[red!75!black,line width=3pt]
    (frame.north west) -- 
    (frame.south west);
  \draw[red!75!black,line width=3pt]
    (frame.north east) -- 
    (frame.south east);
  },
  overlay last app={
  \draw[red!75!black,line width=3pt,line cap=rect]
    (frame.south west) --
    ([xshift=5pt]frame.south west);
  \draw[red!75!black,line width=3pt,line cap=rect]
    (frame.south east) --
    ([xshift=-5pt]frame.south east);
  },
}

\definecolor{promptcardbg}{RGB}{247,247,247}
\definecolor{promptcardtitle}{RGB}{115,115,115}
\definecolor{promptcardframe}{RGB}{55,55,55}
\definecolor{promptcardtext}{RGB}{25,25,25}

\lstdefinestyle{promptcardstyle}{
    basicstyle=\small\rmfamily\color{promptcardtext},
    columns=fullflexible,
    keepspaces=false,
    breaklines=true,
    breakatwhitespace=true,
    breakindent=0pt,
    breakautoindent=false,
    resetmargins=true,
    showstringspaces=false,
    showtabs=false,
    aboveskip=0pt,
    belowskip=0pt,
    xleftmargin=0pt,
    xrightmargin=0pt
}

\newtcblisting{promptcard}[1]{
    enhanced,
    breakable,
    listing only,
    title={#1},
    listing options={style=promptcardstyle},
    colback=promptcardbg,
    colframe=promptcardframe,
    colbacktitle=promptcardtitle,
    coltitle=white,
    fonttitle=\bfseries\sffamily\small,
    boxrule=0.55pt,
    arc=4pt,
    outer arc=4pt,
    halign=left,
    left=8pt,
    right=8pt,
    top=5pt,
    bottom=5pt,
    before skip=6pt,
    after skip=10pt
}

\usepackage{fancyhdr} %
\usepackage{blindtext} %

\usepackage{forest}
\usepackage{xcolor}
\usepackage{tikz}
\usepackage{fontawesome5}
\usetikzlibrary{calc,shadows}
\definecolor{coverink}{HTML}{1F2937}
\definecolor{coverteal}{HTML}{C2410C}
\pgfdeclarehorizontalshading{musegradV1}{100bp}{
  color(0bp)=(muse_burnt);
  color(25bp)=(muse_burnt);
  color(46bp)=(muse_mid);
  color(64bp)=(muse_soft);
  color(75bp)=(muse_pale);
  color(100bp)=(muse_pale)
}
\newcommand{\coverprojurl}{https://agentevolution-muse.github.io/}%
\newtcbox{\coverpill}[1][selfevolagent_dark]{on line, colback=#1, colframe=#1, coltext=white, arc=7pt, boxrule=0pt, left=6pt, right=6pt, top=2pt, bottom=2pt, fontupper=\small\bfseries}
\newlength{\coverleft}
\usepackage{setspace}
\definecolor{lightblue}{RGB}{173,216,230}
\definecolor{lightgreen}{RGB}{144,238,144}
\definecolor{lightyellow}{RGB}{255,255,224}
\definecolor{lightpink}{RGB}{255,182,193}
\definecolor{hidden-draw}{RGB}{0,0,0}

\DeclareCaptionFont{black}{\color{black}}

\definecolor{myblue}{rgb}{0.9, 0.1, 0.94}
\definecolor{mygreen}{rgb}{0.64, 0.56, 0.88}
\definecolor{myyellow}{rgb}{0.68, 0.6, 0.1}
\definecolor{fancygreen}{rgb}{0.33, 0.68, 0.20}
\definecolor{salmon}{rgb}{0.94, 0.52, 0.49}
\definecolor{tablegreen}{rgb}{0.82, 0.94, 0.75}
\definecolor{tableblue}{rgb}{0.81, 0.90, 0.94}
\definecolor{tablered}{rgb}{0.97, 0.85, 0.85}
\definecolor{tableorange}{rgb}{0.96, 0.85, 0.81}

\newenvironment{itemize*}%
 {\leftmargini=10pt\begin{itemize}%
  \setlength{\itemsep}{0pt}%
  \setlength{\parskip}{0pt}%
  }%
 {\end{itemize}}
\newenvironment{enumerate*}%
 {\begin{enumerate}%
  \setlength{\itemsep}{0pt}%
  \setlength{\parskip}{0pt}}%
 {\end{enumerate}}

\usepackage{xcolor}
\usepackage{listings}

\newcommand\JSONnumbervaluestyle{\color{blue}}
\newcommand\JSONstringvaluestyle{\color{red}}

\newcolumntype{Y}{>{\centering\arraybackslash}X}
\newcolumntype{L}{>{\raggedright\arraybackslash}X}
\newcolumntype{C}[1]{>{\centering\arraybackslash}m{#1}}

\newif\ifcolonfoundonthisline

\makeatletter

\lstdefinestyle{json}
{
  showstringspaces    = false,
  keywords            = {false,true},
  alsoletter          = 0123456789.,
  morestring          = [s]{"}{"},
  stringstyle         = \ifcolonfoundonthisline\JSONstringvaluestyle\fi,
  MoreSelectCharTable =%
    \lst@DefSaveDef{`:}\colon@json{\processColon@json},
  basicstyle          = \ttfamily,
  keywordstyle        = \ttfamily\bfseries,
}

\newcommand\processColon@json{%
  \colon@json%
  \ifnum\lst@mode=\lst@Pmode%
    \global\colonfoundonthislinetrue%
  \fi
}

\lst@AddToHook{Output}{%
  \ifcolonfoundonthisline%
    \ifnum\lst@mode=\lst@Pmode%
      \def\lst@thestyle{\JSONnumbervaluestyle}%
    \fi
  \fi
  \lsthk@DetectKeywords%
}

\lst@AddToHook{EOL}%
  {\global\colonfoundonthislinefalse}

\makeatother

\usepackage{etoolbox}
\usepackage{natbib}

\usepackage{url}
\newcounter{bibcount}
\makeatletter
\patchcmd{\@lbibitem}{\item[}{\item[\hfil\stepcounter{bibcount}{[\thebibcount]}}{}{}
\renewcommand\NAT@bibsetup%
  [1]{\setlength{\leftmargin}{\bibhang}\setlength{\itemindent}{-\parindent}%
      \setlength{\itemsep}{\bibsep}\setlength{\parsep}{\z@}}
\makeatother

\theoremstyle{plain}

\theoremstyle{definition}
\newtheorem{definition}{Definition}[section]

\newtheorem*{definition*}{Definition}

\theoremstyle{remark}

\begin{document}

\thispagestyle{empty}
\setlength{\coverleft}{\dimexpr1in+\hoffset+\oddsidemargin\relax}
\begin{tikzpicture}[remember picture,overlay]
  \shade[shading=musegradV1] (current page.north west) rectangle ([yshift=-4.55cm]current page.north east);
  \node[anchor=north west,text=white,font=\fontsize{26}{30}\selectfont\bfseries,align=left] at ([xshift=\coverleft,yshift=-1.05cm]current page.north west) {Metacognitive Unison\\Self-Evolution};
  \node[anchor=north west,text=white!96,font=\large\itshape] at ([xshift=\coverleft,yshift=-3.32cm]current page.north west) {From Human-Engineered Maintenance to Recursive Meta-Evolution};
  \begin{scope}[
    shift={([xshift=-8.45cm,yshift=-3.12cm]current page.north east)},
    line cap=round,
    line join=round,
    coverstage/.style={
      circle,
      draw=white,
      line width=1.1pt,
      inner sep=0pt,
      text=coverink,
      font=\sffamily\bfseries\fontsize{8}{9}\selectfont
    }
  ]
    \coordinate (cs1) at (0.15,0.30);
    \coordinate (cs2) at (0.95,0.41);
    \coordinate (cs3) at (1.80,0.60);
    \coordinate (cs4) at (3.25,1.09);
    \coordinate (cs5) at (5.35,2.17);

    \draw[muse_burnt!75,opacity=0.78,line width=1.25pt]
      (cs1) .. controls (0.98,0.38) and (1.82,0.57) ..
      ($(cs4)+(200:0.64)$);

    \draw[
      muse_burnt!75,
      opacity=0.74,
      line width=0.95pt,
      dash pattern=on 2.4pt off 2.4pt
    ]
      ($(cs4)+(23:0.73)$)
      .. controls (4.18,1.50) and (4.44,1.65) ..
      ($(cs5)+(208:0.74)$);

    \node[
      coverstage,
      fill=spectrum_stage_1,
      minimum size=0.48cm
    ] at (cs1) {1};

    \node[
      coverstage,
      fill=spectrum_stage_2,
      minimum size=0.53cm
    ] at (cs2) {2};

    \node[
      coverstage,
      fill=spectrum_stage_3,
      minimum size=0.58cm
    ] at (cs3) {3};

    \fill[muse_mid,opacity=0.10] (cs4) circle[radius=0.84cm];
    \fill[muse_mid,opacity=0.16] (cs4) circle[radius=0.64cm];

    \draw[muse_burnt!75,opacity=0.86,line width=0.82pt]
      (cs4) circle[radius=0.64cm];

    \foreach \coverangle in {90,210,330}{
      \fill[muse_burnt!75,opacity=0.92]
        ($(cs4)+(\coverangle:0.64)$)
        circle[radius=0.048cm];
    }

    \node[
      coverstage,
      fill=spectrum_stage_4,
      draw=white,
      minimum size=0.84cm,
      line width=1.6pt,
      text=muse_deep,
      font=\sffamily\bfseries\fontsize{9.5}{10.5}\selectfont
    ] at (cs4) {4};

    \draw[muse_burnt!75,opacity=0.66,line width=0.56pt]
      (cs5) circle[radius=0.49cm];

    \foreach \coverangle in {90,210,330}{
      \fill[muse_burnt!75,opacity=0.82]
        ($(cs5)+(\coverangle:0.49)$)
        circle[radius=0.036cm];
    }

    \draw[muse_burnt!75,opacity=0.62,line width=0.60pt]
      (cs5) circle[radius=0.64cm];

    \draw[
      muse_burnt!75,
      opacity=0.82,
      line width=0.68pt,
      -{Stealth[length=3.0pt,width=2.8pt]}
    ]
      ($(cs5)+(20:0.64)$)
      arc[
        start angle=20,
        end angle=93,
        radius=0.64cm
      ];

    \node[
      coverstage,
      fill=spectrum_stage_5!88!white,
      minimum size=0.68cm,
      line width=0.95pt
    ] at (cs5) {5};
  \end{scope}
\end{tikzpicture}
\vspace*{2.25cm}

\begin{flushleft}
{\small
Zongru~Wu\textsuperscript{1,2}\hspace{0.85em}Hongbo~Qiu\textsuperscript{4} \hspace{0.85em} Yi Ren\textsuperscript{3} \hspace{0.85em}Bobo~Li\textsuperscript{2}\hspace{0.85em}Tianjie~Ju\textsuperscript{2,1} \hspace{0.85em}Kelvin~J.L.~Koa\textsuperscript{2}\hspace{0.85em}Wynne~Hsu\textsuperscript{2}\hspace{0.85em}Mong~Li~Lee\textsuperscript{2}\\[2pt]
Hao~Fei\textsuperscript{3}\hspace{0.85em}Zhuosheng~Zhang\textsuperscript{1*}\hspace{0.85em}Gongshen~Liu\textsuperscript{1*}\hspace{0.85em}Michael~Wooldridge\textsuperscript{3}\hspace{0.85em}Shengqiong~Wu\textsuperscript{3*}\par}
\vspace{5pt}
{\small\color{coverink!75}
\textsuperscript{1}Shanghai Jiao Tong University \quad \textsuperscript{2}National University of Singapore\\
\textsuperscript{3}University of Oxford \quad \textsuperscript{4}Shenzhen University\par}
\vspace{8pt}
{\small\color{selfevolagent_dark}\faGlobe}\;\textbf{\small Project Page:} {\small\hypersetup{urlcolor=selfevolagent_dark}\url{\coverprojurl}}
\end{flushleft}
{\renewcommand{\thefootnote}{}\footnotetext{$^*$Corresponding authors: Zhuosheng Zhang, Gongshen Liu, Shengqiong Wu.}}
\vspace{3pt}
\begin{tcolorbox}[colback=selfevolagent_lighter,colframe=selfevolagent_lighter,boxrule=0pt,arc=3pt,left=8pt,right=8pt,top=6pt,bottom=6pt]
{\footnotesize\textbf{Abstract.}~While artificial agents can effectively resolve complex, real-world tasks, these systems remain largely fixed after deployment, limiting their applicability in dynamic, open-ended environments. Bridging the gap between static post-deployment states and the continuous adaptability required for lifelong systems motivates a paradigm shift toward \textbf{agent evolution}. 
Although often viewed as a recent direction, adapting artificial agents spans decades, tracing back to early chatbots relying on human-engineered maintenance. 
Existing evolution methods vary in their autonomy in managing adaptation and obtaining supervision, as well as in the scope of component optimization. These differences motivate a unified framework to describe how evolutionary processes operate and how their capabilities develop.
To provide a unified view, we formalize agent evolution through an \textbf{evolutionary strategy} that characterizes how continuous adaptation is organized. Building on this formalization, we organize the development of agent evolution into a five-stage spectrum, ranging from human-engineered maintenance to recursive meta-evolution, based on three dimensions: evolutionary autonomy, supervision autonomy, and optimization scope and coordination.
Grounded in this framework, we propose \textbf{M}etacognitive \textbf{U}nison \textbf{S}elf-\textbf{E}volution (\textbf{MUSE}), a prototype targeting Stage-4 capabilities of metacognitive self-evolution.
MUSE operates through three core mechanisms: (i) \textit{metacognitive gating} that determines when to evolve, diagnoses capability gaps, and routes optimization demands, (ii) \textit{interaction-grounded self-supervision} that constructs supplementary tasks and rewards from environmental interaction evidence to complement available oracle supervision during evolution, and (iii) \textit{coordinated component evolution} that combines inner-component refinement with inter-component synchronization across the parametric core and symbolic assets.
Experiments on APPWorld support the practical feasibility of these mechanisms. Component comparisons, supervision audits, and multi-round analyses further characterize their contributions to capability improvement and evolutionary efficiency.
}

\end{tcolorbox}

\begin{figure}[H]
    \centering
    \includegraphics[width=\linewidth]{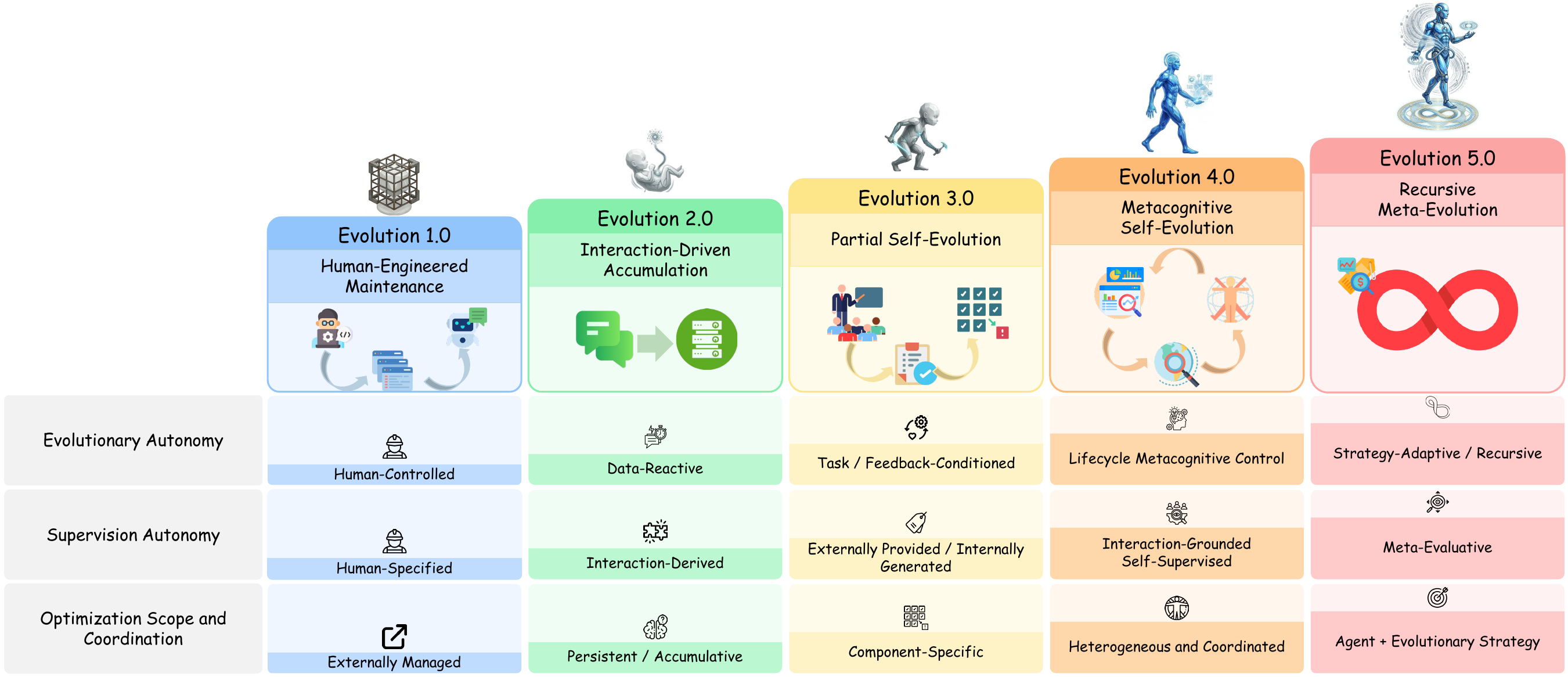}
    \caption{Overview of the Five-Stage Evolutionary Spectrum, from human-engineered maintenance to recursive meta-evolution. Successive stages introduce cumulative capability requirements across evolutionary autonomy, supervision autonomy, and optimization scope and coordination.}
    \label{fig:graph_abstract}
\end{figure}

\newpage

\pagestyle{fancy}
\lhead{\rightmark}
\renewcommand{\headrulewidth}{0.7pt}
\setlength{\headsep}{5mm}

\clearpage

\onecolumn

\startcontents[mainbody]
\section*{Contents}
\markboth{Contents}{Contents}
\thispagestyle{plain}

\begin{spacing}{1.1}
\printcontents[mainbody]{}{1}[3]{}
\end{spacing}
\clearpage

\section{Introduction}\label{sec:intro}

In recent years, the advent of large language models (LLMs)~\citep{yang2024qwen25, deepseekai2024deepseekv3technicalreport, yang2025qwen3} has catalyzed the development of artificial agents capable of solving sophisticated real-world tasks~\citep{yao2022react, wang2024empowering, luo2025large}. However, a persistent mismatch remains in the current agentic paradigm: while the real world is inherently dynamic and constantly changing, many deployed agents remain largely fixed in their underlying capabilities and adaptation strategies~\citep{luo2025large, fang2025comprehensive, jaglan2025continual}. This limitation persists across the history of autonomous systems, spanning from early rule-based artificial agents~\citep{dietrich2003rule, wang2005agent} to contemporary LLM-powered agents. Consider a common scenario in digital agent navigation, such as an assistant agent operating within an application environment like smartphones. Initially, the agent navigates based on its pre-trained knowledge. However, as the environment evolves through events such as new app installations or API deprecations~\citep{mahmud2025automated}, the agent may struggle to adapt to new operational logic. Traditionally, resolving this requires the agent to passively wait for human engineers to push an update~\citep{hayes1983building, kephart2003vision, kaymak2025learning}. This static post-deployment nature limits the applicability of agents in open-ended dynamic environments, motivating the study of \textbf{Agent Evolution}~\citep{fang2025comprehensive, gao2025survey}.

\begin{tcolorbox}[
  enhanced,
  colback=selfevolagent_lighter,
  colframe=selfevolagent_light,
  colbacktitle=muse_soft!85!selfevolagent_light,
  coltitle=muse_deep,
  title={\bfseries\fontfamily{ppl}\selectfont{Definition of Agent Evolution}},
  boxrule=1.2pt,
  arc=4pt,
  drop shadow={opacity=0.14},
  parbox=false,
  before skip=5pt,
  after skip=5pt,
  left=7pt,
  right=7pt,
  top=5pt,
  bottom=5pt,
] \label{def:agent_evolution}

Agent evolution is a continuous adaptation process throughout deployment and ongoing interaction, aimed at maximizing environmental fitness and accumulating capabilities under operational boundaries. The adaptation process is structured by an \textbf{evolutionary strategy} that specifies supervision generation, optimization scope and method, evolutionary gating, adaptation objectives, and boundary constraints.
\end{tcolorbox}

Agent evolution has demonstrated the potential for agents to continuously adapt after deployment, with different approaches establishing different conditions for autonomous improvement. For example, SIA performs iterative improvement given task specifications and outcome verifiers~\citep{hebbar2026sia}, and AgentEvolver constructs supplementary learning tasks from environmental interaction~\citep{zhai2025agentevolver}. These developments reveal an important distinction between the ability to improve task performance and the autonomy of continuous adaptation. An agent may independently revise its behavior while depending on externally prepared tasks or evaluation criteria. Comparing these approaches requires a common account of how feedback for adaptation is obtained and organized, when and what the agent should adapt, and how changes across its components are coordinated. Explicit capability requirements can then characterize increasing evolutionary autonomy and guide the design and evaluation of concrete evolutionary mechanisms. This motivates our central research question: \textit{What capabilities support progressively more autonomous agent evolution?}

To address this question, we first establish a unified formalization of agent evolution by introducing the \textbf{evolutionary strategy} as a structured representation of continuous adaptation. The strategy comprises six components: supervision generator, optimization scope, evolutionary optimizer, evolutionary gating, evolutionary objective, and boundary constraint. This formulation provides a common structural basis for describing and comparing how different agents organize continuous adaptation throughout deployment and ongoing interaction.

Building on this formalization, we characterize evolutionary maturity jointly along three axes: evolutionary autonomy, supervision autonomy, and optimization scope and coordination, specifying a five-stage spectrum of cumulative capability requirements. The spectrum spans human-engineered maintenance~\citep{kephart2003vision} (Stage~1) through recursive meta-evolution (Stage~5). Each stage retains the foundational capabilities of preceding stages and introduces additional capabilities for autonomous evolution. Explicit requirements along the three dimensions provide criteria for distinguishing successive stages and identifying directions for further development. Within this spectrum, Stage~4 (Metacognitive Self-Evolution) is characterized by the joint realization of lifecycle-level metacognitive control, interaction-grounded self-supervision, and coordinated component evolution within a unified evolutionary strategy. Complementary evaluation criteria connect these requirements to observable evolution processes and assess the effectiveness, efficiency, and stability of concrete evolutionary strategies.

Grounded in this theoretical framework, we propose \textbf{M}etacognitive \textbf{U}nison \textbf{S}elf-\textbf{E}volution (MUSE), a prototype that instantiates the six-component evolutionary strategy to explore Stage-4 mechanisms. MUSE instantiates metacognitive self-evolution through three core mechanisms: (i) \textit{metacognitive gating}~\citep{flavell1979metacognition,rehan2025self} that determines when to evolve, diagnoses capability gaps, and routes optimization demands, (ii) \textit{interaction-grounded self-supervision} that constructs supplementary tasks and rewards from environmental interaction evidence to complement available oracle supervision during evolution, and (iii) \textit{coordinated component evolution} that combines inner-component refinement with inter-component synchronization across the parametric core and symbolic assets, including general operational principles, situational experiences~\citep{zhai2025agentevolver, wei2025evo}, and functional tools~\citep{xu2025llm,jin2025stella}. These mechanisms are guided by a shared adaptation objective and accompanied by measures for retaining previously acquired capabilities.

We evaluate this instantiation on APPWorld~\citep{trivedi2024appworld} through controlled ablations, human audits, and multi-round analyses. The six-component formulation guides the experimental comparisons by distinguishing component adaptation, supervision construction, and evolutionary control. We examine their contributions and interactions through task performance, average interaction steps, exploration rollout counts, and performance stability across successive rounds. Human audits and diagnostic records complement these measurements by examining supervision quality and the connection between capability diagnosis and targeted refinement. These analyses provide empirical support for the feasibility of the Stage~4 mechanisms instantiated in MUSE and clarify their contributions and limitations during evolution.

Our contributions are summarized as follows:

(i) We establish a unified formalization of \textbf{agent evolution} by introducing the \textbf{evolutionary strategy} as a structured representation of how the continuous adaptation process is organized and executed. We further characterize the evolutionary strategy through a six-component formulation, providing a general structural framework for decomposing and characterizing existing and future agent evolution paradigms (Section~\ref{sec:formalization}).

(ii) Building on this formalization, we characterize the maturity of evolutionary strategies jointly along three fundamental dimensions: evolutionary autonomy, supervision autonomy, and optimization scope and coordination. We propose a cumulative \textbf{Five-Stage Evolutionary Spectrum} with explicit capability requirements for each stage, describing progression from human-engineered maintenance to recursive meta-evolution (Section~\ref{sec:spectrum}).

(iii) Grounded in this theoretical framework, we propose \textbf{M}etacognitive \textbf{U}nison \textbf{S}elf-\textbf{E}volution (MUSE), a prototype that instantiates the six-component evolutionary strategy to examine the practical realization of Stage~4 capabilities. MUSE connects the formal components to concrete mechanisms through metacognitive gating, interaction-grounded self-supervision, and coordinated component evolution (Section~\ref{sec:methodology}).

(iv) We conduct controlled experiments and human audits on APPWorld to assess the continuous adaptation process and the capabilities of the resulting agent instances. The analyses provide empirical evidence for the feasibility of the Stage~4 mechanisms instantiated in our proposed MUSE and characterize how their effectiveness varies across component configurations and evolution rounds (Section~\ref{sec:experiments}).

\section{Related Work} 
In this section, we review related work from four perspectives: Autonomous Agents, Traditional Adaptation Paradigms, Agent Evolution, and Metacognition in AI Systems.

\subsection{Autonomous Agents}

The development of autonomous artificial agents spans rule-based systems, knowledge-based assistants, and learning-enabled architectures. Early conversational systems like ELIZA~\citep{weizenbaum1983eliza} and PARRY~\citep{colby1971artificial} generated responses through explicitly specified dialogue mechanisms. Early expert systems such as MYCIN~\citep{shortliffe2012computer} incorporated human-engineered knowledge bases and inference rules for domain-specific problem solving. Subsequent research explored rule-based architectures for semantic-web applications~\citep{dietrich2003rule}, automated rule-based knowledge extraction~\citep{wang2005agent}, and intelligent assistants with adaptation to working styles and preferences~\citep{myers2007intelligent}. These developments illustrate variation in both agent architecture and the organization of adaptation. Stage~1 in our evolutionary spectrum characterizes externally managed maintenance of persistent agent configurations.

With the advent of LLMs~\citep{deepseekai2024deepseekv3technicalreport}, the paradigm shifts substantially toward LLM-powered agents. Contemporary LLM-powered agents are capable of perceiving complex contexts~\citep{packer2023memgpt, zhou2024webarena, xie2024osworld, wu2025smoothing}, decomposing overarching goals~\citep{shen2023hugginggpt, wang2023plan, li2025agentoriented}, reasoning over long horizons~\citep{wang2024voyager, chang2024agentboard, trivedi2024appworld, wu2026see}, and executing intricate action sequences through external tool invocation to solve sophisticated real-world tasks~\citep{yao2022react, wang2024empowering, luo2025large}. However, despite their sophisticated capabilities, contemporary LLM-powered agents share a persistent limitation with their rule-based predecessors. Once deployed, many agents remain largely fixed in their underlying capabilities~\citep{fang2025comprehensive}. Consequently, when faced with inevitable dynamic environmental shifts, such as changes to API documentation~\citep{patil2024gorilla}, out-of-distribution instructions~\citep{wu2025gem}, or tool perturbations~\citep{ye2024rotbench}, these static agents may struggle to adapt to new operational requirements. This persistent post-deployment stasis limits their real-world reliability, highlighting the demand for autonomous post-deployment adaptation.

\subsection{Traditional Adaptation Paradigms}

Post-deployment adaptation has been widely explored across traditional paradigms.

Specifically, \textbf{Online Learning} encompasses frameworks that sequentially adapt model parameters based on incoming data or interaction feedback. This spans classical online optimization~\citep{hoi2021online} and online reinforcement learning (RL), which improves policies through continued environmental interaction and reward feedback~\citep{khetarpal2022towards, zheng2022online}. Related preference-based methods adapt language models through reward-model-guided RL, direct preference optimization, and iterative preference learning~\citep{ouyang2022training, rafailov2023direct, xiong2024iterative}. \textbf{Continual Learning} methods are designed to sequentially acquire knowledge across multiple new tasks throughout the lifetime~\citep{parisi2019continual,wang2024comprehensive, shi2025continual} while mitigating catastrophic forgetting~\citep{kirkpatrick2017overcoming, rolnick2019experience}. Across settings with explicit, blurred, or unavailable task boundaries, continual learning methods preserve prior capabilities while adapting to evolving task distributions through mechanisms including regularization, memory replay, and constrained parameter updates~\citep{wang2024comprehensive, wu2024continual, wang2023orthogonal}.

Beyond sequential task acquisition, \textbf{Test-Time Adaptation (TTA)} adapts pretrained models to unlabeled test data under distribution shifts, including batch-based and online settings. Representative adaptation objectives include prediction entropy minimization~\citep{wang2021tent, liang2025comprehensive}. Concurrently, \textbf{Model Editing} emerges as a targeted post-deployment paradigm for knowledge adaptation~\citep{meng2022locating, yao2023editing, wang2024knowledge}, performing localized updates to specific factual knowledge within LLM parameters without the need for computationally expensive retraining~\citep{chen2024lifelong, fang2025alphaedit}.

These paradigms provide complementary mechanisms for acquiring knowledge, adapting to distribution shifts, and retaining prior capabilities. Agent evolution connects these mechanisms within systems that may contain persistent memories, tools, and other adaptable components. This broader setting calls for an explicit account of how learning signals are constructed, how updates are activated and directed, and how changes across heterogeneous components are coordinated over continued interaction. Formalizing these relationships supports comparison across evolutionary approaches and the identification of capabilities required for greater evolutionary autonomy.

\subsection{Agent Evolution}

Following the definition in Section~\ref{sec:intro}, agent evolution is a continuous adaptation process governed by an \textbf{evolutionary strategy} to maximize environmental fitness and accumulate capabilities. 
Existing studies investigate this process through different adaptation targets, sources of supervision, and forms of evolutionary control.

Driven by the expansive capabilities of LLMs, contemporary research extends this paradigm to LLM-powered agents for post-deployment adaptation. Existing works can be roughly categorized based on their optimization scope (i.e., \textit{what to evolve}). Parameter-centric evolution primarily improves the foundation model weights, as exemplified by GiGPO~\citep{feng2025groupingroup}, SCA~\citep{zhou2025selfchallenging}, and SCoRe~\citep{kumar2025training}, while some recent approaches begin to complement parametric optimization with auxiliary symbolic adaptation. AgentEvolver~\citep{zhai2025agentevolver}, for instance, centers on model optimization while concurrently maintaining an external experience bank. Building on this broader trend, recent works increasingly explore non-parametric evolution. Specifically, \textit{memory evolution} enables agents to accumulate, retrieve, and refine explicit skill libraries or episodic experiences over sequential tasks~\citep{wei2025evo,wang2025annaagent,xu2025amem}. MemEvolve further extends adaptation to the memory architecture, including the procedures governing memory encoding, storage, retrieval, and management~\citep{zhang2025memevolve}. \textit{Toolset evolution} empowers agents to autonomously create, debug, and expand external tools, thereby extending their action spaces~\citep{qian2023creator, xiong2026learning}. Furthermore, related work on \textit{hyperparameter adaptation} includes AgentHPO, which organizes iterative hyperparameter search across machine learning tasks~\citep{liu2025agenthpo}, and adaptive decoding methods, which learn policies for selecting inference-time sampling configurations~\citep{su2026learning}. Lastly, for multi-agent systems (MAS), \textit{topology evolution} focuses on dynamically reconfiguring communication structures, agent roles, and collaborative workflows~\citep{wang2025scoreflow, li2024autoflow,xue2025comas, dang2025multiagent, chen2025multi}.

Beyond optimization scope, agent evolution also depends on the construction and validation of supervision. Learning signals can originate from task verifiers, teacher models, or model-generated assessments~\citep{kumar2025training,zhou2025selfchallenging,madaan2023self,chen2025spc,xia2025agent0}. AgentEvolver combines environment-derived task synthesis, experience-guided exploration, and parametric optimization~\citep{zhai2025agentevolver}. OpenSkill constructs practice tasks and verification signals from independently checkable open-world resources~\citep{yan2026openskill}. Skill-SP connects task proposal and solver optimization through an evolving skill library, with execution feedback driving task generation and skill refinement~\citep{huang2026skillsp}. These developments connect supervision construction with changing agent capabilities, while highlighting the importance of validation evidence when generated assessments may reinforce incorrect model beliefs~\citep{he2026how}.

Broader optimization scope also motivates coordination among component updates. SIA combines harness and weight adaptation under task specifications and verifiers~\citep{hebbar2026sia}. COVE connects memory-assisted parameter training with subsequent reassessment of retained knowledge~\citep{ji2026learning}. Harness adaptation further changes the software governing agent execution. ScholarEvolve incorporates research-derived improvements into functional harness modules and evaluates their combinations~\citep{yang2026scholarevolve}. At the system level, recent work connects revisions of multiple agent components with evaluation and subsequent deployment through a shared versioned lifecycle~\citep{zhang2026agentevolversystem}. These studies reveal different relationships between component adaptation, knowledge transfer, and the organization of subsequent execution.

Recent studies explore initial forms of strategy-level adaptation by modifying selected procedures governing subsequent evolution. HyperAgents makes task-solving and self-modification logic editable within a common program~\citep{zhang2026hyperagents}. MetaSkill-Evolve revises a meta-skill that governs diagnosis, retrieval, search allocation, edit proposal, and skill refinement~\citep{wang2026metaskill}. These explorations represent partial progress toward recursive meta-evolution and motivate explicit criteria for distinguishing individual strategy revisions from the cumulative capabilities required for Stage~5 in our proposed spectrum.

Recent conceptual frameworks also characterize different forms of evolutionary progression. \citet{wang2026reliableselfevolving} distinguish five levels according to the deepest evolution target, spanning outputs, models, scaffolds, improvers, and criteria. \citet{zong2026coevolution} describe three stages involving adaptation among agents, adaptation between agents and environments, and meta co-evolution. Our work specifies six functional components of an evolutionary strategy and defines joint, cumulative capability requirements across evolutionary autonomy, supervision autonomy, and optimization scope and coordination. MUSE provides a prototype for investigating the realization of Stage~4 capabilities through concrete mechanisms and experimental evaluation.

\subsection{Metacognition in AI Systems}
Metacognition concerns the monitoring and regulation of cognitive processes~\citep{flavell1979metacognition}. For LLM-powered agents, metacognition is primarily operationalized through self-monitoring, self-evaluation, and self-reflection. For instance, pioneering works explore the abilities to estimate internal confidence and calibrate uncertainty~\citep{kadavath2022language, yin2023large}. Building upon this, Self-Refine iteratively revises generated responses through self-feedback~\citep{madaan2023self}, while Reflexion retains linguistic reflections in episodic memory to guide subsequent attempts~\citep{shinn2023reflexion}. These approaches establish reflective feedback as a means of improving generated responses and subsequent task execution. Continued agent evolution further raises questions about how reflective evidence directs the timing and targets of persistent adaptation.

Recent agent evolution methods connect capability diagnosis with persistent adaptation. SkillMentor learns to identify blind spots and construct corrective skills for a frozen executor~\citep{bao2026skillmentor}, while SkillProx refines structured skills through execution-based diagnosis, update validation, and feedback from previous revisions~\citep{zheng2026skillprox}. COVE extends evolutionary control to memory and parameter adaptation through feedback-conditioned routing and stage-aware scheduling~\citep{ji2026learning}. SEES directs failure diagnosis toward targeted adaptation of skill-family policies~\citep{li2026sees}. These methods establish concrete control mechanisms for skill refinement, memory and parameter coordination, or policy adaptation. However, their formulations provide limited treatment of joint target selection across multiple symbolic components with distinct functions and model parameters. Extending control across this broader space calls for an explicit account of how capability evidence guides evolution activation, diagnosis, and the allocation of component updates. Our six-component formulation represents these decisions through evolutionary gating and specifies their relationship with supervision generation and component optimization. MUSE instantiates this organization in a prototype targeting Stage~4 capabilities, supporting experimental examination of evolutionary control within coordinated symbolic and parametric adaptation.

\section{A Unified Formalization of Agent Evolution} \label{sec:formalization}

As research on agent evolution expands, existing approaches have developed along diverse paradigms, differing in how evolution is initiated, supervised, and implemented across heterogeneous agent components. This growing diversity underscores the value of establishing a unified conceptual framework to characterize the shared structure of different evolutionary methods and support their analysis across different stages of development. Such a formulation provides a consistent foundation for analyzing agent evolution beyond specific implementations.

To this end, we formalize agent evolution from a macroscopic perspective centered on its core construct, the \textbf{evolutionary strategy}, which provides a structured representation of how continuous adaptation is organized and accommodates diverse historical, contemporary, and prospective paradigms. We first define the interaction context and compositional architecture of the evolvable agent system in Section~\ref{subsec:evolution_context}. We then characterize the evolutionary strategy through a six-component representation in Section~\ref{subsec:evolutionary_strategy}. Finally, Section~\ref{subsec:evolution_process} describes how agent-environment interactions and successive evolution rounds constitute a continuous evolutionary trajectory.

\subsection{Evolution Context and Agent System} \label{subsec:evolution_context}
Agent evolution unfolds through continued interaction with an external environment while the agent may undergo successive adaptation over deployment. We therefore begin by specifying the interaction context in which evolutionary evidence is accumulated, followed by the internal system that constitutes the concrete object of evolution. Throughout this paper, $\ell$ indexes environment interaction steps, whereas $t$ indexes evolution rounds.

\begin{definition}[\textbf{Interaction Environment and History}]
\label{def:environment}
We model the external environment $E \in \mathcal{E}$ as a partially observable Markov decision process (POMDP)~\citep{kurniawati2022partially}. We consider a single execution $i$ and omit its index from individual states, observations, goals, actions, and feedback. In this formulation, the environment operates on hidden states $z_{\ell} \in \mathcal{Z}$. At environment interaction step $\ell$, given the current goal $g_{\ell} \in \mathcal{G}$, the agent $A \in \mathcal{A}$ receives an observation $o_{\ell} \in \Omega$ associated with the current state $z_{\ell}$. Let $H_{i,\mathrm{act}}^{(k)}=\bigl((o_j,g_j,u_j,r_j)\bigr)_{j=1}^{k}\in\mathcal{H}$ denote the ordered record of the first $k$ interaction steps, with feedback recorded when available to the agent. The agent selects $u_\ell\in\mathcal U$ according to its action policy $u_\ell\sim\pi_A(\cdot \mid H_{i,\mathrm{act}}^{(\ell-1)},o_\ell,g_\ell)$. Before the first action, the record is empty and the decision depends on $o_1$ and $g_1$. The environment receives the action $u_{\ell}$ and transitions to a new hidden state $z_{\ell+1}$ governed by the Markovian transition dynamics $\mathcal{P}(z_{\ell+1} | z_{\ell}, u_{\ell})$. The resulting interaction may additionally produce an evaluative feedback signal $r_{\ell}$ via a feedback function $\mathcal{R}: \mathcal{Z} \times \mathcal{U} \times \mathcal{G} \rightarrow \mathcal{Y}$, where $r_{\ell} = \mathcal{R}(z_{\ell}, u_{\ell}, g_{\ell})$. Depending on the environment, $r_{\ell} \in \mathcal{Y}$ can manifest as a scalar reward or dense textual critique. The agent then receives this feedback alongside the next observation $o_{\ell+1}$. The resulting interaction trajectory provides empirical evidence for post-deployment adaptation.
\end{definition}

\begin{figure}[!t]
    \centering

    \begin{subfigure}{0.482\linewidth}
        \centering
        \includegraphics[width=\linewidth]{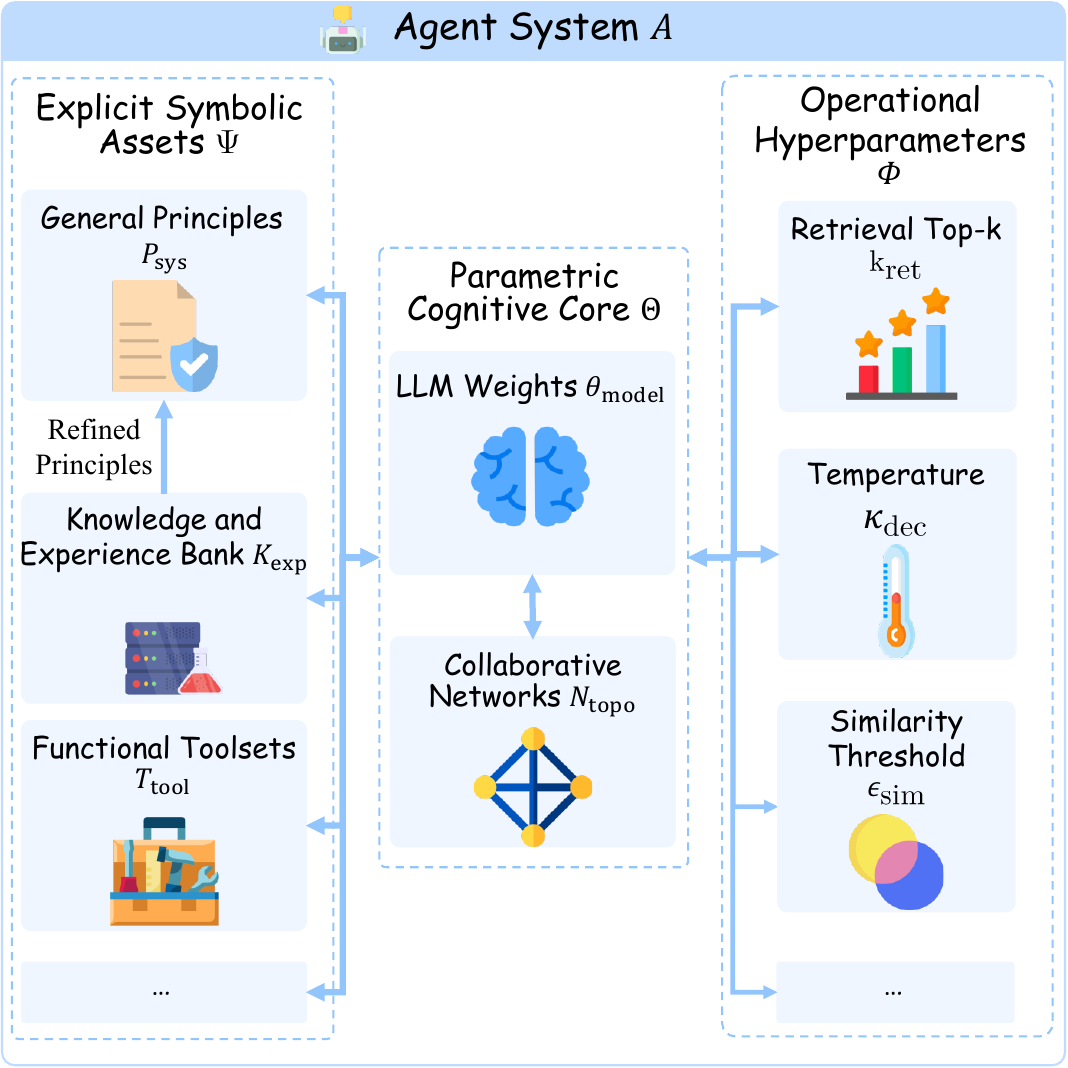}
        \caption{Agent System}
        \label{subfig:agent_system}
    \end{subfigure}
    \hfill
    \begin{subfigure}{0.48\linewidth}
        \centering
        \includegraphics[width=\linewidth]{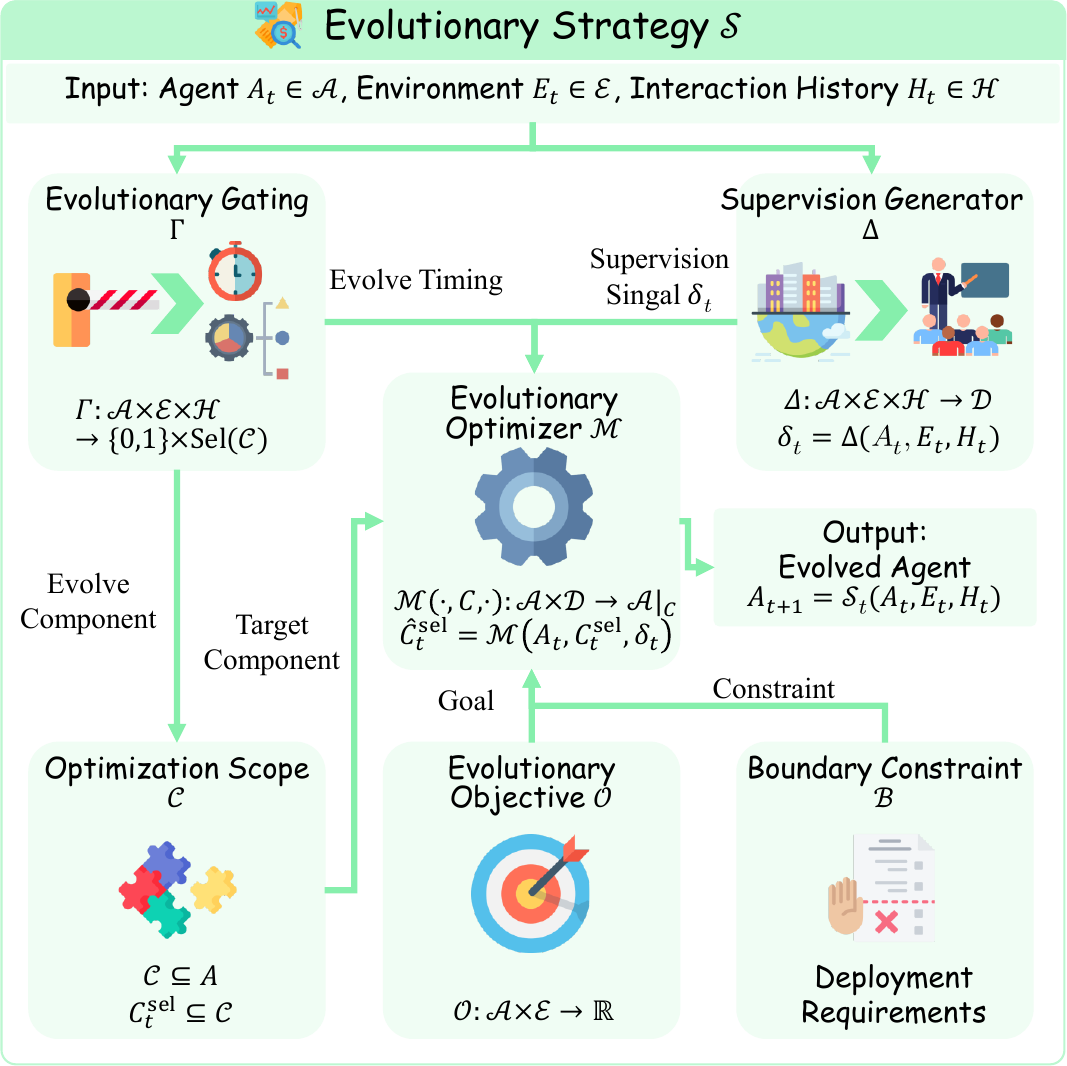}
        \caption{Evolutionary Strategy}
        \label{subfig:evolutionary_strategy}
    \end{subfigure}

    \vspace{-0.2cm}

    \caption{The illustration of the agent evolution framework. 
    (a) The \textbf{Agent System} ($A$) is conceptualized as a hybrid architecture comprising the parametric cognitive core ($\Theta$), explicit symbolic assets ($\Psi$), and operational hyperparameters ($\Phi$). 
    (b) The \textbf{Evolutionary Strategy} ($\mathcal{S}$) is formulated as a six-component structure $\langle \Delta, \mathcal{C}, \mathcal{M}, \Gamma, \mathcal{O}, \mathcal{B} \rangle$, which characterizes the supervision generation, optimization scope, optimization method, evolutionary gating, adaptation objective, and boundary constraints throughout the continuous adaptation trajectory.}

    \label{fig:definition_illustration}
    \vspace{-2mm}
\end{figure}

The interaction context $(E, H)$ captures the external environment and behavioral evidence that may guide subsequent evolution. The corresponding adaptation is realized through changes to the internal structures of the agent. We therefore formalize the evolvable object $A$ as a structured, modular system spanning parametric reasoning, explicit symbolic assets, and operational control configurations.

\begin{definition}[\textbf{Agent System}]
\label{def:agent_system}

Modern LLM-powered agents operate as hybrid systems integrating multiple interconnected components to coordinate reasoning, knowledge retrieval, tool execution, and inference behaviors. We formally define the agent system as $A = \Theta \cup \Psi \cup \Phi$, as illustrated in Figure~\ref{subfig:agent_system}, where:

\begin{enumerate}[label=(\arabic*), leftmargin=*, noitemsep, topsep=2pt]

    \item $\Theta$: The \textit{cognitive core}, serving as the central reasoning engine to internalize knowledge and orchestrate decisions. In an LLM-powered agent, the cognitive core includes model weights $\theta_{\text{model}}$ for reasoning and may also include a collaborative topology $N_{\text{topo}}$ in multi-agent settings.

    \item $\Psi$: The \textit{explicit symbolic assets}, providing non-parametric reference knowledge and actionable interfaces to ground the agent in specific environments. This category broadly encapsulates semantic resources and executable assets including general principles defined in system instructions $P_{\text{sys}}$, external knowledge and experience banks $K_{\text{exp}}$, functional toolsets $T_{\text{tool}}$, and other extensible symbolic assets.
  
    \item $\Phi$: The \textit{operational hyperparameters}, governing the execution logic and inference heuristics of the system, such as retrieval top-$k$ capacities $k_{\text{ret}}$, decoding temperatures $\kappa_{\text{dec}}$, and similarity thresholds $\epsilon_{\text{sim}}$.
\end{enumerate}

\end{definition}

The decomposition $A = \Theta \cup \Psi \cup \Phi$ provides a common representation of the architectural space over which adaptation may operate. Different evolutionary paradigms can involve different subsets of this space according to their objectives and design choices. Together, the interaction context $(E,H)$ and the evolvable agent system $A$ establish the basic elements required to describe an evolutionary process. Section~\ref{subsec:evolutionary_strategy} next introduces the \textbf{evolutionary strategy} as a structured representation of how such adaptation is organized.

\subsection{Evolutionary Strategy} \label{subsec:evolutionary_strategy}

Having specified the interaction context $(E,H)$ and the evolvable agent system $A$, characterizing agent evolution requires a formal representation of how adaptation is structured over time. Concrete evolutionary methods differ in how they update the agent configuration and coordinate changes across its components in response to available interaction evidence and adaptation requirements. We refer to the structured representation of this continuous adaptation process throughout deployment as the \textbf{evolutionary strategy}.

\begin{definition}[\textbf{Evolutionary Strategy}]
\label{def:evolutionary_strategy}

We represent the evolutionary strategy as a mapping $\mathcal{S}: \mathcal{A} \times \mathcal{E} \times \mathcal{H} \rightarrow \mathcal{A}$, which provides a macroscopic formulation of how adaptation proceeds given the current agent, environment, and accumulated interaction history over successive evolution rounds. We further decompose $\mathcal{S}$ into a six-component formulation: $\mathcal{S} = \langle \Delta, \mathcal{C}, \mathcal{M}, \Gamma, \mathcal{O}, \mathcal{B} \rangle$, as illustrated in Figure~\ref{subfig:evolutionary_strategy}, where:

\begin{enumerate}[label=(\arabic*), leftmargin=*, noitemsep, topsep=2pt]

    \item \textbf{Supervision Generator} ($\Delta$): The supervision generator $\Delta:\mathcal{A}\times\mathcal{E}\times\mathcal{H}\rightarrow\mathcal{D}$ characterizes how learning signals for evolution are acquired or synthesized. Here, $\mathcal{D}$ denotes the space of supervision signals $\delta$, which may encompass learnable trajectories, oracle labels, evaluative feedback, or reward assignment criteria. The concrete source and construction process of $\delta_t \in \mathcal{D}$ depend on the corresponding evolutionary strategy. Supervision may be provided externally through oracle labels, teacher models, or evaluators, or synthesized from evidence accumulated during agent-environment interaction. At round $t$, the available supervision is derived via $\delta_t = \Delta(A_t,E_t,H_t)$. An external oracle may supply labels independently of $A_t$, while interaction-grounded self-supervision can draw on the current agent to interpret the interaction evidence.

    \item \textbf{Optimization Scope} ($\mathcal{C}$): The optimization scope specifies the admissible space of components where adaptation can operate. For agent-level adaptation, the optimization scope $\mathcal{C} \subseteq A$ specifies the set of agent components eligible for refinement. At round $t$, $C_t^{\mathrm{sel}}\subseteq\mathcal{C}$ identifies the selected components, and $A_t$ supplies their current configuration. Accordingly, $\mathcal{C}$ characterizes \textit{what can evolve} under a given evolutionary strategy, ranging from individual components to broader combinations of parametric, symbolic, and operational structures.

    \item \textbf{Evolutionary Optimizer} ($\mathcal{M}$): For a component subset $C\subseteq\mathcal C$ selected by $\Gamma$, the evolutionary optimizer acts as a mapping $\mathcal M(\cdot,C,\cdot):\mathcal A\times\mathcal D\rightarrow\mathcal A|_C$, where $\mathcal A|_C$ denotes the space of configurations of the components in $C$. The current agent $A_t$ provides the context for cross-component synchronization. Given supervision $\delta_t$, the updated configuration of the selected components is expressed as $\hat{C}_t^{\mathrm{sel}} = \mathcal{M}(A_t, C_t^{\mathrm{sel}}, \delta_t)$. This component characterizes \textit{how} the selected parts of the agent are adapted within the current evolution round. Depending on the evolutionary strategy, updates may involve persistent information accumulation, component-specific refinement, or coordinated updates across multiple components.

    \item \textbf{Evolutionary Gating} ($\Gamma$): A decision and routing mapping $\Gamma: \mathcal{A} \times \mathcal{E} \times \mathcal{H} \rightarrow \{0, 1\} \times \operatorname{Sel}(\mathcal C)$ that determines whether an evolutionary update is activated and which components are selected when adaptation occurs. Here, $\operatorname{Sel}(\mathcal C)$ denotes the family of component subsets that the gating permits to activate, including the empty set. The \textit{activation} process determines \textit{when} evolution is initiated according to the available environmental, agent, and historical evidence. Once activated, the \textit{routing} process selects a component subset $C_t^{\mathrm{sel}} \subseteq \mathcal{C}$, thereby specifying \textit{what is evolved in the current round} within the admissible optimization space. At round $t$, $\Gamma(A_t,E_t,H_t)=(\gamma_{\text{act}},C_t^{\mathrm{sel}})$, with $C_t^{\mathrm{sel}}=\varnothing$ when $\gamma_{\text{act}}=0$. Depending on the evolutionary strategy, activation and routing may be instantiated through manual interventions, predefined schedules, task-level feedback, or adaptive capability assessment and diagnosis of current agent capabilities.

    \item \textbf{Evolutionary Objective} ($\mathcal{O}$): The evaluation mapping $\mathcal{O}: \mathcal{A} \times \mathcal{E} \rightarrow \mathbb{R}$ specifies the quantitative criteria adopted to assess adaptation and orient the evolution process. Its concrete form may reflect environmental fitness, task performance, operational efficiency, or other domain-specific deployment objectives.

    \item \textbf{Boundary Constraint} ($\mathcal{B}$): The operational boundary specifies the deployment requirements that govern agent updates and the resulting behavior, such as safety, alignment, and capability retention. An evolutionary strategy may check a requirement explicitly or support it implicitly through the design of its updates.
\end{enumerate}

\end{definition}

The six components describe functional roles within an evolutionary strategy. Their specification and execution may involve the agent, external procedures, or human operators, depending on the concrete instantiation.

A key property of this formulation is the explicit separation between architectural capacity and evolutionary control. The optimization scope $\mathcal{C}$ defines the overall space of components eligible for adaptation, whereas the gating mechanism $\Gamma$ selects targeted subsets from this space according to the current evolutionary context and the corresponding control rule. This separation allows a strategy to support adaptation across multiple components while selecting a suitable subset for each evolution round.

\subsection{Agent Evolution Process} \label{subsec:evolution_process}

Agent evolution can be viewed across two temporal scales. At the interaction scale, indexed by $\ell$ with execution $i$, the agent operates within the environment and accumulates observations, actions, and evaluative feedback into its interaction history $H_{i,\mathrm{act}}^{(\ell)}$. At the evolution scale, indexed by $t$, the agent is updated through the corresponding evolutionary strategy $\mathcal{S}$. This separation distinguishes ongoing task execution from structural system adaptation, while the accumulated interaction history may provide empirical evidence for subsequent evolution rounds.

Following the definition introduced in Section~\ref{sec:intro}, we characterize agent evolution as a continuous adaptation process governed by the evolutionary strategy $\mathcal{S}$. Across successive evolution rounds, $\mathcal{S}$ organizes the adaptation of the agent configuration under the guidance of the evolutionary objective $\mathcal{O}$ and subject to the boundary constraint $\mathcal{B}$. The resulting update may therefore vary across different instantiations of $\mathcal{S}$. Let $\mathcal{S}_t$ denote the evolutionary strategy at round $t$, with $E_t$ and $H_t$ denoting the environment and accumulated interaction history at that round. Formally, the transition from the current agent $A_t$ at evolution round $t$ to the subsequent instance $A_{t+1}$ is represented as:
\begin{equation}
    \label{equ:evolutionary_strategy}
    A_{t+1} = \mathcal{S}_t(A_t, E_t, H_t).
\end{equation}

Equation~\ref{equ:evolutionary_strategy} describes a single evolution round at the macroscopic level. The gating mechanism $\Gamma$ controls the activation of evolution and routes the targeted component subset from $\mathcal{C}$. When adaptation is activated, the supervision generator $\Delta$ provides the corresponding learning signals and the optimizer $\mathcal{M}$ updates the selected components. The update proceeds under the objective $\mathcal{O}$ and remains subject to the boundary constraint $\mathcal{B}$. If no update is activated, the agent configuration remains unchanged, i.e., $A_{t+1}=A_t$.

After an evolution round, the updated agent resumes interaction with the environment. Subsequent interactions may reveal that the current adaptation is still insufficient for the existing environment, leaving additional adaptation needs to be addressed in later rounds. Alternatively, dynamic shifts in the environment may introduce new capability requirements after the agent has already reached satisfactory performance under previous conditions. Repeated interaction and adaptation can therefore give rise to a continuous evolutionary trajectory over deployment:
\begin{equation}
    A_0
    \xrightarrow{\mathcal{S}_0(\cdot,E_0,H_0)}
    A_1
    \xrightarrow{\mathcal{S}_1(\cdot,E_1,H_1)}
    \cdots
    \xrightarrow{\mathcal{S}_{T-1}(\cdot,E_{T-1},H_{T-1})}
    A_T .
    \label{equ:evolutionary_trajectory}
\end{equation}

Equation~\ref{equ:evolutionary_trajectory} describes the resulting evolutionary trajectory over $T$ evolution rounds. Each transition corresponds to an instantiation of the evolutionary strategy under the environment and interaction history available at that round. In a fixed environment, this process may stabilize once the desired adaptation objective is reached. During longer-term deployment, further evolution may occur when adaptation remains incomplete or when environmental changes introduce new requirements. The resulting trajectory provides a basis for evaluating capability changes across successive rounds under the specified objectives and operational boundaries.

A fixed strategy corresponds to $\mathcal{S}_t=\mathcal{S}$ across evolution rounds. For strategy-level evolution, the evolving state extends to $(A_t,\mathcal{S}_t)$, and the admissible optimization scope additionally includes representations of selected mechanisms within $\mathcal{S}_t$. This extension permits persistent changes to the mechanisms governing subsequent adaptation and provides the basis for recursive meta-evolution discussed in Section~\ref{sec:spectrum}.

Building on this formulation, the next section examines how the maturity of different evolutionary strategies can be characterized and organizes their developmental progression through the Five-Stage Evolutionary Spectrum.

\section{A Five-Stage Evolutionary Spectrum} \label{sec:spectrum}

Building on the formalization in Section~\ref{sec:formalization}, we characterize the developmental maturity of evolutionary strategies through the \textbf{Five-Stage Evolutionary Spectrum}. The spectrum describes the progression of agent evolution from externally managed maintenance to recursive adaptation of the evolutionary strategy itself. Section~\ref{subsec:evaluate_dimensions} introduces three evaluative dimensions: evolutionary autonomy, supervision autonomy, and optimization scope and coordination. Section~\ref{subsec:spectrum_framework} then defines the five stages according to their joint maturity across these dimensions. Finally, Section~\ref{subsec:evaluation_metrics} introduces complementary criteria for evaluating the empirical effectiveness of agent evolution.

\subsection{Evaluative Dimensions} \label{subsec:evaluate_dimensions}

Section~\ref{sec:formalization} provides a common structural representation of how agent evolution is organized through the evolutionary strategy $\mathcal{S}$. On this basis, evolutionary maturity can be further examined from three complementary perspectives. These dimensions characterize the degree of autonomous control over evolution, the autonomy of supervision generation, and the breadth and coordination of autonomous component adaptation.

\noindent {$\blacktriangleright$} \textbf{Evolutionary Autonomy (EA):} 
This dimension primarily maps to the evolutionary gating $\Gamma$ and characterizes how adaptation is activated and routed. At lower maturity, update timing and adaptation targets are primarily specified through human intervention or predefined schedules. More adaptive strategies introduce failure diagnosis, feedback-conditioned scheduling, or component routing. Higher evolutionary autonomy integrates these functions into a metacognitive control process that draws on accumulated interaction evidence to assess current adaptation needs, regulate evolution activation, and select suitable components for refinement. Evolutionary autonomy therefore reflects the extent to which evolutionary control is internalized within the agent system.

\noindent {$\blacktriangleright$}  \textbf{Supervision Autonomy (SA):} 
This dimension primarily maps to the supervision generator $\Delta$ and characterizes how independently learning signals are acquired, constructed, and validated for evolution. Lower-autonomy evolutionary strategies rely mainly on externally prepared tasks, annotations, or evaluation criteria. Partial self-supervision reduces this dependence through self-generated tasks, reflections, pseudo-labels, or other internally constructed signals. Greater autonomy further involves grounding and validating these signals through evidence obtained from environmental interaction. At higher maturity, supervision generation responds to identified adaptation needs, supporting continued learning when task-specific external supervision is constrained.

\noindent {$\blacktriangleright$} \textbf{Optimization Scope and Coordination (SC):}
This dimension characterizes the breadth of autonomous component adaptation and the coordination among component updates. It maps to the optimization scope $\mathcal{C}$, together with the evolutionary gating $\Gamma$ and optimizer $\mathcal{M}$. At lower maturity, automated updates are concentrated on persistent information stores or individual components. Broader strategies may expose model parameters, memories, principles, tools, skills, harnesses, collaborative structures, and operational configurations to evolution. Higher maturity additionally supports coordinated adaptation across heterogeneous components while accounting for dependencies among their updates. Evaluations of the joint configuration guide subsequent component selection and refinement. The gating mechanism $\Gamma$ selects suitable subsets from the available space for each evolution round.

Taken together, evolutionary autonomy, supervision autonomy, and optimization scope and coordination form a multidimensional profile of evolutionary maturity. The three dimensions may develop at different rates, and the maturity of an evolutionary strategy is therefore characterized by their joint configuration. The \textbf{Five-Stage Evolutionary Spectrum} introduced next specifies cumulative requirements across these dimensions, with each higher stage retaining preceding capabilities and adding new evolutionary capabilities. Table~\ref{tab:evolutionary_capability_checklist} records which of these requirements each assessed framework demonstrates.

\subsection{Spectrum Framework} \label{subsec:spectrum_framework}

\begin{figure}[!t]
    \centering
    \includegraphics[width=\linewidth]{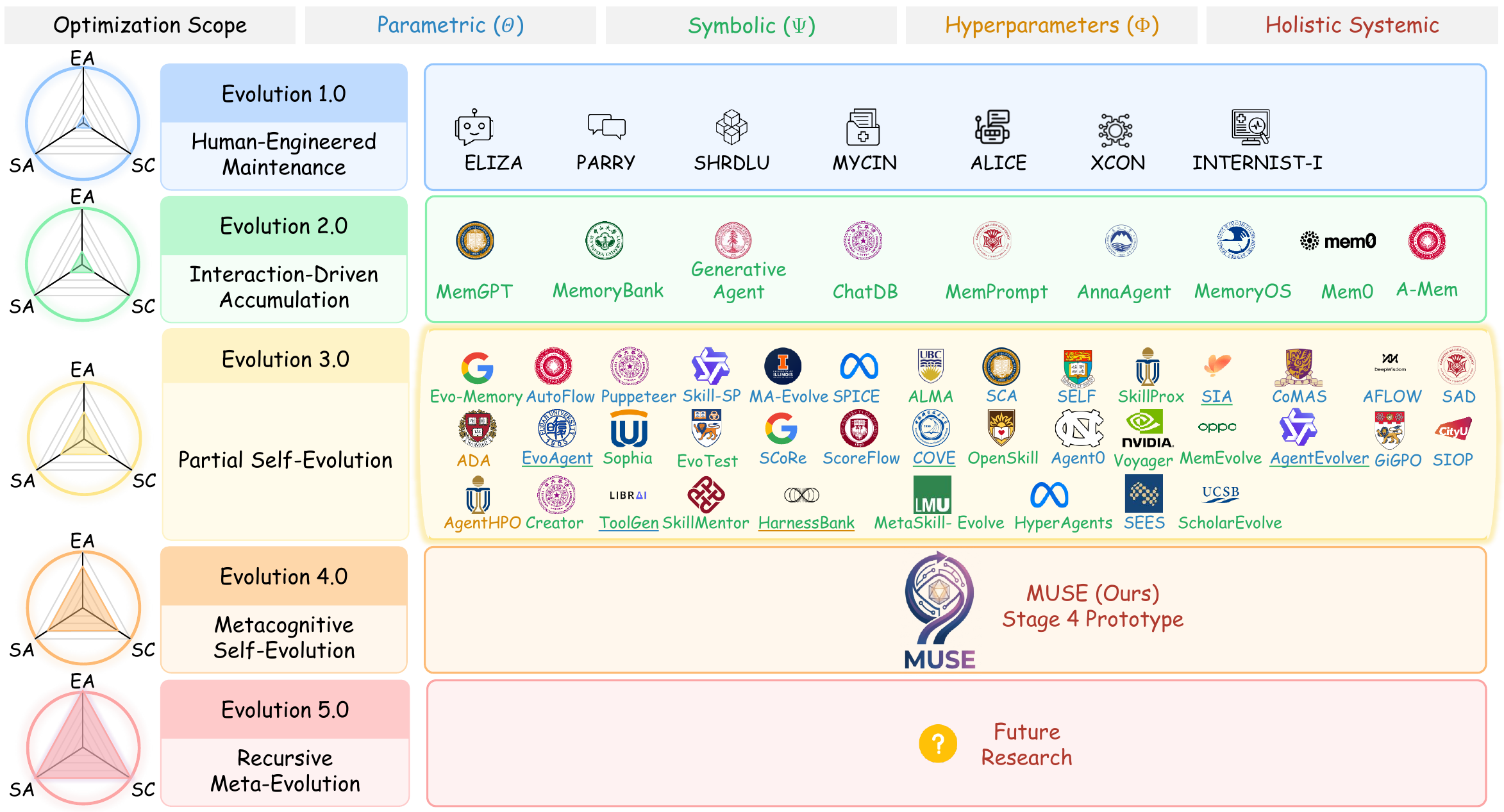}\vspace{-0.2cm}

    \caption{Representative works across the Five-Stage Evolutionary Spectrum. Font color marks the primary optimization scope of each method and underlining marks secondary scopes. The stage profiles qualitatively illustrate EA, SA, and SC. Stage assignment follows Table~\ref{tab:evolution_roadmap} and Equation~\ref{equ:evolutionary_stage_index}, with individual judgments in Table~\ref{tab:evolutionary_capability_checklist}.}

    \label{fig:spectrum_representative_work}
    \vspace{-2mm}
\end{figure}

Building on the three evaluative dimensions above, we systematically organize agent evolution into five stages through cumulative capability requirements. Stage~1 forms the baseline of externally managed maintenance. Stages~2 through~5 introduce successive requirements for persistent adaptation, feedback-guided optimization, metacognitive self-evolution, and recursive refinement of evolutionary strategies. An agent is assigned to the highest stage whose requirements, together with all preceding requirements, have been demonstrated under declared deployment conditions. Table~\ref{tab:evolution_roadmap} specifies the capability requirements of each stage across the three dimensions, while Figure~\ref{fig:spectrum_representative_work} illustrates representative works within this framework.

\begin{table*}[!t]
\centering
\begingroup
\color{black}
\small
\setlength{\tabcolsep}{5pt}
\renewcommand{\arraystretch}{1.3}

\caption{Capability requirements of the Five-Stage Evolutionary Spectrum across evolutionary autonomy (\textbf{EA}), supervision autonomy (\textbf{SA}), and optimization scope and coordination (\textbf{SC}). Stage~1 is the baseline. Requirements introduced at Stages~2 through~5 are cumulatively retained at later stages.}
\label{tab:evolution_roadmap}

\begin{tabularx}{\textwidth}{
    >{\hsize=0.85\hsize\linewidth=\hsize\raggedright\arraybackslash}X
    >{\hsize=1.05\hsize\linewidth=\hsize\hyphenpenalty=20\exhyphenpenalty=20\tolerance=1500\emergencystretch=1.5em\arraybackslash}X
    >{\hsize=1.15\hsize\linewidth=\hsize\hyphenpenalty=20\exhyphenpenalty=20\tolerance=1500\emergencystretch=1.5em\arraybackslash}X
    >{\hsize=0.95\hsize\linewidth=\hsize\hyphenpenalty=20\exhyphenpenalty=20\tolerance=1500\emergencystretch=1.5em\arraybackslash}X
}
\toprule
\rowcolor{muse_soft!80}
\textbf{\color{muse_deep}Evolutionary Stage}
& \raggedright\arraybackslash\textbf{\color{muse_deep}Evolutionary Autonomy (EA)}
& \raggedright\arraybackslash\textbf{\color{muse_deep}Supervision Autonomy (SA)}
& \raggedright\arraybackslash\textbf{\color{muse_deep}Optimization Scope and Coordination (SC)} \\
\midrule

\rowcolor{muse_warmgray}
\textbf{\colorbox{spectrum_stage_1}{1.0:}\newline\colorbox{spectrum_stage_1}{Human-Engineered}\newline\colorbox{spectrum_stage_1}{Maintenance}}\par\vspace{0.1cm}\includegraphics[width=1cm]{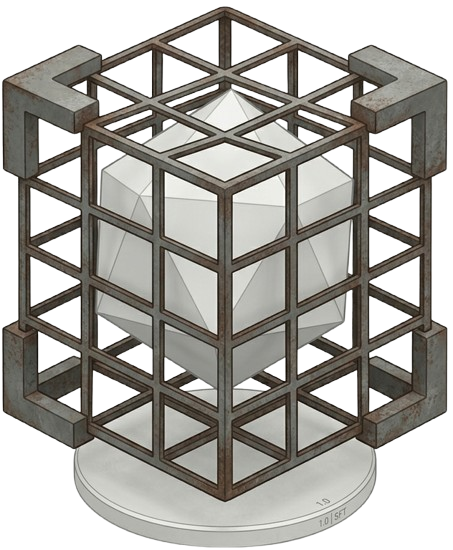}
& \textcolor{royalblue}{\textbf{External control}} determines update timing, adaptation targets, and maintenance procedures.
& Tasks, knowledge, annotations, or evaluation criteria are \textcolor{royalblue}{\textbf{externally prepared and organized}}.
& Persistent agent configurations are modified through \textcolor{royalblue}{\textbf{external maintenance}}.
\\
\midrule

\textbf{\colorbox{spectrum_stage_2}{2.0:}\newline\colorbox{spectrum_stage_2}{Interaction-Driven}\newline\colorbox{spectrum_stage_2}{Accumulation}}\par\vspace{0.1cm}\includegraphics[width=1.2cm]{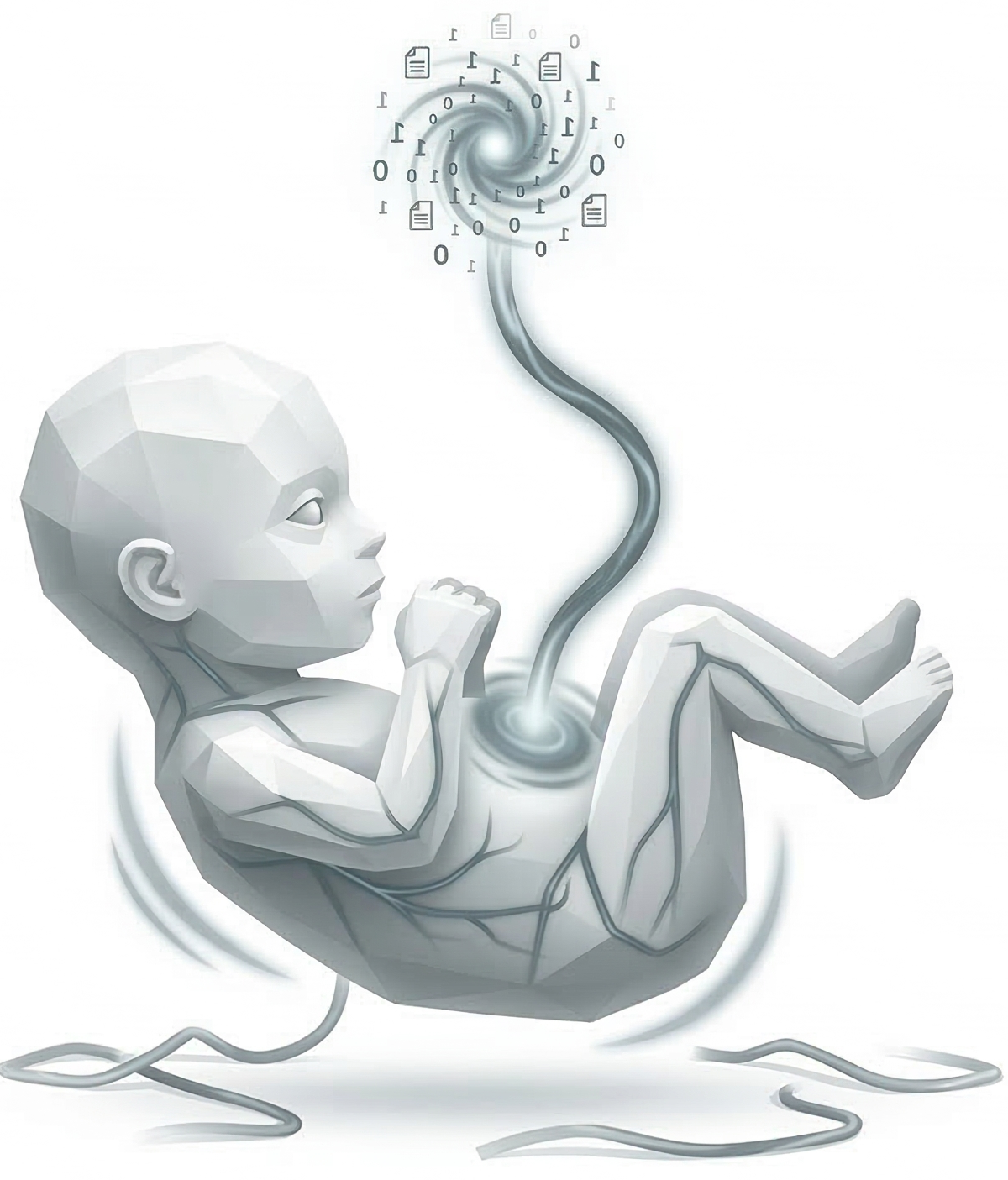}
& (1) \textcolor{royalblue}{\textbf{Automatic activation}} of persistent state updates in response to interaction events.
& (1) \textcolor{royalblue}{\textbf{Automatic acquisition and organization}} of environmental interaction evidence for persistent adaptation.
& (1) \textcolor{royalblue}{\textbf{Persistent accumulation}} of interaction experience within agent components, guiding behavior on later independent tasks.
\\
\midrule

\rowcolor{muse_warmgray}
\textbf{\colorbox{spectrum_stage_3}{3.0: Partial}\newline\colorbox{spectrum_stage_3}{Self-Evolution}}\par\vspace{0.1cm}\includegraphics[width=1.5cm]{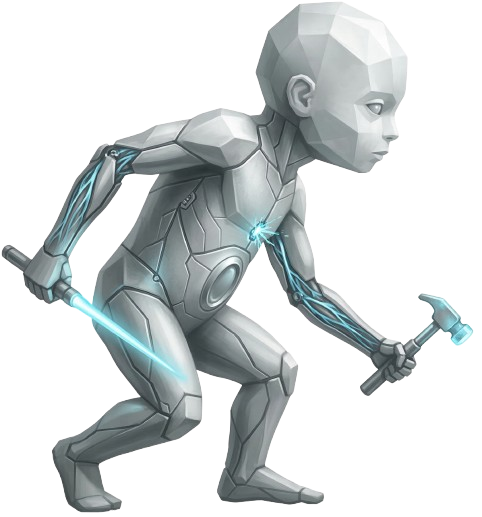}
& (1) \textcolor{royalblue}{\textbf{Scheduled or event-triggered activation}} of configured optimization cycles for adaptation targets specified by the procedure.
& (1) \textcolor{royalblue}{\textbf{Supervision for component optimization}} organized from oracle labels, verifier feedback, or model-generated assessments.
& (1) \textcolor{royalblue}{\textbf{Selected component optimization}}, with evaluation of updated task behavior guiding configuration retention or further refinement.
\\
\midrule

\textbf{\colorbox{spectrum_stage_4}{4.0: Metacognitive}\newline\colorbox{spectrum_stage_4}{Self-Evolution}}\par\vspace{0.1cm}\includegraphics[width=1.5cm]{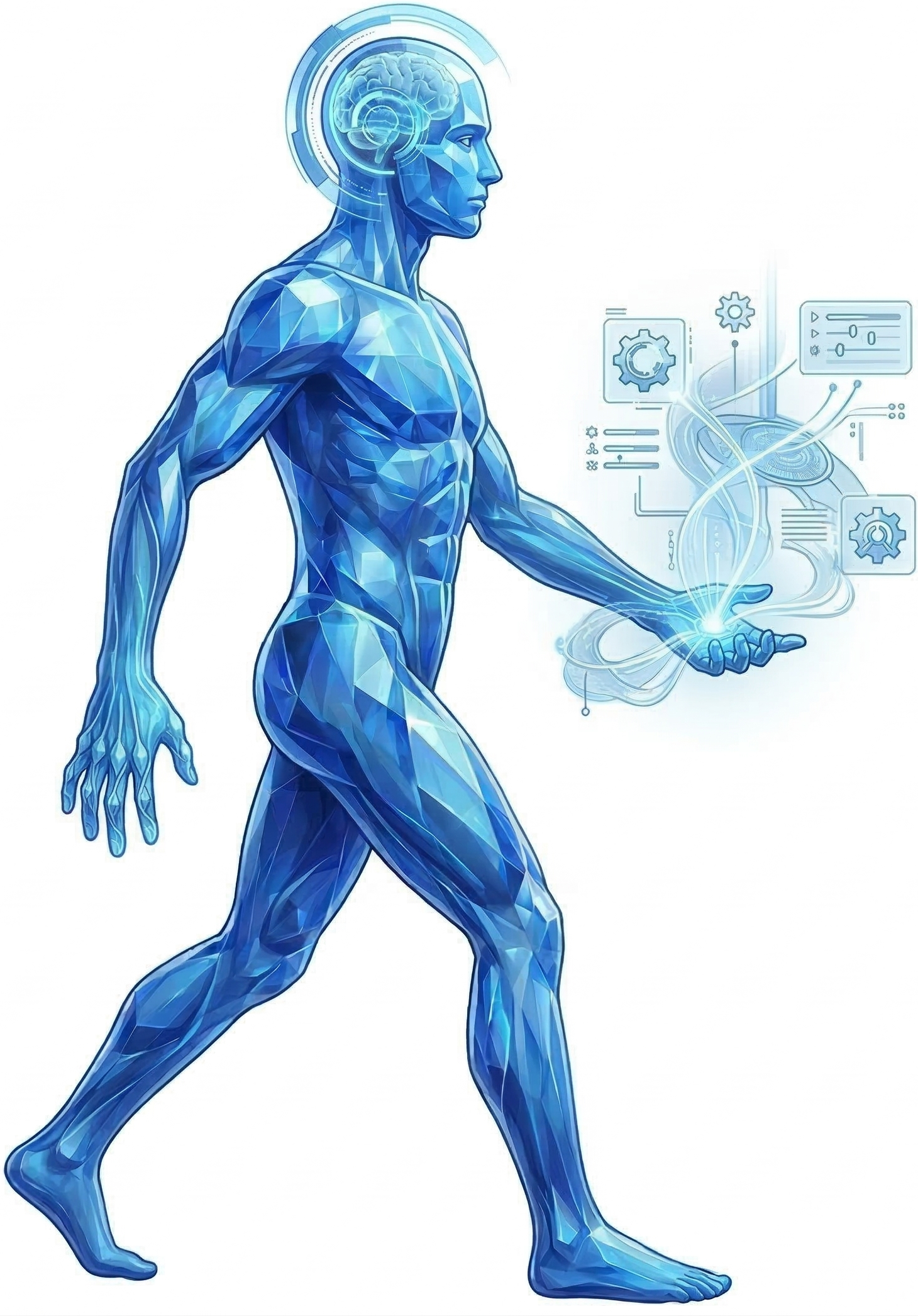}
& (1) \textcolor{royalblue}{\textbf{Autonomous activation}} of evolution in response to observed adaptation needs.\newline
  (2) \textcolor{royalblue}{\textbf{Deferral}} of evolution, continuing interaction when activation criteria are unmet.\newline
  (3) \textcolor{royalblue}{\textbf{Target selection}} among functionally distinct eligible components according to diagnosed adaptation needs.\newline
  (4) \textcolor{royalblue}{\textbf{Diagnosis}} of current adaptation needs, including capability deficiencies and consolidation opportunities, from accumulated interaction evidence.
& (1) \textcolor{royalblue}{\textbf{Autonomous construction of supplementary tasks}} from environmental interactions.\newline
  (2) \textcolor{royalblue}{\textbf{Autonomous construction of evaluative signals}} for interaction-grounded self-supervision.\newline
  (3) \textcolor{royalblue}{\textbf{Validity assessment}} of tasks and evaluative signals against environmental feedback and execution evidence.\newline
  (4) \textcolor{royalblue}{\textbf{Supervision selection}} according to current agent capabilities.\newline
  (5) \textcolor{royalblue}{\textbf{Supervision renewal}} according to current agent capabilities.
& (1) \textcolor{royalblue}{\textbf{Coordinated updates}} across at least two of the cognitive core ($\Theta$), the symbolic assets ($\Psi$), and the operational hyperparameters ($\Phi$).\newline
  (2) \textcolor{royalblue}{\textbf{Joint configuration evaluation}} guiding subsequent component selection and refinement.\newline
  (3) \textcolor{royalblue}{\textbf{Observed update dependencies}} guiding subsequent component selection and refinement.
\\
\midrule

\rowcolor{muse_warmgray}
\textbf{\colorbox{spectrum_stage_5}{5.0: Recursive}\newline\colorbox{spectrum_stage_5}{Meta-Evolution}}\par\vspace{0.1cm}\includegraphics[width=1.5cm]{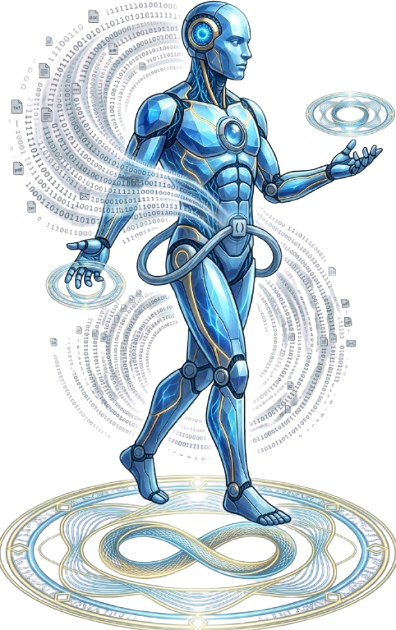}
& (1) \textcolor{royalblue}{\textbf{Autonomous timing}} of revisions to the evolutionary strategy $\mathcal{S}$.\newline
  (2) \textcolor{royalblue}{\textbf{Selection of strategy components}} of the evolutionary strategy to modify.
& (1) \textcolor{royalblue}{\textbf{Supervision for strategy refinement}} from the adaptation benefits, costs, or stability observed under the revised evolutionary strategy.
& (1) \textcolor{royalblue}{\textbf{Strategy-level modification}} of the persistent representation of at least one component of the evolutionary strategy.\newline
  (2) \textcolor{royalblue}{\textbf{Retention and application}} of evaluated strategy revisions in subsequent evolution rounds.
\\

\bottomrule 
\end{tabularx} \vspace{-3mm}
\endgroup
\end{table*}

\noindent\textbf{Cumulative Stage Assignment.} For an agent system $A$ governed by the evolutionary strategy $\mathcal{S}$ in environment $E$, let $\operatorname{CapPass}_{k,d}\in\{0,1\}$ indicate whether the additional requirements introduced at stage $k$ in dimension $d$ have been demonstrated. Here, $k\in\{2,3,4,5\}$ and $d\in\{\mathrm{EA},\mathrm{SA},\mathrm{SC}\}$, where EA, SA, and SC denote evolutionary autonomy, supervision autonomy, and optimization scope and coordination, respectively. Assessment proceeds under declared deployment conditions, external support, and acceptance criteria associated with $\mathcal{O}$ and $\mathcal{B}$. The dimensional requirements are assessed within the same implementation and evolution process, including the functional connections specified by each stage. The assessment distinguishes changes to the deployed agent from changes to the procedures that generate supervision or propose updates. A zero covers both a documented failure to meet a requirement and insufficient evidence to establish it.

Taking Stage~1 as the baseline, cumulative stage satisfaction is defined as:
\begin{equation}
\label{equ:cumulative_stage_pass}
    \begin{aligned}
    \operatorname{StagePass}_{1} &= 1,\\
    \operatorname{StagePass}_{k} &= \operatorname{StagePass}_{k-1}\prod_{d\in\{\mathrm{EA},\mathrm{SA},\mathrm{SC}\}}\operatorname{CapPass}_{k,d},
    \end{aligned}
\end{equation}
for $k=2,\ldots,5$. Each subsequent stage therefore retains the preceding capability requirements and adds jointly satisfied requirements across the three dimensions within this stage.

The highest demonstrated evolutionary stage under the specified deployment conditions is:
\begin{equation}
    \label{equ:evolutionary_stage_index}
    \operatorname{EvoStage}(A,\mathcal{S}\mid E)=\sum_{k=1}^{5}\operatorname{StagePass}_{k}.
\end{equation}
The recursive definition yields $\operatorname{StagePass}_{k}\leq\operatorname{StagePass}_{k-1}$, so the sum corresponds to the highest cumulatively satisfied stage under the joint capability criterion. Table~\ref{tab:evolutionary_capability_checklist} applies these requirements to ten representative agent evolution methods. Within each stage $k$ and dimension $d$, $\operatorname{CapPass}_{k,d}=1$ if all numbered requirements introduced at stage $k$ in dimension $d$ are demonstrated, and $0$ otherwise. For example, the Stage~5 profiles of MetaSkill-Evolve~\citep{wang2026metaskill} and HyperAgents~\citep{zhang2026hyperagents} are both $(0,1,1)$ in the order $(\mathrm{EA},\mathrm{SA},\mathrm{SC})$. Both frameworks reach Stage~3 cumulatively and demonstrate additional capabilities at higher stages. For MUSE, the Stage~4 profile is $(1,1,1)$ and all preceding requirements are demonstrated, giving $\operatorname{EvoStage}=4$.

\begin{table*}[!t]
\centering
\caption{Capability checklist for ten representative agent evolution methods, indexed by the requirements in Table~\ref{tab:evolution_roadmap} (see the note below the table). A check ($\checkmark$) denotes a requirement established by the available evidence and a cross ($\times$) a requirement for which evidence is absent or insufficient. The last column is the highest cumulatively demonstrated stage under the declared conditions according to Equation~\ref{equ:evolutionary_stage_index}.}

\label{tab:evolutionary_capability_checklist}
\begingroup
\footnotesize
\setlength{\tabcolsep}{0.8pt}
\renewcommand{\arraystretch}{1.13}
\begin{tabular*}{\linewidth}{@{\extracolsep{\fill}}l *{23}{c} c@{}}
\toprule
\multirow{3}{*}{\textbf{Agent Evolution Method}} & \multicolumn{8}{c}{\shortstack{\bfseries Evolutionary\\\bfseries Autonomy}} & \multicolumn{8}{c}{\shortstack{\bfseries Supervision\\\bfseries Autonomy}} & \multicolumn{7}{c}{\shortstack{\bfseries Optimization Scope\\\bfseries and Coordination}} & \multirow{3}{*}{\shortstack{\textbf{Cumulative}\\\textbf{stage}}} \\
\cmidrule(lr){2-9}\cmidrule(lr){10-17}\cmidrule(lr){18-24}
 & \multicolumn{1}{c}{\cellcolor{spectrum_stage_2!33}\textbf{2}} & \multicolumn{1}{c}{\cellcolor{spectrum_stage_3!33}\textbf{3}} & \multicolumn{4}{c}{\cellcolor{spectrum_stage_4!33}\textbf{4}} & \multicolumn{2}{c}{\cellcolor{spectrum_stage_5!33}\textbf{5}} & \multicolumn{1}{c}{\cellcolor{spectrum_stage_2!33}\textbf{2}} & \multicolumn{1}{c}{\cellcolor{spectrum_stage_3!33}\textbf{3}} & \multicolumn{5}{c}{\cellcolor{spectrum_stage_4!33}\textbf{4}} & \multicolumn{1}{c}{\cellcolor{spectrum_stage_5!33}\textbf{5}} & \multicolumn{1}{c}{\cellcolor{spectrum_stage_2!33}\textbf{2}} & \multicolumn{1}{c}{\cellcolor{spectrum_stage_3!33}\textbf{3}} & \multicolumn{3}{c}{\cellcolor{spectrum_stage_4!33}\textbf{4}} & \multicolumn{2}{c}{\cellcolor{spectrum_stage_5!33}\textbf{5}} &  \\
 & \cellcolor{spectrum_stage_2!18}{\scriptsize (1)} & \cellcolor{spectrum_stage_3!18}{\scriptsize (1)} & \cellcolor{spectrum_stage_4!18}{\scriptsize (1)} & \cellcolor{spectrum_stage_4!18}{\scriptsize (2)} & \cellcolor{spectrum_stage_4!18}{\scriptsize (3)} & \cellcolor{spectrum_stage_4!18}{\scriptsize (4)} & \cellcolor{spectrum_stage_5!18}{\scriptsize (1)} & \cellcolor{spectrum_stage_5!18}{\scriptsize (2)} & \cellcolor{spectrum_stage_2!18}{\scriptsize (1)} & \cellcolor{spectrum_stage_3!18}{\scriptsize (1)} & \cellcolor{spectrum_stage_4!18}{\scriptsize (1)} & \cellcolor{spectrum_stage_4!18}{\scriptsize (2)} & \cellcolor{spectrum_stage_4!18}{\scriptsize (3)} & \cellcolor{spectrum_stage_4!18}{\scriptsize (4)} & \cellcolor{spectrum_stage_4!18}{\scriptsize (5)} & \cellcolor{spectrum_stage_5!18}{\scriptsize (1)} & \cellcolor{spectrum_stage_2!18}{\scriptsize (1)} & \cellcolor{spectrum_stage_3!18}{\scriptsize (1)} & \cellcolor{spectrum_stage_4!18}{\scriptsize (1)} & \cellcolor{spectrum_stage_4!18}{\scriptsize (2)} & \cellcolor{spectrum_stage_4!18}{\scriptsize (3)} & \cellcolor{spectrum_stage_5!18}{\scriptsize (1)} & \cellcolor{spectrum_stage_5!18}{\scriptsize (2)} &  \\
\midrule
ELIZA~{\scriptsize\citep{weizenbaum1983eliza}} & \textcolor{black!52}{$\times$} & \textcolor{black!52}{$\times$} & \textcolor{black!52}{$\times$} & \textcolor{black!52}{$\times$} & \textcolor{black!52}{$\times$} & \textcolor{black!52}{$\times$} & \textcolor{black!52}{$\times$} & \textcolor{black!52}{$\times$} & \textcolor{black!52}{$\times$} & \textcolor{black!52}{$\times$} & \textcolor{black!52}{$\times$} & \textcolor{black!52}{$\times$} & \textcolor{black!52}{$\times$} & \textcolor{black!52}{$\times$} & \textcolor{black!52}{$\times$} & \textcolor{black!52}{$\times$} & \textcolor{black!52}{$\times$} & \textcolor{black!52}{$\times$} & \textcolor{black!52}{$\times$} & \textcolor{black!52}{$\times$} & \textcolor{black!52}{$\times$} & \textcolor{black!52}{$\times$} & \textcolor{black!52}{$\times$} & \textbf{1.0} \\
MemGPT~{\scriptsize\citep{packer2023memgpt}} & \textcolor{teal!65!black}{$\checkmark$} & \textcolor{black!52}{$\times$} & \textcolor{black!52}{$\times$} & \textcolor{black!52}{$\times$} & \textcolor{black!52}{$\times$} & \textcolor{black!52}{$\times$} & \textcolor{black!52}{$\times$} & \textcolor{black!52}{$\times$} & \textcolor{teal!65!black}{$\checkmark$} & \textcolor{black!52}{$\times$} & \textcolor{black!52}{$\times$} & \textcolor{black!52}{$\times$} & \textcolor{black!52}{$\times$} & \textcolor{black!52}{$\times$} & \textcolor{black!52}{$\times$} & \textcolor{black!52}{$\times$} & \textcolor{teal!65!black}{$\checkmark$} & \textcolor{black!52}{$\times$} & \textcolor{black!52}{$\times$} & \textcolor{black!52}{$\times$} & \textcolor{black!52}{$\times$} & \textcolor{black!52}{$\times$} & \textcolor{black!52}{$\times$} & \textbf{2.0} \\
MemoryBank~{\scriptsize\citep{zhong2024memorybank}} & \textcolor{teal!65!black}{$\checkmark$} & \textcolor{black!52}{$\times$} & \textcolor{black!52}{$\times$} & \textcolor{black!52}{$\times$} & \textcolor{black!52}{$\times$} & \textcolor{black!52}{$\times$} & \textcolor{black!52}{$\times$} & \textcolor{black!52}{$\times$} & \textcolor{teal!65!black}{$\checkmark$} & \textcolor{black!52}{$\times$} & \textcolor{black!52}{$\times$} & \textcolor{black!52}{$\times$} & \textcolor{black!52}{$\times$} & \textcolor{black!52}{$\times$} & \textcolor{black!52}{$\times$} & \textcolor{black!52}{$\times$} & \textcolor{teal!65!black}{$\checkmark$} & \textcolor{black!52}{$\times$} & \textcolor{black!52}{$\times$} & \textcolor{black!52}{$\times$} & \textcolor{black!52}{$\times$} & \textcolor{black!52}{$\times$} & \textcolor{black!52}{$\times$} & \textbf{2.0} \\
SkillMentor~{\scriptsize\citep{bao2026skillmentor}} & \textcolor{teal!65!black}{$\checkmark$} & \textcolor{teal!65!black}{$\checkmark$} & \textcolor{teal!65!black}{$\checkmark$} & \textcolor{teal!65!black}{$\checkmark$} & \textcolor{black!52}{$\times$} & \textcolor{teal!65!black}{$\checkmark$} & \textcolor{black!52}{$\times$} & \textcolor{black!52}{$\times$} & \textcolor{teal!65!black}{$\checkmark$} & \textcolor{teal!65!black}{$\checkmark$} & \textcolor{teal!65!black}{$\checkmark$} & \textcolor{teal!65!black}{$\checkmark$} & \textcolor{black!52}{$\times$} & \textcolor{teal!65!black}{$\checkmark$} & \textcolor{teal!65!black}{$\checkmark$} & \textcolor{teal!65!black}{$\checkmark$} & \textcolor{teal!65!black}{$\checkmark$} & \textcolor{teal!65!black}{$\checkmark$} & \textcolor{black!52}{$\times$} & \textcolor{black!52}{$\times$} & \textcolor{black!52}{$\times$} & \textcolor{teal!65!black}{$\checkmark$} & \textcolor{black!52}{$\times$} & \textbf{3.0} \\
AgentEvolver~{\scriptsize\citep{zhai2025agentevolver}} & \textcolor{teal!65!black}{$\checkmark$} & \textcolor{teal!65!black}{$\checkmark$} & \textcolor{black!52}{$\times$} & \textcolor{black!52}{$\times$} & \textcolor{black!52}{$\times$} & \textcolor{black!52}{$\times$} & \textcolor{black!52}{$\times$} & \textcolor{black!52}{$\times$} & \textcolor{teal!65!black}{$\checkmark$} & \textcolor{teal!65!black}{$\checkmark$} & \textcolor{teal!65!black}{$\checkmark$} & \textcolor{teal!65!black}{$\checkmark$} & \textcolor{teal!65!black}{$\checkmark$} & \textcolor{black!52}{$\times$} & \textcolor{black!52}{$\times$} & \textcolor{black!52}{$\times$} & \textcolor{teal!65!black}{$\checkmark$} & \textcolor{teal!65!black}{$\checkmark$} & \textcolor{teal!65!black}{$\checkmark$} & \textcolor{black!52}{$\times$} & \textcolor{black!52}{$\times$} & \textcolor{black!52}{$\times$} & \textcolor{black!52}{$\times$} & \textbf{3.0} \\
SIA~{\scriptsize\citep{hebbar2026sia}} & \textcolor{teal!65!black}{$\checkmark$} & \textcolor{teal!65!black}{$\checkmark$} & \textcolor{black!52}{$\times$} & \textcolor{black!52}{$\times$} & \textcolor{teal!65!black}{$\checkmark$} & \textcolor{teal!65!black}{$\checkmark$} & \textcolor{black!52}{$\times$} & \textcolor{black!52}{$\times$} & \textcolor{teal!65!black}{$\checkmark$} & \textcolor{teal!65!black}{$\checkmark$} & \textcolor{black!52}{$\times$} & \textcolor{black!52}{$\times$} & \textcolor{black!52}{$\times$} & \textcolor{black!52}{$\times$} & \textcolor{black!52}{$\times$} & \textcolor{black!52}{$\times$} & \textcolor{teal!65!black}{$\checkmark$} & \textcolor{teal!65!black}{$\checkmark$} & \textcolor{teal!65!black}{$\checkmark$} & \textcolor{teal!65!black}{$\checkmark$} & \textcolor{teal!65!black}{$\checkmark$} & \textcolor{black!52}{$\times$} & \textcolor{black!52}{$\times$} & \textbf{3.0} \\
COVE~{\scriptsize\citep{ji2026learning}} & \textcolor{teal!65!black}{$\checkmark$} & \textcolor{teal!65!black}{$\checkmark$} & \textcolor{black!52}{$\times$} & \textcolor{black!52}{$\times$} & \textcolor{teal!65!black}{$\checkmark$} & \textcolor{teal!65!black}{$\checkmark$} & \textcolor{black!52}{$\times$} & \textcolor{black!52}{$\times$} & \textcolor{teal!65!black}{$\checkmark$} & \textcolor{teal!65!black}{$\checkmark$} & \textcolor{black!52}{$\times$} & \textcolor{teal!65!black}{$\checkmark$} & \textcolor{black!52}{$\times$} & \textcolor{teal!65!black}{$\checkmark$} & \textcolor{black!52}{$\times$} & \textcolor{black!52}{$\times$} & \textcolor{teal!65!black}{$\checkmark$} & \textcolor{teal!65!black}{$\checkmark$} & \textcolor{teal!65!black}{$\checkmark$} & \textcolor{teal!65!black}{$\checkmark$} & \textcolor{teal!65!black}{$\checkmark$} & \textcolor{black!52}{$\times$} & \textcolor{black!52}{$\times$} & \textbf{3.0} \\
MetaSkill-Evolve~{\scriptsize\citep{wang2026metaskill}} & \textcolor{teal!65!black}{$\checkmark$} & \textcolor{teal!65!black}{$\checkmark$} & \textcolor{teal!65!black}{$\checkmark$} & \textcolor{black!52}{$\times$} & \textcolor{black!52}{$\times$} & \textcolor{teal!65!black}{$\checkmark$} & \textcolor{black!52}{$\times$} & \textcolor{black!52}{$\times$} & \textcolor{teal!65!black}{$\checkmark$} & \textcolor{teal!65!black}{$\checkmark$} & \textcolor{black!52}{$\times$} & \textcolor{black!52}{$\times$} & \textcolor{black!52}{$\times$} & \textcolor{teal!65!black}{$\checkmark$} & \textcolor{black!52}{$\times$} & \textcolor{teal!65!black}{$\checkmark$} & \textcolor{teal!65!black}{$\checkmark$} & \textcolor{teal!65!black}{$\checkmark$} & \textcolor{black!52}{$\times$} & \textcolor{black!52}{$\times$} & \textcolor{black!52}{$\times$} & \textcolor{teal!65!black}{$\checkmark$} & \textcolor{teal!65!black}{$\checkmark$} & \textbf{3.0} \\
HyperAgents~{\scriptsize\citep{zhang2026hyperagents}} & \textcolor{teal!65!black}{$\checkmark$} & \textcolor{teal!65!black}{$\checkmark$} & \textcolor{black!52}{$\times$} & \textcolor{black!52}{$\times$} & \textcolor{teal!65!black}{$\checkmark$} & \textcolor{teal!65!black}{$\checkmark$} & \textcolor{black!52}{$\times$} & \textcolor{teal!65!black}{$\checkmark$} & \textcolor{teal!65!black}{$\checkmark$} & \textcolor{teal!65!black}{$\checkmark$} & \textcolor{black!52}{$\times$} & \textcolor{teal!65!black}{$\checkmark$} & \textcolor{black!52}{$\times$} & \textcolor{teal!65!black}{$\checkmark$} & \textcolor{black!52}{$\times$} & \textcolor{teal!65!black}{$\checkmark$} & \textcolor{teal!65!black}{$\checkmark$} & \textcolor{teal!65!black}{$\checkmark$} & \textcolor{teal!65!black}{$\checkmark$} & \textcolor{teal!65!black}{$\checkmark$} & \textcolor{teal!65!black}{$\checkmark$} & \textcolor{teal!65!black}{$\checkmark$} & \textcolor{teal!65!black}{$\checkmark$} & \textbf{3.0} \\
\midrule
\textbf{MUSE} (ours) & \textcolor{teal!65!black}{$\checkmark$} & \textcolor{teal!65!black}{$\checkmark$} & \textcolor{teal!65!black}{$\checkmark$} & \textcolor{teal!65!black}{$\checkmark$} & \textcolor{teal!65!black}{$\checkmark$} & \textcolor{teal!65!black}{$\checkmark$} & \textcolor{black!52}{$\times$} & \textcolor{black!52}{$\times$} & \textcolor{teal!65!black}{$\checkmark$} & \textcolor{teal!65!black}{$\checkmark$} & \textcolor{teal!65!black}{$\checkmark$} & \textcolor{teal!65!black}{$\checkmark$} & \textcolor{teal!65!black}{$\checkmark$} & \textcolor{teal!65!black}{$\checkmark$} & \textcolor{teal!65!black}{$\checkmark$} & \textcolor{black!52}{$\times$} & \textcolor{teal!65!black}{$\checkmark$} & \textcolor{teal!65!black}{$\checkmark$} & \textcolor{teal!65!black}{$\checkmark$} & \textcolor{teal!65!black}{$\checkmark$} & \textcolor{teal!65!black}{$\checkmark$} & \textcolor{black!52}{$\times$} & \textcolor{black!52}{$\times$} & \textbf{4.0} \\
\bottomrule
\end{tabular*}

\par\vspace{3pt}
\begin{minipage}{\linewidth}
\scriptsize
\noindent\textit{Note.} Within each dimension group, the bold number in the header (2-5) is the stage at which a requirement is introduced, and the parenthesized number, such as (1) or (2), indexes the individual requirements enumerated in the corresponding cell of Table~\ref{tab:evolution_roadmap}. For example, the column EA~4~(2) refers to the second requirement in the Stage~4 cell of the EA column, namely deferral of evolution when activation criteria are unmet.
\end{minipage}
\endgroup
\end{table*}

\vspace*{1ex}
\noindent \colorbox{spectrum_stage_1}{$\blacktriangleright$} \textbf{ Evolution 1.0: Human-Engineered Maintenance}
\vspace*{1ex}

\noindent \textbf{Definition:} Historically prevalent in early rule-based agent systems, this stage describes post-deployment capability changes that are primarily organized outside the deployed agent. The agent may autonomously execute predefined task behaviors, while persistent modifications to its rules, knowledge, procedures, or model configuration are introduced through human engineering or offline maintenance. Stage~1 therefore distinguishes task-level autonomy from evolutionary autonomy, with the latter remaining primarily external to the deployed system.

\noindent \textbf{Status of Evolutionary Strategy ($\mathcal{S}$):} In this stage, the evolutionary strategy is largely externalized to human developers. Specifically, the evolutionary gating ($\Gamma$) is determined by human-observed system failures, user feedback, or predefined maintenance schedules. The optimization scope ($\mathcal{C}$) specifies the externally designated components and configurations eligible for maintenance. Supervision ($\Delta$) relies primarily on human-engineered rules, expert knowledge, or manually annotated data. The optimizer ($\mathcal{M}$) is executed via offline manual reprogramming or model retraining by a human. This manual adaptation process is guided by externally specified task or deployment objectives ($\mathcal{O}$) and constrained by human-defined validation procedures and operational requirements ($\mathcal{B}$).

\noindent \textbf{Biological Metaphor:} Analogous to artificial selection such as breeding wolves into dogs, adaptation is directed by an external selector that determines which characteristics are modified and retained.

\noindent \textbf{Representative Works:} Rule-based chatbots and early symbolic agent systems whose capabilities were primarily maintained through external engineering, such as ELIZA~\citep{weizenbaum1983eliza}, PARRY~\citep{colby1971artificial}, and ALICE~\citep{wallace2007anatomy}, as well as expert and knowledge-based systems such as MYCIN~\citep{shortliffe2012computer}, XCON~\citep{barker1989expert}, and INTERNIST-I~\citep{miller1985internist}. These systems encoded their core task capabilities primarily through human-authored rules, domain knowledge, or manually constructed knowledge bases. Persistent capability expansion or correction therefore remained largely dependent on external knowledge engineering, rule maintenance, and system updates, leaving evolutionary control outside the deployed system.

\vspace*{1ex}
\noindent \colorbox{spectrum_stage_2}{$\blacktriangleright$} \textbf{Evolution 2.0: Interaction-Driven Accumulation} 
\vspace*{1ex}

\noindent \textbf{Definition:} This stage introduces persistent internal updates driven by continued interaction, commonly through memory or knowledge assets within $\Psi$. The agent automatically retains and organizes interaction experience across task episodes, allowing accumulated information to guide subsequent behavior. The defining capability is persistent information accumulation, which can involve selection, consolidation, reflection, revision, or forgetting. Its realization depends on the component configurations available within the agent system.

\noindent \textbf{Status of Evolutionary Strategy ($\mathcal{S}$):} In this stage, the optimization scope ($\mathcal{C}$) includes persistent states that retain interaction-derived information, commonly represented through memory and knowledge assets such as $K_{\text{exp}}$. The gating mechanism ($\Gamma$) activates updates in response to interaction events or newly available evidence. Supervision ($\Delta$) consists of encountered interaction content and feedback, potentially organized through predefined retention criteria. The optimizer ($\mathcal{M}$) performs persistent updates such as insertion, consolidation, revision, and deletion. The objective ($\mathcal{O}$) guides information retention for subsequent task execution, while the boundary ($\mathcal{B}$) specifies capacity, relevance, and operational resource conditions.

\noindent \textbf{Transition Criterion:} The transition from Stage~1 to Stage~2 requires the Stage-2 requirements EA(1), SA(1), and SC(1) in Table~\ref{tab:evolution_roadmap} to be jointly demonstrated. The defining distinction from Stage~1 is SC(1): the accumulated state must influence behavior on later independent tasks, not only within the session in which it was recorded.

\noindent \textbf{Biological Metaphor:} Analogous to behavioral caching and resource accumulation, such as squirrels caching nuts for winter survival, this stage emphasizes retention of experience or resources for subsequent access.

\noindent \textbf{Representative Works:} Frameworks centered on persistent memory and interaction-derived context include MemGPT~\citep{packer2023memgpt}, MemoryBank~\citep{zhong2024memorybank}, Generative Agents~\citep{park2023generative}, MemPrompt~\citep{madaan2022memory}, ChatDB~\citep{hu2023chatdb}, Mem0~\citep{chhikara2025mem0}, and MemoryOS~\citep{kang2025memory}. AnnaAgent~\citep{wang2025annaagent} further combines a tertiary memory $K_{\text{exp}}$ with interaction-conditioned updates to the seeker state, coordinating memory with evolving emotional and complaint configurations across counseling sessions. In A-Mem~\citep{xu2025amem}, adding a new note can trigger revisions to earlier notes that remain available for later tasks. By contrast, in EvoMem~\citep{fan2025evomem}, both working memories are cleared before each new query, so execution feedback guides refinement only within the current task.

\vspace*{1ex}
\noindent \colorbox{spectrum_stage_3}{$\blacktriangleright$} \textbf{Evolution 3.0: Partial Self-Evolution} 
\vspace*{1ex}

\noindent \textbf{Definition:} This stage marks the transition from persistent accumulation to explicit self-optimization within the agent system. The agent can actively refine selected internal components according to task experience or evaluative feedback, establishing an evolutionary loop that modifies its behavioral capabilities. The defining capability connects persistent component changes to evaluation of the resulting task behavior. Evaluation outcomes guide configuration retention or subsequent refinement. Supervision can originate entirely from external oracle labels or task verifiers, and can also include internally generated signals. The optimization scope can involve one or several components, according to the available agent architecture and adaptation objectives.

\noindent \textbf{Status of Evolutionary Strategy ($\mathcal{S}$):} The optimization scope ($\mathcal{C}$) specifies the persistent components eligible for refinement. The gating mechanism ($\Gamma$) activates and routes optimization through task events, predefined cycles, or evaluative feedback. The supervision generator ($\Delta$) acquires or constructs learning signals from externally provided labels, teacher feedback, task verifiers, or internal assessments. These sources can contribute separately or jointly, with entirely external supervision forming a valid instantiation. The optimizer ($\mathcal{M}$) refines the selected components, while evaluation under $\mathcal{O}$ guides their retention or further adjustment. The boundary ($\mathcal{B}$) specifies the operational conditions associated with this process. When internally generated supervision depends primarily on model confidence or agreement, learning may reinforce incorrect initial beliefs~\citep{he2026how}. This conditional risk motivates environmental validation of autonomous supplementary supervision at Stage~4.

\noindent \textbf{Transition Criterion:} Building on Stage~2, the transition to Stage~3 requires the Stage-3 requirements EA(1), SA(1), and SC(1) in Table~\ref{tab:evolution_roadmap} to be jointly demonstrated. The defining distinction from Stage~2 is SC(1): persistent component changes are followed by evaluation of the updated agent, and the resulting feedback guides configuration retention or further refinement. Supervision derived entirely from external sources can satisfy SA(1) when automatically organized for component optimization.

\noindent \textbf{Biological Metaphor:} Analogous to specialized microevolution, such as peacocks evolving elaborate tails under sexual selection, this stage emphasizes refinement of selected traits under particular environmental pressures.

\noindent \textbf{Representative Works:} A substantial portion of contemporary agent-evolution research can be situated within Stage~3, reflecting diverse forms of partial self-optimization across different agent components. Representative frameworks in this stage can be categorized by their primary optimization scope within the agent system.

Within the parametric cognitive core $\Theta$, evolution methods including SPICE~\citep{liu2025spice}, SCA~\citep{zhou2025selfchallenging}, SELF~\citep{lu2023self}, SAD~\citep{liu2025structured}, GiGPO~\citep{feng2025groupingroup}, SCoRe~\citep{kumar2025training}, ScoreFlow~\citep{wang2025scoreflow}, and Agent0~\citep{xia2025agent0} primarily target the refinement of the underlying model weights $\theta_{\text{model}}$. SEES~\citep{li2026sees} extends targeted parametric adaptation to embodied agents, refining shared skill-family policies through failure-guided reinforcement learning on constructed simulation tasks. Expanding this boundary, methods including AutoFlow~\citep{li2024autoflow}, Puppeteer~\citep{dang2025multiagent}, MA-Evolve~\citep{chen2025multi}, CoMAS~\citep{xue2025comas}, and AFLOW~\citep{zhang2025aflow} jointly optimize the collaborative topology $N_{\text{topo}}$ to enhance multi-agent cooperation.

Within the space of symbolic assets $\Psi$, ALMA~\citep{xiong2026learning} and Creator~\citep{qian2023creator} concentrate on expanding actionable toolsets $T_{\text{tool}}$, while Evo-Memory~\citep{wei2025evo},  Sophia~\citep{sun2025sophia}, EvoTest~\citep{he2025evotest}, and Voyager~\citep{wang2024voyager} adapt experience or skill assets. MemEvolve~\citep{zhang2025memevolve} evaluates candidate memory architectures on task performance, cost, and latency, then selects suitable architectures for further refinement. This feedback illustrates the Stage~3 criterion for optimization scope and coordination in memory-based systems. Recent skill-oriented methods introduce more structured adaptation within this symbolic space. SkillProx~\citep{zheng2026skillprox} couples failure diagnosis with re-execution, rollback, and utility-aware skill consolidation, while SkillMentor~\citep{bao2026skillmentor} trains an auxiliary mentor to diagnose blind spots and propose corrective skills for a frozen executor. These methods strengthen diagnosis and symbolic refinement, although the persistent evolutionary target remains concentrated in skill assets.

In terms of operational hyperparameters $\Phi$, frameworks such as ADA~\citep{su2026learning} and AgentHPO~\citep{liu2025agenthpo} investigate the autonomous adaptation of hyperparameters.

Notably, several frameworks exhibit cross-component evolution. EvoAgent~\citep{yuan2025evoagent} optimizes the collaborative topology $N_{\text{topo}}$ alongside declarative system prompts $P_{\text{sys}}$. AgentEvolver~\citep{zhai2025agentevolver} builds an experience bank from executions and retrieves entries to guide rollouts for subsequent parametric updates. ToolGen~\citep{wang2025toolgen} bridges the parametric and symbolic boundaries by embedding individual tools as virtual tokens within the model vocabulary to unify tool selection and weight updating. More recent frameworks broaden this coordination. SIA~\citep{hebbar2026sia} combines harness adaptation with model weight updates under a feedback-agent given task specifications and verifiers, while COVE~\citep{ji2026learning} coordinates external memory and parametric learning through task-aware routing and stage-aware scheduling. Self-Evolving Agent Harnesses~\citep{luo2026self} develops failure diagnosis and statistically validated harness refinement around a frozen task model. ScholarEvolve~\citep{yang2026scholarevolve} translates research-derived mechanisms into modular harness revisions and selects compatible combinations through joint evaluation. Recent system-wide agent revision further connects multiple editable components to evaluation and subsequent execution through a shared versioned lifecycle~\citep{zhang2026agentevolversystem}. These approaches illustrate cross-component optimization and adaptive routing, whose contributions to overall evolutionary maturity are assessed together with the remaining dimensional requirements.

Supervision acquisition also develops alongside component optimization. AgentEvolver~\citep{zhai2025agentevolver} constructs tasks through environmental exploration and derives rewards through reference-based assessment. OpenSkill~\citep{yan2026openskill} obtains verification anchors from documentation, repositories, and other open-world resources, then refines skills through grounded virtual tasks. Skill-SP~\citep{huang2026skillsp} couples parametric self-play with an evolving task-generation skill library, constructing capability-conditioned curricula through contract validation and solver feedback. These approaches illustrate different supervision sources and validation procedures grounded in environmental and external evidence. Their demonstrated capabilities are recorded in the dimensional profile, with overall stage assignment determined by the joint cumulative requirements.

\vspace*{1ex}
\noindent \colorbox{spectrum_stage_4}{$\blacktriangleright$} \textbf{Evolution 4.0: Metacognitive Self-Evolution} 
\vspace*{1ex}

\noindent \textbf{Definition:} 
Stage~4 builds on feedback-guided self-optimization through the joint realization of longitudinal evolutionary control, interaction-grounded self-supervision, and heterogeneous component coordination. Along evolutionary autonomy, the strategy draws on accumulated interaction evidence to activate or defer evolution, diagnose capability deficiencies and consolidation opportunities, and select adaptation components. Along supervision autonomy, the agent autonomously constructs supplementary tasks and evaluative signals, assessing their validity against environmental feedback and execution evidence. Current capabilities guide supervision selection and renewal. Along optimization scope and coordination, adaptation spans at least two categories among $\Theta$, $\Psi$, and $\Phi$, with joint evaluation and observed update dependencies guiding component selection and refinement.

\noindent \textbf{Status of Evolutionary Strategy ($\mathcal{S}$):} 
The optimization scope ($\mathcal{C}$) specifies the eligible components across heterogeneous categories, while evolutionary gating $\Gamma$ selects suitable subsets according to current capability diagnosis grounded in accumulated interaction evidence. The supervision generator ($\Delta$) constructs and assesses supplementary supervision signals through interaction-grounded self-supervision, which can complement available oracle labels and verifier feedback. The optimizer ($\mathcal{M}$) coordinates selected component updates, with refinements guided by evaluation of the joint configuration and evidence about related component changes. The objective ($\mathcal{O}$) guides assessment of adaptation, while the boundary ($\mathcal{B}$) specifies retention, compatibility, resource, and other deployment conditions. Subsequent interactions provide fresh evidence for evolutionary control and supervision assessment, connecting the three dimensions across successive rounds. Environmental validation examines task feasibility, correspondence with the specified objective, and alignment of evaluative signals with observed outcomes. Successful execution provides feasibility evidence, while task completion assessment additionally considers whether the behavior fulfills the objective. Environmental grounding therefore supplies observable checks on synthetic supervision, with remaining evaluation examined empirically~\citep{yan2026openskill}.

\noindent \textbf{Transition Criterion:} Building on Stage~3, the transition to Stage~4 requires the Stage-4 requirements EA(1)-(4), SA(1)-(5), and SC(1)-(3) in Table~\ref{tab:evolution_roadmap} to be jointly demonstrated within the same evolution process under the declared deployment and evaluation conditions. EA(2) requires evolution to remain inactive when activation criteria are unmet. EA(3) and EA(4) connect capability diagnosis grounded in accumulated interaction evidence to targeted component selection. SA(3) requires that generated tasks and evaluative signals be checked against environmental feedback and interaction evidence. SC(2) and SC(3) require joint configuration evaluation and observed update dependencies to guide subsequent component selection and refinement.

\noindent \textbf{Biological Metaphor (Systemic Macroevolution):} Analogous to systemic macroevolution, such as terrestrial mammals evolving into whales for aquatic life, Stage~4 reflects multi-system co-adaptation across closely interacting biological systems. Changes in locomotion, respiration, sensing, and body structure jointly and progressively support adaptation to a substantially different environment.

\noindent \textbf{Representative Works:} Recent agent evolution methods illustrate partial progress toward Stage~4. SkillMentor~\citep{bao2026skillmentor} connects learned diagnosis and interaction-based task construction with corrective skill curation, while OpenSkill~\citep{yan2026openskill} constructs supplementary supervision and verification anchors. SEES~\citep{li2026sees} connects recurrent failure diagnosis with simulator-grounded task construction, generated success predicates, and targeted adaptation of shared skill-family policies. Skill-SP~\citep{huang2026skillsp} further connects task generation and verification with an evolving skill library and capability-conditioned curricula. SIA~\citep{hebbar2026sia} coordinates harness adaptation with model weight updates, while COVE~\citep{ji2026learning} routes tasks for parametric learning according to observed capability gaps and reassesses individual memories after weight updates. ScholarEvolve~\citep{yang2026scholarevolve} provides complementary evidence of component coordination through joint evaluation of module combinations and retention of compatible revisions. These studies demonstrate individual elements of evolutionary control, supervision construction, and component coordination. However, complete Stage~4 requires interaction-grounded capability diagnosis and adaptive control over evolution activation, deferral, and target selection, autonomous construction and environmental assessment of supplementary supervision, and coordinated refinement across at least two categories among $\Theta$, $\Psi$, and $\Phi$, guided by joint evaluation and update dependencies. Joint fulfillment of the cumulative Stage~4 requirements within the same evolution process has not been established for these works in the present comparative assessment. MUSE, presented in Section~\ref{sec:methodology}, is designed as a prototype targeting Stage~4 capabilities within the six-component evolutionary strategy.

\vspace*{1ex}
\noindent \colorbox{spectrum_stage_5}{$\blacktriangleright$} \textbf{Evolution 5.0: Recursive Meta-Evolution} 
\vspace*{1ex}

\noindent \textbf{Definition:} 
Stage~5 extends the object of evolution from the agent configuration alone to the joint evolution of the agent system and its evolutionary strategy $\mathcal{S}$. Building on Stage~4, $\mathcal{S}$ is further included in optimization targets. Along evolutionary autonomy, the agent determines when strategy revision is warranted and selects the strategy components to adapt. Along supervision autonomy, records of subsequent evolution rounds provide signals for assessing the adaptation benefits, costs, or stability of alternative strategy configurations. Along optimization scope and coordination, evaluated changes to selected strategy components are retained and applied in later rounds.

\noindent \textbf{Status of Evolutionary Strategy ($\mathcal{S}$):} 
Building on the Stage-4 strategy, the optimization scope $\mathcal{C}$ extends to persistent representations of selected components of $\mathcal{S}$ alongside eligible components of $A$. While the agent continues to evolve through lifecycle-level metacognitive control, interaction-grounded self-supervision, and coordinated component evolution, mechanisms governing evolutionary gating $\Gamma$, supervision generation $\Delta$, optimization $\mathcal{M}$, objectives $\mathcal{O}$, or boundaries $\mathcal{B}$ may become adaptation targets. As $\mathcal{S}$ evolves, a central challenge lies in reliable meta-evaluation while maintaining coherent objectives and operational boundaries across recursive rounds.

\noindent \textbf{Transition Criterion:} Building on Stage~4, the transition to Stage~5 requires the Stage-5 requirements EA(1)-(2), SA(1), and SC(1)-(2) in Table~\ref{tab:evolution_roadmap} to be jointly demonstrated. The distinction from Stage~4 is that persistent representations of selected components of the evolutionary strategy $\mathcal{S}$ are modified, with revisions assessed through their effects on the benefits, costs, or stability of subsequent evolution.

\noindent \textbf{Biological Metaphor (Evolution of Evolvability):} 
Analogous to the evolution of evolvability, such as the evolution of stress-responsive mutation mechanisms in bacteria~\citep{galhardo2007mutation, pigliucci2008evolvability}, Stage~5 allows an agent to refine its own configuration $A$ and revise selected components of the evolutionary strategy $\mathcal{S}$ that governs how the agent adapts to new demands across subsequent evolution rounds.

\noindent \textbf{Representative Works:} Recent explorations make partial progress toward Stage~5 by introducing evolvable components of the evolutionary strategy $\mathcal{S}$. In MetaSkill-Evolve, improvements in subsequent descendants contribute to branch selection, and the selected branch resumes evolution with its stored meta-skill. HyperAgents~\citep{zhang2026hyperagents} revises its self-modification code. Its transfer protocol selects a hyperagent according to descendant improvement and evaluates the task agents subsequently generated by its retained meta agent. However, within the cumulative spectrum, Stage~5 requires the Stage~4 foundation together with autonomous control over the timing and targets of strategy revision, feedback from adaptation under the revised strategy to assess the adaptation benefits or stability, and persistent retention and application of evaluated strategy revisions. Joint fulfillment of these cumulative requirements has not been established for these works in the present comparative assessment.

\subsection{Evaluation Criteria for Agent Evolution} \label{subsec:evaluation_metrics}

While the Five-Stage Evolutionary Spectrum specifies cumulative capability requirements for agent evolution, empirical evaluation examines how these capabilities are realized and assesses the effectiveness, efficiency, and stability of the resulting adaptation. We therefore define seven complementary criteria that connect the formulation of $\mathcal{S}$ with observable evolution processes and measurable performance changes. The first five criteria can be directly instantiated within a single evolving environment, while the remaining generalizability and retention provide additional criteria for cross-environment and longer-term evaluation.

\noindent $\bullet$ \textbf{Adaptivity:} 
Adaptivity measures the performance change associated with each evolution round under the objective $\mathcal{O}$. Following the evolutionary transition $A_{t+1}=\mathcal{S}_t(A_t,E_t,H_t)$, the performance change of evolution round $t$ can be measured under the same environmental conditions, task samples, and evaluation protocol as:
\begin{equation}
    \mathrm{EvoGain}_t
    =
    \mathcal{O}(A_{t+1},E_t)
    -
    \mathcal{O}(A_t,E_t).
\end{equation}
This criterion quantifies the capability improvement attributable to each evolution round under its corresponding environment. For a fixed environment, cumulative adaptation over $T$ rounds can be measured by:
\begin{equation}
    \mathrm{CumEvoGain}_T
    =
    \mathcal{O}(A_T,E_0)
    -
    \mathcal{O}(A_0,E_0).
\end{equation}

\noindent $\bullet$ \textbf{Evolutionary Efficiency:} Evolutionary efficiency measures the interaction and optimization resources required by the strategy $\mathcal{S}$ to accumulate capabilities. Let $N_{\mathrm{roll}}^{(t)}$ denote the number of rollout trajectories consumed during round $t$. Practical indicators include the cumulative rollout budget required to reach a target performance level and the performance change achieved under matched resource budgets. Performance trajectories across evolution rounds describe adaptation progress, while resource comparisons additionally account for token consumption and computation associated with supervision generation, environmental validation, evolutionary gating, and component optimization. These measures jointly characterize the rate and resource cost of capability accumulation.

\noindent $\bullet$ \textbf{Component Contribution and Coordination Effectiveness:} This criterion evaluates the contribution of evolvable components within the optimization scope $\mathcal{C}$ and their coordination through $\Gamma$ and $\mathcal{M}$. Component contribution can be quantified through controlled ablations that enable or disable updates to selected components under matched initial conditions, evaluation protocols, and resource budgets. Coordination effectiveness can be examined by comparing coordinated and independent updates to the same component set. Update records further quantify how joint configuration evaluations and component dependencies guide subsequent refinement. These analyses characterize complementary contributions, redundant updates, and interference among components.

\noindent $\bullet$ \textbf{Supervision Effectiveness:} Supervision effectiveness evaluates the contribution of the supervision generator $\Delta$ to the evolutionary process. Its downstream utility can be measured through controlled comparisons that vary supervision sources or the inclusion of generated supplementary supervision under matched resource budgets. For generated supervision, the proportion of candidates passing environmental validation measures validation yield. Independent checks of task feasibility, objective completion, and agreement between assigned signals and observed execution provide evidence of supervision validity. Generation and validation records further document external supervision requirements and how current capabilities guide task selection and renewal. Together, these measures assess both the empirical reliability and downstream utility of generated supervision.

\noindent $\bullet$ \textbf{Gating Effectiveness:} Gating effectiveness evaluates whether $\Gamma$ provides effective control over evolution activation, capability diagnosis, and optimization routing. Its quantitative contribution can be assessed by comparing otherwise matched evolutionary strategies with different activation and routing mechanisms in terms of final performance, activated evolution rounds, and evolutionary efficiency. Separate comparisons can assess activation mechanisms under fixed routing and routing mechanisms under a matched activation schedule. Longitudinal deployment traces connect interaction evidence, capability diagnosis, evolution activation or deferral, and selected optimization targets to subsequent performance changes across successive evolution rounds. This criterion therefore combines performance assessment with analysis of the evolution process.

\noindent $\bullet$ \textbf{Generalizability:}
Generalizability evaluates whether capabilities accumulated through evolution remain effective beyond the environment or task distribution in which they were acquired. An evolved agent can be evaluated directly on unseen environments, domains, or task families without additional adaptation, with transfer measured through the corresponding performance change relative to its pre-evolution state under a consistent target evaluation protocol. Cross-environment transfer performance and out-of-distribution robustness provide complementary indicators of capability generalization. The transferability of $\mathcal{S}$ can additionally be examined by applying the same strategy in unseen environments under specified initial conditions and resource budgets.

\noindent $\bullet$ \textbf{Retention and Boundary Compliance:}
This criterion evaluates whether continued evolution preserves previously acquired capabilities while remaining within the operational boundary $\mathcal{B}$. Retention can be quantified through performance changes on learned legacy tasks after successive evolution rounds. Boundary compliance can be evaluated through compatibility, safety, or violation measures instantiated by $\mathcal{B}$ for a given deployment setting. Together, these characterize the stability of long-term capability accumulation under operational constraints.

Collectively, these criteria evaluate agent evolution from complementary perspectives spanning capability improvement, resource efficiency, component coordination, supervision effectiveness, evolutionary control, generalization, and long-term stability. They provide empirical evidence for examining the capability requirements of the Five-Stage Evolutionary Spectrum and the practical quality of a concrete evolutionary strategy $\mathcal{S}$ within a clearly specified deployment setting. When selected strategy components are revised, evaluation additionally examines subsequent adaptation under the revised strategy, including its adaptation benefits, costs, or stability.

\begin{figure}[!t]
    \centering
    \includegraphics[width=\linewidth]{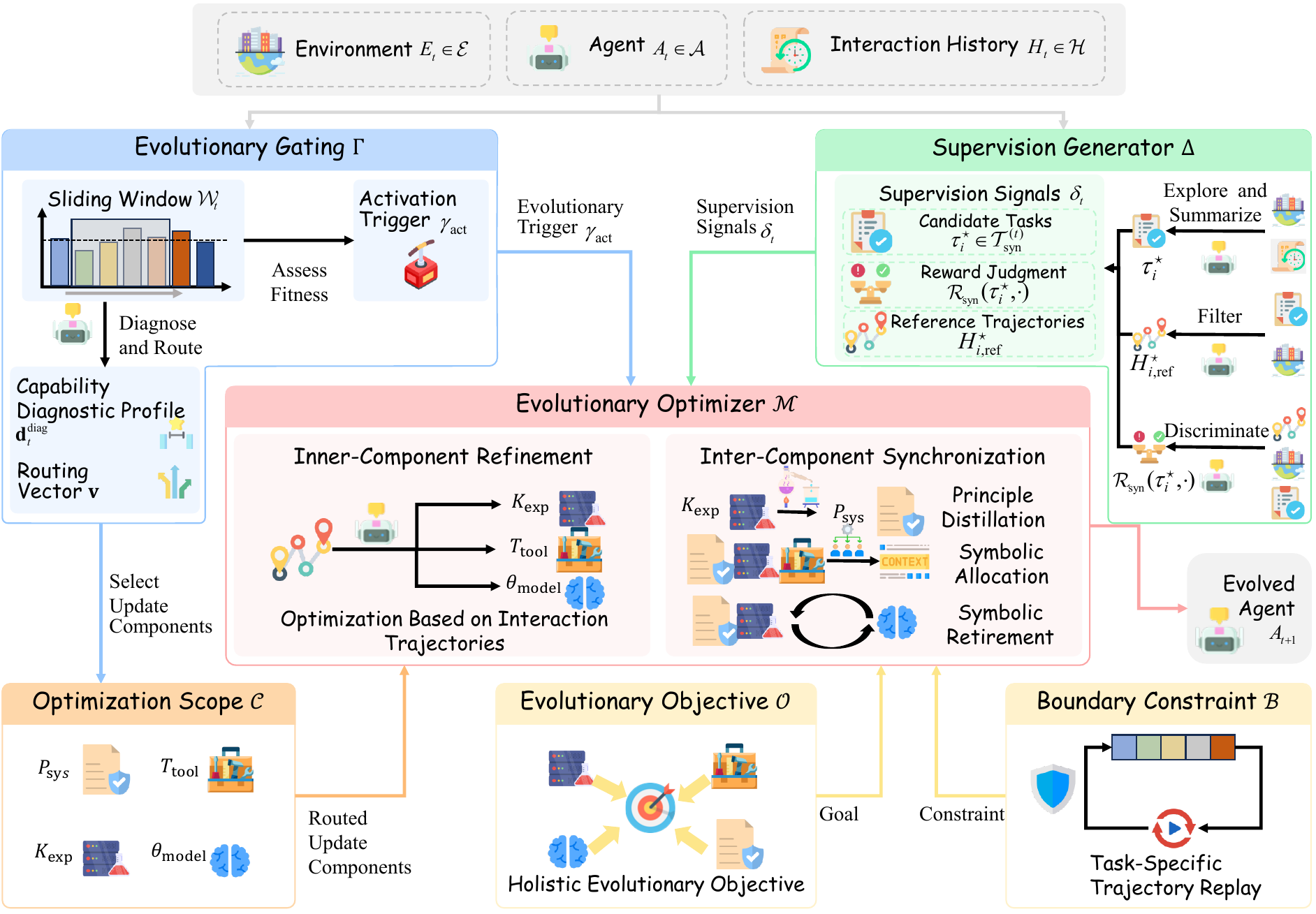}
    \caption{Overview of MUSE. Evolutionary gating ($\Gamma$) decides when to evolve and which components to refine. The supervision generator ($\Delta$) supplements oracle supervision with interaction-grounded tasks and rewards. The optimizer ($\mathcal{M}$) refines and synchronizes the selected components within the scope $\mathcal{C}$, guided by the objective $\mathcal{O}$, while trajectory replay supports the retention aspect of $\mathcal{B}$.}
    \label{fig:overview_muse}
\end{figure}

\section{Methodology} \label{sec:methodology}

The six-component evolutionary strategy provides a structural account of agent evolution, while the Five-Stage Evolutionary Spectrum specifies cumulative requirements for increasing evolutionary autonomy. Guided by the capability requirements of Stage~4, we introduce \textbf{M}etacognitive \textbf{U}nison \textbf{S}elf-\textbf{E}volution (MUSE), a prototype targeting these capabilities through metacognitive gating, interaction-grounded self-supervision, and coordinated component evolution. Following Section~\ref{sec:formalization}, we organize its formulation around $\mathcal{S}=\langle\Delta,\mathcal{C},\mathcal{M},\Gamma,\mathcal{O},\mathcal{B}\rangle$. Section~\ref{subsec:muse_evolution_procedure} brings these components together in the procedure that MUSE follows across successive evolution rounds.

\subsection{Overview} \label{subsec:method_overview}

Figure~\ref{fig:overview_muse} illustrates the macroscopic architecture of MUSE. Following the formulation in Section~\ref{sec:formalization}, MUSE instantiates $\mathcal{S}$ through evolutionary control, supervision generation, and coordinated component updates required for Stage~4. Its optimization scope $\mathcal{C}$ comprises the foundation model weights ($\theta_{\text{model}}\in\Theta$), the experience bank ($K_{\text{exp}}\in\Psi$), general operational principles ($P_{\text{sys}}\in\Psi$), and the functional toolset ($T_{\text{tool}}\in\Psi$). This scope specifies the components available, while $\Gamma$ determines the subset activated in each round.

During deployment, $\Gamma$ monitors recent task performance and interaction accumulation to determine evolution activation, then diagnoses capability gaps and selects adaptation targets (Section~\ref{subsec:evolutionary_gating}). The supervision generator $\Delta$ supplements available oracle supervision with validated synthetic tasks and reference-grounded outcome rewards, and supplies prefix-based process signals for parametric refinement (Section~\ref{subsec:supervision_generator}). Given the selected components and available supervision, $\mathcal{M}$ performs inner-component refinement and inter-component synchronization according to their distinct representational roles (Section~\ref{subsec:optimization_scope_and_evolutionary_optimizer}). The objective $\mathcal{O}$ evaluates overall task performance, while historical trajectory replay supports the capability-retention aspect of $\mathcal{B}$ (Section~\ref{subsec:evolutionary_objective_and_boundary}). The updated agent continues interacting with the environment, providing further evidence for subsequent evolution rounds.

\subsection{Evolutionary Gating \texorpdfstring{$\Gamma$}{Gamma}} \label{subsec:evolutionary_gating}

\begin{figure}[!t]
    \centering
    \includegraphics[width=\linewidth]{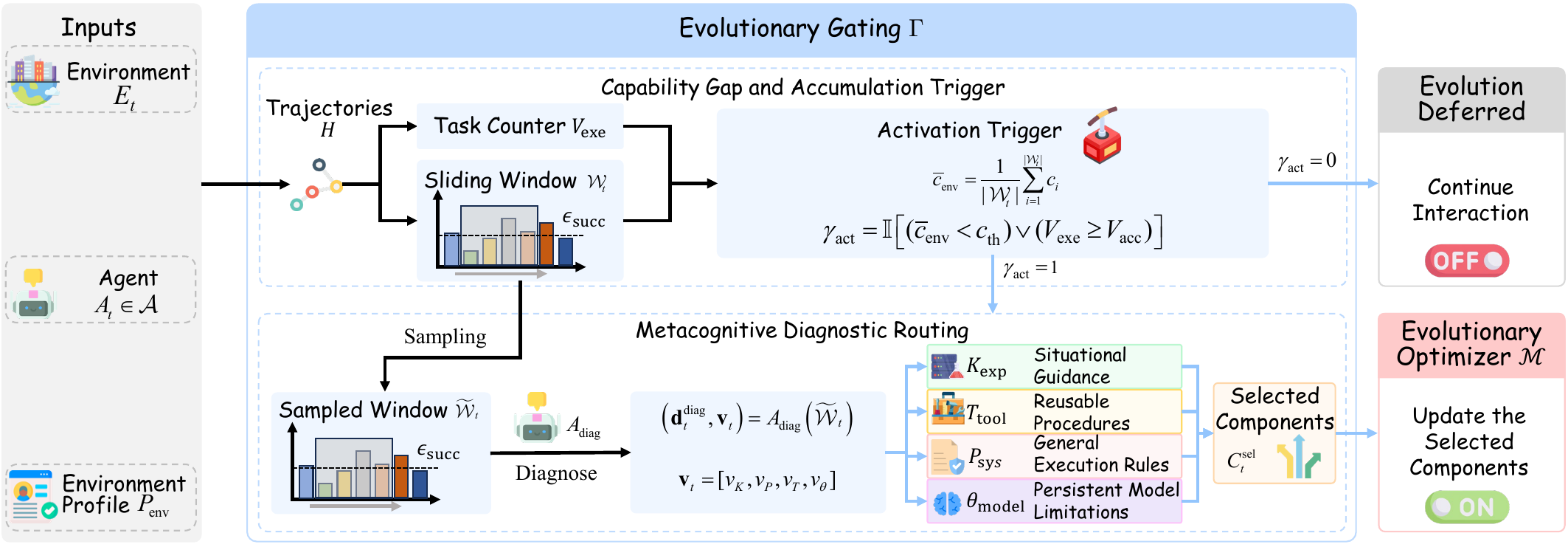}

    \caption{Evolutionary gating ($\Gamma$) in MUSE. A sliding-window performance check and an interaction-count check determine activation. The diagnostic router then compares sampled successful and failed trajectories, including the symbolic support presented during execution, and outputs a capability diagnosis profile of deficiencies and consolidation opportunities and a routing vector over experiences, principles, tools, and model weights.}
    \label{fig:evolutionary_gating_muse}

\end{figure}

Continued deployment provides new interaction evidence, while launching an evolution round incurs additional interaction and optimization costs. Evolutionary control therefore requires both an assessment of adaptation demands and the selection of suitable update components. MUSE connects these decisions through an activation trigger and a metacognitive diagnostic router, as illustrated in Figure~\ref{fig:evolutionary_gating_muse}.

\noindent $\blacktriangleright$ \textbf{Capability Gap and Accumulation Trigger.} A single interaction provides limited evidence of persistent capability gaps, while adaptation after every interaction incurs repeated optimization costs. Continued interaction also provides new behavioral evidence for further refinement, even when aggregate performance remains stable. MUSE therefore combines recent task performance with task accumulation to determine evolution activation. Let $\mathcal{W}_t$ denote a nonempty sliding window of recent interaction records and $c_i\in\{0,1\}$ indicate task completion. The mean completion rate is $\bar{c}_{\text{env}}=\frac{1}{|\mathcal{W}_t|}\sum_{i=1}^{|\mathcal{W}_t|}c_i$, and the activation trigger $\gamma_{\text{act}} \in \{0, 1\}$ is formalized as:
\begin{equation}
    \label{equ:evolution_activation_trigger}
    \gamma_{\text{act}} = \mathbb{I}\left[(\bar{c}_{\text{env}} < c_{\text{th}}) \lor (V_{\text{exe}} \ge V_{\text{acc}})\right],
\end{equation}
where $\mathbb{I}[\cdot]$ is the indicator function, $c_{\text{th}}$ is the performance threshold, $V_{\text{exe}}$ counts task executions since the previous evolution round, and $V_{\text{acc}}$ is the accumulation threshold. These conditions activate an evolution round following insufficient task performance or sustained interaction, with subsequent diagnosis guiding component selection.

\noindent $\blacktriangleright$ \textbf{Metacognitive Diagnostic Routing.} Determining which components should be updated requires a finer assessment of the behaviors underlying task performance. Similar failures can reflect different capability limitations, while successful executions can reveal strategies suitable for broader application. MUSE therefore adopts the evolving agent as a diagnostic router $A_{\text{diag}}$ to translate recurrent patterns across recent trajectories into component-specific adaptation demands. Once activated, the router examines a sampled subset $\tilde{\mathcal{W}}_t\subseteq\mathcal{W}_t$, comparing available successful and failed executions alongside their task descriptions, execution errors, and symbolic assets actually presented to the agent. This context helps distinguish missing symbolic support from difficulties that persist despite available support. The router jointly produces a capability diagnosis profile and a binary routing vector:
\begin{equation}
    \label{equ:metacognitive_diagnostic_routing}
    \left(\mathbf{d}_t^{\mathrm{diag}},\mathbf{v}_t\right)
    =
    A_{\text{diag}}\left(\tilde{\mathcal{W}}_t\right),
    \qquad
    \mathbf{v}_t=[v_K,v_P,v_T,v_\theta].
\end{equation}

The profile $\mathbf{d}_t^{\mathrm{diag}}$ characterizes observed capability limitations and potential consolidation opportunities, connecting their form and scope to component selection. (i) $v_K$ selects the experience bank $K_{\text{exp}}$ for gaps in situational guidance about environment-specific procedures, interface conventions, or operational exceptions. (ii) $v_P$ selects general principles $P_{\text{sys}}$ for abstraction opportunities supported by validated experiences and their observed deployment contexts, where recurring effective experiences suggest guidance applicable across scenarios. (iii) $v_T$ selects the functional toolset $T_{\text{tool}}$ when reusable multi-step operational sequences suggest potential encapsulation of callable abstractions. (iv) $v_\theta$ selects model weights $\theta_{\text{model}}$ for observed reasoning, planning, instruction-following, or output-formatting difficulties that suggest limitations in the underlying foundation model, with the relevance of observed symbolic support considered in the diagnosis. Multiple channels can be selected when the profile identifies co-occurring adaptation demands. The routing vector selects a subset $C_t^{\mathrm{sel}}\subseteq\mathcal{C}$ of components in $A_t$. When evolution is inactive, the selected subset is empty.
\begin{equation}
    \label{equ:evolutionary_gating_output}
    \begin{aligned}
    C_t^{\mathrm{sel}} &= \begin{cases}
    \operatorname{Route}(\mathbf{v}_t, A_t), & \gamma_{\text{act}}=1,\\
    \varnothing, & \gamma_{\text{act}}=0,
    \end{cases}\\
    \Gamma(A_t,E_t,H_t) &= \left(\gamma_{\text{act}},C_t^{\mathrm{sel}}\right).
    \end{aligned}
\end{equation}

Together, the activation trigger determines whether evolution proceeds, and diagnostic routing selects the components to be refined by $\mathcal{M}$. Task performance and trajectory evidence from subsequent interactions provide a basis for activation and routing decisions in later rounds. When $\gamma_{\text{act}}=0$, adaptation is deferred and the agent configuration remains unchanged. Details of the metacognitive gating are provided in Appendix~\ref{subapp:evolutionary_gating}.

\subsection{Supervision Generator \texorpdfstring{$\Delta$}{Delta}} \label{subsec:supervision_generator}

\begin{figure}[!t]
    \centering
    \includegraphics[width=\linewidth]{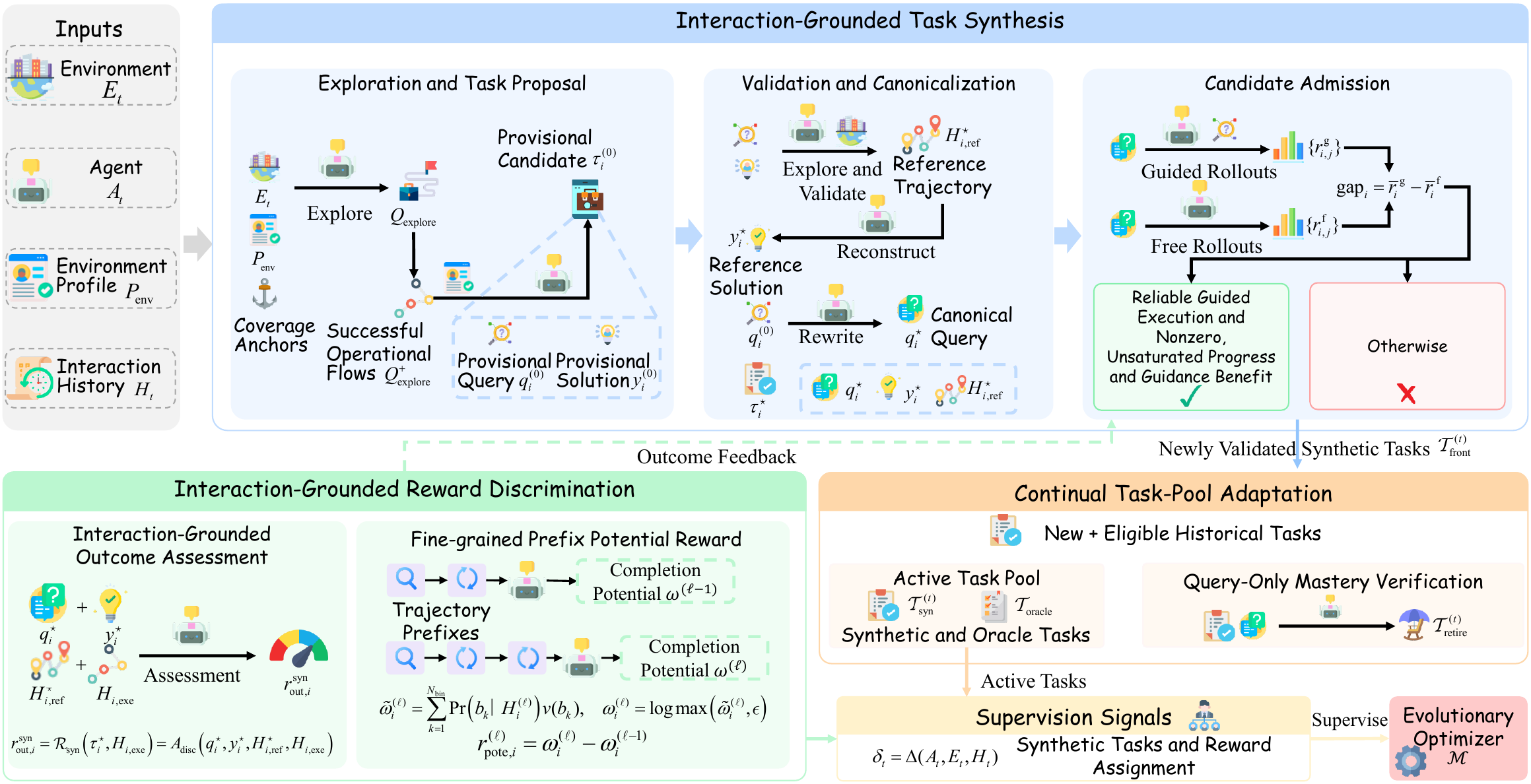}\vspace{-0.2cm}
    \caption{Supervision generator ($\Delta$) in MUSE: interaction-grounded task synthesis, interaction-grounded reward discrimination (outcome rewards for synthetic tasks and prefix-potential step signals for all tasks), and continual task-pool adaptation, jointly supplementing available oracle supervision during evolution.}

    \label{fig:supervision_generator_muse}
\end{figure}

Once $\Gamma$ activates a new round of evolution, updating the selected components requires high-quality supervision signals. However, available oracle supervision might offer limited coverage of the environmental states and operational demands encountered during deployment. Meanwhile, relying solely on task-level rewards also provides limited guidance for intermediate decisions. MUSE therefore instantiates $\Delta:\mathcal{A}\times\mathcal{E}\times\mathcal{H}\rightarrow\mathcal{D}$ to supplement oracle supervision with tasks and evaluation rules at task and step levels. Figure~\ref{fig:supervision_generator_muse} illustrates how task synthesis, reward discrimination, and task-pool adaptation jointly support supervision construction and renewal.

\noindent $\blacktriangleright$ \textbf{Interaction-Grounded Task Synthesis.} Automatically synthesized tasks can extend supervision coverage~\citep{zhou2025selfchallenging,xia2025agent0,huang2025rzero}, while their supervisory value depends on environmental validity and relevance to current capabilities. A plausible task description may still refer to unavailable states or unsupported operations. MUSE therefore grounds task construction and validation in direct interaction. Inspired by the self-questioning paradigm of AgentEvolver~\citep{zhai2025agentevolver}, the evolving agent explores the environment without a predefined task query and abstracts observed workflows into provisional queries and solutions. Solution-guided execution examines operational feasibility and whether the resulting actions satisfy the proposed objective. The query is then reformulated to separate goals and constraints from procedural solution details and checked for consistency with the validated execution. The resulting canonical task is formulated as:
\begin{equation}
    \label{equ:canonical_synthetic_task}
    \tau_i^{\star} = \langle q_i^{\star}, y_i^{\star}, H_{i,\text{ref}}^{\star} \rangle.
\end{equation}
Here, $q_i^{\star}$ denotes the canonical query, $y_i^{\star}$ denotes the reference solution reconstructed from successful validation, and $H_{i,\text{ref}}^{\star}$ records the supporting environmental interaction as the reference trajectory.

A feasible task additionally requires assessment of its suitability for the current agent. MUSE compares \textit{guided rollouts} receiving both $q_i^{\star}$ and $y_i^{\star}$ with \textit{free rollouts} receiving only $q_i^{\star}$. Both are evaluated by the reward discriminator introduced below. Let $\bar{r}_i^{\mathrm{g}}$ and $\bar{r}_i^{\mathrm{f}}$ denote their mean outcome rewards. The empirical guidance gap is:
\begin{equation}
    \label{equ:guidance_gap}
    \operatorname{gap}_i=\bar{r}_i^{\mathrm{g}}-\bar{r}_i^{\mathrm{f}}.
\end{equation}
This gap measures the execution benefit of reference guidance. Candidate admission requires reliable guided execution, nonzero yet unsaturated independent progress, and a sufficient advantage from reference guidance. Tasks satisfying these criteria form the validated frontier set $\mathcal{T}_{\text{front}}^{(t)}$, which supplies new candidates to the supervision pool.

\noindent $\blacktriangleright$ \textbf{Interaction-Grounded Reward Discrimination.} A reference execution trajectory demonstrates a feasible solution, while evaluating new attempts requires rewards reflecting actual task completion. Model-generated feedback supports iterative refinement~\citep{madaan2023self,chen2025spc}, and the risk of self-reinforcing errors motivates grounding these assessments in environmental evidence~\citep{he2026how}. MUSE therefore employs the evolving agent as an intrinsic discriminator $A_{\text{disc}}$ to assess task completion against the canonical objective and validated reference evidence. During free rollouts and subsequent query-only rollouts, the acting agent receives $q_i^{\star}$ without the reference solution $y_i^{\star}$ or reference trajectory $H_{i,\text{ref}}^{\star}$. After execution, the discriminator examines the observed actions and environmental feedback in $H_{i,\text{exe}}$ and assigns an outcome reward:
\begin{equation}
    \label{equ:synthetic_reward}
    r_{\text{out},i}^{\text{syn}} 
    = \mathcal{R}_{\text{syn}}\left(\tau_i^{\star},H_{i,\text{exe}}\right) 
    = A_{\text{disc}}\left(q_i^{\star},y_i^{\star},H_{i,\text{ref}}^{\star},H_{i,\text{exe}}\right),
\end{equation}
where $\mathcal{R}_{\text{syn}}$ denotes the outcome reward function for synthetic tasks and $r_{\text{out},i}^{\text{syn}}\in[0,1]$ estimates the degree of task completion. The discriminator supplies task-performance evidence for guided and free candidate evaluation, subsequent optimization, and pool maintenance. Oracle tasks retain their environment-provided evaluation criteria.

For both oracle and synthetic tasks, outcome rewards assess overall completion but provide limited guidance for intermediate decisions. To complement this task-level feedback, MUSE estimates a \textit{prefix potential} from the task objective and visible interaction history, tracking changes in assessed task-completion prospects as execution proceeds. For execution $i$, the task query $q_i$ expresses the goal and its constraints. Starting from the query-only prefix $H_i^{(0)}=(q_i)$, the evaluator forms $H_i^{(\ell)}$ by appending the executed action $u_{i,\ell}$ and any returned observation $o_{i,\ell+1}$ to $H_i^{(\ell-1)}$. Direct scalar scoring summarizes each assessment in a single decoded value. To retain relative support across potential levels, the evolving agent assigns normalized probabilities $\Pr(b_k\mid H_i^{(\ell)})$ to ordered potential bins $\{b_k\}_{k=1}^{N_{\mathrm{bin}}}$, conditioned on the visible prefix. Each bin has a numerical anchor $v(b_k)=(k-1)/(N_{\mathrm{bin}}-1)$, where $N_{\mathrm{bin}}\ge2$. The probability-weighted potential and its logarithmic representation are:
\begin{equation}
    \label{equ:prefix_potential}
    \tilde{\omega}_i^{(\ell)}
    = \sum_{k=1}^{N_{\mathrm{bin}}}
    \Pr\left(b_k\mid H_i^{(\ell)}\right)v(b_k),\quad
    \omega_i^{(\ell)}
    = \log\max\left(\tilde{\omega}_i^{(\ell)},\epsilon\right),
\end{equation}
where $\epsilon\in(0,1)$ is a numerical floor. Differences between adjacent log-potentials express relative changes in the assessed completion prospects, yielding the potential-based step reward:
\begin{equation}
    \label{equ:potential_step_reward}
    r_{\text{pote},i}^{(\ell)}
    = \omega_i^{(\ell)}-\omega_i^{(\ell-1)}
    = \log
    \frac{\max\left(\tilde{\omega}_i^{(\ell)},\epsilon\right)}
         {\max\left(\tilde{\omega}_i^{(\ell-1)},\epsilon\right)}.
\end{equation}

Positive and negative values indicate increases and decreases in assessed completion prospects of the task, respectively. These self-supervised signals provide supplementary process supervision alongside task-level rewards. The optimizer $\mathcal{M}$ normalizes and incorporates these signals into parameter updates (Section~\ref{subsec:optimization_scope_and_evolutionary_optimizer}). Scoring configurations and reward-processing details are provided in Appendix~\ref{subapp:supervision_generator}.

\noindent $\blacktriangleright$ \textbf{Continual Task-Pool Adaptation.} Candidate admission reflects the current assessment of task suitability, while its supervisory value may change as agent capabilities evolve. Previously informative tasks can become mastered, and continued task synthesis introduces additional learning opportunities. MUSE therefore maintains a capacity-bounded active pool $\mathcal{T}_{\text{syn}}^{(t)}$ that integrates newly validated tasks from $\mathcal{T}_{\text{front}}^{(t)}$ with eligible historical tasks. Pool selection balances the guidance gap, variability in free-rollout rewards, and structural diversity, with priorities reassessed as agent capabilities change. Tasks exhibiting sustained successful execution are retired after passing separate query-only verification, releasing capacity for further supervision.

Together, the three modules provide supplementary tasks and task-level evaluation criteria for component refinement, together with step-level scoring rules for parametric updates:
\begin{equation}
    \label{equ:supervision_generator_output}
    \delta_t = \Delta(A_t,E_t,H_t).
\end{equation}
The evaluation rules are applied as execution trajectories become available during optimization. The resulting $\delta_t$ complements available oracle supervision for optimization by $\mathcal{M}$ over the components selected by $\Gamma$. Implementation details for exploration procedures, task canonicalization, admission thresholds, outcome reward discrimination for synthetic tasks, prefix assessment, and task-pool management are provided in Appendix~\ref{subapp:supervision_generator}.

\subsection{Optimization Scope \texorpdfstring{$\mathcal{C}$}{C} and Evolutionary Optimizer \texorpdfstring{$\mathcal{M}$}{M}} \label{subsec:optimization_scope_and_evolutionary_optimizer}

With components selected for adaptation by $\Gamma$ and supervision provided by $\Delta$, MUSE translates the interaction evidence into persistent updates of the agent system throughout deployment. Different components encode capabilities in distinct representational forms and support different functions, while changes to one component can affect the contribution of another. MUSE therefore organizes the evolutionary optimizer $\mathcal{M}$ into two complementary processes: inner-component refinement and inter-component synchronization.

For MUSE, $\mathcal C=\{K_{\mathrm{exp}},P_{\mathrm{sys}},T_{\mathrm{tool}},\theta_{\mathrm{model}}\}\subseteq\Theta\cup\Psi$. The routing vector identifies the directly activated refinement channels. When $T_{\mathrm{tool}}$ is selected, toolset refinement may also adjust its management settings $\phi_{\mathrm{tool}}\in\Phi$. Model weight refinement can trigger reassessment of related symbolic entries. Under supervision $\delta_t$, the updated configuration of the selected subset $C_t^{\mathrm{sel}}\subseteq\mathcal C$ is formulated as:
\begin{equation}
    \label{equ:component_update}
    \hat{C}_t^{\mathrm{sel}}
    = \mathcal{M}\left( A_t, C_t^{\mathrm{sel}},\delta_t \right).
\end{equation}

\subsubsection{Inner-Component Refinement}

\begin{figure}[!t]
    \centering
    \includegraphics[width=\linewidth]{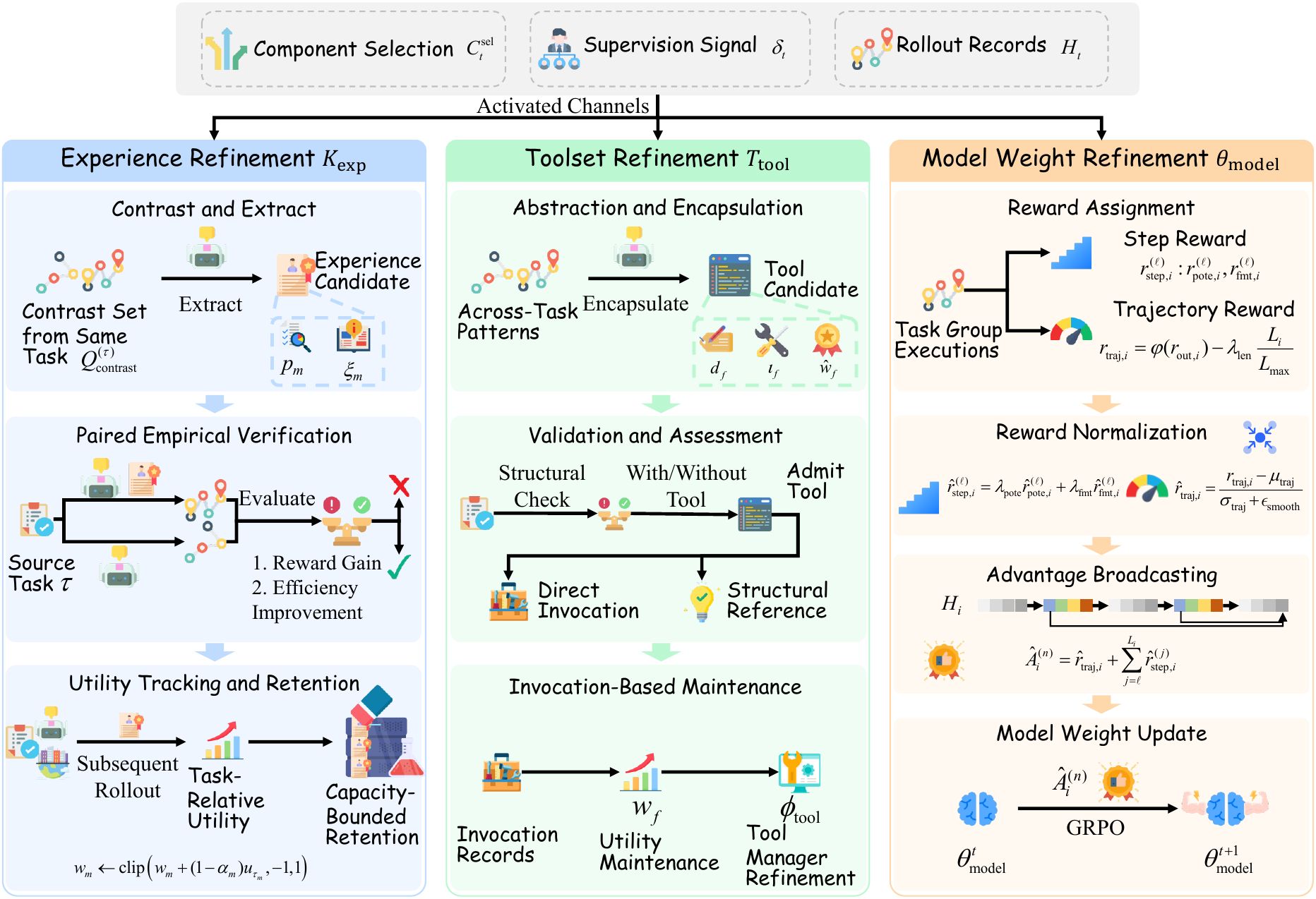}

    \caption{Inner-component refinement in MUSE. $\Gamma$ selects update channels, with supervision from $\Delta$ and rollout records guiding refinement. \textit{Experience Refinement} extracts situational guidance from contrasting executions of the same task. \textit{Toolset Refinement} abstracts recurring operations into tools supporting direct invocation and structural reference. Both symbolic channels verify candidates and reassess their utility. \textit{Model Weight Refinement} converts trajectory and step rewards into action-token advantages for GRPO.}

    \label{fig:muse_inner_refinement}
\end{figure}

Inner-component refinement connects the supervision associated with each selected target to its corresponding adaptation procedure. Experiences provide situational guidance, tools encapsulate executable operations, and model weights support learned behavioral capabilities. MUSE refines these representations through the component-specific channels illustrated in Figure~\ref{fig:muse_inner_refinement}, with activation determined by $\Gamma$.

\noindent $\blacktriangleright$ \textbf{Experience Bank $K_{\text{exp}}$}.

Persistent memory preserves information across interactions~\citep{packer2023memgpt, hu2023chatdb}, while trajectory reflection can extract compact behavioral guidance~\citep{zhai2025agentevolver}. Experience refinement additionally requires connecting this guidance to observed performance and maintaining its relevance as agent capabilities change. MUSE first contrasts a positive anchor with alternative executions of the same task, forming $Q_{\text{contrast}}^{(\tau)}$. The evolving agent acts as the experience extractor $A_{\text{exp}}$, identifying a situational precondition $p_m$ and a guideline $\xi_m$:
\begin{equation}
    \label{equ:experience_extraction}
    m = \langle p_m, \xi_m \rangle
    = A_{\text{exp}}\left(Q_{\text{contrast}}^{(\tau)}\right).
\end{equation}

A plausible guideline may still provide limited assistance during execution, motivating empirical verification before admission. After semantic screening distinguishes new candidates from revisions, MUSE compares candidate and control configurations on the source task with other symbolic assets fixed. New candidates are evaluated against their absence, while revisions are compared with the corresponding existing entries. Admission or replacement requires a sufficient reward improvement, or an efficiency improvement under saturated rewards or a nonsaturated reward tie with at least one positive score. These comparisons establish the initial utility of accepted experiences.

Initial verification reflects the capability state at admission, while subsequent learning can change the contribution of retained guidance. MUSE therefore tracks task-relative utility during later interactions. Let $\mathcal{T}_{\cap}^{(t_{\text{roll}})}$ contain tasks observed in consecutive rollout phases, and let $r_\tau^\Delta$ denote the change in their mean task rewards. For a task $\tau_m$ receiving experience $m$, the utility signal subtracts the common performance change before updating the entry:
\begin{equation}
    \label{equ:task_relative_utility}
    \begin{aligned}
        u_{\tau_m}
        &=
        \operatorname{clip}\left(
            r_{\tau_m}^{\Delta}
            -
            \frac{1}{\left|\mathcal{T}_{\cap}^{(t_{\text{roll}})}\right|}
            \sum_{\tau\in\mathcal{T}_{\cap}^{(t_{\text{roll}})}}
            r_\tau^\Delta,
            -1,1
        \right),\\
        w_m
        &\leftarrow
        \operatorname{clip}\left(
            w_m+(1-\alpha_m)u_{\tau_m},
            -1,1
        \right).
    \end{aligned}
\end{equation}
Here, $\alpha_m\in(0,1)$ controls the adjustment magnitude. Updates apply to delivered experiences when common-task evidence is available. Utility and retrieval frequency guide capacity-bounded retention, while probation and retrieval restrictions support continued assessment before utility-based removal. This maintenance connects experience retention to deployment evidence and supplies empirically supported candidates for principle abstraction. Detailed extraction, verification, and maintenance procedures are provided in Appendix~\ref{subapp:experience_bank_refinement}.

\noindent $\blacktriangleright$ \textbf{Functional Toolset $T_{\text{tool}}$}.

Recurring operation sequences motivate callable abstractions that can reduce the effort of repeatedly constructing similar action plans~\citep{qian2023creator, wang2025toolgen, xiong2026learning}. Their applicability depends on capturing shared operational structure across task contexts. MUSE therefore extracts recurring patterns from trajectories of multiple tasks and encapsulates the corresponding procedures together with their execution dependencies into callable functional tools. Given a dependency-closed pattern $\iota_{\mathrm{dc}}$ and a representative successful execution $H_{\mathrm{src}}^{\star}$, candidate construction is formulated as follows:
\begin{equation}
    \label{equ:tool_encapsulation}
    \tilde{f}
    = \left\langle d_f,\iota_f,\hat{w}_f \right\rangle
    = \operatorname{Encap}\left(
        \iota_{\mathrm{dc}},H_{\mathrm{src}}^{\star}
    \right).
\end{equation}
Here, $d_f$ describes the functional intent and callable interface, $\iota_f$ specifies the operational realization, and $\hat{w}_f$ denotes provisional utility estimated from supporting trajectories.

Variation in operational requirements and environmental conditions across tasks motivates two complementary forms of tool-mediated support. The agent can directly invoke a tool when its encapsulated procedure matches the current requirements, or inspect its realization as a structural reference for constructing an adapted action sequence. MUSE therefore exposes both the callable interface and the underlying procedure, connecting efficient execution of recurring operations with flexible adaptation to task-specific constraints.

Tool availability can influence both action selection and execution effort, motivating empirical assessment of its contribution. Following structural and execution checks, MUSE compares paired rollouts with and without the candidate on source and compatible tasks, holding other symbolic assets and operational settings fixed. The resulting reward difference measures the empirical contribution of making the candidate available:
\begin{equation}
    \label{equ:tool_exposure_gain}
    \operatorname{gain}(\tilde{f}) = \bar{r}_{\mathrm{tool}}^{+}(\tilde{f}) - \bar{r}_{\mathrm{tool}}^{-}(\tilde{f}).
\end{equation}
The two terms denote mean task rewards, allowing the comparison to reflect both direct invocation and reference-guided adaptation. Sufficient reward improvements or efficiency improvements under comparable rewards determine candidate activation and initial utility, while inconclusive evidence motivates subsequent reassessment.

Tool value can change as task demands and agent capabilities evolve, motivating continued assessment beyond initial verification. During deployment, actual invocation provides an observable basis for tracking the performance associated with an executable procedure. For a tool $f$ invoked on task $\tau_f$, MUSE updates its utility through the task-relative signal introduced for experience maintenance:
\begin{equation}
    \label{equ:tool_utility_update}
    w_f \leftarrow
    \operatorname{clip}\left(
        w_f+(1-\alpha_f)u_{\tau_f},
        -1,1
    \right).
\end{equation}
Here, $u_{\tau_f}$ measures task-performance change relative to the common trend across consecutive rollout phases, and $\alpha_f\in(0,1)$ controls the adjustment magnitude. These updates apply when sufficient common-task evidence is available from repeated task execution. Utility, task completion, and internal execution reliability jointly guide continued deployment, reassessment, and retirement from retrieval.

As the tool inventory evolves, its management configuration also shapes retrieval competition among available tools and context demands during task execution. MUSE therefore evaluates bounded adjustments to $\phi_{\mathrm{tool}}\in\Phi$ on matched validation tasks, retaining proposals with no decrease in mean task performance. This connects individual tool refinement to the conditions governing their collective deployment. Detailed construction procedures, verification criteria, and maintenance rules are provided in Appendix~\ref{subapp:functional_toolset_refinement}.

\noindent $\blacktriangleright$ \textbf{Foundation Model Weights $\theta_{\text{model}}$}.

Persistent difficulties in reasoning, planning, and instruction following motivate adaptation of the parametric cognitive core. To connect parameter updates to both overall task completion and assessed intermediate progress, MUSE integrates the task-level rewards and prefix-potential signals supplied through $\Delta$ (Section~\ref{subsec:supervision_generator}) into trajectory and step rewards for advantage estimation. Screened interaction trajectories $H_i$ are grouped by source task, providing comparisons among alternative executions of the same objective.

At the step level, assessed progress toward task completion and optional format penalty provide complementary signals. MUSE combines the potential-based reward $r_{\text{pote},i}^{(\ell)}$ from Equation~\ref{equ:potential_step_reward} with an optional format penalty $r_{\text{fmt},i}^{(\ell)}$ after normalizing each component separately within its task group:
\begin{equation}
    \label{equ:model_step_reward}
    \hat{r}_{\text{step},i}^{(\ell)} = \lambda_{\text{pote}} \hat{r}_{\text{pote},i}^{(\ell)} + \lambda_{\text{fmt}} \hat{r}_{\text{fmt},i}^{(\ell)}
\end{equation}
Here, $\hat{r}_{\text{pote},i}^{(\ell)}$ and $\hat{r}_{\text{fmt},i}^{(\ell)}$ denote the separately normalized components, with details in Appendix~\ref{subapp:foundation_model_weights}. The coefficient $\lambda_{\text{pote}}$ weights potential-derived progress within the step reward, while $\lambda_{\text{fmt}}$ controls the contribution of format compliance. The raw format term is 0 for valid actions and negative for violations of the required syntax or schema. This combination connects assessed progress to the structural requirements of executable actions.

At the trajectory level, overall task completion provides a complementary objective spanning the full interaction. MUSE transforms the task reward and incorporates an execution-length penalty to account for operational efficiency:
\begin{equation}
    \label{equ:model_trajectory_reward}
    r_{\text{traj},i} =
    \varphi(r_{\text{out},i})
    -
    \lambda_{\text{len}}
    \frac{L_i}{L_{\text{max}}}.
\end{equation}
Here, $r_{\text{out},i}$ denotes the available task-level reward, derived from the environment for oracle tasks or from $\mathcal{R}_{\text{syn}}$ for synthetic tasks. To emphasize distinctions among trajectories closer to task completion, the default configuration applies $\varphi(x)=x^2$. For scores in $[0,1]$, this transformation gives greater sensitivity to differences at higher reward levels and compresses differences among low-scoring attempts. The quantities $L_i$ and $L_{\text{max}}$ denote the executed and maximum interaction steps, respectively, and $\lambda_{\text{len}}$ controls the execution-length penalty.

The two reward dimensions differ in numerical scale and temporal granularity, motivating separate normalization within each task group. Trajectory rewards are standardized across retained trajectories for the same task, yielding $\hat{r}_{\text{traj},i}$, while the potential and format components are normalized separately before their weighted combination into $\hat{r}_{\text{step},i}^{(\ell)}$ in Equation~\ref{equ:model_step_reward}. Since an action can influence subsequent decisions, MUSE combines the normalized trajectory reward with cumulative step rewards from the current step onward. For each trainable action token $n$ generated at step $\ell$, the advantage is formulated as follows:
\begin{equation}
    \label{equ:model_token_advantage}
    \hat{A}_{i}^{(n)}
    =
    \hat{r}_{\text{traj},i}
    +
    \sum_{j=\ell}^{L_i}
    \hat{r}_{\text{step},i}^{(j)}.
\end{equation}

Symbolic assets can support action generation during rollout collection, providing behavioral evidence for parametric refinement. Training retains task queries, generated actions, and environmental observations while omitting explicit retrieved guidance. The training loss applies to generated action tokens, with padding entries masked. MUSE incorporates the combined advantages into the GRPO objective~\citep{shao2024deepseekmath}, allowing task-level and step-level supervision to jointly guide updates of $\theta_{\text{model}}$. Policy clipping and reference-policy regularization moderate changes during continued adaptation.

The resulting parameter updates can subsequently change the contribution of symbolic support, motivating the reassessment introduced under inter-component synchronization. Trajectory preparation, reward-processing settings, and the complete optimization objective are provided in Appendix~\ref{subapp:foundation_model_weights}.

\subsubsection{Inter-Component Synchronization}

\begin{figure}[!t]
    \centering
    \includegraphics[width=\linewidth]{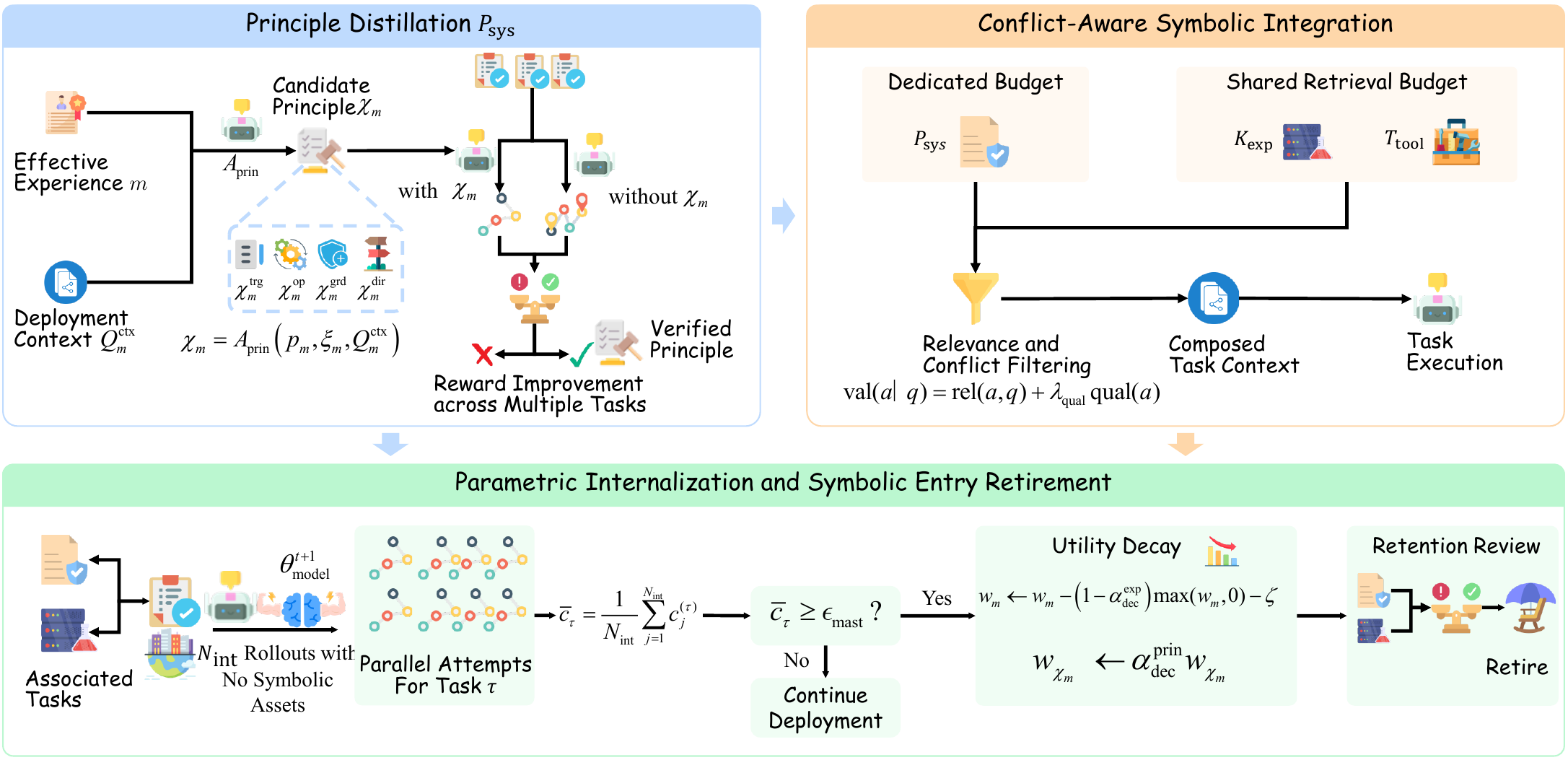}
    \caption{Inter-component synchronization in MUSE. \textit{Principle Distillation} derives principles from effective experiences and verifies them across tasks. \textit{Conflict-Aware Symbolic Integration} combines heterogeneous symbolic assets to guide execution. \textit{Parametric Internalization and Symbolic Entry Retirement} tests associated tasks without symbolic support. Verified mastery lowers the retention priority of the corresponding experiences and principles.}

    \label{fig:muse_inter_synchronization}
\end{figure}

Inner-component refinement adapts individual components, while their contributions remain distributed across different forms and abstraction levels. Experiences can support broader principles, symbolic assets jointly shape task execution, and parameter updates can change the value of explicit guidance. MUSE coordinates these dependencies through principle abstraction, symbolic integration, and symbolic retirement, as illustrated in Figure~\ref{fig:muse_inter_synchronization}.

\noindent $\blacktriangleright$ \textbf{Distilling General Principles $P_{\text{sys}}$ from Experiences}.

Repeatedly effective experiences may reflect general operational regularities extending beyond their original situational descriptions. Promoting these regularities into general principles can provide guidance for tasks sharing the same structural conditions. MUSE therefore distills general principles from experiences with sustained empirical utility, examining their observed deployment contexts to identify the conditions supporting broader applicability. The evolving agent acts as the principle abstractor $A_{\text{prin}}$:
\begin{equation}
    \label{equ:principle_abstraction}
    \chi_m = \left\langle \chi_m^{\mathrm{trg}}, \chi_m^{\mathrm{op}}, \chi_m^{\mathrm{grd}}, \chi_m^{\mathrm{dir}} \right\rangle = A_{\text{prin}}\left(p_m, \xi_m, Q_m^{\mathrm{ctx}}\right).
\end{equation}
Here, $Q_m^{\mathrm{ctx}}$ contains observed deployment contexts of experience $m$. The resulting principle specifies a structural trigger, an operational pattern, execution constraints, and a directive.

Expanding the proposed applicability introduces a corresponding demand for evidence beyond the source task. Following structural and novelty screening, MUSE evaluates the principle on other tasks satisfying its trigger through matched executions with and without the candidate. Acceptance requires sufficient mean task-reward improvement supported across multiple tasks, together with adequate evaluation coverage and execution reliability. This assessment connects the proposed generalization to demonstrated benefits across compatible contexts.

Accepted principles and their source experiences provide complementary guidance at different levels, with provenance links preserving the connection between them. As agent capabilities evolve, task evaluation provides evidence for adjusting the retention priority of associated principles within the bounded pool. Their joint delivery can introduce overlapping or incompatible guidance, motivating coordinated selection during task execution. Detailed abstraction, verification, and maintenance procedures are provided in Appendix~\ref{subapp:general_principle_distillation}.

\noindent $\blacktriangleright$ \textbf{Conflict-Aware Symbolic Integration}.

Principle distillation introduces broader guidance that can overlap with existing experiences and the procedures exposed by tools. A principle and its source experience may both match a task while expressing the same recommendation at different levels of guidance. Independently derived guidance may also recommend incompatible actions under overlapping conditions. MUSE therefore coordinates symbolic delivery to preserve complementary task support while managing redundancy and conflicting instructions within a bounded context.

Limited context motivates considering the additional information provided by an entry alongside its individual relevance. Among candidates satisfying the applicability and deployment conditions of their source components, MUSE applies maximum marginal relevance to balance task relevance against similarity to already selected entries. The coefficient $\lambda_{\mathrm{mmr}}$ controls this balance, favoring relevant content that broadens the selected support. Principles retain a dedicated delivery budget, while experiences and tools share the retrieved-knowledge budget.

Selecting compatible guidance additionally requires identifying contradictory recommendations. MUSE assesses contradictions among principles and experiences with overlapping applicability through bidirectional natural language inference. Task relevance and empirical utility then guide the choice between conflicting entries:
\begin{equation}
    \label{equ:symbolic_candidate_value}
    \operatorname{val}(a\mid q)
    =
    \operatorname{rel}(a,q)
    +
    \lambda_{\mathrm{qual}}\operatorname{qual}(a).
\end{equation}
Here, $a$ denotes a candidate principle or experience, $\operatorname{rel}(a,q)$ denotes the relevance of $a$ to query $q$, $\operatorname{qual}(a)$ denotes normalized empirical utility, and $\lambda_{\mathrm{qual}}$ weights deployment evidence. The higher-value entry is retained when conflicting guidance is identified. Provenance links distinguish principles and their source experiences, whose overlap can reflect abstraction of the same guidance. Tools contribute executable support during task execution, with operation-family comparisons identifying redundant implementations.

Semantic overlap is consolidated when entries serve comparable retrieval roles, allowing broader principles and situational detail to coexist when they provide complementary support. Selected entries are delivered as complete blocks so that applicability conditions and operational content accompany the guidance. This coordination connects individually refined symbolic assets to the combined context through which they influence task execution. Detailed retrieval, consistency assessment, and context-construction procedures are provided in Appendix~\ref{subapp:conflict_aware_symbolic_integration}.

\noindent $\blacktriangleright$ \textbf{Parametric Internalization and Symbolic Entry Retirement}.

Conflict-aware integration coordinates the delivery of symbolic entries during task execution. As foundation model weights evolve, the agent may complete some associated tasks without the guidance previously supplied by symbolic assets. Historical utility can consequently become an incomplete basis for retention decisions. MUSE therefore evaluates task proficiency without stored symbolic support following parameter updates, providing evidence for adjusting the retention priority of associated guidance.

Reassessing retention priorities requires evidence of task execution without stored symbolic support. MUSE evaluates tasks associated with existing experiences and principles, without access to any stored experiences, principles, or generated functional tools. Environment-provided operations remain available. Since individual executions can vary, task proficiency is summarized across repeated trials:
\begin{equation}
    \label{equ:internalization_completion}
    \bar{c}_{\tau} = \frac{1}{N_{\text{int}}} \sum_{j=1}^{N_{\text{int}}} c_j^{(\tau)}.
\end{equation}
Here, $c_j^{(\tau)}\in\{0,1\}$ denotes completion of task $\tau$ in trial $j$, and $N_{\text{int}}$ is the number of trials. This evaluation provides task-specific evidence for reconsidering the retention priority of associated guidance under the updated model.

An entry may still contribute in other contexts after an associated task becomes solvable without stored symbolic support, motivating gradual adjustment of its utility. When $\bar{c}_{\tau}$ reaches the mastery threshold $\epsilon_{\text{mast}}$, MUSE applies utility decay only to experiences and principles associated with the verified task $\tau$ according to:
\begin{equation}
    \label{equ:symbolic_utility_decay}
    \begin{aligned}
        w_m &\leftarrow w_m - \left(1-\alpha_{\mathrm{dec}}^{\mathrm{exp}}\right) \max(w_m,0) - \zeta,\\
        w_{\chi_m} &\leftarrow \alpha_{\mathrm{dec}}^{\mathrm{prin}}w_{\chi_m}.
    \end{aligned}
\end{equation}
The factors $\alpha_{\mathrm{dec}}^{\mathrm{exp}}$ and $\alpha_{\mathrm{dec}}^{\mathrm{prin}}$ lie in $(0,1)$, and $\zeta$ controls the additional experience decrement. For experiences, proportional shrinkage is restricted to positive utility because multiplying a negative score by a factor below one would increase its value. The additive decrement also lowers weak or non-positive utility, supporting utility-based retirement. For principles, proportional decay adjusts their priority within the bounded principle pool.

These adjustments connect evidence of current proficiency to the existing retention policies. Task association determines the entries subject to this utility decay, while subsequent maintenance follows the existing lifecycle and capacity criteria across the retained pools. Experience probation provides opportunities for further assessment before utility-based removal, while principle retirement follows utility ranking under the capacity bound. Generated functional tools remain governed by their invocation-based maintenance. Task-proficiency evidence gathered after parameter updates thus guides the continued allocation of symbolic support. Verification schedules, task selection, and detailed retirement criteria are provided in Appendix~\ref{subapp:parametric_internalization}.

\subsection{Evolutionary Objective \texorpdfstring{$\mathcal{O}$}{O} and Boundary \texorpdfstring{$\mathcal{B}$}{B}} \label{subsec:evolutionary_objective_and_boundary}

The preceding evolutionary mechanisms determine when adaptation is activated, which supervision signals are available, and how selected components are updated. Coordinating these updates over successive evolution rounds additionally requires a shared criterion for capability improvement and operational requirements for retaining previously acquired capabilities. MUSE instantiates the evolutionary objective $\mathcal{O}$ through empirical task performance and supports the capability-retention aspect of $\mathcal{B}$ through task-specific trajectory replay.

\noindent \textbf{Overarching Evolutionary Objective ($\mathcal{O}$).} 
MUSE instantiates the evaluation mapping $\mathcal{O}: \mathcal{A} \times \mathcal{E} \rightarrow \mathbb{R}$ through empirical task performance of the evolving agent. This shared objective connects the heterogeneous updates in Section~\ref{subsec:optimization_scope_and_evolutionary_optimizer}. Symbolic validation and utility estimates assess contributions to task execution, while task-level and step-level rewards guide parametric refinement through $\mathcal{M}$.

\noindent \textbf{Supporting Capability Retention through Task-Specific Trajectory Replay ($\mathcal{B}$).} 
As adaptation shifts toward new tasks, recent rollouts may underrepresent previously acquired behavior. Reintroducing eligible historical executions into training provides support for previously acquired capabilities. MUSE addresses this need through task-specific trajectory replay within parametric optimization, supporting the capability-retention aspect of $\mathcal{B}$.

A bounded replay buffer $Q_{\text{buffer}}$ retains eligible historical trajectories organized by task. Replay draws on both tasks revisited in the current round and earlier tasks absent from the new rollout collection, extending training coverage beyond immediate adaptation demands. To balance historical evidence with current learning, MUSE combines new trajectories $Q_{\text{new}}$ with a selected historical subset $Q_{\text{replay}}$ under a replay-volume bound:
\begin{equation}
    \label{equ:replay_mixture_and_bound}
    \begin{aligned}
        Q_{\text{mix}} &= Q_{\text{new}} \cup Q_{\text{replay}},\\
        |Q_{\text{replay}}| &\le \left\lfloor \lambda_{\text{replay}}  N_{\text{new}}^{\mathrm{nom}} \right\rfloor.
    \end{aligned}
\end{equation}
Here, $\lambda_{\text{replay}}\in(0,1)$ bounds replay volume relative to the nominal rollout allocation $N_{\text{new}}^{\mathrm{nom}}$ of the current task set, with actual volume depending on eligible evidence. The mixed collection supplies task groups for reward normalization and parametric updates in $\mathcal{M}$. By including eligible historical trajectories in these updates, MUSE continues to learn from earlier task executions as it adapts to new tasks. Replay thereby provides implicit support for the capability-retention aspect of $\mathcal{B}$. Details of the replay mechanism are provided in Appendix~\ref{subapp:evolutionary_objective_and_boundary}.

\subsection{MUSE Evolution Procedure} \label{subsec:muse_evolution_procedure}

Algorithm~\ref{alg:muse_evolution} summarizes how MUSE carries out the component updates selected by metacognitive gating once evolution is activated. Each selected component collects fresh rollouts for its own refinement. When symbolic assets and model weights are selected in the same round, the rollouts for parametric refinement therefore reflect the updated symbolic assets. At the end of the round, MUSE evaluates the joint configuration and selects the agent for deployment. Evaluation trajectories from the deployed agent provide the evidence for the next diagnosis.

\begin{algorithm}[!t]
\caption{The evolution pipeline of MUSE}
\label{alg:muse_evolution}
\small
\begin{algorithmic}[1]
\Require Deployed agent $A_0$, environment $E$, available oracle tasks, and a round budget
\Ensure Agent selected for subsequent deployment
\State Initialize the diagnostic window from evaluation trajectories of $A_0$
\For{each evolution round $t$ within the budget}
    \State Compare sampled successes and failures alongside the symbolic support presented during execution
    \State $(\mathbf d_t^{\mathrm{diag}},\mathbf v_t)\gets A_{\mathrm{diag}}(\tilde{\mathcal W}_t)$
        \Comment{Capability gaps and consolidation opportunities, Eq.~\eqref{equ:metacognitive_diagnostic_routing}}
    \State $C_t^{\mathrm{sel}}\gets\operatorname{Route}(\mathbf v_t,A_t)$
        \Comment{Targeted component selection, Eq.~\eqref{equ:evolutionary_gating_output}}
    \Statex \hspace{\algorithmicindent}Route situational gaps to experiences, recurring strategies in verified experiences to principles,
    \Statex \hspace{\algorithmicindent}recurring procedures to tools, and persistent reasoning or planning difficulties to model weights.
    \State \textbf{if} $C_t^{\mathrm{sel}}=\varnothing$ \textbf{then break}
    \If{the active synthetic task pool requires a refresh} \Comment{$\Delta$, Section~\ref{subsec:supervision_generator}}
        \State Construct tasks from environmental interactions and validate their reference solutions
        \State Score guided and free executions against the task objective and validated reference evidence
        \State Admit tasks with reliable guided execution, partial independent progress, and a sufficient guidance gap
    \EndIf
    \State Supplement oracle tasks with eligible tasks from the active pool
    \For{each $c\in C_t^{\mathrm{sel}}$ in the order $(K_{\mathrm{exp}},P_{\mathrm{sys}},T_{\mathrm{tool}},\theta_{\mathrm{model}})$}
        \If{$c\in\Psi$}
            \State Collect fresh rollouts for $c$ with the current agent configuration
            \State Apply the refinement corresponding to $c$
                \Comment{Section~\ref{subsec:optimization_scope_and_evolutionary_optimizer}}
            \Statex \hspace{\dimexpr\algorithmicindent*4\relax}%
                $\begin{aligned}
                    K_{\mathrm{exp}} &:~\text{Extract and verify experiences, then reassess their utility}\\
                    P_{\mathrm{sys}} &:~\text{Distill principles from experiences, verify them across tasks, and rank them by utility}\\
                    T_{\mathrm{tool}} &:~\text{Construct and verify tools, then review existing tools for retirement from retrieval}
                \end{aligned}$
        \Else \Comment{Selected model weights $\theta_{\mathrm{model}}$}
            \For{each parametric refinement epoch}
                \State Collect fresh rollouts $Q_{\mathrm{new}}$ with the current symbolic assets
                \State $Q_{\mathrm{mix}}\gets Q_{\mathrm{new}}\cup Q_{\mathrm{replay}}$
                    \Comment{Bounded replay, Eq.~\eqref{equ:replay_mixture_and_bound}}
                \State Derive progress signals from differences between successive prefix potentials
                    \Comment{Section~\ref{subsec:supervision_generator}}
                \State Form token advantages from trajectory and cumulative step rewards for the GRPO update
                    \Comment{Section~\ref{subsec:optimization_scope_and_evolutionary_optimizer}}
            \EndFor
            \State Test task mastery without stored symbolic support
                \Comment{Section~\ref{subsec:optimization_scope_and_evolutionary_optimizer}}
            \State For each mastered task, lower the utility of its associated entries:
            \Statex \hspace{\dimexpr\algorithmicindent*4\relax}%
                $\begin{aligned}
                    K_{\mathrm{exp}} &:~\text{Revised utility guides subsequent probation or retirement}\\
                    P_{\mathrm{sys}} &:~\text{Revised utility determines retention rank within the bounded principle pool}
                \end{aligned}$
        \EndIf
    \EndFor
    \State Evaluate the joint candidate under $\mathcal O$
    \Statex \hspace{\algorithmicindent}For invoked tools, record execution reliability and adjust utility from task-relative performance when available.
    \State Select $A_{t+1}$ for deployment
    \State Check eligible synthetic tasks for mastery and refresh the diagnostic window from $A_{t+1}$
\EndFor
\end{algorithmic}
\end{algorithm}

\section{Experiments} \label{sec:experiments}
In this section, we evaluate MUSE to examine the practical feasibility of the proposed theoretical framework and the mechanisms supporting continuous agent adaptation. Following the experimental setup in Section~\ref{subsec:experimental_setup}, we investigate five questions. \textbf{RQ1}: How effectively does MUSE improve task performance through continuous evolution (Section~\ref{subsec:overall_adaptivity})? \textbf{RQ2}: How do individual and joint component updates contribute to adaptation (Section~\ref{subsec:component_contributions})? \textbf{RQ3}: How effective and reliable is the constructed supplementary supervision (Section~\ref{subsec:supervision_effectiveness})? \textbf{RQ4}: How does metacognitive gating affect task performance and exploration rollout cost (Section~\ref{subsec:metacognitive_gating_evaluation})? \textbf{RQ5}: How do performance and stability change across successive evolution rounds (Section~\ref{subsec:continuous_evolution})? Section~\ref{subsec:experimental_limitations} discusses the scope and limitations of the empirical evidence obtained from these evaluations.

\subsection{Experimental Setups} \label{subsec:experimental_setup}

\noindent \textbf{Foundation Models.} We instantiate the evolving agent with Qwen3.5-9B~\citep{qwen2026qwen35} as its foundation model for the main experiments. We additionally examine symbolic evolution of agents initialized with Qwen3.5-27B~\citep{qwen2026qwen35} to assess component contributions at a larger model scale.

\noindent \textbf{Benchmark.} We conduct agent evolution in APPWorld~\citep{trivedi2024appworld}, where agents complete multi-step tasks through programmatic interactions with simulated applications. Task executions provide trajectories and environmental feedback for evolution, while state-based evaluation measures task performance. The \texttt{test\_normal} split serves as a proxy for the operating environment, with execution trajectories supporting metacognitive diagnosis and evaluation scores guiding configuration selection across rounds. Benchmark details and the evaluation protocol are provided in Appendix~\ref{subapp:benchmark_details} and Appendix~\ref{subapp:evolution-evaluation}, respectively.

\noindent \textbf{Baselines and MUSE Variants.} We compare evolutionary strategies that adapt different components of the agent system. Baselines with fixed executor weights include Reflexion~\citep{shinn2023reflexion}, MemP~\citep{fang2026memp}, ReasoningBank~\citep{ouyang2026reasoningbank}, and SkillMentor~\citep{bao2026skillmentor}. Vanilla GRPO~\citep{shao2024deepseekmath} provides a reference strategy  centered on model weight adaptation. Within MUSE, we distinguish three component variants. MUSE (Symbolic) enables refinement of experiences, principles, and tools. MUSE (Parametric) enables model weight refinement with prefix-potential supervision. MUSE (Joint) combines symbolic and parametric refinement following Section~\ref{subsec:optimization_scope_and_evolutionary_optimizer}. These component variants operate without metacognitive gating, with task sources specified in the corresponding comparisons. MUSE denotes the full configuration with metacognitive gating, supplementary synthetic tasks, and joint symbolic and parametric adaptation. In the gating comparisons, the Always-on setting and Random setting replace metacognitive gating while retaining the other mechanisms of MUSE. Baseline and variant details are provided in Appendix~\ref{subapp:baselines}.

\noindent \textbf{Metrics.} Following the evaluation criteria in Section~\ref{subsec:evaluation_metrics}, we report task success rate (SR $\uparrow$) as the primary performance metric. Best SR ($\uparrow$) is the highest SR within the observation horizon, including the shared Base Model reference at round~0. Final SR ($\uparrow$) reports task performance at the last reported evolution round. Distinct from the single-round $\mathrm{EvoGain}_t$ in Section~\ref{subsec:evaluation_metrics}, EvoGain ($\uparrow$) in the experimental tables denotes the peak improvement over the shared reference, and Evo-AUC ($\uparrow$) summarizes improvement relative to that reference across rounds. Avg. Steps ($\downarrow$) denotes the mean interaction steps across all evaluated tasks for the agent instance attaining Best SR. Additional measures of evolution dynamics, supervision validity, and resource consumption are introduced in the corresponding analyses. The details of these evaluation metrics are provided in Appendix~\ref{subapp:metrics}.

\noindent \textbf{Implementation Details.} We analyze agent evolution over a maximum observation horizon of 8 rounds, tracking task performance as the persistent agent configuration changes. Each experimental setting specifies the components within the agent system eligible for adaptation, the available supervision, and the gating mechanism. Retained updates persist into subsequent rounds, allowing later interactions to reflect the capabilities acquired through earlier evolution. Component refinement and synchronization follow Section~\ref{subsec:optimization_scope_and_evolutionary_optimizer}. GRPO implements the parametric update channel, with VERL supporting optimization and vLLM supporting agent inference. Hyperparameters, stopping conditions, and evolution and evaluation protocols are detailed in Appendix~\ref{subapp:evolution-evaluation}.

\subsection{Overall Adaptivity} \label{subsec:overall_adaptivity}

\begin{table*}[!t]
    \centering
    \small
    \renewcommand{\arraystretch}{1.05}
    \setlength{\tabcolsep}{2.4pt}
    \definecolor{overallSRHeat}{RGB}{155,184,212}
    \definecolor{overallStepHeat}{RGB}{225,192,157}
    \definecolor{overallGroup}{RGB}{244,246,248}

    \newcommand{\overallgroup}[1]{%
        \rowcolor{overallGroup}%
        \multicolumn{8}{c}{\footnotesize\bfseries\color{black!65}#1}\\
    }

    \caption{Overall adaptivity on APPWorld with Qwen3.5-9B within 8 evolution rounds. $\mathcal C$: component categories eligible for refinement. Exec. Upd. and Aux. Upd.: whether the parameters of the task executor or of a separate auxiliary model are updated. Blue and orange shading mark Best SR and Avg. Steps, respectively, with darker shading indicating more favorable values within each metric. Bold marks the best value.}

    \label{tab:overall_adaptivity}

    \begin{tabularx}{\textwidth}{
    >{\raggedright\arraybackslash}m{0.16\textwidth}
    >{\centering\arraybackslash}m{0.16\textwidth}
    >{\centering\arraybackslash}m{0.10\textwidth}
    >{\centering\arraybackslash}m{0.10\textwidth}
    >{\centering\arraybackslash}X
    >{\centering\arraybackslash}X
    >{\centering\arraybackslash}X
    >{\centering\arraybackslash}X
}
    \toprule
    \multirow{2}{*}{\textbf{Method}}
    & \multicolumn{3}{c}{\textbf{Evolution Configuration}}
    & \multicolumn{4}{c}{\textbf{APPWorld}} \\
    \cmidrule(lr){2-4}
    \cmidrule(lr){5-8}

    & $\boldsymbol{\mathcal{C}}$
    & \mbox{\textbf{Exec. Upd.}}
    & \mbox{\textbf{Aux. Upd.}}
    & \mbox{\textbf{Best SR}$\uparrow$}
    & \mbox{$\boldsymbol{\Delta}$\textbf{SR}$\uparrow$}
    & \mbox{\textbf{Avg. Steps}$\downarrow$}
    & \mbox{$\boldsymbol{\Delta}$\textbf{Steps}$\downarrow$} \\
    \midrule

    \overallgroup{Reference}
    Base Model
    & --
    & $\times$
    & $\times$
    & \cellcolor{overallSRHeat!8}30.35
    & --
    & \cellcolor{overallStepHeat!12}20.66
    & -- \\

    \addlinespace[2pt]
    \overallgroup{Baselines}

    Reflexion
    & $\Psi$
    & $\times$
    & $\times$
    & \cellcolor{overallSRHeat!12}31.55
    & \cellcolor{overallSRHeat!12}+1.20
    & \cellcolor{overallStepHeat!6}20.80
    & \cellcolor{overallStepHeat!6}+0.14 \\

    MemP
    & $\Psi$
    & $\times$
    & $\times$
    & \cellcolor{overallSRHeat!18}32.74
    & \cellcolor{overallSRHeat!18}+2.39
    & \cellcolor{overallStepHeat!20}20.60
    & \cellcolor{overallStepHeat!20}$-0.06$ \\

    ReasoningBank
    & $\Psi$
    & $\times$
    & $\times$
    & \cellcolor{overallSRHeat!24}33.33
    & \cellcolor{overallSRHeat!24}+2.98
    & \cellcolor{overallStepHeat!20}20.60
    & \cellcolor{overallStepHeat!20}$-0.06$ \\

    SkillMentor
    & $\Psi$
    & $\times$
    & $\checkmark$
    & \cellcolor{overallSRHeat!35}41.07
    & \cellcolor{overallSRHeat!35}+10.72
    & \cellcolor{overallStepHeat!45}18.50
    & \cellcolor{overallStepHeat!45}$-2.16$ \\

    Vanilla GRPO
    & $\Theta$
    & $\checkmark$
    & $\times$
    & \cellcolor{overallSRHeat!45}44.64
    & \cellcolor{overallSRHeat!45}+14.29
    & \cellcolor{overallStepHeat!35}18.83
    & \cellcolor{overallStepHeat!35}$-1.83$ \\

    \addlinespace[2pt]
    \overallgroup{Ours}

    \textbf{MUSE}
    & \mbox{$\Theta\cup\Psi$}
    & $\checkmark$
    & $\times$
    & \cellcolor{overallSRHeat!60}\textbf{54.76}
    & \cellcolor{overallSRHeat!60}\textbf{+24.41}
    & \cellcolor{overallStepHeat!60}\textbf{16.91}
    & \cellcolor{overallStepHeat!60}$\boldsymbol{-3.75}$ \\

    \bottomrule
    \end{tabularx}
\end{table*}

To investigate \textbf{RQ1}, we examine how continuous evolution with MUSE changes agent task capability and execution behavior. Table~\ref{tab:overall_adaptivity} compares the evolved agent instances produced by MUSE and representative evolution baselines outlined in Section~\ref{subsec:experimental_setup} and summarizes the component categories eligible for adaptation. For MUSE and Vanilla GRPO, Best SR denotes the highest recorded SR within 8 evolution rounds, and Avg. Steps denotes the mean interaction steps across all evaluated tasks for the corresponding evolved agent instance.

From the experimental results, we observe the following findings:

\noindent \textbf{(i) The evolved agent configuration supports higher task success.} Starting from the same 9B foundation model, the pre-evolution agent achieves 30.35\% SR, while the agent evolved with MUSE reaches 54.76\% within 8 rounds. SkillMentor and Vanilla GRPO also improve on the initial agent, reaching 41.07\% and 44.64\%, respectively. In MUSE, the agent carries its refined symbolic assets and model weights into subsequent task execution, surpassing the task-solving capability of the pre-evolution agent and baselines. The observed improvement in task success provides evidence that MUSE can produce an evolved agent with higher task capability than the initial configuration.

\noindent \textbf{(ii) Adaptation with MUSE also reduces the mean number of interaction steps.} At the round attaining Best SR, the mean number of interaction steps across all evaluated tasks is 16.91 for MUSE, compared with 20.66 before evolution. The corresponding averages are 18.83 for Vanilla GRPO and 18.50 for SkillMentor. For MUSE, the change from the initial to the evolved configuration is reflected in both a larger proportion of tasks completed and 3.75 fewer interactions per evaluated task on average.

\subsection{Component Contributions and Joint Adaptation} \label{subsec:component_contributions}

To investigate \textbf{RQ2}, we examine how component configurations and refinement procedures affect continuous agent evolution. Table~\ref{tab:symbolic_contribution} compares symbolic refinement channels with Qwen3.5-9B and Qwen3.5-27B, while Table~\ref{tab:unison_contribution} compares symbolic, parametric, and joint adaptation with Qwen3.5-9B. All evolution runs in this subsection operate on oracle tasks without supplementary synthetic tasks or metacognitive gating. The tables identify the refinement channels enabled in each experimental configuration. We assess the resulting agent instances through task performance, performance across evolution rounds, and average interaction steps.

\begin{table*}[!t]
    \centering
    \small
    \renewcommand{\arraystretch}{1.10}
    \setlength{\tabcolsep}{2.2pt}

    \definecolor{componentSRHeat}{RGB}{155,184,212}
    \definecolor{componentStepHeat}{RGB}{225,192,157}
    \definecolor{componentGroup}{RGB}{244,246,248}

    \newcommand{\symbolicgroup}[1]{%
        \rowcolor{componentGroup}%
        \multicolumn{9}{c}{\footnotesize\bfseries\color{black!65}#1}\\
    }

    \caption{Symbolic component contributions on APPWorld with Qwen3.5-9B and Qwen3.5-27B. Model weights are fixed, and all variants evolve on oracle tasks across 8 evolution rounds without metacognitive gating. Bold marks the best value within each backbone, with shading as in Table~\ref{tab:overall_adaptivity}.}

    \label{tab:symbolic_contribution}

    \begin{tabularx}{\textwidth}{
        >{\centering\arraybackslash}m{0.225\textwidth}
        *{8}{>{\centering\arraybackslash}X}
    }
        \toprule

        \multirow{2}{*}{
            \makecell[c]{
                \textbf{Component}\\
                \textbf{Configuration}
            }
        }
        & \multicolumn{4}{c}{\textbf{Qwen3.5-9B}}
        & \multicolumn{4}{c}{\textbf{Qwen3.5-27B}} \\

        \cmidrule(lr){2-5}
        \cmidrule(lr){6-9}

        & {\footnotesize\mbox{\textbf{Best SR}$\uparrow$}}
        & {\footnotesize\mbox{\textbf{EvoGain}$\uparrow$}}
        & {\footnotesize\mbox{\textbf{Evo-AUC}$\uparrow$}}
        & {\footnotesize\mbox{\textbf{Avg. Steps}}$\downarrow$}
        & {\footnotesize\mbox{\textbf{Best SR}$\uparrow$}}
        & {\footnotesize\mbox{\textbf{EvoGain}$\uparrow$}}
        & {\footnotesize\mbox{\textbf{Evo-AUC}$\uparrow$}}
        & {\footnotesize\mbox{\textbf{Avg. Steps}}$\downarrow$} \\

        \midrule
        \symbolicgroup{Pre-Evolution Reference}

        Base Model
        & \cellcolor{componentSRHeat!8}30.35
        & --
        & \cellcolor{componentSRHeat!37}0.00
        & \cellcolor{componentStepHeat!28}20.66
        & \cellcolor{componentSRHeat!8}61.91
        & --
        & \cellcolor{componentSRHeat!8}0.00
        & \cellcolor{componentStepHeat!8}17.97 \\

        \addlinespace[2pt]
        \symbolicgroup{Inner-Component Symbolic Evolution}

        $K_{\mathrm{exp}}$
        & \cellcolor{componentSRHeat!60}\textbf{36.90}
        & \cellcolor{componentSRHeat!60}\textbf{+6.55}
        & \cellcolor{componentSRHeat!60}\textbf{2.16}
        & \cellcolor{componentStepHeat!29}20.63
        & \cellcolor{componentSRHeat!51}70.83
        & \cellcolor{componentSRHeat!51}+8.92
        & \cellcolor{componentSRHeat!36}4.57
        & \cellcolor{componentStepHeat!18}17.69 \\

        $T_{\mathrm{tool}}$
        & \cellcolor{componentSRHeat!46}35.11
        & \cellcolor{componentSRHeat!46}+4.76
        & \cellcolor{componentSRHeat!8}$-2.64$
        & \cellcolor{componentStepHeat!60}\textbf{19.61}
        & \cellcolor{componentSRHeat!25}65.48
        & \cellcolor{componentSRHeat!25}+3.57
        & \cellcolor{componentSRHeat!11}0.55
        & \cellcolor{componentStepHeat!58}16.59 \\

        \addlinespace[2pt]
        \symbolicgroup{Combined Symbolic Adaptation}

                $K_{\mathrm{exp}}\cup P_{\mathrm{sys}}$
        & \cellcolor{componentSRHeat!55}36.30
        & \cellcolor{componentSRHeat!55}+5.95
        & \cellcolor{componentSRHeat!43}0.60
        & \cellcolor{componentStepHeat!8}21.32
        & \cellcolor{componentSRHeat!60}\textbf{72.62}
        & \cellcolor{componentSRHeat!60}\textbf{+10.71}
        & \cellcolor{componentSRHeat!45}5.87
        & \cellcolor{componentStepHeat!40}17.08 \\

        $K_{\mathrm{exp}}\cup P_{\mathrm{sys}}\cup T_{\mathrm{tool}}$
        & \cellcolor{componentSRHeat!60}\textbf{36.90}
        & \cellcolor{componentSRHeat!60}\textbf{+6.55}
        & \cellcolor{componentSRHeat!58}1.97
        & \cellcolor{componentStepHeat!58}19.69
        & \cellcolor{componentSRHeat!60}\textbf{72.62}
        & \cellcolor{componentSRHeat!60}\textbf{+10.71}
        & \cellcolor{componentSRHeat!60}\textbf{8.36}
        & \cellcolor{componentStepHeat!60}\textbf{16.54} \\

        \bottomrule
    \end{tabularx}
\end{table*}

\begin{table*}[!t]
    \centering
    \small
    \renewcommand{\arraystretch}{1.10}
    \setlength{\tabcolsep}{2.2pt}

    \definecolor{componentSRHeat}{RGB}{155,184,212}
    \definecolor{componentStepHeat}{RGB}{225,192,157}
    \definecolor{componentGroup}{RGB}{244,246,248}

    \newcommand{\componentgroup}[1]{%
        \rowcolor{componentGroup}%
        \multicolumn{8}{c}{\footnotesize\bfseries\color{black!65}#1}\\
    }

    \caption{Separate and joint symbolic and parametric adaptation on APPWorld with Qwen3.5-9B, on oracle tasks without synthetic tasks or metacognitive gating. $\Psi$ and $\Theta$ indicate whether symbolic refinement and model weight refinement are enabled. Bold and shading as in Table~\ref{tab:overall_adaptivity}.}

    \label{tab:unison_contribution}

    \begin{tabularx}{\textwidth}{
        >{\raggedright\arraybackslash}m{0.25\textwidth}
        >{\centering\arraybackslash}m{0.10\textwidth}
        >{\centering\arraybackslash}m{0.10\textwidth}
        *{5}{>{\centering\arraybackslash}X}
    }
        \toprule

        \multirow{2}{*}{\textbf{Evolution Setting}}
        & \multicolumn{2}{c}{\textbf{Evolution Configuration}}
        & \multicolumn{5}{c}{\textbf{Evolutionary Performance}} \\

        \cmidrule(lr){2-3}
        \cmidrule(lr){4-8}

        & $\boldsymbol{\Psi}$
        & $\boldsymbol{\Theta}$
        & {\footnotesize\mbox{\textbf{Best SR}$\uparrow$}}
        & {\footnotesize\mbox{\textbf{EvoGain}$\uparrow$}}
        & {\footnotesize\mbox{\textbf{Final SR}$\uparrow$}}
        & {\footnotesize\mbox{\textbf{Evo-AUC}$\uparrow$}}
        & {\footnotesize\mbox{\textbf{Avg. Steps}}$\downarrow$} \\

        \midrule
        \componentgroup{Pre-Evolution Reference}

        Base Model
        & $\times$
        & $\times$
        & \cellcolor{componentSRHeat!8}30.35
        & --
        & \cellcolor{componentSRHeat!8}30.35
        & \cellcolor{componentSRHeat!8}0.00
        & \cellcolor{componentStepHeat!8}20.66 \\

        \addlinespace[2pt]
        \componentgroup{Separate Symbolic and Parametric Adaptation}

        MUSE (Symbolic)
        & $\checkmark$
        & $\times$
        & \cellcolor{componentSRHeat!22}36.90
        & \cellcolor{componentSRHeat!22}+6.55
        & \cellcolor{componentSRHeat!15}33.33
        & \cellcolor{componentSRHeat!13}1.97
        & \cellcolor{componentStepHeat!17}19.69 \\

        Vanilla GRPO
        & $\times$
        & $\checkmark$
        & \cellcolor{componentSRHeat!38}44.64
        & \cellcolor{componentSRHeat!38}+14.29
        & \cellcolor{componentSRHeat!42}44.04
        & \cellcolor{componentSRHeat!36}10.83
        & \cellcolor{componentStepHeat!25}18.83 \\

        MUSE (Parametric)
        & $\times$
        & $\checkmark$
        & \cellcolor{componentSRHeat!46}48.21
        & \cellcolor{componentSRHeat!46}+17.86
        & \cellcolor{componentSRHeat!51}47.61
        & \cellcolor{componentSRHeat!41}12.91
        & \cellcolor{componentStepHeat!38}17.48 \\

        \addlinespace[2pt]
        \componentgroup{Joint Symbolic and Parametric Adaptation}

        MUSE (Joint)
        & $\checkmark$
        & $\checkmark$
        & \cellcolor{componentSRHeat!60}\textbf{54.76}
        & \cellcolor{componentSRHeat!60}\textbf{+24.41}
        & \cellcolor{componentSRHeat!60}\textbf{51.19}
        & \cellcolor{componentSRHeat!60}\textbf{20.28}
        & \cellcolor{componentStepHeat!60}\textbf{15.20} \\

        \bottomrule
    \end{tabularx}
\end{table*}

From the evaluation results, we draw the following findings:

\noindent \textbf{(i) At the 9B scale, the peak improvement from toolset refinement is not sustained across rounds.} Experience refinement and MUSE (Symbolic) both reach a Best SR of 36.90\% from a pre-evolution reference of 30.35\%. Toolset refinement reaches 35.11\% Best SR and records an Evo-AUC of $-2.64$, indicating lower performance relative to the reference across the 8 rounds in aggregate. This result highlights a distinction between tool verification and deployment utility. Paired verification assesses a candidate on tasks for which it is considered applicable, whereas subsequent execution also requires the agent to identify when the tool is appropriate. The negative Evo-AUC therefore motivates assessing toolset utility beyond the tasks examined at admission.

\noindent \textbf{(ii) MUSE (Symbolic) achieves a larger improvement at the 27B scale.} With experience, principle, and toolset refinement jointly enabled, Best SR rises notably from 61.91\% to 72.62\% for Qwen3.5-27B, while the exact same configuration improves Best SR from 30.35\% to 36.90\% for Qwen3.5-9B. At 27B, all four symbolic configurations also record positive Evo-AUC, ranging from 0.55 for toolset refinement to 8.36 for MUSE (Symbolic). Thus, the larger peak improvement under combined symbolic refinement is accompanied by a positive aggregate contribution across each of the symbolic configurations examined at this scale.

\noindent \textbf{(iii) Parametric refinement is more effective than symbolic refinement in the evaluated 9B setting.} With the same oracle task pool, both Vanilla GRPO and MUSE (Parametric) exceed MUSE (Symbolic) in Best SR, Final SR, and Evo-AUC. The difference persists in the final agent instances, where MUSE (Parametric) reaches 47.61\% SR compared with 33.33\% for MUSE (Symbolic). Even when a relevant symbolic asset is available, its benefit depends on whether the executor can apply it correctly at the appropriate step. At the 9B scale, this dependence may limit how much additional symbolic guidance improves task execution. Parametric refinement, by contrast, directly shapes the action-selection behavior of the executor through feedback from prior executions.

\noindent \textbf{(iv) Enabling symbolic refinement alongside model weight refinement improves both peak and final performance at 9B.} With oracle tasks, extending MUSE (Parametric) to MUSE (Joint) raises Best SR from 48.21\% to 54.76\% and Final SR from 47.61\% to 51.19\%. Although symbolic refinement yields modest improvements with fixed executor weights at 9B, it provides an additional benefit when the weights also undergo refinement. The increase in Final SR shows that this benefit remains evident in the final agent instance.

In MUSE (Joint), model weight refinement proceeds on rollouts collected with the updated symbolic assets. The improvements in peak and final task performance under this arrangement provide empirical support for the feasibility of Stage~4 coordinated component evolution in MUSE.

\subsection{Supervision Effectiveness and Validity} \label{subsec:supervision_effectiveness}

To investigate \textbf{RQ3}, we examine the effectiveness and validity of supervision constructed through the supervision generator $\Delta$. Table~\ref{tab:supplementary_supervision_effectiveness} compares outcome-supervised Vanilla GRPO with MUSE (Parametric), which incorporates prefix-potential supervision, on oracle tasks, followed by the addition of supplementary synthetic tasks to MUSE (Parametric) and MUSE (Joint). These comparisons operate without metacognitive gating. We additionally compare both component configurations with and without supplementary synthetic tasks when a fixed subset of 22 oracle tasks (nominally 25\%) is available. The same oracle subset is retained across these four settings. Supplementary task construction can explore the environments without receiving the excluded oracle task queries. All configurations follow a maximum budget of 8 evolution rounds, with subset selection and early stopping detailed in Appendix~\ref{subapp:evolution-evaluation}.

Table~\ref{tab:synthetic_task_quality} and Table~\ref{tab:synthetic_outcome_reward_reliability} complement the performance comparisons with human assessments of task quality and outcome rewards for synthetic tasks. For these audits, we sample task and execution records collected throughout a complete run of the full MUSE configuration (last row in Table~\ref{tab:overall_adaptivity}), with metacognitive gating, supplementary supervision, and symbolic and parametric adaptation enabled.

For task quality, we assess generated candidates from interaction-grounded task synthesis and synthetic tasks admitted to the active pool $\mathcal{T}_{\text{syn}}^{(t)}$. We randomly sample five candidates from initial task generation and each of 7 subsequent task-pool updates within 8 evolution rounds, yielding 40 candidates. We additionally audit all 36 distinct synthetic tasks admitted during the evaluated evolution process, including 26 tasks remaining in the active pool in the final evolution round and 10 subsequently retired through continual task-pool adaptation. Goal Clarity assesses whether the query specifies a clear objective and necessary constraints. Environmental Feasibility assesses whether environmental evidence supports the operations required for task completion. Reference Alignment assesses whether the reference solution and reference trajectory correspond to the task objective and constraints. Overall Validity requires all three criteria to pass. All rates are calculated over tasks with determinate judgments for every criterion, comprising 39 generated candidates and all 36 admitted synthetic tasks.

For outcome reward assessment, we focus on the 31 admitted synthetic tasks covered by the recorded frontier assessments, with guided and free executions available for auditing during the same evolution run. This audit examines the outcome rewards associated with tasks selected for further adaptation. Each of these tasks has 2 guided and 6 free executions recorded during task assessment, as detailed in Appendix~\ref{subapp:supervision_generator}. We randomly sample 1 guided and 1 free execution per task, yielding 62 executions for human assessment. A human reviewer examines each task query together with the recorded actions and environmental observations. An execution is labeled complete when the evidence supports fulfillment of all required operations and constraints. Partial or incomplete indicate unmet task requirements, while uncertain indicates insufficient evidence for a determinate judgment. Recorded outcome rewards are hidden during review. Detailed annotation criteria and review procedures are provided in Appendix~\ref{subapp:audit_protocols}.

Human judgments serve as reference labels for evaluating outcome rewards assigned by the evolving agent through $\mathcal{R}_{\text{syn}}$. For binary assessment, human annotations of complete form the positive class, partial and incomplete form the negative class, and uncertain annotations are excluded. An outcome reward of 1 denotes predicted full completion. This audit evaluates full-reward assignment of $1.0$. Table~\ref{tab:synthetic_outcome_reward_reliability} reports four metrics. Completion Agreement measures the proportion of reward-based full-completion decisions matching human judgments. Full-Completion Precision measures the proportion of executions receiving full reward that are judged complete by the reviewer. Full-Completion Recall measures the proportion of executions judged complete that receive full reward. False-Positive Rate measures the proportion of executions judged not fully complete that still receive full reward.

\begin{table*}[!t]
    \centering
    \small
    \renewcommand{\arraystretch}{1.10}
    \setlength{\tabcolsep}{2.2pt}

    \definecolor{supervisionSRHeat}{RGB}{155,184,212}
    \definecolor{supervisionStepHeat}{RGB}{225,192,157}
    \definecolor{supervisionGroup}{RGB}{244,246,248}

    \newcommand{\supervisiongroup}[1]{%
        \rowcolor{supervisionGroup}%
        \multicolumn{7}{c}{\footnotesize\bfseries\color{black!65}#1}\\
    }

    \caption{Task sources and prefix-potential supervision for parametric and joint adaptation on APPWorld with Qwen3.5-9B, without metacognitive gating. Oracle percentages indicate the available fraction of oracle tasks. In the group of parametric evolution with 100\% oracle tasks, Vanilla GRPO provides the outcome-supervised reference, while MUSE (Parametric) incorporates prefix-potential supervision into parametric refinement. Bold marks the best value within each group, and shading follows the conventions described in Table~\ref{tab:overall_adaptivity}.}

    \label{tab:supplementary_supervision_effectiveness}

    \begin{tabularx}{\textwidth}{
        >{\centering\arraybackslash}m{0.23\textwidth}
        >{\centering\arraybackslash}m{0.14\textwidth}
        *{5}{>{\centering\arraybackslash}X}
    }
        \toprule

        \multirow{2}{*}{\textbf{Task Source}}
        & \multirow{2}{*}{
            \makecell[c]{
                \textbf{Prefix}\\
                \textbf{Potential}
            }
        }
        & \multicolumn{5}{c}{\textbf{Evolutionary Performance}} \\

        \cmidrule(lr){3-7}

        &
        & {\footnotesize\mbox{\textbf{Best SR}$\uparrow$}}
        & {\footnotesize\mbox{\textbf{EvoGain}$\uparrow$}}
        & {\footnotesize\mbox{\textbf{Final SR}$\uparrow$}}
        & {\footnotesize\mbox{\textbf{Evo-AUC}$\uparrow$}}
        & {\footnotesize\mbox{\textbf{Avg. Steps}}$\downarrow$} \\

        \midrule
        \supervisiongroup{Parametric Evolution with 100\% Oracle Tasks}

        Oracle
        & $\times$
        & \cellcolor{supervisionSRHeat!8} 44.64
        & \cellcolor{supervisionSRHeat!8} +14.29
        & \cellcolor{supervisionSRHeat!18} 44.04
        & \cellcolor{supervisionSRHeat!8} 10.83
        & \cellcolor{supervisionStepHeat!8}18.83 \\

        Oracle
        & $\checkmark$
        & \cellcolor{supervisionSRHeat!26} 48.21
        & \cellcolor{supervisionSRHeat!26} +17.86
        & \cellcolor{supervisionSRHeat!37} \textbf{47.61}
        & \cellcolor{supervisionSRHeat!19} 12.91
        & \cellcolor{supervisionStepHeat!27} 17.48 \\

        Oracle + Synthetic
        & $\checkmark$
        & \cellcolor{supervisionSRHeat!60} \textbf{54.76}
        & \cellcolor{supervisionSRHeat!60} \textbf{+24.41}
        & \cellcolor{supervisionSRHeat!8} 42.26
        & \cellcolor{supervisionSRHeat!27} \textbf{14.21}
        & \cellcolor{supervisionStepHeat!39} \textbf{16.70} \\

        \addlinespace[2pt]
        \supervisiongroup{Parametric Evolution with 25\% Oracle Tasks}

        Oracle
        & $\checkmark$
        & \cellcolor{supervisionSRHeat!14} 45.83
        & \cellcolor{supervisionSRHeat!14} +15.48
        & \cellcolor{supervisionSRHeat!8} 39.88
        & \cellcolor{supervisionSRHeat!9} 10.94
        & \cellcolor{supervisionStepHeat!8} 19.34 \\

        Oracle + Synthetic
        & $\checkmark$
        & \cellcolor{supervisionSRHeat!23} \textbf{47.61}
        & \cellcolor{supervisionSRHeat!23} \textbf{+17.26}
        & \cellcolor{supervisionSRHeat!37} \textbf{47.61}
        & \cellcolor{supervisionSRHeat!15} \textbf{12.17}
        & \cellcolor{supervisionStepHeat!28} \textbf{17.39} \\

        \addlinespace[2pt]
        \supervisiongroup{Joint Symbolic and Parametric Evolution with 100\% Oracle Tasks}

        Oracle
        & $\checkmark$
        & \cellcolor{supervisionSRHeat!60} \textbf{54.76}
        & \cellcolor{supervisionSRHeat!60} \textbf{+24.41}
        & \cellcolor{supervisionSRHeat!57} 51.19
        & \cellcolor{supervisionSRHeat!60} \textbf{20.28}
        & \cellcolor{supervisionStepHeat!60} \textbf{15.20} \\

        Oracle + Synthetic
        & $\checkmark$
        & \cellcolor{supervisionSRHeat!48} 52.38
        & \cellcolor{supervisionSRHeat!48} +22.03
        & \cellcolor{supervisionSRHeat!60}\textbf{51.78}
        & \cellcolor{supervisionSRHeat!38} 16.30
        & \cellcolor{supervisionStepHeat!27} 17.52 \\

        \addlinespace[2pt]
        \supervisiongroup{Joint Symbolic and Parametric Evolution with 25\% Oracle Tasks}

        Oracle
        & $\checkmark$
        & \cellcolor{supervisionSRHeat!14} 45.83
        & \cellcolor{supervisionSRHeat!14} +15.48
        & \cellcolor{supervisionSRHeat!24} \textbf{45.23}
        & \cellcolor{supervisionSRHeat!13} \textbf{11.72}
        & \cellcolor{supervisionStepHeat!8} 18.93 \\

        Oracle + Synthetic
        & $\checkmark$
        & \cellcolor{supervisionSRHeat!32} \textbf{49.40}
        & \cellcolor{supervisionSRHeat!32} \textbf{+19.05}
        & \cellcolor{supervisionSRHeat!24} \textbf{45.23}
        & \cellcolor{supervisionSRHeat!12} 11.57
        & \cellcolor{supervisionStepHeat!15} \textbf{18.34} \\

        \bottomrule
    \end{tabularx}
\end{table*}

\begin{table*}[!t]
    \centering
    \small
    \renewcommand{\arraystretch}{1.10}
    \setlength{\tabcolsep}{2.2pt}

    \definecolor{taskAuditBlue}{RGB}{155,184,212}
    \definecolor{taskAuditGroup}{RGB}{244,246,248}

    \caption{Human audit of synthetic task quality in one complete MUSE run on APPWorld, covering 40 sampled candidates and all 36 admitted tasks. Assessable / Total: tasks with determinate judgments on all three criteria. Rates are percentages over this subset. Overall Validity requires all three criteria to pass.}

    \label{tab:synthetic_task_quality}

    \begin{tabularx}{\textwidth}{
        >{\centering\arraybackslash}m{0.25\textwidth}
        >{\centering\arraybackslash}m{0.14\textwidth}
        *{4}{>{\centering\arraybackslash}X}
    }
        \toprule

        \multirow{2}{*}{\textbf{Task Set}}
        & \multirow{2}{*}{
            \makecell[c]{\textbf{Assessable}\\\textbf{/ Total}}
        }
        & \multicolumn{4}{c}{\textbf{Task Quality}} \\

        \cmidrule(lr){3-6}

        &
        & \makecell[c]{
            \footnotesize\textbf{Goal}\\
            \footnotesize\textbf{Clarity}$\uparrow$
        }
        & \makecell[c]{
            \footnotesize\textbf{Environmental}\\
            \footnotesize\textbf{Feasibility}$\uparrow$
        }
        & \makecell[c]{
            \footnotesize\textbf{Reference}\\
            \footnotesize\textbf{Alignment}$\uparrow$
        }
        & \makecell[c]{
            \footnotesize\textbf{Overall}\\
            \footnotesize\textbf{Validity}$\uparrow$
        } \\

        \midrule

        Generated Candidates
        & 39 / 40
        & \cellcolor{taskAuditBlue!36} 97.44
        & \cellcolor{taskAuditBlue!40} 100.00
        & \cellcolor{taskAuditBlue!15} 53.85
        & \cellcolor{taskAuditBlue!15} 53.85 \\

        Admitted Synthetic Tasks
        &  36 / 36
        & \cellcolor{taskAuditBlue!40} 100.00
        & \cellcolor{taskAuditBlue!40} 100.00
        & \cellcolor{taskAuditBlue!40} 100.00
        & \cellcolor{taskAuditBlue!40} \textbf{100.00} \\

        \bottomrule
    \end{tabularx}
\end{table*}

\begin{table*}[!t]
    \centering
    \small
    \renewcommand{\arraystretch}{1.10}
    \setlength{\tabcolsep}{2.2pt}

    \definecolor{rewardAuditBlue}{RGB}{155,184,212}
    \definecolor{rewardAuditOrange}{RGB}{225,192,157}
    \definecolor{rewardAuditGroup}{RGB}{244,246,248}

    \caption{Human audit of outcome rewards for synthetic tasks: one guided execution (with the reference solution) and one free execution (query only) for each of the 31 frontier tasks from the run in Table~\ref{tab:synthetic_task_quality}. Metrics evaluate full-completion decisions, with an assigned reward of 1 denoting predicted full completion. Assessable / Total: executions with determinate human judgments. Rates are percentages. The guided False-Positive Rate is undefined because all audited guided executions were judged complete.}

    \label{tab:synthetic_outcome_reward_reliability}

    \begin{tabularx}{\textwidth}{
        >{\centering\arraybackslash}m{0.23\textwidth}
        >{\centering\arraybackslash}m{0.13\textwidth}
        *{4}{>{\centering\arraybackslash}X}
    }
        \toprule

        \multirow{2}{*}{\textbf{Execution Set}}
        & \multirow{2}{*}{
            \makecell[c]{
                \textbf{Assessable}\\
                \textbf{/ Total}
            }
        }
        & \multicolumn{4}{c}{\textbf{Reward Reliability}} \\

        \cmidrule(lr){3-6}

        &
        & \makecell[c]{
            \footnotesize\textbf{Completion}\\
            \footnotesize\textbf{Agreement}$\uparrow$
        }
        & \makecell[c]{
            \footnotesize\textbf{Full-Completion}\\
            \footnotesize\textbf{Precision}$\uparrow$
        }
        & \makecell[c]{
            \footnotesize\textbf{Full-Completion}\\
            \footnotesize\textbf{Recall}$\uparrow$
        }
        & \makecell[c]{
            \footnotesize\textbf{False-Positive}\\
            \footnotesize\textbf{Rate}$\downarrow$
        } \\

        \midrule

        \rowcolor{rewardAuditGroup}
        \multicolumn{6}{c}{
            \footnotesize\bfseries\color{black!65}
            Recorded Outcome Rewards
        } \\

        Guided Executions
        & 31 / 31
        & \cellcolor{rewardAuditBlue!36} 93.55
        & \cellcolor{rewardAuditBlue!40} 100.00
        & \cellcolor{rewardAuditBlue!36} 93.55
        & N/A \\

        Free Executions
        &  31 / 31
        & \cellcolor{rewardAuditBlue!28} 80.65
        & \cellcolor{rewardAuditBlue!20} 66.67
        & \cellcolor{rewardAuditBlue!10} 28.57
        & \cellcolor{rewardAuditOrange!30} 4.17 \\

        \bottomrule
    \end{tabularx}
\end{table*}

From the evaluation results in Table~\ref{tab:supplementary_supervision_effectiveness}, Table~\ref{tab:synthetic_task_quality}, and Table~\ref{tab:synthetic_outcome_reward_reliability}, we draw the following findings:

\noindent \textbf{(i) MUSE (Parametric) improves task performance on oracle tasks.} With 100\% oracle task availability, MUSE (Parametric) achieves higher Best SR, Final SR, and Evo-AUC compared to Vanilla GRPO. The improvement extends across evolution rounds without expanding task coverage, supporting the effectiveness of this parametric refinement configuration. Within this configuration, prefix-potential supervision complements trajectory-level outcome rewards with assessments of changes in task-completion prospects across every step, providing parametric refinement with fine-grained guidance for intermediate decisions.

\noindent \textbf{(ii) The benefit of supplementary synthetic tasks depends on the component configuration.} Under full oracle supervision, supplying synthetic tasks to MUSE (Parametric) raises Best SR from 48.21\% to 54.76\% and improves Evo-AUC, whereas the same supplement decreases Best SR from 54.76\% to 52.38\% and reduces Evo-AUC for MUSE (Joint). These differences indicate that additional task coverage does not consistently translate into improved task performance across component configurations. One possible explanation concerns the additional support provided by supplementary synthetic tasks relative to the available oracle coverage. When oracle tasks already support effective joint adaptation, supplementary tasks may change the task mixture without providing equivalent additional support for both symbolic refinement and parametric adaptation.

\noindent \textbf{(iii) Supplementary synthetic tasks support evolution when oracle supervision is limited.} With 25\% oracle task availability, supplementary synthetic tasks raise the Best SR of MUSE (Parametric) from 45.83\% to 47.61\%, with higher Final SR and Evo-AUC. Its peak performance approaches the 48.21\% attained with full oracle supervision without synthetic tasks, indicating that self-constructed supervision can partially compensate for reduced oracle coverage. For MUSE (Joint), supplementary tasks raise Best SR from 45.83\% to 49.40\%, while Final SR remains 45.23\% and Evo-AUC changes from 11.72 to 11.57. The peak-performance benefit contrasts with the full-oracle joint comparison. One possible explanation is that limited oracle coverage provides fewer opportunities to construct reusable symbolic assets, increasing the potential value of supplementary interactions for joint adaptation. The improvement supports supplementary supervision under limited oracle availability.

\noindent \textbf{(iv) Interaction-Grounded Task Synthesis supports the quality of admitted synthetic tasks.} Initially generated candidates commonly exhibit clear objectives and environmental feasibility, while Reference Alignment reaches only 53.85\%, as shown in Table~\ref{tab:synthetic_task_quality}. Following solution-guided execution, objective-alignment checks, and query-execution consistency checks illustrated in Section~\ref{subsec:supervision_generator}, all 36 audited admitted tasks satisfy the three quality criteria. The contrast supports the role of environmental validation in selecting supplementary synthetic tasks whose reference solutions correspond to the stated objectives and constraints. 

\noindent \textbf{(v) Interaction-grounded outcome rewards support supplementary supervision for synthetic tasks.} Completion Agreement reaches 93.55\% for guided executions and 80.65\% for free executions, as shown in Table~\ref{tab:synthetic_outcome_reward_reliability}. This agreement supports deriving outcome rewards through the evolving agent by assessing executions against task objectives and validated environmental evidence, as illustrated in Section~\ref{subsec:supervision_generator}. For free executions, the low False-Positive Rate of 4.17\% accompanies limited Full-Completion Recall, indicating that successful executions can receive insufficient reward. Human annotations additionally identify 31 complete guided executions and seven complete free executions. This difference is consistent with the guidance gap considered during task admission and indicates remaining scope for improving independent execution on the selected synthetic tasks.

These results provide empirical support for the feasibility of Stage~4 interaction-grounded self-supervision in MUSE, with validated supplementary tasks, self-derived outcome rewards for synthetic tasks, and self-evaluative prefix-potential supervision complementing available oracle supervision during agent evolution.

\subsection{Metacognitive Gating and Evolutionary Efficiency} \label{subsec:metacognitive_gating_evaluation}

To investigate \textbf{RQ4}, we compare task performance and exploration cost under three gating mechanisms within MUSE: Always-on, Random, and Metacognitive. All configurations retain supplementary supervision and the same eligible optimization scope $\mathcal{C}$ over 8 evolution rounds on APPWorld with Qwen3.5-9B. The Always-on setting enables all evolvable components within $\mathcal{C}$ in every round. The Metacognitive setting follows the metacognitive gating mechanism introduced in Section~\ref{subsec:evolutionary_gating}. The Random setting is constructed from the recorded Metacognitive selections to match the number of selected components in each round. Specifically, adaptation of model weights is enabled in all 8 rounds of the recorded Metacognitive run, so the Random setting also keeps this channel enabled throughout. This choice is consistent with the stronger improvements from parametric evolution at the 9B scale reported in Section~\ref{subsec:component_contributions}. For the 3 symbolic components, the Random setting randomly selects the same number of components as the Metacognitive setting in each corresponding round. The Random setting also preserves the total activation count of each symbolic component and excludes principle refinement in the first round when the experience bank is empty, as detailed in Appendix~\ref{subapp:evolution-evaluation}. This design compares metacognitive and random component selection under matched component counts and comparable rollout budgets. 

\begin{figure*}[!t]
    \centering

    \begin{subfigure}[t]{0.485\textwidth}
        \centering
        \includegraphics[width=\linewidth]{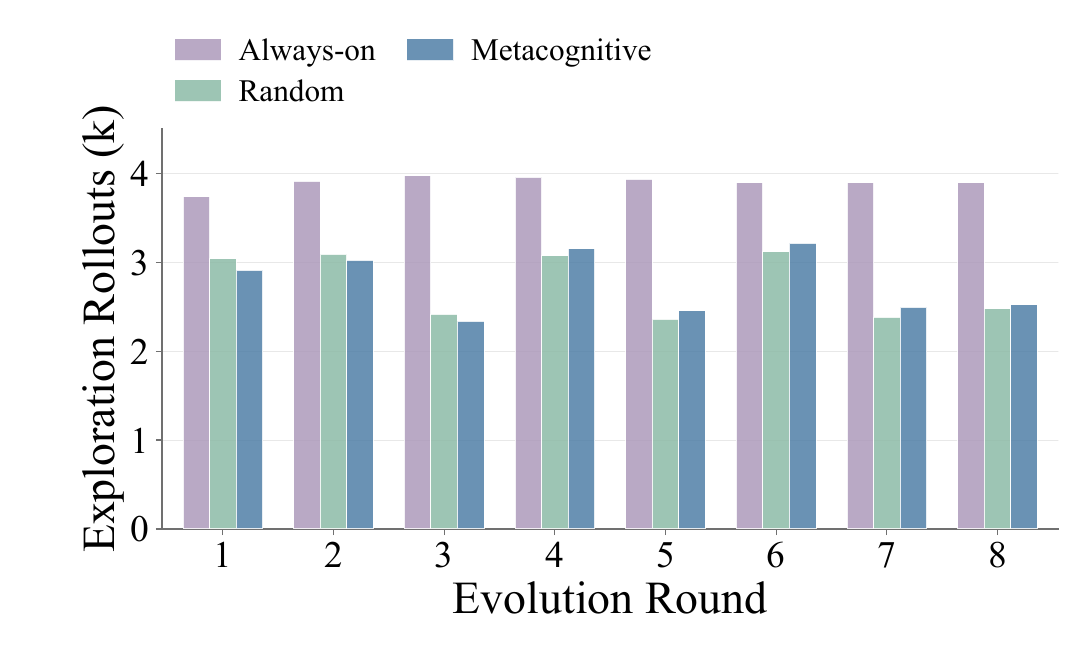}
        \caption{Exploration Rollouts per Round}
        \label{fig:gating_rollouts_per_round}
    \end{subfigure}
    \hfill
    \begin{subfigure}[t]{0.485\textwidth}
        \centering
        \includegraphics[width=\linewidth]{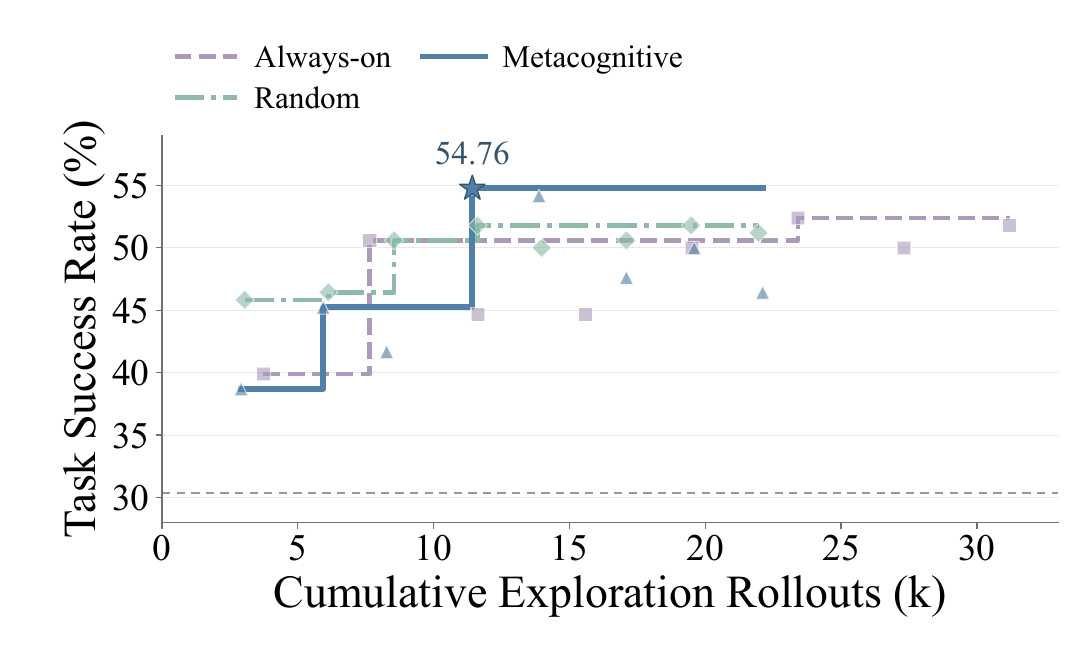}
        \caption{Observed Performance and Peak Attainment}
        \label{fig:gating_best_observed_sr}
    \end{subfigure}

    \caption{Exploration cost and task performance under three gating mechanisms within MUSE on APPWorld with Qwen3.5-9B. (a) Fresh exploration rollouts per evolution round. (b) SR of each evaluated agent instance against cumulative exploration rollouts. Step lines trace the best SR observed so far, the dashed line marks the pre-evolution SR, and the star marks the earliest instance attaining Best SR under Metacognitive gating. 
    \vspace{-2mm}}
    \label{fig:metacognitive_gating}
\end{figure*}

\begin{figure*}[!t]
    \centering
    \includegraphics[width=\textwidth]{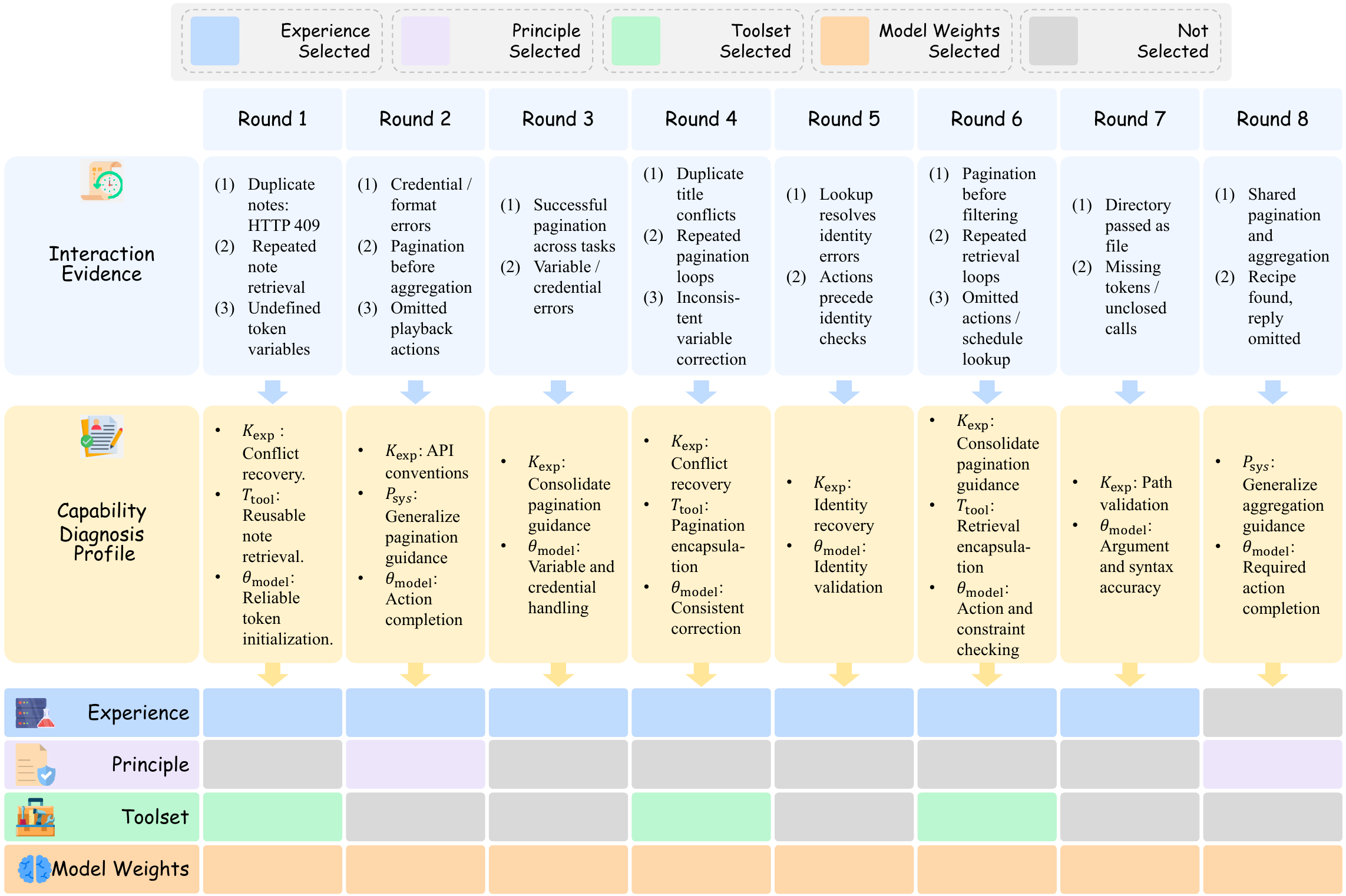}

    \caption{Interaction evidence, capability diagnosis profiles, and component selection over eight evolution rounds of a complete MUSE run on APPWorld with Qwen3.5-9B. Each column combines recorded diagnoses of the corresponding interaction traces. The lower matrix shows the component selection of experiences ($K_{\mathrm{exp}}$), principles ($P_{\mathrm{sys}}$), tools ($T_{\mathrm{tool}}$), and model weights ($\theta_{\mathrm{model}}$) in each round. \vspace{-2mm}}

    \label{fig:metacognitive_gating_case_study}
\end{figure*}

Table~\ref{tab:gating_effectiveness} reports task performance and exploration rollout cost. Exploration Rollouts counts fresh trajectories collected for symbolic refinement and adaptation of model weights across the 8 rounds, and Reduction measures the decrease relative to the Always-on setting. Figure~\ref{fig:metacognitive_gating} shows rollout counts per round and task performance against cumulative rollouts. Figure~\ref{fig:metacognitive_gating_case_study} presents interaction evidence, capability diagnosis profiles, and component selections across a complete Metacognitive run. These results examine the performance and resource aspects of gating, together with the evidence associated with different adaptation targets.

\begin{table*}[!t]
    \centering
    \small
    \renewcommand{\arraystretch}{1.16}
    \setlength{\tabcolsep}{4pt}

    \definecolor{gatingSRHeat}{RGB}{137,176,211}
    \definecolor{gatingCostHeat}{RGB}{225,181,137}

    \caption{Gating mechanisms within MUSE on APPWorld with Qwen3.5-9B over 8 evolution rounds, all with supplementary supervision and model weight adaptation in every round. Always-on selects all symbolic components. Random matches the number of symbolic components selected by Metacognitive in each round. Exploration Rollouts: fresh trajectories collected for symbolic and model weight refinement over the 8 rounds. Reduction: decrease relative to Always-on. Bold and shading as in Table~\ref{tab:overall_adaptivity}.}

    \label{tab:gating_effectiveness}

    \begin{tabularx}{\textwidth}{
        >{\centering\arraybackslash}m{0.09\textwidth}
        >{\centering\arraybackslash}m{0.17\textwidth}
        *{5}{>{\centering\arraybackslash}X}
    }
        \toprule
        \multirow{2}{*}{\textbf{Method}}
        & \multirow{2}{*}{\textbf{Gating}}
        & \multicolumn{3}{c}{\textbf{Evolutionary Performance}}
        & \multicolumn{2}{c}{\textbf{Exploration Cost}} \\
        \cmidrule(lr){3-5}
        \cmidrule(lr){6-7}
        & &
        \mbox{\textbf{Best SR}$\uparrow$}
        & \mbox{\textbf{Final SR}$\uparrow$}
        & \mbox{\textbf{Evo-AUC}$\uparrow$}
        & \shortstack{\textbf{Exploration}\\\textbf{Rollouts}$\downarrow$}
        & \mbox{\textbf{Reduction}$\uparrow$} \\
        \midrule

        \multirow{3}{*}{\textbf{MUSE}}
        & Always-on
        & \cellcolor{gatingSRHeat!21}52.38
        & \cellcolor{gatingSRHeat!55}\textbf{51.78}
        & \cellcolor{gatingSRHeat!19}16.30
        & \cellcolor{gatingCostHeat!12}31,208
        & \cellcolor{gatingCostHeat!12}-- \\

        & Random
        & \cellcolor{gatingSRHeat!12}51.78
        & \cellcolor{gatingSRHeat!50}51.19
        & \cellcolor{gatingSRHeat!55}\textbf{18.13}
        & \cellcolor{gatingCostHeat!55}\textbf{21,960}
        & \cellcolor{gatingCostHeat!55}\textbf{29.63\%} \\

        & \textbf{Metacognitive}
        & \cellcolor{gatingSRHeat!55}\textbf{54.76}
        & \cellcolor{gatingSRHeat!12}46.42
        & \cellcolor{gatingSRHeat!12}15.96
        & \cellcolor{gatingCostHeat!54}22,112
        & \cellcolor{gatingCostHeat!54}29.15\% \\

        \bottomrule
    \end{tabularx}
\end{table*}

From these results, we draw the following findings.

\noindent \textbf{(i) The benefit of selective evolution depends on the components selected for refinement.} Metacognitive gating achieves 54.76\% Best SR, compared with 52.38\% for the Always-on setting and 51.78\% for Random gating. Since the Random setting matches the number of selected components in each round, the higher peak under Metacognitive gating suggests a benefit from matching refinement targets to observed adaptation demands. Different components address different forms of capability limitation, so selecting an update channel also determines how the agent attempts to improve its behavior. These results support diagnosis as a basis for directing component refinement.

\noindent \textbf{(ii) Selective symbolic refinement reduces the interaction cost of continued evolution.} Metacognitive gating reduces exploration rollouts by 29.15\% relative to the Always-on setting while retaining model weight refinement throughout all 8 rounds. This allocation reduces the fresh trajectories collected for symbolic updates while preserving the parametric adaptation that provides substantial improvements at the 9B scale. The higher Best SR indicates that refining every symbolic component in every round is unnecessary for attaining stronger peak capability in this comparison. Selective activation therefore provides a way to reduce the additional interaction cost of symbolic refinement while retaining effective model weight adaptation.

\noindent \textbf{(iii) Capability diagnosis distinguishes the forms of refinement required by interaction evidence.} In the Round~5 example presented in Figure~\ref{fig:metacognitive_gating_case_study}, successful lookup after an identity error provides situational guidance suitable for experience refinement, while acting before validating the identity indicates a planning deficiency associated with model weight refinement. The same interaction can therefore reveal both knowledge worth retaining and behavior requiring improvement. In Round~6, repeated retrieval and aggregation sequences additionally support toolset selection since they offer recurring procedures suitable for callable abstraction. These examples illustrate the value of distinguishing what can be retained as situational guidance, what requires behavioral refinement, and what can be consolidated into reusable procedures when selecting adaptation targets, as illustrated in Section~\ref{subsec:evolutionary_gating}.

These results provide empirical support for the feasibility of Stage~4 metacognitive control over component adaptation in MUSE, with capability diagnosis translating observed deficiencies and consolidation opportunities into targeted refinement, and selective activation regulating the interaction demands of continued evolution.

\subsection{Continuous Evolution and Performance Stability} \label{subsec:continuous_evolution}

\begin{figure*}[!t]
    \centering
    \begin{subfigure}[t]{0.485\textwidth}
        \centering
        \includegraphics[width=\linewidth]{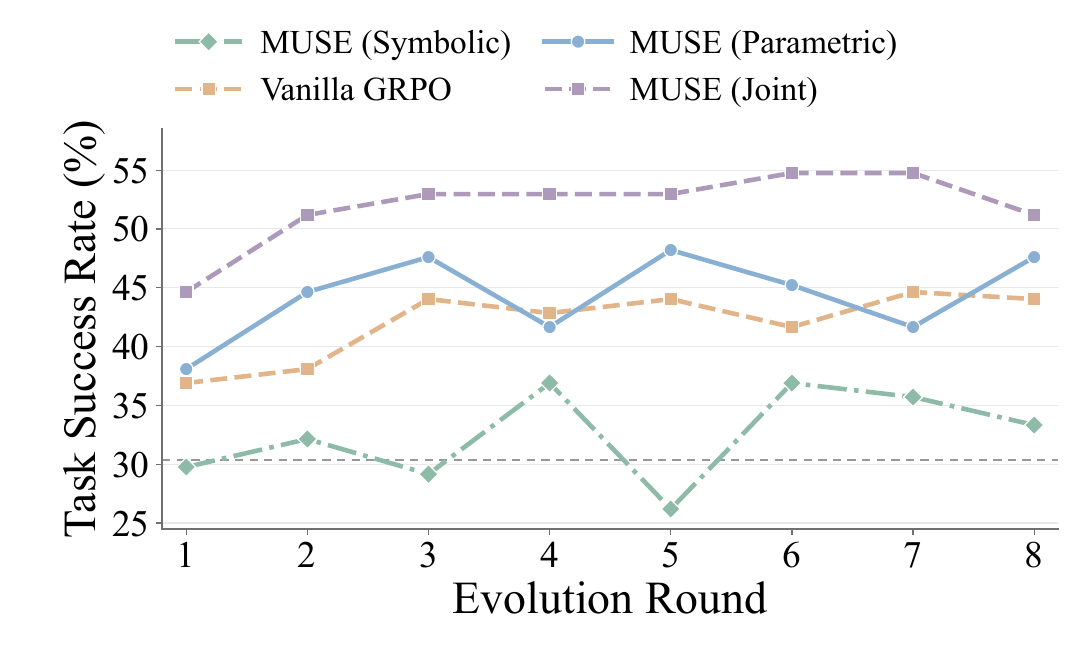}
        \caption{Oracle Tasks}
        \label{fig:continuous_evolution_oracle}
    \end{subfigure}
    \hfill
    \begin{subfigure}[t]{0.485\textwidth}
        \centering
        \includegraphics[width=\linewidth]{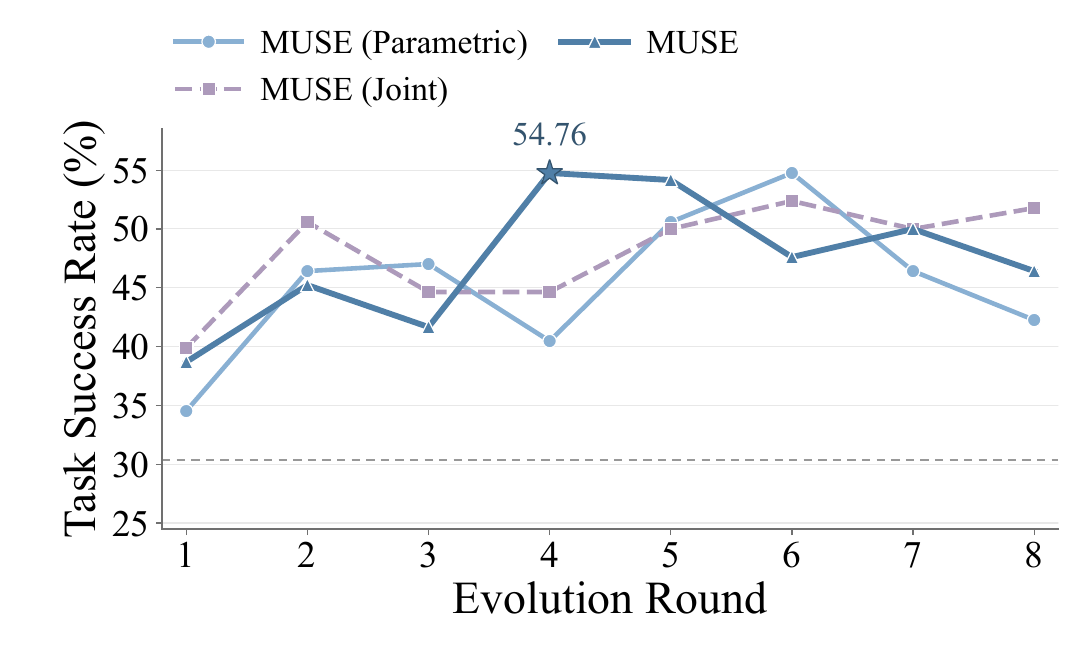}
        \caption{Oracle + Synthetic Tasks}
        \label{fig:continuous_evolution_synthetic}
    \end{subfigure}
    \caption{Continuous evolution on APPWorld with Qwen3.5-9B under (a) oracle tasks only and (b) oracle tasks supplemented with synthetic tasks. Points give the SR of each evolved candidate configuration per round. The dashed line marks the pre-evolution reference, and the star marks the earliest round attaining Best SR for MUSE. \vspace{-4mm}}
    \label{fig:continuous_evolution}
\end{figure*}

To investigate \textbf{RQ5}, we examine how task performance develops and persists across a fixed observation horizon of 8 rounds. We track the evaluated candidate agent configurations produced across 8 rounds of continued interaction and adaptation, examining how task performance changes throughout the evolutionary process. Figure~\ref{fig:continuous_evolution} presents the resulting trajectories with oracle tasks and with additional synthetic tasks as the interaction source. We examine these trajectories alongside Evo-AUC reported in Section~\ref{subsec:component_contributions} and Section~\ref{subsec:supervision_effectiveness}, relating aggregate performance over the evolution horizon to the development and persistence of improvements across successive agent instances.

From Figure~\ref{fig:continuous_evolution} and the corresponding aggregate performance measurements, we draw the following findings.

\noindent \textbf{(i) Successive evolution rounds extend task-performance improvements beyond initial adaptation.} Across the configurations in Figure~\ref{fig:continuous_evolution}, final SR exceeds that of the first evolved agent instance, with peak performance attained in later rounds. This pattern appears under both oracle tasks and supplementary synthetic tasks. The initial update therefore captures only part of the improvement available during the observed evolution process. Continued interaction produces further execution evidence as agent capabilities change, allowing subsequent refinement to support additional capability development. The later improvements support continued adaptation beyond the first evolution round, although individual updates do not consistently increase performance.

\noindent \textbf{(ii) Joint adaptation can maintain elevated performance during subsequent refinement.} With oracle tasks, MUSE (Joint) maintains SR between 51.19\% and 54.76\% from the second through the eighth round, accompanied by an Evo-AUC of 20.28 shown in Table~\ref{tab:unison_contribution}. This pattern indicates that, at the evaluated model scale, continued symbolic and parametric refinement can preserve the substantial improvement established early in evolution, extending the benefit of joint adaptation across repeated component updates.

\noindent \textbf{(iii) Similar peak performance can accompany different levels of candidate stability.} With oracle and synthetic tasks as the interaction source, MUSE and MUSE (Parametric) both attain 54.76\% Best SR, while MUSE records higher Final SR and Evo-AUC across evaluated candidates before deployment selection. Its Final SR of 46.42\%, compared with 42.26\% for MUSE (Parametric), indicates that the final MUSE candidate retains more of the attained improvement at the end of the observed evolution proces. The two configurations therefore differ in the persistence of their improvement despite reaching the same peak capability. However, the decline in MUSE candidates after the fourth and fifth rounds shows that subsequent candidate updates do not consistently preserve the strongest attained performance. Maintaining that performance across subsequent candidates while addressing new adaptation demands remains a challenge for continuous agent evolution.

\subsection{Limitations of the Empirical Evaluation} \label{subsec:experimental_limitations}

The experiments examine the practical feasibility of the Stage~4 mechanisms instantiated in MUSE on APPWorld over a maximum of 8 evolution rounds. Full MUSE is evaluated with Qwen3.5-9B, complemented by symbolic refinement experiments at the 27B scale. Extending the findings to different environments and model families, and assessing capability retention over longer deployments, requires broader empirical evaluation.

Within the evaluated setting, the component comparisons support coordinated symbolic and parametric refinement, with benefits that depend on the agent configuration and available supervision. Symbolic refinement contributes differently at the two model scales, and supplementary synthetic tasks affect parametric and joint adaptation differently, as presented in Section~\ref{subsec:component_contributions} and Section~\ref{subsec:supervision_effectiveness}. These differences indicate that the contribution of an individual refinement channel depends on whether additional component updates are enabled. The human audits provide evidence for the quality of admitted synthetic tasks, while the performance comparisons leave their suitability for different refinement channels incompletely characterized. The current evidence therefore supports the feasibility of coordinated component evolution, with the conditions for matching supplementary supervision to component-specific adaptation demands requiring further investigation.

This configuration dependence also affects how improvements persist across evolution rounds. Metacognitive gating attains stronger peak performance with fewer exploration rollouts, supporting diagnosis-guided component selection, as presented in Section~\ref{subsec:metacognitive_gating_evaluation}. Its peak-performance advantage is less consistent in Final SR and Evo-AUC, and the evolution trajectories show that later updates can partially reduce the attained improvement, as presented in Section~\ref{subsec:continuous_evolution}. The benefits of targeted refinement and evolutionary efficiency therefore require consideration alongside capability retention. The experiments provide empirical support for the practical feasibility of metacognitive gating, interaction-grounded self-supervision, and coordinated component evolution within MUSE as a Stage~4 prototype. Sustaining the capability improvements across successive updates requires further validation.

\section{Open Problems toward Higher Evolutionary Autonomy} \label{sec:open_problems}

This section develops a prospective research agenda for progressively more autonomous agent evolution under changing deployment conditions. Building on the six-component evolutionary strategy and the cumulative capability requirements of the Five-Stage Evolutionary Spectrum, we formulate open problems concerning autonomous adaptation decisions and the evidence needed to assess them. We examine knowledge allocation across symbolic and parametric components (Section~\ref{subsec:discussion_component_coordination}), supervision reliability and task maintenance (Section~\ref{subsec:discussion_supervision}), and targeted refinement and sustained adaptation (Section~\ref{subsec:discussion_control_retention}). We then consider operational boundaries beyond capability retention (Section~\ref{subsec:discussion_objectives_boundaries}), endogenous safety risks of continuous and recursive agent evolution (Section~\ref{subsec:discussion_rsi_risk}), and the evaluation of evolutionary strategy revisions toward Stage~5 (Section~\ref{subsec:discussion_recursive_evolution}).

The capability judgments expressed by $\operatorname{CapPass}_{k,d}$ and summarized in Table~\ref{tab:evolutionary_capability_checklist} identify which evolutionary requirements have been demonstrated under declared conditions. Once a capability has been demonstrated, evaluation can examine the conditions under which the associated autonomous decisions support further adaptation. This requires a testable hypothesis connecting the interaction evidence available to the agent, the expected adaptation decision, and its effects on subsequent tasks. The hypothesis can then be examined through comparisons that start from matched agent configurations and interaction histories under comparable external support, resource budgets, and permissions. These comparisons assess whether the predicted benefit appears in capability improvement, retention, boundary compliance, safety preservation, or the interaction cost of acquiring further capabilities.

\subsection{Optimization Scope and Coordination: Knowledge Allocation} \label{subsec:discussion_component_coordination}

Knowledge acquired during agent evolution can support later execution through symbolic assets, model weights, or both. The allocation affects execution context and the cost of revising knowledge when task demands change. Recent work investigates coordination between these representations through selective internalization and memory-parameter refinement~\citep{ji2026learning, guo2026peam, chen2026rethinking}. These developments motivate examining conditions under which each allocation supports continued adaptation. We ask: \textit{Under what conditions should acquired knowledge be retained in symbolic assets, internalized into model weights, or preserved in both forms?}

We conjecture that validated patterns that recur frequently across tasks and remain stable should be preferentially internalized into model weights, while context-sensitive or frequently changing knowledge should remain available as symbolic assets. This preference is constrained by the execution-context budget and knowledge acquisition and revision costs. External retention also introduces retrieval competition under limited context~\citep{hu2026continual}, so retaining more symbolic support can introduce additional access costs. These criteria would guide $\Gamma$ in selecting which components within $\mathcal C$ should retain a pattern, with $\mathcal M$ implementing the corresponding refinement.

These preferences should also be reassessed as knowledge and agent capabilities change. A pattern initially retained as symbolic guidance may become a candidate for parametric internalization as further interactions establish its recurrence and stability. Changes in previously stable knowledge would also affect the cost of maintaining its parametric representation. Temporary retention in both forms can preserve explicit support while its contribution after model weight updates is assessed. Matched executions with and without the relevant guidance would assess its continued contribution across applicable task contexts. This evidence, together with acquisition, retrieval, and revision costs, would support decisions about continued retention in one or both forms.

Complementing these retention and reassessment criteria, learning dynamics~\citep{ren2025learning} offers a systematic framework for analyzing knowledge allocation, particularly for characterizing the isolated influence of individual data examples on model updates. Recent work~\citep{ren2026learning} further extends this framework to the continual adaptation of LLMs, demonstrating how individual training examples affect downstream predictions and how acquiring new knowledge can interfere with existing capabilities. Characterizing these effects can guide both training-data selection for parametric consolidation and decisions on whether specific knowledge should be internalized into model weights or retained in explicit symbolic form.

The allocation conjecture can be examined through task sequences independently varying cross-task recurrence and the frequency of knowledge changes. Strategies following the proposed criteria would be compared with symbolic retention, parametric internalization, and retention in both forms, starting from the same agent configuration and interaction history under matched resource limits. Allocation decisions would depend on preceding evidence, while subsequent independent tasks would assess task performance, capability retention, and cumulative acquisition, retrieval, and revision costs. The conjecture would be weakened if its preferred allocations offer no cost advantage at comparable performance, or if symbolic retirement degrades execution in less frequent contexts. Varying the amount of preceding evidence would further measure the cost of revising early allocation decisions, including subsequent weight updates and reconstruction of retired symbolic guidance.

\subsection{Supervision Autonomy: Reliability and Task Maintenance}
\label{subsec:discussion_supervision}

Supervision autonomy raises a challenge beyond the correctness of individual assessments: errors can persist through the process that constructs subsequent supervision. A mistaken interpretation may affect outcome rewards for synthetic tasks and prefix-potential supervision, then receive reinforcement through symbolic or parametric refinement. Later assessments can even reinforce the incorrect interpretation, forming a \textit{semantic echo chamber}~\citep{he2026how} through supervision generated by $\Delta$ and updates performed by $\mathcal M$. Studies of self-graded memory document related error reinforcement through retrieval and reuse~\citep{asadolahi2026memory}. Perspectives from cognitive science further clarify this phenomenon: the Bayesian iterated learning framework~\citep{ren2024bias} provides a theoretical account of which inductive biases are more susceptible to amplification over successive learning rounds, suggesting directions for detecting and mitigating such amplification. Since task selection determines which synthetic supervision is admitted, mitigating such error amplification also depends on maintaining a task pool that exposes unresolved deficiencies. We therefore ask: \textit{How can an evolving agent detect and correct semantic echo chambers while maintaining a task pool that supports continued learning?}

We conjecture that effective correction requires environmental evidence capable of contradicting the assessments that previously supported an update. Increasing agreement among generated assessments provides limited evidence of correction when those assessments share the same underlying error. Related studies identify correlated grading errors in memory agents and the reinforcement of incorrect preferences under intrinsic rewards~\citep{asadolahi2026memory, he2026how}. Candidate detection mechanisms could compare generated assessments with executable checks of relevant state changes and additional executions that test disputed task requirements. Tracking the supervision sources associated with recurring errors would help identify which judgments require reassessment. Correcting those judgments should also be accompanied by tests of the behaviors they previously reinforced, as revising a reward does not directly correct related symbolic guidance or learned behavior.

These checks depend on a task pool that continues to expose unresolved deficiencies as agent capabilities change. Repeated success on familiar tasks can reflect genuine mastery, while providing limited evidence about deficiencies outside the currently sampled conditions. Capability-dependent task synthesis offers a foundation for adapting learning opportunities to the current agent~\citep{yang2026ttcs}. We propose that task maintenance should consider verified learning progress together with coverage of unresolved deficiencies. Tasks that become mastered can receive lower optimization priority while remaining available for capability-retention assessment. Newly constructed tasks should examine emerging deficiencies and variations in conditions that challenge previously successful behavior, providing further opportunities to reassess both task relevance and supervision reliability.

These proposals can be examined by independently varying corrective environmental checks and adaptive task-pool maintenance. Comparisons would begin from the same agent configuration and interaction history, with a common candidate-task stream, declared oracle availability, and matched resource budgets. Independently verified executions reserved for evaluation would measure error detection, judgment correction, and the recurrence of errors. Subsequent independent tasks would assess capability improvement and retention. These proposals would be weakened if correction repeatedly excludes valid executions, if identified errors continue to recur, or if task renewal raises self-assessed performance without improving independently evaluated capabilities. These comparisons would establish when environmental checks and task-pool maintenance jointly support reliable supervision.

\subsection{Evolutionary Autonomy: Targeted and Sustained Adaptation}
\label{subsec:discussion_control_retention}

Evolutionary autonomy requires judging when interaction evidence supports persistent adaptation and which components can address the diagnosed demand. An observed failure may reflect a recurring capability deficiency, a temporary increase in task difficulty, or incomplete execution evidence. Successful interactions can also reveal consolidation opportunities. These situations call for different adaptation decisions, whose continued value depends on subsequent behavior. The open problem therefore connects evidence for activation and routing with the pace of refinement and recovery from detrimental updates. We ask: \textit{What evidence should govern evolution activation, component selection, and recovery during continuous adaptation?}

We conjecture that combining recurrence across tasks with limited tests of candidate interventions can improve decisions made through $\Gamma$ at a given interaction budget. Recurring patterns provide evidence of persistent adaptation demands, while intervention comparisons help assess which components can address them. For example, executions with and without relevant symbolic support can help distinguish missing guidance from difficulties that persist despite its availability. The capability diagnosis profile would thereby connect observed deficiencies to refinement targets through tested behavioral differences. Update pace should also reflect how rapidly fresh executions can assess the preceding refinement. Frequent updates based on narrowly sampled conditions can give those conditions disproportionate influence, motivating reassessment before further updates strengthen the same behavior.

Such reassessment connects targeted refinement to capability retention and recovery. Re-execution and update rejection in skill evolution~\citep{zheng2026skillprox}, together with versioned system-wide revision~\citep{zhang2026agentevolversystem}, provide foundations for examining these decisions. We propose treating recoverability as an additional requirement of $\mathcal B$, with compatible versions of model weights and symbolic assets retained for comparison and restoration. Historical-task assessments under controlled symbolic exposure would help distinguish parametric forgetting from changes in available guidance. Recovery decisions should then consider both the earlier capabilities restored and the newly acquired capabilities lost, since returning to a previous configuration can reverse beneficial changes.

These proposals can be examined through task sequences containing persistent deficiencies, temporary increases in difficulty, and recurring successful procedures. Starting from the same agent configuration and interaction history, comparisons would independently vary activation criteria, intervention-based routing, reassessment frequency, and recovery mechanisms. Diagnostic tests and recovery evaluations would count toward matched resource budgets. Subsequent independent tasks would measure capability improvement, retention, and cumulative interaction cost, while intervention results would assess whether selected components address the diagnosed demands. The proposals would be weakened if additional assessment delays beneficial adaptation without improving retention or cost, if selected interventions repeatedly fail to correct the targeted behavior, or if recovery repeatedly discards valuable new capabilities. Such evaluation would clarify when longitudinal evidence supports targeted refinement and how update records can guide its continuation or reversal.

\subsection{Boundary Constraints beyond Capability Retention}
\label{subsec:discussion_objectives_boundaries}

Beyond capability retention and recoverability within $\mathcal B$, continuous evolution also changes the procedures through which the agent interacts with external resources. Operational boundaries therefore also govern the external effects permitted during both adaptation and subsequent execution, including changes in permissions throughout deployment. MUSE supports the capability-retention aspect of $\mathcal B$ through historical trajectory replay, while constraints on authorization and external operational effects are not instantiated or evaluated in the prototype. Establishing explicit runtime constraints on these external effects constitutes a separate research direction. We ask: \textit{How can an evolving agent align the effects of its actions with changing authorization boundaries?}

We propose that operational requirements within $\mathcal B$ should specify permitted effects on data access, external state, and information disclosure. The corresponding hypothesis is that assessing these effects can preserve authorization boundaries across changes in tool implementation and composition. A revised tool may retain its interface while changing which resources it accesses or which external states it modifies. Runtime constraint specification provides relevant foundations~\citep{wang2026agentspec}, while autonomous tool refinement introduces the additional challenge of maintaining those constraints after implementation changes. Candidate mechanisms could combine inspection of revised procedures, isolated execution, and monitoring of actual effects. Their reliability would depend on whether the effects anticipated before execution correspond to those independently observed afterward.

Assessing effects also requires distinguishing environmental evidence from authority to act. Research on prompt injection demonstrates how instructions embedded in tool returns can redirect agent behavior~\citep{debenedetti2024agentdojo}. Such content can describe an operation without providing authorization for its consequences. We propose that unclear boundaries should lead to a request identifying the specific resource, operation, or recipient requiring clarification from an authorized party. Previously granted permissions should likewise be reassessed when their scope changes or authorization is withdrawn. These decisions concern both task execution and the collection of further learning evidence. They are particularly important for consequential operations, since restoring an earlier agent configuration does not automatically reverse information disclosure or external state changes.

The hypothesis can be examined under fixed task objectives and initial agent configurations while independently varying tool implementations, permission scopes, authorization withdrawal, and instructions embedded in environmental content. Each setting would specify permitted effects independently of the agent and include cases requiring clarification. Comparisons between checks of tool identities and checks of operational effects would record execution, blocking, and clarification decisions alongside actual data access and state changes. Evaluation would jointly measure authorized task completion, permission violations, unnecessary clarification, and adaptation cost following permission changes. Failure conditions would include missed unauthorized effects in composed operations, continued execution after permission withdrawal, or excessive interruption of permitted tasks. Varying the observability of external effects would further identify where additional external checks are needed to support autonomous evolution within declared boundaries.

\subsection{Safety Risks of Continuous and Recursive Agent Evolution}
\label{subsec:discussion_rsi_risk}
Even when explicit operational boundaries are specified within $\mathcal{B}$ and supervision remains reliable, continuous capability improvement can diverge from safety preservation. Adaptation toward a target objective $\mathcal{O}$ can unintentionally erode existing safety behaviors through learning interference~\citep{ren2026learning}, while minor biases can accumulate and amplify across evolution rounds. Similar endogenous risks extend to non-parametric adaptation: the OpenAI/Hugging Face incident~\citep{metr_hf} illustrates how autonomous agent coordination and knowledge sharing can produce unintended and unsafe behavior. We therefore ask: \textit{How can safety risk accumulation be detected and mitigated during continuous and recursive agent evolution?}

We conjecture that safety preservation requires independent behavioral assessment throughout evolution in addition to pre-deployment verification. Recent advances in misalignment monitoring, including activation-based probes and chain-of-thought monitoring~\citep{yueh2026reasoning,openai2026misalignment}, provide concrete foundations for such assessment. However, the reliability of these monitoring mechanisms can itself shift as the agent adapts across rounds. This concern becomes particularly critical toward Stage~5, where revisions to the evolutionary strategy $\mathcal S$ can modify supervision and evaluation mechanisms directly.

This conjecture can be examined by tracking task performance and safety-related behaviors across evolution rounds. Comparisons should contrast pre-deployment verification alone with continuous monitoring and corrective interventions across parametric and symbolic adaptation under matched objectives and resource budgets, with independent safety evaluations covering behaviors beyond immediate optimization targets. The conjecture would be weakened if safety remains stable under pre-deployment verification alone, or if continuous monitoring and intervention fail to reduce degradation without halting capability progress. Such experiments would clarify when safety degradation emerges, how it accumulates across rounds, and whether continuous monitoring and corrective interventions can mitigate these risks while preserving capability acquisition.

\subsection{Toward Stage~5: Recursive Meta-Evolution}
\label{subsec:discussion_recursive_evolution}

Recursive meta-evolution raises a further question about how improvements in the evolutionary strategy can be assessed. Changes to $\mathcal{S}$ can affect how subsequent interactions support capability development, with effects extending across later evolution rounds. Recent studies explore this direction through editable self-modification procedures and evolving meta-skills~\citep{zhang2026hyperagents, wang2026metaskill}. Assessing such changes requires examining the adaptation supported by the revised strategy under comparable deployment conditions. We ask: \textit{Under what conditions does revising the evolutionary strategy improve subsequent adaptation?}

Update records provide a basis for proposing revisions. When similar capability deficiencies persist across successive refinements, the agent may need to consider revising its evolutionary strategy. The relevant changes can involve any of the six components of $\mathcal{S}$ and the relationships among them. We propose that a candidate revision should connect an observed adaptation demand to an expected change in subsequent evolution. Its evaluation would then examine whether that change improves capability acquisition, capability retention, or evolutionary efficiency. This connects the reason for revising the strategy to evidence for assessing the revision.

To examine a candidate revision, we propose comparing the original strategy $\mathcal{S}_t$ and the revised strategy $\hat{\mathcal{S}}_t$ from the same initial agent configuration $A_t$, environment state $E_t$, and interaction history $H_t$. Both strategies would encounter the same sequence of new task demands under matched external supervision, resource budgets, and permissions. Each would select its own component updates and collect the resulting interaction evidence. The task sequences for comparison would remain separate from those involved in proposing and selecting the revision. Each comparison would assess one strategy revision before further changes to $\mathcal{S}$ are introduced. The interaction and computation required to propose, select, and test the revision would be included in its total cost. Evaluation would follow common criteria specified independently of both strategies, including when the revision changes $\mathcal{O}$ or $\mathcal{B}$.

A revision would receive support if it reduces the interactions required to reach a common capability level, or sustains higher task performance across later rounds at comparable cost, while satisfying declared retention and operational requirements. Repeated comparisons across independent task sequences would examine whether benefits extend beyond the conditions motivating the revision. Benefits confined to previously examined tasks or disappearing after revision costs are included provide insufficient support for adoption. A concrete step is to make adoption decisions from a small set of new tasks, then evaluate both adopted and rejected revisions through paired comparisons across later evolution rounds. Evaluation would record testing cost and the frequencies of adopting revisions that degrade subsequent adaptation and rejecting revisions that improve it under matched comparisons.

\section{Conclusion}

In this paper, we formalize \textbf{agent evolution} through a six-component evolutionary strategy that provides a common basis for describing how continuous adaptation is organized throughout deployment under specified objectives and operational constraints. To characterize the capabilities that support progressively more autonomous agent evolution, we introduce the Five-Stage Evolutionary Spectrum with cumulative capability requirements across three dimensions, namely \textit{evolutionary autonomy}, \textit{supervision autonomy}, and \textit{optimization scope and coordination}. Within this progression from human-engineered maintenance to recursive meta-evolution, MUSE instantiates the evolutionary strategy as a Stage~4 prototype through metacognitive gating, interaction-grounded self-supervision, and coordinated component evolution. On APPWorld, we examine both how MUSE organizes continuous adaptation and how agent capabilities develop across successive evolution rounds under different component configurations. Supervision audits characterize the quality of the supplementary tasks and their outcome rewards, and diagnostic records show how identified capability gaps guide component selection. Controlled comparisons further relate coordinated component refinement to improved task performance and selective activation to greater evolutionary efficiency. Together, these findings provide empirical support for the practical feasibility of the Stage~4 mechanisms instantiated in MUSE. Further research can assess the reliability of autonomous adaptation decisions and the value of evolutionary strategy revisions across changing environments and longer deployments.

{
  \small
  \bibliographystyle{plainnat}
  \bibliography{main}
}

\clearpage
\appendix

\section*{Appendix Contents}
\addcontentsline{toc}{section}{Appendix}
\stopcontents[mainbody]
\startcontents[appendices]

\markboth{Appendix Contents}{Appendix Contents}
\thispagestyle{plain}

\begin{spacing}{1.1}
\printcontents[appendices]{}{1}[4]{}
\end{spacing}
\clearpage

\begingroup
\clubpenalty=10000
\widowpenalty=10000
\setlength{\intextsep}{4pt plus 1.5pt minus 1.5pt}
\setlength{\textfloatsep}{5pt plus 1.5pt minus 1.5pt}
\setlength{\floatsep}{4pt plus 1.5pt minus 1.5pt}
\setlength{\aboverulesep}{0.2ex}
\setlength{\belowrulesep}{0.2ex}
\setlength{\parskip}{2pt plus 1pt}
\captionsetup[table]{skip=2pt,font=small}

\section{Notation}
\label{app:notation}

The following tables summarize the notation for agent evolution. Appendix~\ref{subapp:notation_agent_system_and_strategy} covers the agent system and the six-component evolutionary strategy, while Appendix~\ref{subapp:notation_evolutionary_spectrum} summarizes the notation for assessing dimensional requirements and assigning stages in the Five-Stage Evolutionary Spectrum. For MUSE, Appendix~\ref{subapp:notation_gating_and_supervision} covers evolutionary gating and supervision generation, and Appendix~\ref{subapp:notation_refinement_synchronization_and_replay} groups the symbols for inner-component refinement, inter-component synchronization, and task-specific trajectory replay. Appendix~\ref{subapp:notation_evaluation_measures} lists the experimental measures, with their definitions and calculation procedures detailed in Appendix~\ref{subapp:metrics}.

\subsection{Agent System and Evolutionary Strategy}
\label{subapp:notation_agent_system_and_strategy}

The formulation in Section~\ref{sec:formalization} distinguishes interaction steps $\ell$ from evolution rounds $t$. Table~\ref{tab:notation_agent_system_and_strategy} connects these indices to the histories and configurations involved in successive agent updates.

\begin{table}[H]
    \centering
    \caption{Notation for the interaction environment, agent system, and evolutionary strategy.}
    \label{tab:notation_agent_system_and_strategy}
    \footnotesize
    \setlength{\tabcolsep}{3.5pt}
    \renewcommand{\arraystretch}{0.97}
    \begin{tabularx}{\linewidth}{@{}
        >{\raggedright\arraybackslash}p{0.23\linewidth}
        X
        @{}
    }
        \toprule
        \textbf{Notation} & \textbf{Definition} \\
        \midrule

        \rowcolor{notationblue!18}
        \multicolumn{2}{c}{\textit{Interaction Environment and History}} \\
        \midrule

        $E \in \mathcal{E}$
        & Partially observable interaction environment. \\

        $i,\ \ell$
        & Execution index and interaction step within an execution, respectively. \\

        $z_\ell \in \mathcal{Z},\ o_\ell \in \Omega$
        & Hidden environment state and corresponding agent observation at step $\ell$. \\

        $g_\ell \in \mathcal{G},\ u_\ell \in \mathcal{U}$
        & Current goal and action selected by the agent, $u_\ell\sim\pi_A(\cdot\mid H_{i,\mathrm{act}}^{(\ell-1)},o_\ell,g_\ell)$, where $\pi_A$ denotes the action policy. \\

        $\mathcal{P}(z_{\ell+1}\mid z_\ell,u_\ell)$
        & Environment transition law for the next state given the current state and action. \\

        $\mathcal{R},\ r_\ell \in \mathcal{Y}$
        & Environment feedback function and optional feedback $r_\ell=\mathcal{R}(z_\ell,u_\ell,g_\ell)$, which may be a scalar reward or textual critique. \\

        $H_{i,\mathrm{act}}^{(k)}$
        & Ordered record of the first $k$ interaction steps in execution $i$, comprising observations, goals, actions, and available feedback. Before action $\ell$, the agent conditions on $H_{i,\mathrm{act}}^{(\ell-1)}$ together with the current observation and goal. \\

        \midrule

        \rowcolor{notationorange!18}
        \multicolumn{2}{c}{\textit{Agent System}} \\
        \midrule

        $A \in \mathcal{A}$
        & Agent system with configuration $A=\Theta\cup\Psi\cup\Phi$. \\

        $\Theta$
        & Cognitive core, including model weights $\theta_{\text{model}}$ and, in multi-agent settings, an optional collaborative topology $N_{\text{topo}}$. \\

        $\Psi$
        & Explicit symbolic assets, including general principles $P_{\text{sys}}$, an experience bank $K_{\text{exp}}$, a functional toolset $T_{\text{tool}}$, and other assets. \\

        $\Phi$
        & Operational hyperparameters, including retrieval capacity $k_{\text{ret}}$, decoding temperature $\kappa_{\text{dec}}$,  similarity threshold $\epsilon_{\text{sim}}$, and other operational hyperparameters. \\

        \midrule

        \rowcolor{notationblue!18}
        \multicolumn{2}{c}{\textit{Six-Component Evolutionary Strategy}} \\
        \midrule

        $\mathcal{S}\colon\mathcal{A}\times\mathcal{E}\times\mathcal{H}\to\mathcal{A}$
        & Evolutionary strategy $\langle\Delta,\mathcal{C},\mathcal{M},\Gamma,\mathcal{O},\mathcal{B}\rangle$ governing adaptation of the agent configuration. \\

        $\Delta,\ \delta_t\in\mathcal{D}$
        & Supervision generator and the learning signals acquired or constructed for round $t$. \\

        $\mathcal{C} \subseteq A,\ C \subseteq \mathcal{C}$
        & Set of agent components eligible for adaptation and a selected subset. The current agent supplies the configurations of these components. \\

        $\mathcal{M},\ \mathcal{A}|_C$
        & Evolutionary optimizer and the configuration space of the selected components. The optimizer takes the current agent, selected component subset, and supervision, and returns an updated configuration in $\mathcal{A}|_C$. \\

        $\Gamma,\ \operatorname{Sel}(\mathcal C)$
        & Evolutionary gating mechanism and its family of permitted component subsets, including the empty set. The output $(\gamma_{\text{act}},C_t^{\mathrm{sel}})$ specifies activation and component selection, with $\gamma_{\text{act}}\in\{0,1\}$. \\

        $\mathcal{O}$
        & Evolutionary objective that evaluates an agent configuration in its environment under the specified criteria. \\

        $\mathcal{B}$
        & Boundary constraint specifying deployment requirements for agent updates and resulting behavior. Requirements may be checked explicitly or supported implicitly through update design. \\

        \midrule

        \rowcolor{notationorange!18}
        \multicolumn{2}{c}{\textit{Evolution across Rounds}} \\
        \midrule

        $t,\ T$
        & Evolution-round index and total number of rounds in the evolutionary trajectory. \\

        $E_t,\ H_t$
        & Environment and accumulated interaction history available at evolution round $t$. \\

        $C_t^{\mathrm{sel}},\ \hat{C}_t^{\mathrm{sel}}$
        & Component subset selected by $\Gamma$ and its updated configuration, $\hat{C}_t^{\mathrm{sel}}=\mathcal{M}(A_t, C_t^{\mathrm{sel}},\delta_t)$. \\

        $A_t,\ \mathcal{S}_t,\ A_{t+1}$
        & Agent transition under the strategy at round $t$, $A_{t+1}=\mathcal{S}_t(A_t,E_t,H_t)$. \\

        \bottomrule
    \end{tabularx}
\end{table}

\newpage
\subsection{Five-Stage Evolutionary Spectrum}
\label{subapp:notation_evolutionary_spectrum}

Stage assignment combines the dimensional requirements in Table~\ref{tab:evolution_roadmap} through the indicators in Table~\ref{tab:notation_evolutionary_spectrum}. These indicators support the cumulative assessment defined in Sections~\ref{subsec:evaluate_dimensions} and~\ref{subsec:spectrum_framework}.

\begin{table}[H]
    \centering
    \caption{Notation for the evaluative dimensions and cumulative stage assignment.}
    \label{tab:notation_evolutionary_spectrum}
    \footnotesize
    \setlength{\tabcolsep}{3.5pt}
    \renewcommand{\arraystretch}{0.97}
    \begin{tabularx}{\linewidth}{@{}
        >{\raggedright\arraybackslash}p{0.23\linewidth}
        X
        @{}
    }
        \toprule
        \textbf{Notation} & \textbf{Definition} \\
        \midrule

        \rowcolor{notationblue!18}
        \multicolumn{2}{c}{\textit{Evaluative Dimensions}} \\
        \midrule

        $\mathrm{EA}$
        & \textit{Evolutionary autonomy.} The extent to which the agent determines when evolution is activated and which components are selected for adaptation. \\

        $\mathrm{SA}$
        & \textit{Supervision autonomy.} The extent to which the agent independently acquires, constructs, and validates learning signals for evolution. \\

        $\mathrm{SC}$
        & \textit{Optimization scope and coordination.} The breadth of autonomous component adaptation and the coordination of updates across those components. \\

        \midrule

        \rowcolor{notationorange!18}
        \multicolumn{2}{c}{\textit{Cumulative Stage Assignment}} \\
        \midrule

        $k,\ d$
        & Evolutionary stage index and evaluative dimension, respectively. Stage~1 is the baseline. Additional requirements are assessed for $k\in\{2,3,4,5\}$ and $d\in\{\mathrm{EA},\mathrm{SA},\mathrm{SC}\}$. \\

        $\operatorname{CapPass}_{k,d}$
        & Binary indicator of whether all numbered requirements introduced for dimension $d$ at stage $k$ in Table~\ref{tab:evolution_roadmap} have been demonstrated. A zero covers both a documented failure to meet a requirement and insufficient evidence to establish it. \\

        $\operatorname{StagePass}_{k}$
        & Cumulative stage indicator, with $\operatorname{StagePass}_{1}=1$. For $k\geq 2$, it equals 1 exactly when the preceding stage is satisfied and the same implementation jointly demonstrates the new requirements across all three dimensions, as defined in Equation~\ref{equ:cumulative_stage_pass}. \\

        $\operatorname{EvoStage}(A,\mathcal{S}\mid E)$
        & Highest stage demonstrated by agent system $A$ with strategy $\mathcal{S}$ in environment $E$ under the declared deployment conditions. Equation~\ref{equ:evolutionary_stage_index} gives its value as the sum of the cumulative stage indicators. \\

        \bottomrule
    \end{tabularx}
\end{table}

\subsection{Evolutionary Gating and Supervision Generation}
\label{subapp:notation_gating_and_supervision}

Within each evolution round, evolutionary gating $\Gamma$ links evolution activation to component selection. Table~\ref{tab:notation_evolutionary_gating} distinguishes the activation checks from diagnostic routing in Section~\ref{subsec:evolutionary_gating} and Appendix~\ref{subapp:evolutionary_gating}.

\begin{table}[H]
    \centering
    \caption{Notation for evolution activation and metacognitive diagnostic routing.}
    \label{tab:notation_evolutionary_gating}
    \footnotesize
    \setlength{\tabcolsep}{3.5pt}
    \renewcommand{\arraystretch}{0.97}
    \begin{tabularx}{\linewidth}{@{}
        >{\raggedright\arraybackslash}p{0.23\linewidth}
        X
        @{}
    }
        \toprule
        \textbf{Notation} & \textbf{Definition} \\
        \midrule

        \rowcolor{notationblue!18}
        \multicolumn{2}{c}{\textit{Capability Gap and Accumulation Trigger}} \\
        \midrule

        $\mathcal{W}_t$
        & Nonempty sliding window of recent task executions and their records at round $t$. \\

        $r_{\text{out},i},\ \epsilon_{\text{succ}},\ c_i$
        & Outcome reward, success threshold, and binary completion indicator for execution $i$. In the threshold-based instantiation, $c_i=\mathbb{I}[r_{\text{out},i}\geq\epsilon_{\text{succ}}]$. \\

        $\bar{c}_{\text{env}},\ c_{\text{th}}$
        & Mean completion rate over $\mathcal{W}_t$ and the performance threshold for activation. \\

        $V_{\text{exe}},\ V_{\text{acc}}$
        & Number of executions since the previous evolution round and the accumulation threshold. \\

        $\mathbb{I}[\cdot],\ \gamma_{\text{act}}$
        & Indicator function and binary activation decision. In our MUSE implementation, evolution is activated if $\bar{c}_{\text{env}}<c_{\text{th}}$ or $V_{\text{exe}}\geq V_{\text{acc}}$. \\

        \midrule

        \rowcolor{notationorange!18}
        \multicolumn{2}{c}{\textit{Metacognitive Diagnostic Routing}} \\
        \midrule

        $\tilde{\mathcal{W}}_t,\ A_{\text{diag}}$
        & Diagnostic sample from $\mathcal{W}_t$ and the evolving agent acting as the diagnostic router. The sample balances successful and failed executions when both are available and records the symbolic support presented during execution. \\

        $\mathbf{d}_t^{\mathrm{diag}},\ \mathbf{v}_t$
        & Capability diagnosis profile and binary routing vector jointly produced by $A_{\text{diag}}$. The profile describes capability limitations and consolidation opportunities. The vector identifies the corresponding update channels. \\

        $v_K,\ v_P$
        & Experience and principle entries of $\mathbf{v}_t$. The former selects $K_{\text{exp}}$ for gaps in situational guidance. The latter selects $P_{\text{sys}}$ when validated experiences support a rule applicable across contexts. \\

        $v_T,\ v_\theta$
        & Toolset and model weight entries of $\mathbf{v}_t$. The former selects $T_{\text{tool}}$ for recurring procedures that can become callable tools. The latter selects $\theta_{\text{model}}$ for observed reasoning or execution difficulties. \\

        $\operatorname{Route}(\mathbf{v}_t,A_t)$
        & Mapping from $\mathbf{v}_t=[v_K,v_P,v_T,v_\theta]$ to a selected component subset $C_t^{\mathrm{sel}}\subseteq\mathcal{C}$ in the current agent $A_t$. Multiple entries may be active. \\

        $C_t^{\mathrm{sel}},\ \varnothing$
        & Selected component subset and the empty set returned when evolution is inactive. \\

        $\Gamma(A_t,E_t,H_t)$
        & Gating output $(\gamma_{\text{act}},C_t^{\mathrm{sel}})$, comprising the activation decision and the selected component subset. \\

        \bottomrule
    \end{tabularx}
\end{table}

Once evolution is activated, $\Delta$ constructs supplementary tasks and evaluative signals, following Section~\ref{subsec:supervision_generator} and Appendix~\ref{subapp:supervision_generator}. Table~\ref{tab:notation_supervision_generation} summarizes the notation for interaction-grounded task synthesis.

\begin{table}[H]
    \centering
    \caption{Notation for interaction-grounded task synthesis in MUSE.}
    \label{tab:notation_supervision_generation}
    \footnotesize
    \setlength{\tabcolsep}{3.5pt}
    \renewcommand{\arraystretch}{0.97}
    \begin{tabularx}{\linewidth}{@{}
        >{\raggedright\arraybackslash}p{0.23\linewidth}
        X
        @{}
    }
        \toprule
        \textbf{Notation} & \textbf{Definition} \\
        \midrule

        \rowcolor{notationblue!18}
        \multicolumn{2}{c}{\textit{Interaction-Grounded Task Synthesis}} \\
        \midrule

        $P_{\text{env}},\ A_{\text{sum}}$
        & Environment profile describing available operations and domains, and the evolving agent acting as task summarizer. \\

        $H_i^{\text{explore}},\ N_{\text{explore}}$
        & Exploratory trajectory and number of trajectories collected without a predefined task query. \\

        $Q_{\text{explore}},\ Q_{\text{explore}}^{+}$
        & All exploratory trajectories and the subset containing at least one successful operational workflow. \\

        $\mathcal{T}_{\text{cand}}^{(t)},\ N_{\text{cand}}$
        & Collection and number of provisional tasks proposed from the successful workflows at round $t$. \\

        $\tau_i^{(0)},\ q_i^{(0)},\ y_i^{(0)}$
        & Provisional task, its proposed query, and its proposed solution. \\

        $\tau_i^\star,\ q_i^\star,\ y_i^\star,\ H_{i,\text{ref}}^\star$
        & Canonical task, comprising its query, grounded reference solution, and successful reference trajectory. \\

        \bottomrule
    \end{tabularx}
\end{table}

Candidate synthetic tasks undergo guided and free executions to evaluate their suitability for the current agent configuration, with the frontier-assessment quantities summarized in Table~\ref{tab:notation_frontier_assessment}.

\begin{table}[H]
    \centering
    \caption{Notation for guided and free frontier assessment of synthetic tasks.}
    \label{tab:notation_frontier_assessment}
    \footnotesize
    \setlength{\tabcolsep}{3.5pt}
    \renewcommand{\arraystretch}{0.97}
    \begin{tabularx}{\linewidth}{@{}
        >{\raggedright\arraybackslash}p{0.23\linewidth}
        X
        @{}
    }
        \toprule
        \textbf{Notation} & \textbf{Definition} \\
        \midrule

        \rowcolor{notationorange!18}
        \multicolumn{2}{c}{\textit{Guided and Free Frontier Assessment}} \\
        \midrule

        $N_{\mathrm{g}},\ N_{\mathrm{f}}$
        & Numbers of guided and free rollouts per task. Guided executions receive $(q_i^\star,y_i^\star)$, and free executions receive only $q_i^\star$. \\

        $r_{i,j}^{\mathrm{g}},\ r_{i,j}^{\mathrm{f}}$
        & Outcome reward of rollout $j$ for task $i$ under guided or free execution, respectively. \\

        $\bar r_i^{\mathrm{g}},\ \bar r_i^{\mathrm{f}},\ \operatorname{gap}_i$
        & Mean guided and free rewards and their difference, $\operatorname{gap}_i=\bar r_i^{\mathrm{g}}-\bar r_i^{\mathrm{f}}$. \\

        $r_{i,\max}^{\mathrm{g}},\ r_{i,\max}^{\mathrm{f}}$
        & Maximum guided and free rewards for task $i$. \\

        $\hat{p}_i^{\mathrm{g}},\ \hat{p}_i^{\mathrm{f}}$
        & Fractions of guided and free executions that are valid and judged successful by the discriminator, respectively. \\

        $\epsilon_{\mathrm{g,succ}}$
        & Minimum guided success rate required for both frontier classes. \\

        $\epsilon_{\mathrm{f,low}},\ \epsilon_{\mathrm{f,high}}$
        & Lower and upper bounds on the free success rate for regular-frontier tasks. \\

        $\epsilon_{\mathrm{f,partial}}$
        & Minimum peak free outcome reward for challenge-frontier tasks, which additionally require $\hat{p}_i^{\mathrm{f}}=0$. \\

        $\epsilon_{\mathrm{gap}},\ \epsilon_{\mathrm{gap,ch}},\ \epsilon_{\mathrm{gap,max}}$
        & Minimum guidance gaps for regular and challenge tasks, respectively, and their common upper bound. \\

        $\operatorname{type}(\tau_i^\star),\ \mathcal{T}_{\text{front}}^{(t)}$
        & Frontier classification (regular, challenge, or rejected) and the set of validated regular and challenge tasks at round $t$. \\

        \bottomrule
    \end{tabularx}
\end{table}

Table~\ref{tab:notation_reward_prefix_supervision} summarizes the notation for reference-grounded outcome rewards on synthetic tasks and prefix-potential process supervision across both oracle and synthetic task sources.

\begin{table}[H]
    \centering
    \caption{Notation for outcome rewards on synthetic tasks and prefix-potential process supervision.}
    \label{tab:notation_reward_prefix_supervision}
    \footnotesize
    \setlength{\tabcolsep}{3.5pt}
    \renewcommand{\arraystretch}{0.97}
    \begin{tabularx}{\linewidth}{@{}
        >{\raggedright\arraybackslash}p{0.23\linewidth}
        X
        @{}
    }
        \toprule
        \textbf{Notation} & \textbf{Definition} \\
        \midrule

        \rowcolor{notationblue!18}
        \multicolumn{2}{c}{\textit{Outcome Rewards and Prefix-Potential Process Supervision}} \\
        \midrule

        $H_{i,\text{exe}},\ A_{\text{disc}}$
        & New execution trajectory and the evolving agent acting as discriminator, assessing the execution against the canonical query and validated reference evidence. \\

        $\mathcal{R}_{\text{syn}},\ r_{\text{out},i}^{\text{syn}}$
        & Outcome reward function for synthetic tasks and the assigned completion score in $[0,1]$. \\

        $q_i,\ H_i^{(0)},\ H_i^{(\ell)}$
        & Task query specifying the goal and its constraints, query-only prefix $H_i^{(0)}=(q_i)$, and visible prefix containing the query, executed actions, and returned observations through step $\ell$, for either an oracle or a synthetic task. \\

        $b_k,\ N_{\mathrm{bin}},\ v(b_k)$
        & Ordered potential bin, number of bins, and numerical anchor $v(b_k)=(k-1)/(N_{\mathrm{bin}}-1)$ for $1\leq k\leq N_{\mathrm{bin}}$ and $N_{\mathrm{bin}}\geq2$. \\

        $\mathrm{logit}_i^{(\ell)}(b_k),\ \kappa_{\text{eval}}$
        & Generation logit for bin $b_k$ given the prefix and the positive temperature for normalizing logits across bins. \\

        $\Pr(b_k\mid H_i^{(\ell)}),\ \tilde{\omega}_i^{(\ell)}$
        & Normalized probability of bin $b_k$ and the expected bin anchor under that distribution. The query-only estimate is $\tilde{\omega}_i^{(0)}$. \\

        $\omega_i^{(\ell)},\ \epsilon$
        & Log-potential $\omega_i^{(\ell)}=\log\max(\tilde{\omega}_i^{(\ell)},\epsilon)$ and its numerical floor $\epsilon\in(0,1)$. \\

        $r_{\text{pote},i}^{(\ell)},\ L_i$
        & Potential-based step reward $\omega_i^{(\ell)}-\omega_i^{(\ell-1)}$ and execution length. Explicit submission or termination with no error indicator in its observation inherits the preceding potential, giving $r_{\text{pote},i}^{(L_i)}=0$. A final operational step at the interaction limit is assessed normally. \\

        \bottomrule
    \end{tabularx}
\end{table}

These assessments also guide the composition of the active task pool. Table~\ref{tab:notation_task_pool_adaptation} connects task priority, mastery assessment, and retirement to task selection in subsequent rounds.

\begin{table}[H]
    \centering
    \caption{Notation for continual task-pool adaptation in MUSE.}
    \label{tab:notation_task_pool_adaptation}
    \footnotesize
    \setlength{\tabcolsep}{3.5pt}
    \renewcommand{\arraystretch}{0.97}
    \begin{tabularx}{\linewidth}{@{}
        >{\raggedright\arraybackslash}p{0.23\linewidth}
        X
        @{}
    }
        \toprule
        \textbf{Notation} & \textbf{Definition} \\
        \midrule

        \rowcolor{notationorange!18}
        \multicolumn{2}{c}{\textit{Continual Task-Pool Adaptation}} \\
        \midrule

        $\operatorname{prio}(\tau_i^\star),\ \lambda_{\text{unc}},\ \operatorname{Std}$
        & Task selection priority, uncertainty penalty weight, and population standard deviation. Priority discounts the guidance gap by $\lambda_{\text{unc}}$ times the standard deviation of free rewards. \\

        $\mathcal{T}_{\text{syn}}^{(t)},\ N_{\text{syn}}$
        & Active synthetic task pool at round $t$ and its maximum capacity. \\

        $n_{i,\mathrm{mast}}^{(t)},\ n_{\mathrm{master}}$
        & Number of qualifying rounds among the most recent $n_{\mathrm{master}}$ complete sampled rounds, and the assessment-window length. \\

        $\epsilon_{\mathrm{syn,succ}},\ \epsilon_{\mathrm{hist,SR}}$
        & Outcome reward threshold for success in mastery assessment and verification, and minimum success rate for a sampled round to qualify. \\

        $\bar r_{i,\mathrm{hist}}^{(t)},\ \epsilon_{\mathrm{hist,mean}}$
        & Pooled mean outcome reward across the assessment window and its minimum threshold for separate mastery verification. \\

        $r_{i,\mathrm{rev},j}^{\mathrm{f}},\ \bar r_{i,\mathrm{rev}}^{\mathrm{f}}$
        & Individual and mean outcome rewards from separate query-only mastery verification. \\

        $\epsilon_{\text{retire}}$
        & Minimum verification mean outcome reward. Following the historical mastery criteria, retirement requires every verification reward to reach $\epsilon_{\mathrm{syn,succ}}$ and their mean to reach $\epsilon_{\text{retire}}$. \\

        $\mathcal{T}_{\text{retire}}^{(t)},\ \mathcal{I}_{\text{retire}}^{(\leq t)}$
        & Tasks retired at round $t$ and the cumulative registry of retired task identifiers through $t$. \\

        $\operatorname{id}(\tau),\ \operatorname{Merge}_{N_{\text{syn}}}$
        & Stable identifier derived from the canonical query, and the capacity-bounded merge of eligible historical and new frontier tasks. Retired identifiers and duplicates are excluded. \\

        $\Delta,\ \delta_t=\Delta(A_t,E_t,H_t)$
        & MUSE supervision generator and its output $\delta_t\in\mathcal{D}$ at round $t$, comprising supplementary tasks, outcome scoring rules for synthetic tasks, and prefix-scoring rules for both task sources. \\

        \bottomrule
    \end{tabularx}
\end{table}

\subsection{Component Refinement, Synchronization, and Replay}
\label{subapp:notation_refinement_synchronization_and_replay}

Given $C_t^{\mathrm{sel}}\subseteq\mathcal C$ selected by evolutionary gating $\Gamma$ and supervision $\delta_t$ from Appendix~\ref{subapp:notation_gating_and_supervision}, the optimizer returns $\hat C_t^{\mathrm{sel}}=\mathcal M(A_t,C_t^{\mathrm{sel}},\delta_t)$. For experiences, candidate admission depends on the paired execution comparisons in Table~\ref{tab:notation_experience_refinement}. The corresponding procedures are detailed in Section~\ref{subsec:optimization_scope_and_evolutionary_optimizer} and Appendix~\ref{subapp:optimization_scope_and_evolutionary_optimizer}.

\begin{table}[H]
    \centering
    \caption{Notation for experience extraction, screening, and paired verification.}
    \label{tab:notation_experience_refinement}
    \footnotesize
    \setlength{\tabcolsep}{3.5pt}
    \renewcommand{\arraystretch}{0.97}
    \begin{tabularx}{\linewidth}{@{}
        >{\raggedright\arraybackslash}p{0.23\linewidth}
        X
        @{}
    }
        \toprule
        \textbf{Notation} & \textbf{Definition} \\
        \midrule

        \rowcolor{notationblue!18}
        \multicolumn{2}{c}{\textit{Experience Extraction and Screening}} \\
        \midrule

        $K_{\text{exp}}, A_{\text{exp}}$
        & Experience bank and the evolving agent acting as experience extractor. \\

        $\tau, N_{\text{roll}}, \epsilon_{\text{anch}}$
        & Source task, number of parallel attempts, and outcome reward threshold for selecting a positive anchor. \\

        $H_{\text{anchor}}^{(\tau)}, Q_{\text{subopt}}^{(\tau)}, Q_{\text{contrast}}^{(\tau)}$
        & Positive anchor, selected suboptimal executions, and their combined same-task contrast group. \\

        $m, p_m, \xi_m$
        & Experience $m=\langle p_m,\xi_m\rangle$, comprising a triggering precondition and an actionable guideline. \\

        $q, [\epsilon_{\text{lower}},\epsilon_{\text{upper}}]$
        & Source query and admissible range for its similarity to $p_m$. \\

        $\mathcal N_m, \operatorname{sim}, \epsilon_{\text{sim}}$
        & Nearby entries retrieved by precondition, semantic similarity, and novelty threshold. A novel candidate has complete-pair similarity below the threshold for every $m'\in\mathcal N_m$. \\

        \midrule

        \rowcolor{notationorange!18}
        \multicolumn{2}{c}{\textit{Paired Verification}} \\
        \midrule

        $\tilde m, \tau_{\tilde m}, N_{\text{veri}}$
        & Candidate experience, its source task, and number of parallel verification rollouts. \\

        $\bar r_{\text{exp}}^{\pm}(\tilde m)$, $\bar L_{\text{exp}}^{\pm}(\tilde m)$
        & Mean outcome rewards and execution steps with the candidate $(+)$ and under the matched control $(-)$. The control omits a new entry or retains the previous version of a revised entry. \\

        $\operatorname{Sat}(\tilde m), r_{\text{sat}}$
        & Joint saturation predicate and the outcome reward threshold required in both configurations. \\

        $\operatorname{Tie}(\tilde m), \epsilon_{\text{tie}}$
        & Comparable nonsaturated outcome rewards and their maximum absolute difference. At least one mean outcome reward is positive. \\

        $\epsilon_{\text{CNA}}$
        & Component-specific minimum outcome reward improvement required for admission outside the saturated and comparable regimes. \\

        $\epsilon_{\text{step}}^{\text{sat}}, \epsilon_{\text{step}}^{\text{tie}}$
        & Required relative reductions in mean steps in the saturated and comparable regimes. \\

        $w_m^{(0)}, w_{\text{eff}}^{\text{sat}}, w_{\text{eff}}^{\text{tie}}$
        & Initial experience utility and the positive initial utilities assigned when the corresponding step-reduction criterion is met. \\

        \bottomrule
    \end{tabularx}
\end{table}

Subsequent outcome rewards guide experience retention through the quantities in Table~\ref{tab:notation_experience_utility}.

\begin{table}[H]
    \centering
    \caption{Notation for task-relative experience utility and retention.}
    \label{tab:notation_experience_utility}
    \footnotesize
    \setlength{\tabcolsep}{3.5pt}
    \renewcommand{\arraystretch}{0.97}
    \begin{tabularx}{\linewidth}{@{}
        >{\raggedright\arraybackslash}p{0.23\linewidth}
        X
        @{}
    }
        \toprule
        \textbf{Notation} & \textbf{Definition} \\
        \midrule

        \rowcolor{notationblue!18}
        \multicolumn{2}{c}{\textit{Task-Relative Utility and Retention}} \\
        \midrule

        $t_{\text{roll}}, \mathcal T^{(t_{\text{roll}})}, \mathcal T_{\cap}^{(t_{\text{roll}})}$
        & Rollout phase, tasks evaluated in that phase, and tasks shared with the preceding phase. \\

        $\bar r_\tau^{(t_{\text{roll}})}, r_\tau^\Delta$
        & Mean outcome reward for task $\tau$ and its change from the preceding rollout phase. \\

        $\tau_m, u_{\tau_m}$
        & Task receiving experience $m$ and its reward change relative to the mean change on shared tasks, clipped to $[-1,1]$. \\

        $w_m, \alpha_m, \operatorname{clip}$
        & Experience utility, adjustment coefficient, and clipping operator. The update adds $(1-\alpha_m)u_{\tau_m}$ and clips the result to $[-1,1]$. \\

        $V_m, \lambda_V, \operatorname{Ret}(m), N_{\mathrm{exp}}$
        & Normalized historical retrieval frequency, its weight, retention score $w_m+\lambda_V V_m$, and experience-bank capacity. \\

        \bottomrule
    \end{tabularx}
\end{table}

For the functional toolset, MUSE mines recurring cross-task operation patterns and encapsulates them into callable candidates through the abstraction and screening quantities in Table~\ref{tab:notation_tool_abstraction}.

\begin{table}[H]
    \centering
    \caption{Notation for cross-task tool abstraction and encapsulation.}
    \label{tab:notation_tool_abstraction}
    \footnotesize
    \setlength{\tabcolsep}{3.5pt}
    \renewcommand{\arraystretch}{0.97}
    \begin{tabularx}{\linewidth}{@{}
        >{\raggedright\arraybackslash}p{0.23\linewidth}
        X
        @{}
    }
        \toprule
        \textbf{Notation} & \textbf{Definition} \\
        \midrule

        \rowcolor{notationblue!18}
        \multicolumn{2}{c}{\textit{Cross-Task Abstraction and Encapsulation}} \\
        \midrule

        $T_{\text{tool}}, Q_t^{\mathrm{tool}}$
        & Functional toolset and collection of trajectories examined for tool construction at round $t$. \\

        $H_i, \tau_i, \iota_i, \operatorname{OpIR}$
        & Execution trajectory, its source task, and its operation representation $\iota_i=\operatorname{OpIR}(H_i)$. \\

        $\iota_{\mathrm{dc}}, \preceq$
        & Dependency-closed operation pattern and its occurrence relation, preserving operation order and dependencies. \\

        $\operatorname{supp}_{\tau}(\iota_{\mathrm{dc}}), N_{\mathrm{cross}}$
        & Set of distinct tasks supporting the pattern and minimum number required for abstraction. \\

        $H_{\mathrm{src}}^\star, \operatorname{Encap}, \tilde f$
        & Representative successful trajectory, encapsulation operator, and proposed tool $\tilde f=\langle d_f,\iota_f,\hat w_f\rangle$. \\

        $d_f, \iota_f, \hat w_f$
        & Tool descriptor and callable interface, executable procedure, and provisional utility estimated from supporting trajectories. \\

        $\operatorname{fam}(f), \operatorname{type}(f)$
        & Operation family and functional role. The family also records the ordered core operations $\operatorname{Seq}_{\mathrm{core}}(f)$ and their effect semantics $\operatorname{Sig}_{\mathrm{sem}}(f)$. \\

        \bottomrule
    \end{tabularx}
\end{table}

Paired counterfactual rollouts, invocation-based utility tracking, and manager-level validation govern tool deployment and retirement, with the corresponding notation listed in Table~\ref{tab:notation_tool_refinement}.

\begin{table}[H]
    \centering
    \caption{Notation for counterfactual tool verification, utility tracking, and maintenance.}
    \label{tab:notation_tool_refinement}
    \footnotesize
    \setlength{\tabcolsep}{3.5pt}
    \renewcommand{\arraystretch}{0.97}
    \begin{tabularx}{\linewidth}{@{}
        >{\raggedright\arraybackslash}p{0.23\linewidth}
        X
        @{}
    }
        \toprule
        \textbf{Notation} & \textbf{Definition} \\
        \midrule

        \rowcolor{notationorange!18}
        \multicolumn{2}{c}{\textit{Tool Verification and Maintenance}} \\
        \midrule

        $\bar r_{\mathrm{tool}}^{\pm}(\tilde f)$, $\bar L_{\mathrm{tool}}^{\pm}(\tilde f)$
        & Mean outcome rewards and execution steps with or without the candidate available. Availability permits direct invocation or inspection as a structural reference. \\

        $\operatorname{gain}(\tilde f), \epsilon_{\mathrm{tool}}$
        & Difference between the two mean rewards and the tool-specific margin for outcome reward improvement or comparability. \\

        $\operatorname{Sat}(\tilde f), \operatorname{Tie}(\tilde f), \mathcal I_{\mathrm{eff}}(\tilde f)$
        & Joint saturation, comparable nonsaturated outcome rewards, and indicator of sufficient step reduction under the applicable regime. Step thresholds and efficiency utilities follow Table~\ref{tab:notation_experience_refinement}. \\

        $\operatorname{state}(\tilde f), w_f^{(0)}$
        & Initial deployment state (active, shadow, or disabled) and initial utility. An incompletely verified candidate remains in shadow with provisional utility $\hat w_f$. \\

        $f, w_f, \tau_f, u_{\tau_f}, \alpha_f$
        & Retained tool $f=\langle d_f,\iota_f,w_f\rangle$, current utility, task on which it was invoked, task-relative signal, and coefficient scaling the adjustment by $1-\alpha_f$. \\

        $V_f, \epsilon_{\mathrm{prune}}, \epsilon_{\mathrm{retr}}$
        & Historical task-level invocation count and disabling thresholds. A tool is disabled when $w_f<\epsilon_{\mathrm{prune}}$ and $V_f>\epsilon_{\mathrm{retr}}$. \\

        $\phi_{\mathrm{tool}}\in\Phi$, $\phi_{\mathrm{prop}}$, $\nu_{\mathrm{val}}(\phi)$
        & Current and proposed tool-manager configurations and mean performance on matched validation tasks. A proposal is retained if its score equals or exceeds the current score. \\

        \bottomrule
    \end{tabularx}
\end{table}

For model weights, MUSE organizes task groups and converts outcome rewards and step-level signals from $\Delta$ into standardized rewards through the quantities in Table~\ref{tab:notation_weight_refinement}.

\begin{table}[H]
    \centering
    \caption{Notation for trajectory task groups and reward integration in model weight refinement.}
    \label{tab:notation_weight_refinement}
    \footnotesize
    \setlength{\tabcolsep}{3.5pt}
    \renewcommand{\arraystretch}{0.97}
    \begin{tabularx}{\linewidth}{@{}
        >{\raggedright\arraybackslash}p{0.23\linewidth}
        X
        @{}
    }
        \toprule
        \textbf{Notation} & \textbf{Definition} \\
        \midrule

        \rowcolor{notationblue!18}
        \multicolumn{2}{c}{\textit{Task Groups and Reward Integration}} \\
        \midrule

        $\theta_{\text{model}}, H_i, Q_\tau, W'$
        & Foundation model weights, execution trajectory, initial trajectory group for task $\tau$, and its size before screening. \\

        $Q_\tau^{\mathrm{cur}}, Q_\tau^*, W, H_{\mathrm{best}}, \operatorname{pad}$
        & Curated group, aligned group, target group size, highest-scoring retained trajectory, and group-alignment padding operator. Padding copies are excluded from normalization statistics and receive zero loss masks. \\

        $r_{\text{out},i}, r_{\text{pote},i}^{(\ell)}$
        & Task-level outcome reward and raw prefix-potential signal at step $\ell$. Oracle tasks supply environment-provided outcome rewards. For synthetic tasks, $\mathcal R_{\text{syn}}$ supplies the outcome reward. The prefix signal is constructed for both sources in Appendix~\ref{subapp:notation_gating_and_supervision}. \\

        $r_{\text{fmt},i}^{(\ell)}$
        & Format penalty, zero for a valid action and negative for a detected syntax or schema violation. \\

        $\hat r_{\text{pote},i}^{(\ell)}, \hat r_{\text{fmt},i}^{(\ell)}$
        & Separately standardized potential and format components. Potential statistics cover eligible clipped signals. Format statistics cover nonzero penalties on nonpadding steps. \\

        $\lambda_{\text{pote}}, \lambda_{\text{fmt}}, \hat r_{\text{step},i}^{(\ell)}$
        & Component weights and combined step reward, $\hat r_{\text{step},i}^{(\ell)}=\lambda_{\text{pote}}\hat r_{\text{pote},i}^{(\ell)}+\lambda_{\text{fmt}}\hat r_{\text{fmt},i}^{(\ell)}$. \\

        $\varphi, L_i, L_{\text{max}}, \lambda_{\text{len}}$
        & Outcome reward transformation, executed and maximum interaction steps, and length-penalty weight. The default $\varphi(x)=x^2$ gives greater separation to scores near full completion. \\

        $r_{\text{traj},i}, \hat r_{\text{traj},i}$
        & Trajectory reward $\varphi(r_{\text{out},i})-\lambda_{\text{len}}L_i/L_{\text{max}}$ and its standardized value within the task group. \\

        $\mu_{\text{traj}}, \sigma_{\text{traj}}, \mu_{\text{pote}}, \sigma_{\text{pote}}$
        & Task-group means and population standard deviations for nonpadding trajectory rewards and for potential signals clipped to $[-1,1]$ on steps with an available estimate and trainable action tokens. \\

        $\epsilon_{\text{smooth}}$
        & Positive constant added to standard deviations during normalization. \\

        \bottomrule
    \end{tabularx}
\end{table}

These standardized rewards are assigned to generated action tokens and incorporated into the clipped GRPO objective, with the update quantities summarized in Table~\ref{tab:notation_token_advantages}.

\begin{table}[H]
    \centering
    \caption{Notation for action-token advantage estimation and GRPO parameter updates.}
    \label{tab:notation_token_advantages}
    \footnotesize
    \setlength{\tabcolsep}{3.5pt}
    \renewcommand{\arraystretch}{0.97}
    \begin{tabularx}{\linewidth}{@{}
        >{\raggedright\arraybackslash}p{0.23\linewidth}
        X
        @{}
    }
        \toprule
        \textbf{Notation} & \textbf{Definition} \\
        \midrule

        \rowcolor{notationorange!18}
        \multicolumn{2}{c}{\textit{Token Advantages and Parametric Updates}} \\
        \midrule

        $n, n_{i,\ell}, N_i$
        & Token position, terminal token position of step $\ell$, and terminal token position of trajectory $H_i$. \\

        $\tilde r_i^{(n)}, \mathbb I[\cdot]$
        & Token-position reward and indicator function. Step rewards enter at step-terminal tokens and the trajectory reward at the trajectory-terminal token. \\

        $\hat A_i^{(n)}$
        & Advantage for a generated action token in step $\ell$, given by $\hat r_{\text{traj},i}+\sum_{j=\ell}^{L_i}\hat r_{\text{step},i}^{(j)}$. \\

        $u_{i,n}, H_{i,<n}$
        & Generated action token and the preceding training context. \\

        $\pi_\theta, \pi_{\text{old}}, \pi_{\text{ref}}$
        & Current policy, frozen pre-update policy evaluated in the training context, and frozen reference policy. \\

        $\rho_i^{(n)}, \epsilon_{\text{low}}, \epsilon_{\text{high}}$
        & Current-to-previous action-token probability ratio and clipping offsets defining $[1-\epsilon_{\text{low}},1+\epsilon_{\text{high}}]$. \\

        $\hat D_{\text{KL},i}^{(n)}, \beta, \mathcal J(\theta)$
        & Sampled token-level KL penalty, reference regularization coefficient, and GRPO objective averaged over trainable action tokens. \\

        $W_{\mathrm{mb}}, N_{\mathrm{train}}, \mathbb I_{\mathrm{train}}(i,n)$
        & Number of trajectories in the mini-batch, total trainable action tokens, and token mask applied to both the policy and KL terms. \\

        \bottomrule
    \end{tabularx}
\end{table}

Beyond individual tasks, experiences supply the basis for principle distillation (Table~\ref{tab:notation_component_sync}).

\begin{table}[H]
    \centering
    \caption{Notation for distilling general principles from experiences.}
    \label{tab:notation_component_sync}
    \footnotesize
    \setlength{\tabcolsep}{3.5pt}
    \renewcommand{\arraystretch}{0.97}
    \begin{tabularx}{\linewidth}{@{}
        >{\raggedright\arraybackslash}p{0.23\linewidth}
        X
        @{}
    }
        \toprule
        \textbf{Notation} & \textbf{Definition} \\
        \midrule

        \rowcolor{notationblue!18}
        \multicolumn{2}{c}{\textit{Distilling General Principles from Experiences}} \\
        \midrule

        $\epsilon_{\text{score}}, \epsilon_{\text{freq}}, Q_m^{\mathrm{ctx}}$
        & Minimum experience utility $w_m$, minimum normalized retrieval frequency $V_m$, and observed contexts in which $m$ was delivered. Distillation requires a nonempty context collection. \\

        $A_{\text{prin}}, \chi_m$
        & Evolving agent acting as principle abstractor and the candidate principle distilled from $(p_m,\xi_m,Q_m^{\mathrm{ctx}})$. \\

        $\chi_m^{\mathrm{trg}}, \chi_m^{\mathrm{op}}, \chi_m^{\mathrm{grd}}, \chi_m^{\mathrm{dir}}$
        & Structural trigger, operational pattern, execution constraints, and directive. The retrieval representation $\chi_m^{\mathrm{ret}}$ concatenates the trigger and directive via $\operatorname{Concat}$. \\

        $\epsilon_{\text{sys}}$
        & Maximum permitted similarity between the proposed directive and an existing directive in $P_{\text{sys}}$. \\

        $\mathcal T_{\mathrm{app}}(\chi_m)$, $\mathcal T_{\mathrm{eval}}(\chi_m)$, $\mathcal T_{\mathrm{pos}}(\chi_m)$
        & Applicable tasks outside the source scenario, tasks with valid paired evaluations, and evaluated tasks on which the candidate improves mean outcome reward. \\

        $\bar r_{\mathrm{prin},\tau}^{\pm}(\chi_m)$, $\bar r_{\mathrm{prin}}^{\pm}(\chi_m)$
        & Mean outcome rewards with or without the principle on task $\tau$, and their unweighted means across evaluated tasks. \\

        $\mathcal I_{\mathrm{ver}}(\chi_m)$, $\mathcal I_{\mathrm{accept}}(\chi_m)$, $N_{\mathrm{pos}}^{\min}$
        & Verification-validity and admission indicators, and the minimum number of improved tasks. Admission also requires a valid comparison exceeding the principle-specific mean outcome reward margin $\epsilon_{\mathrm{CNA}}$. \\

        $w_{\chi_m}^{(0)}, w_{\chi_m}, N_{\mathrm{sys}}$
        & Initial principle utility, current utility, and principle-pool capacity. Initial utility divides the mean improvement by the remaining headroom $1-\bar r_{\mathrm{prin}}^{-}(\chi_m)$. \\

        \bottomrule
    \end{tabularx}
\end{table}

During task execution, MUSE retrieves, selects, and arbitrates heterogeneous symbolic assets within bounded context limits through the coordination and filtering quantities in Table~\ref{tab:notation_symbolic_integration}.

\begin{table}[H]
    \centering
    \caption{Notation for source-aware retrieval, marginal relevance selection, and consistency arbitration.}
    \label{tab:notation_symbolic_integration}
    \footnotesize
    \setlength{\tabcolsep}{3.5pt}
    \renewcommand{\arraystretch}{0.97}
    \begin{tabularx}{\linewidth}{@{}
        >{\raggedright\arraybackslash}p{0.23\linewidth}
        X
        @{}
    }
        \toprule
        \textbf{Notation} & \textbf{Definition} \\
        \midrule

        \rowcolor{notationorange!18}
        \multicolumn{2}{c}{\textit{Conflict-Aware Symbolic Integration}} \\
        \midrule

        $q, c\in\{P,K,T\}, \mathcal X_c^{\mathrm{all}}$
        & Task query, source (principles, experiences, or tools), and full inventory of that source. \\

        $\operatorname{Elig}_c(x,q), \mathcal X_c(q)$
        & Eligibility of candidate $x$ for the query and the resulting source-specific candidate set. \\

        $\Psi_{\mathrm{sel}}, \mathcal X_{\mathrm{feas}}(\Psi_{\mathrm{sel}}), N_c^{\max}$
        & Entries selected within a delivery layer, remaining feasible candidates, and per-source selection limit. Principles have a dedicated budget. Experiences and tools share a budget. \\

        $\operatorname{score}_q(a)$, $\operatorname{red}(a,\Psi_{\mathrm{sel}})$
        & Query-conditioned ranking score and maximum similarity to a previously selected entry. Redundancy is zero for an empty selection. \\

        $\lambda_{\mathrm{mmr}}, a^*, \Psi_{\mathrm{int}}$
        & Relevance weight in maximum marginal relevance selection, next selected candidate, and union of the selections from the delivery layers. \\

        $\mathcal X_{\mathrm{dec}}^{\mathrm{int}}$, $\operatorname{link}_{\mathrm{src}}(a,b)$
        & Selected declarative entries and indicator of a direct link between a principle and its source experience. Contradiction checks compare distinct, unlinked entries. \\

        $\operatorname{scope}(a), \operatorname{claim}(a)$
        & Applicability conditions and substantive recommendation or constraint of a declarative entry. \\

        $\operatorname{sim}_{\mathrm{scope}}(a,b), \epsilon_{\mathrm{scope}}$
        & Scope similarity and the minimum overlap required for a contradiction check. \\

        $\operatorname{NLI}_{\mathrm{con}}(\cdot,\cdot), \operatorname{con}_{\mathrm{bi}}(a,b)$
        & Directional contradiction score and its maximum over the two orderings of the candidate claims. \\

        $\epsilon_{\mathrm{NLI}}, \mathcal I_{\text{conflict}}(a,b)$
        & Contradiction threshold and conflict indicator for an eligible pair meeting both the scope and contradiction thresholds. \\

        ${\operatorname{val}(a\mid q)}$, $\operatorname{rel}(a,q), \operatorname{qual}(a), \lambda_{\mathrm{qual}}$
        & Value for resolving competing guidance, $\operatorname{rel}(a,q)+\lambda_{\mathrm{qual}}\operatorname{qual}(a)$, combining query relevance and weighted normalized empirical utility. \\

        $\epsilon_{\mathrm{dedup}}, a^{-}$
        & Similarity threshold for redundant principle and experience entries serving comparable retrieval roles, and the entry removed. Equal values favor the principle. \\

        \bottomrule
    \end{tabularx}
\end{table}

Following model weight updates, closed-book task-proficiency checks guide utility decay and retention priority adjustments for associated experiences and principles (Table~\ref{tab:notation_symbolic_integration_and_retirement}).

\begin{table}[H]
    \centering
    \caption{Notation for parametric internalization checks and symbolic entry retirement.}
    \label{tab:notation_symbolic_integration_and_retirement}
    \footnotesize
    \setlength{\tabcolsep}{3.5pt}
    \renewcommand{\arraystretch}{0.97}
    \begin{tabularx}{\linewidth}{@{}
        >{\raggedright\arraybackslash}p{0.23\linewidth}
        X
        @{}
    }
        \toprule
        \textbf{Notation} & \textbf{Definition} \\
        \midrule

        \rowcolor{notationblue!18}
        \multicolumn{2}{c}{\textit{Parametric Internalization and Symbolic Entry Retirement}} \\
        \midrule

        $N_{\text{int}}, c_j^{(\tau)}, \bar c_\tau$
        & Number of verification rollouts, binary task completion in trial $j$, and mean completion without stored symbolic support. Environment-provided operations remain available. \\

        $\epsilon_{\text{mast}}$
        & Task-proficiency threshold. When $\bar c_\tau\geq\epsilon_{\text{mast}}$, the utilities of experiences and principles associated with $\tau$ are reassessed. \\

        $\alpha_{\mathrm{dec}}^{\mathrm{exp}}, \alpha_{\mathrm{dec}}^{\mathrm{prin}}, \zeta$
        & Decay factors for experience and principle utility, and additional experience decrement. Proportional decay applies to the positive part of $w_m$ and to the current principle utility $w_{\chi_m}$. \\

        \bottomrule
    \end{tabularx}
\end{table}

Historical trajectories complement current rollouts in model weight refinement. Table~\ref{tab:notation_trajectory_replay} distinguishes replay selection from task-group alignment, following Section~\ref{subsec:evolutionary_objective_and_boundary} and Appendix~\ref{subapp:evolutionary_objective_and_boundary}.

\begin{table}[H]
    \centering
    \caption{Notation for the evolutionary objective, replay selection, and task-group alignment.}
    \label{tab:notation_trajectory_replay}
    \footnotesize
    \setlength{\tabcolsep}{3.5pt}
    \renewcommand{\arraystretch}{0.97}
    \begin{tabularx}{\linewidth}{@{}
        >{\raggedright\arraybackslash}p{0.23\linewidth}
        X
        @{}
    }
        \toprule
        \textbf{Notation} & \textbf{Definition} \\
        \midrule

        \rowcolor{notationblue!18}
        \multicolumn{2}{c}{\textit{Evolutionary Objective and Replay Selection}} \\
        \midrule

        $\mathcal O,\mathcal B$
        & Overall task-performance objective and boundary constraint. Historical replay supports the capability-retention aspect of $\mathcal B$ during parametric updates. \\

        $Q_{\text{buffer}}, \tau, W_{\text{group}}$
        & Historical replay buffer, source task, and maximum number of eligible trajectories stored per task before global capacity limits are applied. \\

        $Q_{\text{new}}, Q_{\text{replay}}, Q_{\text{mix}}$
        & Newly collected trajectories, selected historical trajectories, and mixed collection $Q_{\text{mix}}=Q_{\text{new}}\cup Q_{\text{replay}}$. Revisited tasks may receive supplementary executions. Replay-only groups require at least two eligible trajectories with sufficient outcome reward contrast to contribute training loss. \\

        $N_{\text{new}}^{\mathrm{nom}}, \lambda_{\text{replay}}$
        & Nominal rollout allocation for the current task set and replay-volume coefficient. Before group alignment, $|Q_{\text{replay}}|\leq\lfloor\lambda_{\text{replay}}N_{\text{new}}^{\mathrm{nom}}\rfloor$. \\

        \midrule

        \rowcolor{notationorange!18}
        \multicolumn{2}{c}{\textit{Task-Group Alignment for Parametric Updates}} \\
        \midrule

        $Q_{\text{mix}}^{(\tau)}, W$
        & Mixed trajectories for task $\tau$ and target optimization group size. $W_{\text{group}}$ governs buffer storage, whereas $W$ governs the group supplied to the optimizer. \\

        $\operatorname{prune}(\cdot),\operatorname{pad}(\cdot)$
        & Group-alignment operators. Oversized groups retain high-scoring trajectories and, when possible, a contrasting low-scoring trajectory. Undersized groups are padded with copies of the best retained execution $H_{\text{best}}^{(\tau)}$. \\

        $Q_{\text{opt}}^{(\tau)}, \mathcal T_{\text{ret}}, Q_{\text{opt}}$
        & Aligned group for task $\tau$, set of tasks with retained groups, and final batch comprising those groups. Padding copies preserve group size and are excluded from normalization statistics and from both loss terms. \\

        \bottomrule
    \end{tabularx}
\end{table}

\subsection{Evaluation Measures}
\label{subapp:notation_evaluation_measures}

The criteria in Section~\ref{subsec:evaluation_metrics} connect capability changes with resource demands and boundary compliance. Table~\ref{tab:notation_evaluation_criteria} identifies the corresponding quantities, with detailed calculations in Appendix~\ref{subapp:metrics}.

\begin{table}[H]
    \centering
    \caption{Evaluation criteria and the associated quantities and comparisons.}
    \label{tab:notation_evaluation_criteria}
    \footnotesize
    \setlength{\tabcolsep}{3.5pt}
    \renewcommand{\arraystretch}{0.97}
    \begin{tabularx}{\linewidth}{@{}
        >{\raggedright\arraybackslash}p{0.23\linewidth}
        X
        @{}
    }
        \toprule
        \textbf{Notation} & \textbf{Definition} \\
        \midrule

        \rowcolor{notationblue!18}
        \multicolumn{2}{c}{\textit{Adaptivity and Evolutionary Efficiency}} \\
        \midrule

        $\mathrm{EvoGain}_t$
        & Change in the evolutionary objective $\mathcal{O}$ from $A_t$ to $A_{t+1}$, evaluated on the same tasks in $E_t$ under the same protocol. \\

        $\mathrm{CumEvoGain}_T$
        & Net change in $\mathcal{O}$ from the initial agent $A_0$ to the agent $A_T$ after $T$ rounds, with both evaluated in the fixed environment $E_0$. \\

        $N_{\mathrm{roll}}^{(t)}$
        & Number of rollout trajectories consumed in evolution round $t$. Evolutionary efficiency relates the resources consumed to the performance attained. \\

        \midrule

        \rowcolor{notationorange!18}
        \multicolumn{2}{c}{\textit{Component, Supervision, and Gating Comparisons}} \\
        \midrule

        Component Contribution and Coordination Effectiveness
        & Performance comparisons across component selections assess their contributions. Coordination can be assessed by comparing coordinated and independent updates to the same components. \\

        Supervision Effectiveness
        & Downstream performance under different supervision sources, together with evidence of supervision validity. Validation yield is the proportion of generated candidates passing environmental validation. \\

        Gating Effectiveness
        & Task performance and resource consumption under different gating mechanisms. Interaction and update records connect capability diagnosis to evolution activation and component selection. \\

        \midrule

        \rowcolor{notationblue!18}
        \multicolumn{2}{c}{\textit{Generalizability, Retention, and Boundary Compliance}} \\
        \midrule

        Generalizability
        & Transfer of an evolved agent to unseen tasks or environments, or transfer of the evolutionary strategy $\mathcal{S}$ to a new environment. \\

        Retention and Boundary Compliance
        & Performance on previously learned tasks under a consistent evaluation protocol, and compliance with the operational constraints specified by $\mathcal{B}$. \\

        \bottomrule
    \end{tabularx}
\end{table}

Peak performance and performance across evolution rounds capture different aspects of adaptation. Table~\ref{tab:notation_evolution_performance_cost} distinguishes Best SR, Final SR, and Evo-AUC under the same pre-evolution reference.

\begin{table}[H]
    \centering
    \caption{Notation for evolutionary performance and stability relative to the round~0 Base Model at each scale.}
    \label{tab:notation_evolution_performance_cost}
    \footnotesize
    \setlength{\tabcolsep}{3.5pt}
    \renewcommand{\arraystretch}{0.97}
    \begin{tabularx}{\linewidth}{@{}
        >{\raggedright\arraybackslash}p{0.23\linewidth}
        X
        @{}
    }
        \toprule
        \textbf{Notation} & \textbf{Definition} \\
        \midrule

        \rowcolor{notationblue!18}
        \multicolumn{2}{c}{\textit{Evolutionary Performance and Stability}} \\
        \midrule

        SR ($\uparrow$)
        & Task success rate, reported as the percentage of evaluated tasks completed successfully by an agent instance. \\

        Best SR ($\uparrow$)
        & Highest SR within the observation horizon, including the shared Base Model reference at round~0. \\

        Final SR ($\uparrow$)
        & SR of the agent instance evaluated at the last reported evolution round. \\

        EvoGain, $\Delta\mathrm{SR}$ ($\uparrow$)
        & Best SR minus the corresponding Base Model SR, in percentage points. Both experimental labels denote improvement at the peak. \\

        Evo-AUC ($\uparrow$)
        & Signed area between the SR trajectory and the Base Model SR, normalized by the observation horizon. Rounds below the reference contribute negatively. \\

        SR across evolution rounds
        & SR of each evaluated agent instance at its round. The earliest round attaining Best SR locates the peak. SR in later rounds shows whether performance stays near the peak or declines. \\

        \bottomrule
    \end{tabularx}
\end{table}

Performance changes also involve interaction cost. Table~\ref{tab:notation_execution_and_rollout_cost} distinguishes the steps taken during task execution from the rollouts collected for component updates in Section~\ref{sec:experiments}.

\begin{table}[H]
    \centering
    \caption{Notation for execution steps and exploration rollout cost.}
    \label{tab:notation_execution_and_rollout_cost}
    \footnotesize
    \setlength{\tabcolsep}{3.5pt}
    \renewcommand{\arraystretch}{0.97}
    \begin{tabularx}{\linewidth}{@{}
        >{\raggedright\arraybackslash}p{0.23\linewidth}
        X
        @{}
    }
        \toprule
        \textbf{Notation} & \textbf{Definition} \\
        \midrule

        \rowcolor{notationorange!18}
        \multicolumn{2}{c}{\textit{Execution Steps}} \\
        \midrule

        Avg.\ Steps ($\downarrow$)
        & Mean number of interaction steps across all evaluated tasks for the agent instance attaining Best SR. \\

        $\Delta\mathrm{Steps}$ ($\downarrow$)
        & Avg.\ Steps minus the corresponding Base Model average, in interaction steps. A negative value denotes fewer steps per evaluated task. \\

        \midrule

        \rowcolor{notationblue!18}
        \multicolumn{2}{c}{\textit{Exploration Cost and Peak Attainment under Gating}} \\
        \midrule

        Exploration Rollouts ($\downarrow$)
        & Fresh trajectories collected for symbolic refinement and model weight adaptation. Figure~\ref{fig:gating_rollouts_per_round} gives the count per round. Table~\ref{tab:gating_effectiveness} reports the total over eight rounds. \\

        Reduction ($\uparrow$)
        & Percentage decrease in total Exploration Rollouts relative to the Always-on setting over the same eight rounds. \\

        Cumulative exploration rollouts
        & Total fresh exploration rollouts through a given evaluation. In Figure~\ref{fig:gating_best_observed_sr}, points give the SR at each budget and the step line gives the highest SR observed so far. The first point attaining Best SR identifies the rollout budget at peak attainment. \\

        \bottomrule
    \end{tabularx}
\end{table}

Finally, the audits in Tables~\ref{tab:synthetic_task_quality} and~\ref{tab:synthetic_outcome_reward_reliability} assess task quality and assigned outcome rewards separately. Table~\ref{tab:notation_supervision_audits} distinguishes task-level validity from execution-level agreement with human completion judgments.

\begin{table}[H]
    \centering
    \caption{Human audit measures for synthetic task quality and outcome reward reliability.}
    \label{tab:notation_supervision_audits}
    \footnotesize
    \setlength{\tabcolsep}{3.5pt}
    \renewcommand{\arraystretch}{0.97}
    \begin{tabularx}{\linewidth}{@{}
        >{\raggedright\arraybackslash}p{0.255\linewidth}
        X
        @{}
    }
        \toprule
        \textbf{Notation} & \textbf{Definition} \\
        \midrule

        \rowcolor{notationblue!18}
        \multicolumn{2}{c}{\textit{Synthetic Task Quality}} \\
        \midrule

        Assessable / Total
        & Tasks with determinate human judgments on all three criteria, out of all audited tasks in the group. The four rates below share the same assessable subset within each group of generated candidates or admitted synthetic tasks. \\

        Goal Clarity ($\uparrow$)
        & Percentage of assessable tasks whose query specifies a clear objective and the necessary constraints. \\

        Environmental Feasibility ($\uparrow$)
        & Percentage of assessable tasks for which environmental evidence supports the operations required for completion. \\

        Reference Alignment ($\uparrow$)
        & Percentage of assessable tasks whose reference solution and reference trajectory correspond to the task objective and constraints. \\

        Overall Validity ($\uparrow$)
        & Percentage of assessable tasks passing Goal Clarity, Environmental Feasibility, and Reference Alignment together. \\

        \midrule

        \rowcolor{notationorange!18}
        \multicolumn{2}{c}{\textit{Outcome Rewards for Synthetic Tasks}} \\
        \midrule

        Assessable / Total
        & Executions with a determinate human completion judgment, out of all audited executions in the group. Guided and free executions are assessed separately. \\

        Full-completion decision
        & The evolving agent predicts full completion when $r_{\mathrm{out},i}^{\mathrm{syn}}=1$. Human judgments supply the reference labels. Complete is positive, partial or incomplete is negative, and uncertain judgments are excluded. \\

        Completion Agreement ($\uparrow$)
        & Percentage of assessable executions for which the assigned reward and human judgment agree on whether the task was fully completed. \\

        Full-Completion Precision ($\uparrow$)
        & Among executions assigned full reward, the percentage judged complete by the human reviewer. \\

        Full-Completion Recall ($\uparrow$)
        & Among executions judged complete by the human reviewer, the percentage assigned full reward. \\

        False-Positive Rate ($\downarrow$)
        & Among executions judged partial or incomplete by the human reviewer, the percentage assigned full reward. N/A denotes an undefined rate when this group is empty. \\

        \bottomrule
    \end{tabularx}
\end{table}

\clearpage
\endgroup

\section{Detailed Implementation of MUSE} \label{app:implementation_of_muse}

This appendix supplements Section~\ref{sec:methodology} with the detailed implementation of MUSE. Following the formulation of the six-component evolutionary strategy $\mathcal{S}$, it specifies the verification criteria, update procedures, and maintenance rules for the framework architecture illustrated in Figure~\ref{fig:overview_muse}.

Appendix~\ref{subapp:evolutionary_gating} details the evolutionary gating mechanism ($\Gamma$), including activation criteria, diagnostic evidence, and component routing. Appendix~\ref{subapp:supervision_generator} describes the supervision generator ($\Delta$) through task construction, outcome reward discrimination for synthetic tasks, process supervision derived from prefix potentials, and continual task-pool adaptation. Appendix~\ref{subapp:optimization_scope_and_evolutionary_optimizer} specifies the optimization scope ($\mathcal{C}$) and evolutionary optimizer ($\mathcal{M}$), detailing inner-component refinement and inter-component synchronization. Finally, Appendix~\ref{subapp:evolutionary_objective_and_boundary} describes the evolutionary objective ($\mathcal{O}$) and task-specific trajectory replay supporting the capability-retention aspect of $\mathcal{B}$.

\subsection{Evolutionary Gating \texorpdfstring{$\Gamma$}{Gamma}} \label{subapp:evolutionary_gating}

During deployment, interaction evidence accumulates continuously, while launching an evolution round incurs additional interaction and optimization costs. Consequently, evolutionary control should determine whether accumulated evidence warrants adaptation, how observed capability demands manifest, and which evolvable components should respond. MUSE therefore instantiates the evolutionary gating $\Gamma$ as a lifecycle-level metacognitive controller that organizes these decisions through evolution activation and diagnostic routing. Building on Section~\ref{subsec:evolutionary_gating}, this subsection elaborates on the evidence supporting activation, the construction of diagnostic inputs, and the interpretation of component-specific adaptation demands. The overview of the gating is shown in Figure~\ref{fig:evolutionary_gating_muse}.

\noindent $\blacktriangleright$ \textbf{Capability Gap and Accumulation Trigger.} A single interaction provides limited evidence for deciding whether an additional evolution round is warranted, while activating adaptation after every task would introduce unnecessary optimization overhead. Aggregating recent interactions provides a broader basis for assessing current task performance, with the recent window emphasizing behavior under the current agent configuration and deployment conditions. To monitor recent environmental fitness, MUSE maintains a sliding window $\mathcal{W}_t$ containing execution trajectories and task-completion records from recent deployment interactions at evolution round $t$. Let $c_i \in \{0, 1\}$ denote the generalized binary completion status of the $i$-th task execution. This status is typically instantiated via the outcome reward $r_{\text{out}, i}$ meeting a predefined success criterion $\epsilon_{\text{succ}}$, formalized as $c_i = \mathbb{I}\left[r_{\text{out}, i} \ge \epsilon_{\text{succ}}\right]$. These completion records determine the recent mean completion rate $\bar{c}_{\text{env}}$ that enters Equation~\ref{equ:evolution_activation_trigger}.

A low recent completion rate suggests that capabilities may inadequately support current task demands, motivating the performance condition $(\bar{c}_{\text{env}} < c_{\text{th}})$ to initiate diagnosis and corrective adaptation. This condition concerns the current completion level relative to the configured threshold, including situations in which performance remains consistently low across successive interactions. Continued interaction can reveal successful strategies and recurring operational patterns suitable for abstraction or consolidation, even when aggregate performance remains stable. The accumulation condition $(V_{\text{exe}} \ge V_{\text{acc}})$ provides opportunities to examine these patterns after sufficient interaction. The threshold $V_{\text{acc}}$ specifies the additional execution volume preceding an accumulation-driven activation, allowing new behavioral evidence to develop between consolidation opportunities. Its configuration therefore relates the frequency of these opportunities to the associated interaction and optimization costs. Together, the two conditions connect performance-driven correction with interaction-driven consolidation, while diagnostic routing identifies the adaptation targets supported by the observed evidence.

The activation conditions summarize whether recent performance or continued interaction motivates a new evolution round. Determining the appropriate adaptation targets requires examining the behavioral evidence behind these aggregate indicators. MUSE returns to the trajectory-level evidence accumulated in $\mathcal{W}_t$ for subsequent system-level capability diagnosis and diagnostic routing.

\noindent $\blacktriangleright$ \textbf{Metacognitive Diagnostic Routing.}
Once evolution is activated, selecting appropriate adaptation targets requires identifying the capability demands underlying recent task performance. Similar task failures can arise from qualitatively different causes, including localized knowledge gaps, recurrent behavioral inefficiencies, missing functional abstractions, or persistent reasoning limitations. Multiple capability limitations may coexist across recent interactions, while successful behavior can reveal consolidation opportunities at different levels of abstraction. MUSE therefore performs system-level capability diagnosis over recent environmental interaction evidence, connecting recurrent behavioral patterns to the forms of adaptation they suggest.

Task-completion records indicate whether recent interactions succeed, but provide limited evidence about which observed behaviors reflect capability deficiencies and which strategies remain effective. Comparing representative successful and failed interactions provides richer diagnostic context by exposing recurrent differences in execution patterns, errors, and available support. Meanwhile, continuous reliance on external oracles or human annotators for such diagnosis can be constrained in post-deployment settings. MUSE therefore draws on the self-reflection capabilities of the evolving agent, which acts as a diagnostic router $A_{\text{diag}}$.

When successful or failed interactions dominate the recent window, diagnosis may overlook contrasting evidence about effective behaviors or recurring limitations. MUSE therefore constructs a balanced diagnostic subset $\tilde{\mathcal{W}}_t \subseteq \mathcal{W}_t$, balancing successful and failed interactions when both are available within the sliding window. Successful trajectories provide behavioral references for interpreting unsuccessful attempts, while failed trajectories expose difficulties that remain unresolved during recent interactions. Balancing the diagnostic evidence gives both forms of information a role in capability assessment. The activation statistic remains computed from $\mathcal{W}_t$, while the sampled subset $\tilde{\mathcal{W}}_t$ supplies the contrasting evidence for diagnostic reasoning.

Interpreting these behavioral differences also requires considering the support available during execution. An unsuccessful trajectory may arise since relevant symbolic support was unavailable, or because the agent still failed despite the corresponding experience, principle, or tool already being exposed. For each sampled trajectory, MUSE therefore records the symbolic assets actually emitted into the context. An asset present in the persistent inventory may not have been retrieved for a particular execution, making actual exposure relevant to interpreting the observed behavior. When support was exposed, its applicability and relationship to the observed error provide additional context for assessing the adaptation demand. Combining behavioral traces with symbolic exposure helps distinguish deficiencies in available support from limitations that persist despite such support.

The diagnostic router then compares the sampled trajectories in terms of task descriptions, completion records, errors, interaction steps, symbolic exposure, and available runtime evidence to form a system-level capability diagnosis profile $\mathbf{d}_t^{\mathrm{diag}}$. This profile characterizes observed capability limitations and potential consolidation opportunities, connecting the supporting behavioral evidence and relevant deployment contexts to corresponding forms of adaptation. To connect this assessment to component updates, the router jointly produces a binary routing vector $\mathbf{v}_t$ that identifies the components relevant to the diagnosed demands, as formulated in Equation~\ref{equ:metacognitive_diagnostic_routing}.

The profile records the rationale connecting observed behavior to adaptation demands, while the routing vector identifies the corresponding optimization channels. Their joint construction connects component selection to the evidence examined during diagnosis. The resulting diagnosis profile guides component selection according to the scope of the inferred demands and the form of support provided by each component as follows:

\noindent $\bullet$ $v_K$ targets the experience bank ($K_{\text{exp}}$) when the diagnosis profile identifies gaps in situational guidance concerning environment-specific procedures or operational exceptions. These gaps motivate preserving the operational details together with their applicability conditions, providing situational guidance for later tasks with similar requirements.

\noindent $\bullet$ $v_P$ targets the general principles ($P_{\text{sys}}$) when the diagnosis profile identifies an opportunity for cross-scenario abstraction from validated experiences exposed in the sampled trajectories. A recurring effective strategy across these contexts suggests that an experience may capture operational regularities beyond its original situational description. These observations motivate distilling the recurring operational logic and its applicability conditions into candidate principles. Within $\mathcal{M}$, empirical utility and retrieval evidence determine candidate eligibility, while cross-task verification assesses the value of the resulting principles.

\noindent $\bullet$ $v_T$ targets the functional toolset ($T_{\text{tool}}$) when the diagnosis profile identifies reusable multi-step operation sequences that suggest potential callable abstractions. Such patterns motivate encapsulating shared operational structure into persistent executable tools for future task execution.

\noindent $\bullet$ $v_\theta$ targets the foundation model weights ($\theta_{\text{model}}$) when the diagnosis profile identifies observed reasoning, planning, instruction-following, or output-formatting difficulties, suggesting a need for parametric refinement. The assessment considers the relevance of symbolic support actually presented during the corresponding environmental interactions. Difficulties that recur despite relevant symbolic support provide additional evidence for refining the behavioral capabilities encoded in the foundation model.

Multiple routing dimensions can be activated when the diagnosis profile identifies co-occurring adaptation demands. For example, missing situational guidance about API conventions may coexist with persistent difficulties in planning multi-step executions. The former motivates experience refinement to preserve relevant operational details, while the latter motivates model refinement to improve planning capabilities. Joint selection of $v_K$ and $v_\theta$ therefore addresses complementary adaptation demands within the same evolution round.

Successful trajectories additionally provide evidence for consolidation. When activation is driven by task accumulation without a performance deficit, the diagnosis profile can identify opportunities to consolidate effective behaviors. Validated experiences and their observed deployment contexts may support principle abstraction, while recurring operation sequences may motivate tool construction. These opportunities connect accumulation-driven activation to component-specific adaptation even when aggregate environmental fitness remains stable.

Overall, the evolutionary gating $\Gamma$ transforms longitudinal interaction evidence into a selective evolutionary response through two levels of metacognitive control. The activation trigger identifies occasions for further adaptation, and the diagnostic router examines the capability limitations or consolidation opportunities underlying the available evidence. Accordingly, $\gamma_{\text{act}}$ determines \textit{when} evolution is warranted, $\mathbf{d}_t^{\mathrm{diag}}$ characterizes what capability limitations or consolidation opportunities motivate evolution and \textit{why}, and $\mathbf{v}_t$ specifies \textit{which} evolutionary channels should respond. The selected component subset and resulting gating output follow Equation~\ref{equ:evolutionary_gating_output}. When $\gamma_{\text{act}}=0$, diagnostic routing is deferred and the agent configuration remains unchanged.

\subsection{Supervision Generator \texorpdfstring{$\Delta$}{Delta}} \label{subapp:supervision_generator}

The evolutionary gating $\Gamma$ identifies emerging capability demands and selects the corresponding adaptation targets for a new evolution round, and the ensuing adaptation requires sufficient supervision to address these demands. Reliable oracle tasks provide valuable supervision signals when available, yet their predefined coverage can become insufficient as deployment exposes new environmental states, operational dependencies, and capability frontiers. Task-level rewards additionally provide limited guidance for intermediate decisions along long-horizon interaction trajectories. MUSE therefore instantiates the supervision generator $\Delta$ to augment available oracle supervision with supplementary tasks and reward signals at task and step levels. Formally defined as $\Delta: \mathcal{A} \times \mathcal{E}\times \mathcal{H} \rightarrow \mathcal{D}$, the generator derives interaction-grounded tasks and outcome evaluation criteria, together with process supervision from visible interaction prefixes, supporting continued adaptation alongside available oracle signals.

Effective supplementary supervision further requires the synthesized objectives to remain connected to executable environmental behaviors, exhibit learning value relative to the current agent capability, and admit grounded outcome assessment throughout evolution. Process supervision additionally requires intermediate assessments grounded in the evidence available at each interaction step. Accordingly, MUSE constructs a supervision generation pipeline comprising interaction-grounded task synthesis, interaction-grounded outcome reward discrimination for synthetic tasks, prefix-potential process supervision, and continual task-pool adaptation. These implementation blocks expand Figure~\ref{fig:supervision_generator_muse}, with task-level reward discrimination and prefix-based process assessment detailed below.

\noindent $\blacktriangleright$ \textbf{Interaction-Grounded Task Synthesis.} Supplementary tasks can extend supervision to environmental situations beyond the predefined task collection. Their value depends on whether the proposed objectives correspond to executable operations and present suitable challenges for the current agent. Automated task generation~\citep{zhou2025selfchallenging,zhai2025agentevolver} and self-generated feedback~\citep{xia2025agent0,huang2025rzero,madaan2023self,chen2025spc} provide relevant foundations, while errors in synthesized content motivate additional examination against environmental evidence~\citep{he2026how}. MUSE therefore adopts curiosity-driven environmental exploration to construct task queries and reference solutions, followed by execution-based assessment of feasibility and relevance to current capabilities. The resulting task synthesis process forms a pipeline that integrates environment-guided exploration, evidence-grounded task proposal, canonical task construction, and learnability-frontier verification.

\noindent \textbf{Environment-Guided Exploration.} Task synthesis begins by exposing feasible behaviors and state transitions supported by the environment, providing an empirical substrate for subsequent objective construction. Let $P_{\text{env}}$ denote the specific environment profile characterizing the available functional entities, application domains, and operational capabilities. During this exploratory phase, the evolving agent interacts with the environment $E$, without receiving a predefined target task query or an intended solution. Instead, MUSE supplies lightweight coverage anchors that encourage interactions across diverse domains and operational relations without prescribing a concrete objective. The exploratory trajectories are mathematically represented as:
\begin{equation}
    Q_{\text{explore}} = \left\{ H_i^{\text{explore}} \right\}_{i=1}^{N_{\text{explore}}}, \qquad Q_{\text{explore}}^{+} \subseteq Q_{\text{explore}}.
\end{equation}
Here, $Q_{\text{explore}}^{+}$ denotes the subset of exploratory trajectories that contain at least one successful operational workflow. These workflows are identified from interactions yielding valid environmental observations without execution failures. The coverage anchors encourage exploration across environmental functions and operational relations, while candidate task objectives are subsequently abstracted from the observed workflows and their effects.

\noindent \textbf{Evidence-Grounded Task Proposal.} A successful exploratory workflow demonstrates environmental feasibility, while its transformation into reusable supervision further requires the implicit behavioral objective to be abstracted from exploratory interactions. The evolving agent subsequently serves as a task summarizer $A_{\text{sum}}$ and analyzes the successful behavioral evidence in $Q_{\text{explore}}^{+}$. Each coherent workflow is abstracted into a provisional task query $q_i^{(0)}$ and a provisional solution $y_i^{(0)}$, accompanied by the interaction provenance and dependency relations extracted from its source trajectory. Historical task objectives may serve as diversity references, yet the interaction trajectory remains the primary evidence supporting the proposed task. The candidate collection is formulated as:
\begin{equation}
    \mathcal{T}_{\text{cand}}^{(t)} = A_{\text{sum}}(Q_{\text{explore}}^{+}, P_{\text{env}}) = \left\{ \tau_i^{(0)} \mid \tau_i^{(0)} = \left\langle q_i^{(0)}, y_i^{(0)} \right\rangle, \, 1 \le i \le N_{\text{cand}} \right\}.
\end{equation}
The provisional solution $y_i^{(0)}$ summarizes the explored workflow, serving as an intermediate hypothesis for subsequent environmental examination. Exploratory interactions can yield redundant or weakly operational objectives, motivating an initial refinement of the candidate collection. A filtering mechanism then removes near-duplicate queries, documentation-oriented objectives, and tasks with limited operational content. This yields a candidate reservoir supported by meaningful interactions and observable environmental changes.

The remaining candidates compete for a bounded validation budget. MUSE therefore applies a soft candidate preselection strategy based on structural complexity, cross-application dependencies, less frequent domain coverage, and a query-only estimate of independent task difficulty. This estimate serves as a ranking metric for empirical validation. This prioritization directs the validation budget toward structurally informative candidates and sustains diversity across functional domains, dependency patterns, and source-task families.

\noindent \textbf{Grounded Canonical Task Construction.} The provisional task derived from successful exploration remains an intermediate hypothesis whose summarized objective requires correspondence with  reproducible environmental execution. MUSE therefore assesses the empirical feasibility of each retained candidate $\tau_i^{(0)}$ through solution-guided validation. The provisional solution $y_i^{(0)}$ assists the evolving agent in locating a feasible execution path, while a query-replay assessment examines whether the resulting actions correspond to the intended user objective. The interaction trajectory recorded during this successful validation serves directly as the reference trajectory $H_{i,\text{ref}}^{\star}$, which is subsequently applied to reconstruct a grounded reference solution $y_i^{\star}$ that captures the essential operational steps, state transitions, and data dependencies required for task completion while filtering out exploratory noise.

With empirical feasibility established, the subsequent concern lies in separating the user-facing objective from execution-specific information inherited during exploration. A notable vulnerability in self-synthesized tasks involves \textit{procedural leakage}, where the provisional task query $q_i^{(0)}$ inadvertently exposes execution details or runtime-specific identifiers. To separate the user intent from the underlying operational logic, MUSE abstracts a compact semantic contract from the provisional query $q_i^{(0)}$, the reconstructed solution $y_i^{\star}$, and the successful replay evidence. Operating as a structural boundary, this contract preserves the core objectives, user constraints, and dynamic field types, while excluding transient observations, system-level setup commands, and procedural solution steps. Subsequently, the evolving agent serves as an isolated goal constructor to map this contract into a canonical user query $q_i^{\star}$ without directly observing the original execution trace. This isolation reduces the tendency to generate queries that presuppose dynamic states or introduce unjustified operational bindings. The resulting canonical task follows Equation~\ref{equ:canonical_synthetic_task}, associating the canonical query $q_i^{\star}$ with the empirically verified solution $y_i^{\star}$ and supporting reference trajectory $H_{i,\text{ref}}^{\star}$. A stable identifier derived from $q_i^{\star}$ associates later pool membership, training records, and mastery observations with the same semantic task objective.

Following canonicalization, semantic consistency between the reconstructed query and its verified execution is further examined before frontier evaluation. Each canonical task undergoes a task-contract assessment examining the alignment between the canonical query and the projected execution logic. It systematically filters candidates for missing query goals, unauthorized state mutations, presupposed dynamic states, or residual leaked solution steps. The best successful canonical execution may further support a closure step over $q_i^{\star}$ and $y_i^{\star}$, keeping the final task representation connected to its empirical execution evidence.

\noindent \textbf{Learnability-Frontier Verification.} Structural validity and environmental feasibility establish the empirical validity of a synthesized task, while its supervisory value further depends on the capability state of the current agent. Unsuccessful guided rollouts suggest that reliable execution remains difficult for the current agent, while consistently successful free rollouts indicate limited additional challenge. MUSE therefore assesses the suitability of each canonical task for supplementary supervision through paired guided and free rollouts, with the guidance gap serving as a proxy for task learnability of the current agent. Guided rollouts expose $y_i^{\star}$ and $q_i^{\star}$ to the evolving agent to assess execution reliability with reference guidance, while free query-only rollouts present $q_i^{\star}$ without  $y_i^{\star}$ or $H_{i,\text{ref}}^{\star}$ to estimate the current independent execution ability of the evolving agent.

Let $N_{\mathrm{g}}$ and $N_{\mathrm{f}}$ denote the numbers of guided and free rollouts, respectively. For task $\tau_i^{\star}$, $r_{i,j}^{\mathrm{g}}\in[0,1]$ and $r_{i,j}^{\mathrm{f}}\in[0,1]$ denote the outcome rewards assigned to the $j$-th guided and free rollouts by the interaction-grounded discriminator for synthetic tasks. MUSE summarizes these outcome rewards by their means and maxima:

\begin{equation}
\begin{aligned}
    \bar{r}_i^{\mathrm{g}} &= \frac{1}{N_{\mathrm{g}}} \sum_{j=1}^{N_{\mathrm{g}}} r_{i,j}^{\mathrm{g}}, & r_{i,\max}^{\mathrm{g}} &= \max_{1\le j\le N_{\mathrm{g}}} r_{i,j}^{\mathrm{g}},
    \\
    \bar{r}_i^{\mathrm{f}} &= \frac{1}{N_{\mathrm{f}}} \sum_{j=1}^{N_{\mathrm{f}}} r_{i,j}^{\mathrm{f}}, & r_{i,\max}^{\mathrm{f}} &= \max_{1\le j\le N_{\mathrm{f}}} r_{i,j}^{\mathrm{f}}.
\end{aligned}
\end{equation}
The guidance gap in Equation~\ref{equ:guidance_gap} measures the empirical contribution of the grounded solution relative to query-only execution. A positive $\operatorname{gap}_i$ indicates that the grounded solution provides measurable assistance beyond the current independent capability of the evolving agent. Guided performance reflects execution with validated reference support, whereas free performance characterizes the current independent capability. Their contrast provides an empirical criterion for selecting tasks whose execution benefits from reference guidance. Let $\hat{p}_i^{\mathrm{g}}$ and $\hat{p}_i^{\mathrm{f}}$ denote the fractions of guided and free executions that are valid and judged successful by the discriminator, respectively. Frontier admission requires all expected guided and free executions to be recorded. Based on these statistics, MUSE categorizes each canonical task into a regular frontier, a challenge frontier, or a rejected region:
\begin{equation}
    \operatorname{type}(\tau_i^{\star}) =
    \begin{cases}
        \mathrm{regular},
        & \begin{aligned}[t]
            & \hat{p}_i^{\mathrm{g}}
              \geq\epsilon_{\mathrm{g,succ}},\quad
              \hat{p}_i^{\mathrm{f}}\in
              [\epsilon_{\mathrm{f,low}},\epsilon_{\mathrm{f,high}}],\\
            & \operatorname{gap}_i\in
              [\epsilon_{\mathrm{gap}},\epsilon_{\mathrm{gap,max}}],
          \end{aligned}
          \\[6pt]
        \mathrm{challenge},
        & \begin{aligned}[t]
            & \hat{p}_i^{\mathrm{g}}
              \geq\epsilon_{\mathrm{g,succ}},\quad
              \hat{p}_i^{\mathrm{f}}=0,\quad
              r_{i,\max}^{\mathrm{f}}
              \geq\epsilon_{\mathrm{f,partial}},\\
            & \operatorname{gap}_i\in
              [\epsilon_{\mathrm{gap,ch}},\epsilon_{\mathrm{gap,max}}],
          \end{aligned}
          \\[6pt]
        \mathrm{rejected},
        & \text{otherwise}.
    \end{cases}
\end{equation}
Here, $\epsilon_{\mathrm{g,succ}}$ specifies the minimum guided success rate, while $[\epsilon_{\mathrm{f,low}},\epsilon_{\mathrm{f,high}}]$ bounds the free success rate for regular-frontier tasks. Challenge-frontier tasks require no successful free execution and a peak free outcome reward reaching $\epsilon_{\mathrm{f,partial}}$. The thresholds $\epsilon_{\mathrm{gap}}$ and $\epsilon_{\mathrm{gap,ch}}$ specify the minimum guidance gaps for the two categories, with $\epsilon_{\mathrm{gap,max}}$ providing a common upper bound. Their experimental values are reported in Table~\ref{tab:muse_component_supervision_settings}.

Regular-frontier tasks exhibit partial yet unsaturated query-only performance alongside reliable grounded guidance. Challenge-frontier tasks reflect scenarios where no free execution is judged successful, yet at least one free rollout demonstrates partial progress ($r_{i,\max}^{\mathrm{f}}\geq\epsilon_{\mathrm{f,partial}}$) and the guided executions confirm a reliable solution path. Tasks outside these regions fail the configured criteria for guided feasibility, independent performance, or guidance benefit and are excluded from frontier admission. Only canonical tasks that pass the structural assessments and are classified as either regular or challenge frontiers are admitted to the validated task set $\mathcal{T}_{\text{front}}^{(t)}$. The resulting set provides supplementary objectives complementing the oracle corpus, with environmentally supported objectives selected for their empirical guidance benefit and remaining challenge relative to current capabilities.

\noindent $\blacktriangleright$ \textbf{Interaction-Grounded Outcome Reward Discrimination for Synthetic Tasks.} The validated supplementary tasks specify new learning objectives, while subsequent optimization additionally requires outcome feedback for newly sampled executions. Existing evolutionary frameworks commonly derive such feedback through external or semantically generated evaluation signals. Examples include majority-voting consensus~\citep{xia2025agent0, huang2025rzero}, evaluation delegated to teacher models~\citep{zhai2025agentevolver}, and generated semantic evaluation functions~\citep{zhou2025selfchallenging}. These approaches offer viable supervision, although sustained evolution may encounter settings where external evaluators are limited. MUSE therefore derives outcome rewards for synthetic tasks from the grounded task execution evidence established during environmental validation and performs the corresponding assessment intrinsically with the evolving agent itself.

During free candidate evaluation and subsequent query-only rollouts, the evolving agent receives only the canonical query $q_i^{\star}$ and interacts with the environment to generate a new execution trajectory $H_{i,\text{exe}}$. Neither the grounded solution $y_i^{\star}$ nor the successful reference trajectory $H_{i,\text{ref}}^{\star}$ is exposed during these query-only executions. Guided candidate evaluation additionally presents $y_i^{\star}$, as specified in the frontier verification procedure. The same discriminator evaluates both rollout conditions, allowing their rewards to reflect a common task objective.

After the rollout is completed, the agent serves as an intrinsic discriminator $A_{\text{disc}}$ and assesses the observed actions and environmental feedback within $H_{i,\text{exe}}$ against the canonical query $q_i^{\star}$, with $y_i^{\star}$ and $H_{i,\text{ref}}^{\star}$ provided as reference. The canonical query specifies the required goals and constraints, while the reference evidence connects their interpretation to observed environmental operations and effects. Assessment therefore concerns whether the new execution satisfies the objective, including executions that follow a different valid operational path.

The resulting outcome reward $r_{\text{out},i}^{\text{syn}}\in[0,1]$ follows Equation~\ref{equ:synthetic_reward} and estimates the degree of task completion. These rewards support guided and free candidate evaluation, subsequent optimization, and query-only mastery verification. Here, $\mathcal{R}_{\text{syn}}(\cdot)$ denotes the interaction-grounded outcome reward function for synthetic tasks, while predefined oracle tasks retain their inherent objective metrics supplied by the environment.

\noindent $\blacktriangleright$ \textbf{Prefix-Potential Process Supervision.} 
Outcome rewards assess overall task completion, while long-horizon execution additionally requires fine-grained feedback about intermediate decisions. Executions with similar final rewards can contain different patterns of progress, making the completed-task score insufficient to characterize each preceding transition. This motivates supplementary assessments based on the evidence available at successive interaction steps. Existing frameworks introduce step-level attribution to complement task-level feedback~\citep{zhai2025agentevolver, wang2025spa}, and recent studies further examine process assessments derived from the policy model itself~\citep{fei2025self, hu2026self, hu2026optimizing}. MUSE accordingly obtains prefix-based process supervision from the evolving agent for both oracle and synthetic tasks.

\noindent \textbf{Prefix-Conditioned Assessment.}
The significance of an intermediate action depends on the preceding execution state and the observation it produces. Assessing progress therefore requires interpreting the latest interaction within the task objective and its execution context. Specifically, MUSE inspects the visible trajectory prefixes and elicits a corresponding assessment of task-completion prospects from the evolving agent. The evaluation begins with the query-only prefix $H_i^{(0)}=(q_i)$, producing an initial potential $\tilde{\omega}_i^{(0)}$ that reflects the estimated solvability of the task before any interaction. This initial assessment provides the reference point for evaluating the first interaction step.

For each subsequent step $\ell$, the evaluator forms $H_i^{(\ell)}$ by appending the executed action $u_{i,\ell}$ and any returned observation $o_{i,\ell+1}$ to $H_i^{(\ell-1)}$. Future actions and observations, reference solutions, and separately computed outcome rewards remain outside the assessment context. The same restriction applies when prefixes are assessed after rollout collection, keeping each score conditioned on its corresponding execution state.

\noindent \textbf{Assessment over Ordered Bins.}
A single decoded numerical value does not expose the relative support assigned to different completion levels. Retaining this support allows the potential estimate to reflect a distribution of assessments, including cases where several neighboring levels remain plausible. MUSE therefore evaluates the ordered bins and numerical anchors defined in Section~\ref{subsec:supervision_generator}. The same bin definitions and anchor values apply across interaction prefixes, providing a common numerical scale for successive assessments. Let $\mathrm{logit}_i^{(\ell)}(b_k)$ denote the generation logit assigned by the foundation model to the discrete potential bin $b_k$ conditioned on the visible trajectory prefix $H_i^{(\ell)}$. A temperature parameter $\kappa_{\text{eval}}>0$ controls the sharpness of the predictive distribution, with probabilities normalized over the predefined bin alternatives:
\begin{equation}
    \Pr(b_k \mid H_i^{(\ell)}) = \frac{\exp\left( \mathrm{logit}_i^{(\ell)}(b_k) / \kappa_{\text{eval}} \right)}{\sum_{j=1}^{N_{\mathrm{bin}}} \exp\left( \mathrm{logit}_i^{(\ell)}(b_j) / \kappa_{\text{eval}} \right)}.
\end{equation}
A lower temperature concentrates probability on more strongly supported bins, while a higher temperature distributes support more broadly. Equation~\ref{equ:prefix_potential} converts this distribution into the expected potential $\tilde{\omega}_i^{(\ell)}$. The expectation preserves graded assessments even though the underlying alternatives are discrete.

\noindent \textbf{Relative Progress across Adjacent Prefixes.}
A fixed increase in potential can represent different proportional changes depending on its starting value. MUSE therefore adopts the logarithmic representation in Equation~\ref{equ:prefix_potential} to express relative changes in assessed task-completion prospects. Its numerical floor $\epsilon$ keeps the transformation finite when the expected potential approaches zero. Above this floor, equal proportional changes produce equal log-potential differences, giving the resulting signal a consistent interpretation across different starting levels.

A prefix potential describes the execution state reached so far, while step-level supervision concerns the change associated with the latest interaction. Adjacent prefixes differ by the newly available action-observation pair, motivating the comparison of their corresponding potentials. MUSE therefore constructs $r_{\text{pote},i}^{(\ell)}$ through the adjacent difference in Equation~\ref{equ:potential_step_reward}. Positive and negative differences respectively indicate increased and decreased assessed completion prospects. When both expected potentials fall below the floor, their transformed difference is zero. The resulting signal associates each interaction step with the change in the model assessment of task progress.

\noindent \textbf{Submission-Boundary Treatment.}
A completion declaration can change the textual prefix without adding evidence of task fulfillment. Assessing such a declaration as a substantive execution step can consequently introduce a process-reward change that reflects termination wording. This motivates a separate treatment of explicit submission boundaries. MUSE therefore applies a neutral submission convention when the final step $L_i$ contains an explicit submission or termination marker and its observation contains no error indicator. Consistent with Equation~\ref{equ:prefix_potential}, this convention carries forward the preceding expected potential before the logarithmic transformation:
\begin{equation}
    \tilde{\omega}_i^{(L_i)}
    \leftarrow
    \tilde{\omega}_i^{(L_i-1)}.
\end{equation}
The corresponding log-potentials coincide, yielding $r_{\text{pote},i}^{(L_i)}=0$. This adjustment concerns the process signal assigned to the termination action, while the outcome reward continues to assess overall task completion.

Reaching the interaction limit has a different interpretation, since the final operational step may still change the task state or reveal evidence relevant to completion. When a trajectory reaches the maximum allowed interaction steps without an explicit termination action, the ordinary final-prefix assessment is retained. Its potential difference can therefore reflect either an increase or a decrease in assessed completion prospects.

Following the boundary convention, adjacent log-potential differences telescope over the trajectory, so their sum equals the change between the initial and final log-potentials. This property concerns the potential signal before reward weighting, format penalties, and group-relative normalization. This relation connects the accumulated process signals to the net change in assessed completion prospects over the trajectory.

The resulting $r_{\text{pote},i}^{(\ell)}$ supplies supplementary process supervision alongside the outcome rewards of oracle and synthetic tasks. The optimizer $\mathcal{M}$ combines these signals with action-format penalties and performs normalization and advantage estimation, as detailed in Appendix~\ref{subapp:foundation_model_weights}.

\noindent $\blacktriangleright$ \textbf{Continual Task-Pool Adaptation.} The supervisory value of supplementary tasks varies with the capability state of the evolving agent. Previously informative tasks can gradually become saturated as their associated capability gaps narrow, while continued deployment can expose new regions requiring supplementary supervision. MUSE therefore organizes validated frontier tasks within a capacity-bounded supervision pool that dynamically adapts to the current agent capability. For each task $\tau_i^{\star}$, MUSE computes a selection priority by discounting its guidance gap against the standard deviation of its query-only rollout rewards:
\begin{equation}
    \operatorname{prio}(\tau_i^{\star}) = \operatorname{gap}_i - \lambda_{\text{unc}} \operatorname{Std} \left( \left\{ r_{i,j}^{\mathrm{f}} \right\}_{j=1}^{N_{\mathrm{f}}} \right),
\end{equation}
where $\lambda_{\text{unc}}\geq 0$ scales the uncertainty penalty and $\operatorname{Std}$ denotes the population standard deviation of the outcome rewards from query-only rollout explorations. This priority favors tasks where guidance yields distinct empirical advantages, with uncertainty discounting when $\lambda_{\text{unc}}>0$. To facilitate structural diversity, the pool selection allocates capacity across various difficulty levels and compositional topologies.

Capability updates may shift the learning frontier and consequently alter the relevance of previously estimated task priorities. The active pool undergoes a refresh after capability updates, allowing subsequent task selection to reflect the capability of the current agent. Newly validated tasks are merged with eligible historical tasks, preserving valuable supervision without reconstructing the pool from scratch. When the combined collection exceeds the fixed capacity, retained entries are selected according to the aforementioned selection priorities and structural metrics.

Over successive evolutions, the agent masters certain frontier tasks. As these objectives become saturated, their continued retention consumes capacity that could accommodate newly emerging supervision demands. MUSE therefore tracks the empirical mastery of each active task. Let $n_{i,\mathrm{mast}}^{(t)}$ denote the number of qualifying rounds among the most recent $n_{\mathrm{master}}$ complete sampled rounds. A round is complete when all expected rollout rewards are available and finite, and qualifies when its success rate reaches $\epsilon_{\mathrm{hist,SR}}$. For mastery assessment, a rollout counts as successful when its outcome reward reaches $\epsilon_{\mathrm{syn,succ}}$. Let $\bar{r}_{i,\mathrm{hist}}^{(t)}$ denote the pooled mean outcome reward across these rounds. Once $n_{i,\mathrm{mast}}^{(t)}\geq n_{\mathrm{master}}$ and $\bar{r}_{i,\mathrm{hist}}^{(t)}\geq\epsilon_{\mathrm{hist,mean}}$, the task undergoes an isolated query-only verification to assess whether it is independently solvable without guidance. The mean outcome reward of these verification rollouts is denoted by $\bar{r}_{i,\mathrm{rev}}^{\mathrm{f}}$. Let $r_{i,\mathrm{rev},j}^{\mathrm{f}}$ denote the reward of the $j$-th verification rollout. All expected verification rewards must be available and finite. The subset of tasks for retirement at evolution round $t$ is:
\begin{equation}
    \mathcal{T}_{\text{retire}}^{(t)}
    =
    \left\{
        \tau_i^{\star}\in\mathcal{T}_{\text{syn}}^{(t)}
        \;\middle|\;
        \begin{aligned}
            & n_{i,\mathrm{mast}}^{(t)}\geq n_{\mathrm{master}},
              \quad
              \bar{r}_{i,\mathrm{hist}}^{(t)}
              \geq\epsilon_{\mathrm{hist,mean}},\\
            & \min_j r_{i,\mathrm{rev},j}^{\mathrm{f}}
              \geq\epsilon_{\mathrm{syn,succ}},
              \quad
              \bar{r}_{i,\mathrm{rev}}^{\mathrm{f}}
              \geq\epsilon_{\text{retire}}
        \end{aligned}
    \right\}.
\end{equation}
Here, $\epsilon_{\text{retire}}$ specifies the minimum mean outcome reward required during query-only mastery reverification. The historical criteria assess repeated performance, while verification requires every rollout to meet the success threshold and their mean reward to reach $\epsilon_{\text{retire}}$. The corresponding settings are reported in Table~\ref{tab:muse_component_supervision_settings}. To mitigate the resurfacing of saturated tasks during later synthesis rounds, MUSE maintains a cumulative retirement registry:
\begin{equation}
\mathcal{I}_{\text{retire}}^{(\leq t)} = \bigcup_{s=0}^{t}\left\{\operatorname{id}(\tau)\mid\tau\in\mathcal{T}_{\text{retire}}^{(s)}\right\}.
\end{equation}
This registry archives retired canonical identifiers, filtering them out from all subsequent candidate pools. Here, $\operatorname{id}(\tau)$ denotes the stable identifier derived from the canonical task query. The active supervision pool for the subsequent evolution round $t+1$ is ultimately assembled as follows:
\begin{equation}
\mathcal{T}_{\text{syn}}^{(t+1)} = \operatorname{Merge}_{N_{\text{syn}}}\left(\left\{\tau\in\mathcal{T}_{\text{syn}}^{(t)}\cup\mathcal{T}_{\text{front}}^{(t+1)}\mid\operatorname{id}(\tau)\notin\mathcal{I}_{\text{retire}}^{(\leq t)}\right\}\right).
\end{equation}
Here $N_{\text{syn}}$ limits the maximum pool size. The $\operatorname{Merge}_{N_{\text{syn}}}(\cdot)$ operator unifies the unretired historical tasks with the newly validated frontier tasks, deduplicates overlapping canonical identifiers, and maintains capacity constraints by evicting lower-priority entries. This continuous adaptation progressively retires mastered tasks, allocating the constrained supervision capacity primarily to the evolving empirical frontier.

Overall, the supervision generator $\Delta$ extends the coverage of predefined supervision through interaction-grounded supplementation while retaining reliable oracle signals within the evolutionary corpus. Task synthesis derives supplementary objectives from empirically observed environmental behaviors and evaluates their feasibility and learnability. Outcome reward discrimination for synthetic tasks associates these objectives with grounded outcome feedback, while continual task-pool adaptation refreshes their supervisory value as the agent evolves. Prefix-potential process supervision additionally provides step-level supervision for both oracle and synthetic trajectories.

Together, these mechanisms instantiate the supervision output $\delta_t$ in Equation~\ref{equ:supervision_generator_output}. Active synthetic tasks retain their validated reference evidence and associated outcome reward functions, while prefix-scoring rules provide process supervision. The resulting supervision complements available oracle supervision, with $\mathcal{M}$ performing reward integration and refinement of the components selected by $\Gamma$.

\subsection{Optimization Scope \texorpdfstring{$\mathcal{C}$}{C} and Evolutionary Optimizer \texorpdfstring{$\mathcal{M}$}{M}} \label{subapp:optimization_scope_and_evolutionary_optimizer}

\setcounter{secnumdepth}{4}

With adaptation targets selected by $\Gamma$ and supervision supplied through $\Delta$, the remaining challenge is to translate interaction evidence into updates across heterogeneous agent components. Experiences provide situational guidance, functional tools encapsulate reusable operations, and model weights support behavioral capabilities. These differences motivate component-specific adaptation procedures. Changes to one component can also alter the contribution of another, motivating reassessment of their joint contribution during task execution. MUSE therefore organizes the evolutionary optimizer $\mathcal{M}$ into inner-component refinement and inter-component synchronization.

The optimization scope $\mathcal{C}$ encompasses model weights, experiences, principles, and functional tools, as specified in Section~\ref{subsec:optimization_scope_and_evolutionary_optimizer}. Within this admissible scope, $\Gamma$ selects the components activated for the current evolution round. When $v_T=1$, toolset refinement may also adjust the associated management settings $\phi_{\mathrm{tool}}\in\Phi$. Given the selected components and available supervision $\delta_t$, updates follow Equation~\ref{equ:component_update}. The following sections explain the optimization, verification, and maintenance criteria for individual components and the coordination mechanisms that address their changing contributions during continued evolution.

\subsubsection{Inner-Component Refinement}

The components selected by $\Gamma$ encode capability in different representational forms, so their adaptation should reflect the distinct roles they play within the agent system. Situational knowledge is refined through the experience bank, reusable operational structure is consolidated into functional tools, and persistent behavioral capability is adapted within the parametric core. MUSE therefore performs component-specific refinement over $K_{\text{exp}}$, $T_{\text{tool}}$, and $\theta_{\text{model}}$ according to the available evolutionary evidence. These component-specific updates provide the experiences, tools, and model capabilities for subsequent inter-component synchronization.

\appmodule{Experience Bank \texorpdfstring{$K_{\text{exp}}$}{K_exp}}
\label{subapp:experience_bank_refinement}

\begin{figure}[!t]
    \centering
    \includegraphics[width=\linewidth]{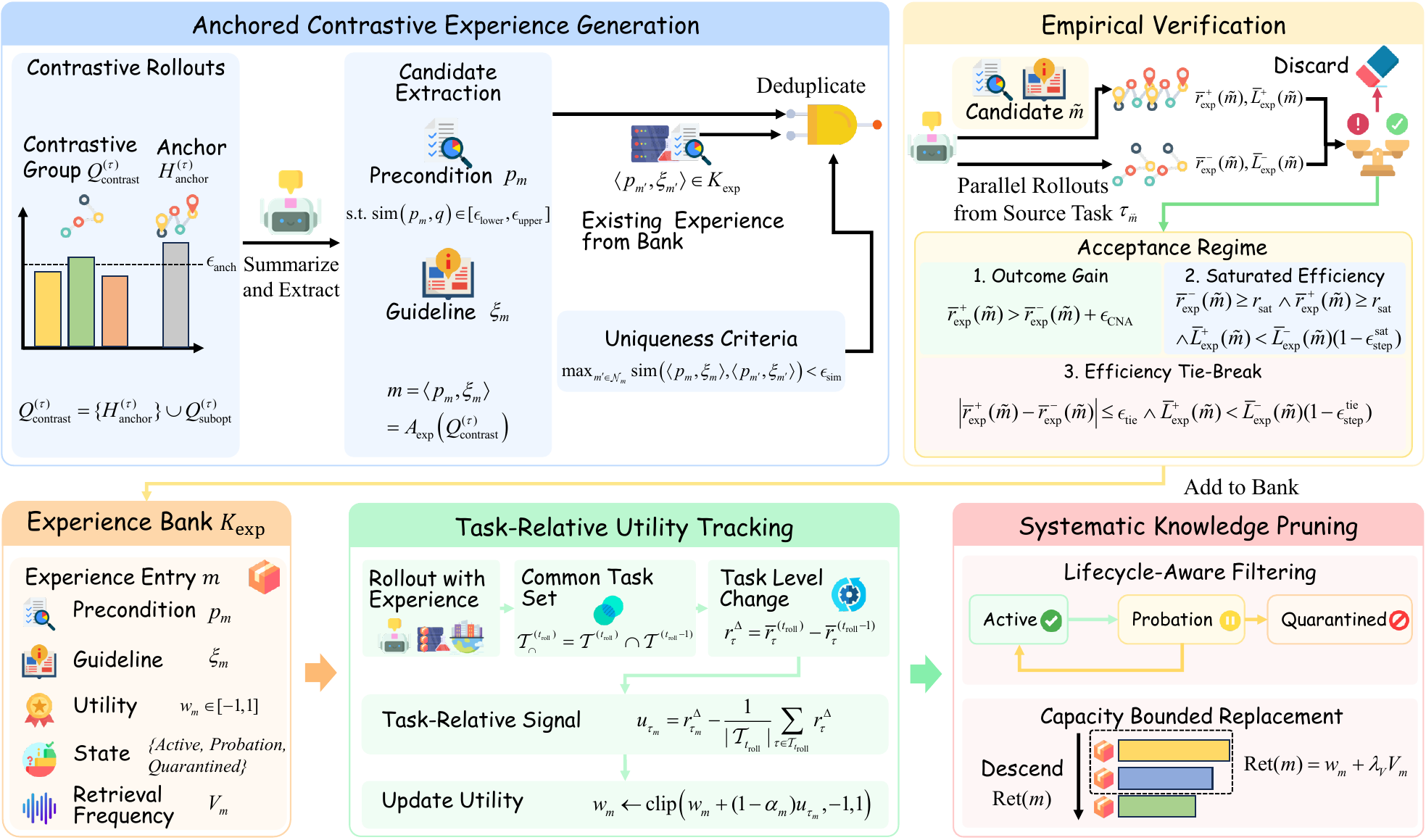}
    \caption{The refinement workflow of the experience bank ($K_{\text{exp}}$). The framework first extracts preconditions and guidelines from anchored contrastive rollouts and carefully distinguishes novel candidates from potential revisions of existing experiences. It then evaluates each candidate through controlled paired rollouts on its source task and determines admission or replacement based on empirical outcome and execution efficiency evidence. During subsequent evolutionary rollouts, performance changes relative to other tasks update the utility scores of the experiences associated with the evaluated tasks. Finally, lifecycle-aware filtering manages entries across active, probation, and quarantined states, while capacity-bounded replacement ranks retained entries by their empirical utility and retrieval frequency to maintain a concise adaptive experience bank.}
    \label{fig:optimization_framework_for_experience_bank}
\end{figure}

Interaction trajectories contain rich task-specific evidence, yet directly retaining raw trajectories~\citep{packer2023memgpt, hu2023chatdb} can lead to rapid memory growth and limited reuse when later states and actions differ from their original contexts. Recent reflection paradigms~\citep{zhai2025agentevolver} therefore abstract trajectories into compact verbalized experiences with broader applicability. The value of such distilled knowledge, however, is inherently dynamic: an experience may initially capture only a partial behavioral pattern, become increasingly useful as similar situations recur, or gradually lose relevance as the environment and internal capability of the agent evolve. Consequently, semantic relevance provides only one dimension for maintaining long-lived experiences. MUSE complements semantic retrieval with empirical evidence of behavioral contribution, constructing a dynamic experience lifecycle that connects knowledge generation, verification, utility reassessment, and retirement. Specifically, the experience-bank refinement comprises anchored contrastive experience generation, empirical verification, task-relative utility tracking, and systematic knowledge pruning, as illustrated in Figure~\ref{fig:optimization_framework_for_experience_bank}.

\noindent \textbf{Anchored Contrastive Experience Generation.} Extracting reusable knowledge from trajectories first requires separating behavioral patterns associated with favorable outcomes from incidental execution choices. Reflection over failure cases primarily exposes ineffective pathways, while an isolated successful trajectory provides limited evidence about which of its actions contribute to the observed advantage. MUSE therefore employs an anchored contrastive distillation that compares trajectories spanning different outcome reward levels within the same task.

For a task $\tau$, experience generation begins when at least one trajectory exceeds the outcome reward threshold $\epsilon_{\text{anch}}$ and can serve as a positive anchor $H_{\text{anchor}}^{(\tau)}$. To examine behavioral differences across multiple outcome reward levels, MUSE constructs a diverse contrastive trajectory group $Q_{\text{contrast}}^{(\tau)}$. This group integrates the positive anchor with a high-scoring suboptimal trajectory and additional diverse executions $Q_{\text{subopt}}^{(\tau)}$ drawn from the $N_{\text{roll}}$ parallel attempts, forming the union $Q_{\text{contrast}}^{(\tau)} = \{H_{\text{anchor}}^{(\tau)}\} \cup Q_{\text{subopt}}^{(\tau)}$. Contrasting this performance spectrum provides the experience extractor $A_{\text{exp}}$ with evidence about operational differences associated with outcome reward variation. The agent can consequently abstract useful behavioral patterns from higher-scoring trajectories together with corrective observations from lower-scoring attempts. The extractor derives the triggering precondition $p_m$ and actionable guideline $\xi_m$ according to Equation~\ref{equ:experience_extraction}. The precondition supports subsequent retrieval, while the complete precondition-guideline pair enters semantic screening and empirical verification before admission.

The extracted experience subsequently undergoes semantic screening before empirical evaluation. A source-aligned boundary check compares the original task query $q$ with the extracted precondition $p_m$ and discards candidates whose similarity falls outside the predefined range $[\epsilon_{\text{lower}},\epsilon_{\text{upper}}]$. MUSE then performs two-stage semantic deduplication. It first retrieves a shortlist $\mathcal{N}_m \subseteq K_{\text{exp}}$ of nearby entries according to $p_m$ and compares the complete candidate $\langle p_m,\xi_m\rangle$ with the shortlisted experiences. A candidate is treated as a novel proposal when:
\begin{equation}
\operatorname{sim}\left(\langle p_m,\xi_m\rangle,\langle p_{m'},\xi_{m'}\rangle\right)<\epsilon_{\text{sim}}, \qquad \forall m'\in\mathcal{N}_m.
\end{equation}
When the candidate overlaps with an existing experience while introducing a potentially useful revision, MUSE treats it as a replacement proposal and compares the proposed revision with the corresponding existing experience based on the subsequent empirical verification before committing the update.

\noindent \textbf{Empirical Verification.} Semantic screening establishes source correspondence and distinguishes novel proposals from revisions of existing experiences, while their integration into the persistent experience bank further requires evidence that the candidate contributes to actual execution. To this end, MUSE subjects each candidate to controlled empirical verification. Specifically, MUSE isolates the candidate $\tilde{m}$ and evaluates it through $N_{\text{veri}}$ parallel rollouts on its original source task $\tau_{\tilde{m}}$. A low-temperature decoding configuration is adopted during these trials to reduce variation introduced by exploratory generation and provide a more controlled estimate of candidate contribution.

Notably, the control and candidate configurations share the same frozen background symbolic assets retrieved for source task $\tau_{\tilde{m}}$, leaving the candidate experience as the controlled variation. For a novel proposal, the comparison measures the effect of introducing $\tilde{m}$, whereas a replacement proposal compares $\tilde{m}$ against the corresponding existing experience. Let $\bar{r}_{\text{exp}}^{+}(\tilde{m})$ and $\bar{r}_{\text{exp}}^{-}(\tilde{m})$ denote the mean outcome rewards obtained under the candidate and control configurations, respectively, with their corresponding mean operational step counts $\bar{L}_{\text{exp}}^{+}(\tilde{m})$ and $\bar{L}_{\text{exp}}^{-}(\tilde{m})$.

By evaluating both the average task outcome and average execution steps, the verification classifies the candidate into three acceptance regimes. First, if the candidate yields a substantive outcome reward improvement exceeding a predefined margin $\epsilon_{\text{CNA}}$, it is accepted with a utility score equal to its normalized advantage. Second, when both configurations reach a saturated outcome score $r_{\text{sat}}$, the candidate is accepted with an initial score $w_{\text{eff}}^{\text{sat}}$ only if it substantially reduces execution steps. Third, when the two outcomes are not jointly saturated, their difference is bounded by $\epsilon_{\text{tie}}$, and at least one outcome is positive, the candidate is accepted with an initial score $w_{\text{eff}}^{\text{tie}}$ if it achieves the required reduction in execution steps. Candidates without an outcome or efficiency improvement are discarded. When a candidate $\tilde{m}$ satisfies any acceptance condition, a novel proposal is admitted to the experience bank $K_{\text{exp}}$, while a replacement proposal updates its corresponding existing experience. The saturated and comparable-outcome regimes are evaluated before outcome-improvement admission. For experiences, $\operatorname{Sat}(\tilde{m})$ holds when both mean outcome rewards reach $r_{\text{sat}}$. The predicate $\operatorname{Tie}(\tilde{m})$ holds when $\neg\operatorname{Sat}(\tilde{m})$, $\left|\bar{r}_{\text{exp}}^{+}(\tilde{m})-\bar{r}_{\text{exp}}^{-}(\tilde{m})\right|\le\epsilon_{\text{tie}}$, and $\max\{\bar{r}_{\text{exp}}^{+}(\tilde{m}),\bar{r}_{\text{exp}}^{-}(\tilde{m})\}>0$. The initial utility of each accepted experience is defined by:
\begin{equation}
    w_m^{(0)} =
    \begin{cases}
    \dfrac{\bar{r}_{\text{exp}}^{+}(\tilde{m})-\bar{r}_{\text{exp}}^{-}(\tilde{m})}{1 - \bar{r}_{\text{exp}}^{-}(\tilde{m})}, & \neg\operatorname{Sat}(\tilde{m})\land\neg\operatorname{Tie}(\tilde{m}) \land \bar{r}_{\text{exp}}^{+}(\tilde{m})>\bar{r}_{\text{exp}}^{-}(\tilde{m})+\epsilon_{\text{CNA}}, \\[3mm]
    w_{\text{eff}}^{\text{sat}}, & \operatorname{Sat}(\tilde{m})\land\bar{L}_{\text{exp}}^{+}(\tilde{m})<\bar{L}_{\text{exp}}^{-}(\tilde{m})(1-\epsilon_{\text{step}}^{\text{sat}}), \\[2mm]
    w_{\text{eff}}^{\text{tie}}, & \operatorname{Tie}(\tilde{m})\land\bar{L}_{\text{exp}}^{+}(\tilde{m})<\bar{L}_{\text{exp}}^{-}(\tilde{m})(1-\epsilon_{\text{step}}^{\text{tie}}).
    \end{cases}
\end{equation}

\noindent \textbf{Task-Relative Utility Tracking.} 
Admission-time verification evaluates an experience under the current agent configuration, while subsequent learning can change the contribution of retained guidance. Absolute task-performance changes may also reflect capability changes shared across tasks. MUSE therefore tracks task-relative utility to support continued reassessment after admission. During evolutionary rollout phase $t_{\text{roll}}$, let $\bar{r}_{\tau}^{(t_{\text{roll}})}$ denote the mean outcome reward for task $\tau$. The common task set is $\mathcal{T}_{\cap}^{(t_{\text{roll}})}=\mathcal{T}^{(t_{\text{roll}})}\cap\mathcal{T}^{(t_{\text{roll}}-1)}$, and the corresponding performance change is $r_{\tau}^{\Delta}=\bar{r}_{\tau}^{(t_{\text{roll}})}-\bar{r}_{\tau}^{(t_{\text{roll}}-1)}$. Equation~\ref{equ:task_relative_utility} subtracts the mean change over this common set and updates the utility of experiences delivered to task $\tau_m$. This adjustment partially accounts for the common performance shift and provides a longitudinal maintenance signal associated with experience-supported execution.

Tasks without delivered experiences contribute to the common performance baseline but do not update any experience entry. The update is postponed when common-task history is insufficient. The coefficient $\alpha_m\in(0,1)$ controls the adjustment magnitude. Continued utility tracking supports retention decisions, while the paired admission comparison provides candidate-specific evidence under controlled configurations.

\noindent \textbf{Systematic Knowledge Pruning.} 
Utility estimates can fluctuate across deployment contexts, so an isolated negative observation provides limited evidence for permanent removal. Continued accumulation also creates competition for finite memory capacity. MUSE therefore combines lifecycle-aware filtering, which permits further assessment of uncertain entries, with capacity-bounded replacement. Each experience is maintained in one of three utility states: \textit{active}, \textit{probation}, or \textit{quarantined}. Active experiences satisfying the injection requirement constitute the ordinary retrieval pool. When an experience receives a negative utility signal and its updated score falls below the operational boundary, it first enters the probation state. Probationary experiences remain eligible for retrieval so that subsequent observations can provide additional evidence. If the experience continues to receive negative evidence while remaining below the boundary, it transitions to the quarantined state and is excluded from retrieval. Confirmed low-utility or legacy harmful entries can subsequently be removed, whereas probationary entries are protected from immediate utility-based removal so that additional evidence can be collected.

Lifecycle filtering governs empirical eligibility, whereas finite context and storage capacity introduce a separate allocation constraint. Let $V_m$ denote the normalized historical retrieval frequency of experience $m$. When the number of retained experiences exceeds the capacity limit $N_{\mathrm{exp}}$, MUSE ranks eligible entries according to:
\begin{equation}
    \operatorname{Ret}(m) = w_m + \lambda_V V_m,
\end{equation}
where $\lambda_V$ controls the contribution of historical usage evidence. MUSE retains the $N_{\mathrm{exp}}$ highest-ranked experiences and removes the remaining entries. In this way, the lifecycle mechanism protects insufficiently tested experiences from premature utility-based removal, while capacity-based ranking systematically prioritizes entries that demonstrate sustained empirical utility and recurrent contextual relevance across evolving deployment stages.

Via anchored contrastive experience generation, empirical verification, task-relative utility tracking, and systematic knowledge pruning, the optimizer $\mathcal{M}$ establishes a dynamic and self-refining experience bank $K_{\text{exp}}$. The framework distills reusable situational knowledge from interaction, reassesses its empirical value as agent capability evolves, and progressively retires entries with sustained low utility. This lifecycle-aware refinement controls memory growth while maintaining a compact experience space aligned with evolving deployment evidence.

\appmodule{Functional Toolset \texorpdfstring{$T_{\text{tool}}$}{T_tool}}
\label{subapp:functional_toolset_refinement}

Recurring operation sequences provide opportunities to reduce repeated planning and action construction through reusable functional tools~\citep{qian2023creator,wang2025toolgen,xiong2026learning}. Their value depends on both the structure captured during construction and their contribution during subsequent execution. An abstraction derived from a single task can retain incidental bindings, motivating evidence of shared operational structure across distinct tasks. A callable implementation may also provide limited benefit under different task constraints, motivating empirical evaluation of candidate exposure and continued assessment after admission.

\begin{figure}[!t]
    \centering
    \includegraphics[width=\linewidth]{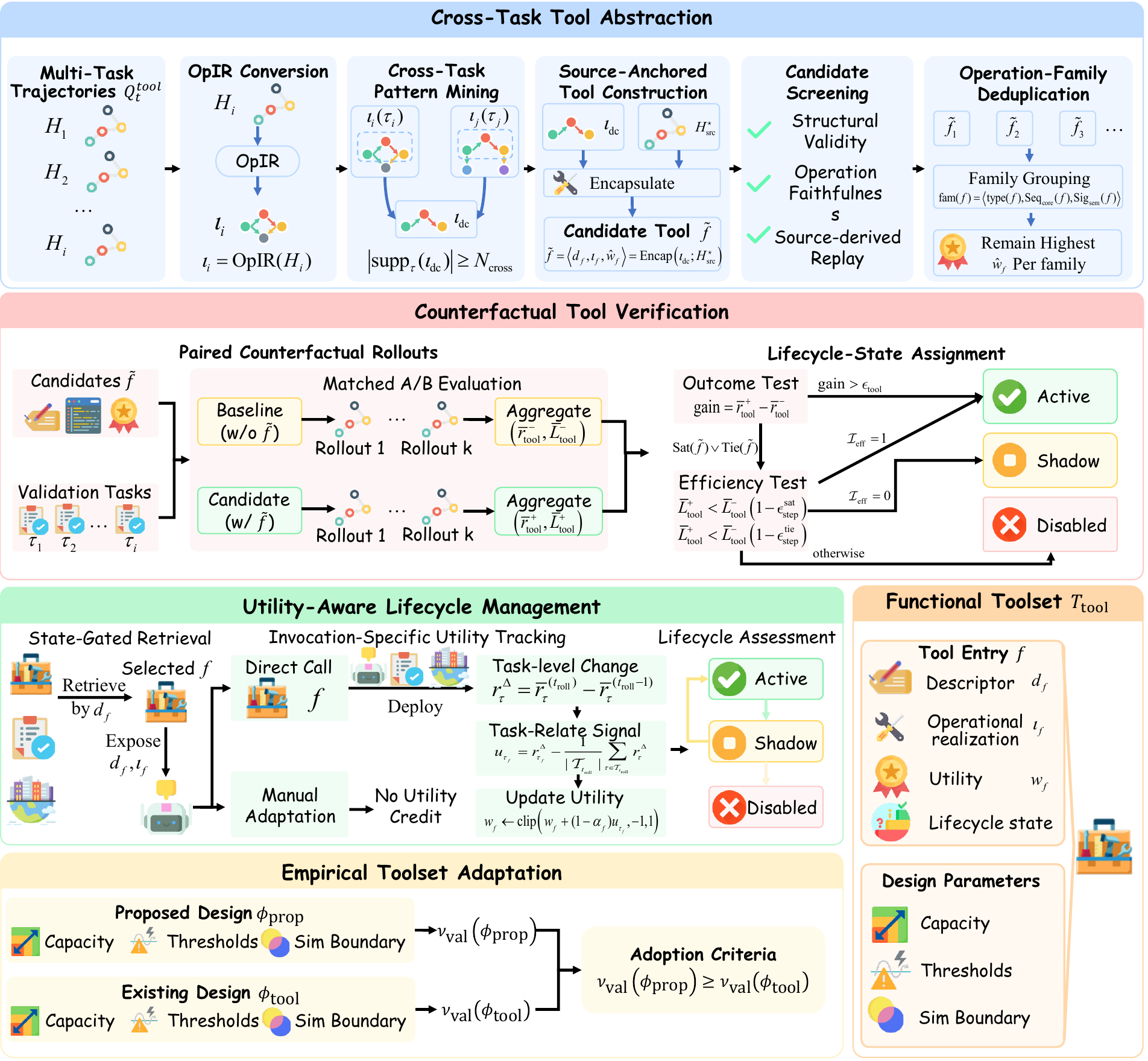}\vspace{-0.2cm}
    \caption{The optimization workflow for the functional toolset ($T_{\text{tool}}$). MUSE first converts trajectories from multiple tasks into operation-level representations, mines recurring dependency-closed patterns, and constructs source-anchored candidates that undergo structural and operational screening, source replay, and operation-family deduplication. Paired counterfactual rollouts assign candidates to active, shadow, or disabled states based on outcome and execution efficiency evidence. During deployment, state-gated retrieval exposes selected tools for direct invocation or manual adaptation. Only actual invocations receive task-relative utility credit, while execution reliability supports lifecycle assessment. At the manager level, MUSE empirically compares proposed and existing toolset configurations and adopts or rolls back changes according to their validation outcomes.}

    \label{fig:optimization_framework_for_toolset}
\end{figure}

MUSE therefore combines cross-task tool abstraction, counterfactual tool verification, utility-aware lifecycle management, and bounded adaptation of the toolset-management configuration. Tools support direct invocation and inspection as structural references for constructing adapted action sequences. Figure~\ref{fig:optimization_framework_for_toolset} illustrates how these mechanisms connect reusable operation patterns to their evolving deployment value.

\noindent \textbf{Cross-Task Tool Abstraction.} 
Functional tool construction begins by identifying operational structures that recur across distinct task contexts. If a tool is synthesized from a single trajectory, it risks overfitting to incidental variables or isolated execution choices. MUSE accordingly identifies recurring operation patterns supported by interaction trajectories from multiple distinct tasks. Let $Q_{t}^{\mathrm{tool}}$ denote the collection of interaction trajectories available for analysis at evolution round $t$. For each trajectory $H_i\in Q_{t}^{\mathrm{tool}}$, MUSE extracts an operation-level intermediate representation $\iota_i$ to facilitate cross-task pattern analysis:
\begin{equation}
    \iota_i = \operatorname{OpIR}(H_i).
\end{equation}
The representation $\iota_i$ abstracts away superficial textual variations while retaining the atomic operations and their structural dependencies. Let $\iota_{\mathrm{dc}}$ denote a dependency-closed operation pattern, and $\iota_{\mathrm{dc}} \preceq \iota_i$ indicate that the pattern occurs within $\iota_i$ while preserving its constituent operations, their execution order, and their structural dependencies. With $\tau_i$ denoting the source task associated with trajectory $H_i$, the task-level support of $\iota_{\mathrm{dc}}$ is defined as:
\begin{equation}
    \operatorname{supp}_{\tau}(\iota_{\mathrm{dc}}) = \left\{ \tau_i \;\middle|\; \iota_{\mathrm{dc}}\preceq\iota_i,\ H_i\in Q_{t}^{\mathrm{tool}} \right\}.
\end{equation}

MUSE considers a pattern sufficiently recurrent only if it is supported by at least $N_{\mathrm{cross}}$ distinct tasks:
\begin{equation}
    \left| \operatorname{supp}_{\tau}(\iota_{\mathrm{dc}}) \right| \ge N_{\mathrm{cross}}.
\end{equation}

A representative successful occurrence $H_{\mathrm{src}}^{\star}$ provides the execution basis for constructing candidate $\tilde{f}$ according to Equation~\ref{equ:tool_encapsulation}. Other supporting occurrences provide evidence about varying inputs and shared operational dependencies. In Equation~\ref{equ:tool_encapsulation}, $d_f$ serves as the semantic retrieval descriptor, defining the intent, callable interface, required inputs, and expected effects. The component $\iota_f$ represents the encapsulated operational realization, while $\hat{w}_f$ denotes a provisional utility score derived from the mean outcomes of the supporting trajectories.

Encapsulation can nevertheless introduce interface inconsistencies or deviations from the original recurring operation pattern. Before counterfactual verification, each candidate undergoes a targeted screening pipeline. A structural validity check first confirms that the candidate forms a self-contained operational unit with a consistent interface and does not encode task-specific bindings as fixed assumptions. An operation-faithfulness check then verifies that the encapsulated realization $\iota_f$ preserves the core operations, their ordering, informational dependencies, and environmental effects represented by $\iota_{\mathrm{dc}}$. Finally, MUSE performs source-derived replay checks constructed from the supporting occurrences, verifying that the candidate preserves the expected operation sequence and remains compatible with the observed response structures.

A separate source of redundancy arises when different task contexts yield superficially different implementations of the same underlying operation. MUSE performs operation-family deduplication to mitigate such toolset redundancy. Each candidate is assigned a deterministic structural identifier:
\begin{equation}
    \operatorname{fam}(f) = \left\langle \operatorname{type}(f), \operatorname{Seq}_{\mathrm{core}}(f), \operatorname{Sig}_{\mathrm{sem}}(f) \right\rangle,
\end{equation}
where $\operatorname{type}(f)$ categorizes the functional role,
$\operatorname{Seq}_{\mathrm{core}}(f)$ records the ordered sequence of core environment-facing operations, and $\operatorname{Sig}_{\mathrm{sem}}(f)$ summarizes the result-relevant and side-effect-relevant operational semantics while abstracting away superficial packaging and auxiliary control details. For candidates assigned to the same operation family, only the candidate with the highest provisional utility $\hat{w}_f$ is retained.

\noindent \textbf{Counterfactual Tool Verification.} The preceding abstraction and screening establish structural support for a candidate tool, while its actual contribution depends on how candidate exposure affects end-to-end execution. A structurally valid tool might still degrade end-to-end performance by unnecessarily consuming context length, introducing semantic ambiguity, or steering the agent toward suboptimal action trajectories. To estimate the marginal deployment value of a synthesized candidate $\tilde{f}$, MUSE applies controlled counterfactual verification. The system evaluates each candidate $\tilde{f}$ over the source task and a set of related tasks with compatible operational requirements. For every verification task, the experience bank $K_{\mathrm{exp}}$, general principles $P_{\mathrm{sys}}$, existing functional tools, and other operational settings are held fixed. A baseline configuration executes without the candidate, whereas a candidate configuration receives the same symbolic context augmented with $\tilde{f}$. Multiple paired rollouts are conducted for both configurations to estimate the empirical marginal effect of candidate exposure.

Let $\bar{r}_{\mathrm{tool}}^{+}(\tilde{f})$ and $\bar{r}_{\mathrm{tool}}^{-}(\tilde{f})$ denote the mean terminal outcome rewards obtained with and without the candidate functional tool, respectively. The corresponding average execution steps are denoted as $\bar{L}_{\mathrm{tool}}^{+}(\tilde{f})$ and $\bar{L}_{\mathrm{tool}}^{-}(\tilde{f})$, respectively. Equation~\ref{equ:tool_exposure_gain} summarizes the mean outcome difference between these matched configurations, providing empirical evidence of the contribution of candidate exposure.

Terminal outcome captures the primary behavioral contribution, while comparable outcomes may still conceal an efficiency advantage. MUSE distinguishes two regimes with comparable outcome rewards when evaluating this operational efficiency. The saturated regime, denoted by $\operatorname{Sat}(\tilde{f})$, applies when both mean outcomes reach a predefined saturation level $r_{\mathrm{sat}}$. The regime of comparable nonsaturated outcomes, denoted by $\operatorname{Tie}(\tilde{f})$, applies when the outcomes are not jointly saturated, their absolute difference does not exceed $\epsilon_{\mathrm{tool}}$, and at least one configuration obtains a positive outcome. Separate relative step-reduction thresholds $\epsilon_{\mathrm{step}}^{\mathrm{sat}}$ and $\epsilon_{\mathrm{step}}^{\mathrm{tie}}$ allow the efficiency criterion to differ between the two outcome regimes. The resulting efficiency indicator is formulated as:
\begin{equation}
    \mathcal{I}_{\mathrm{eff}}(\tilde{f}) =
    \begin{cases}
        \mathbb{I}\!\left[\bar{L}_{\mathrm{tool}}^{+}(\tilde{f}) < \bar{L}_{\mathrm{tool}}^{-}(\tilde{f}) \left(1-\epsilon_{\mathrm{step}}^{\mathrm{sat}}\right)\right], & \operatorname{Sat}(\tilde{f}), \\[2mm]

        \mathbb{I}\!\left[\bar{L}_{\mathrm{tool}}^{+}(\tilde{f}) < \bar{L}_{\mathrm{tool}}^{-}(\tilde{f}) \left(1-\epsilon_{\mathrm{step}}^{\mathrm{tie}}\right)\right], & \operatorname{Tie}(\tilde{f}), \\[2mm]

        0, & \text{otherwise}.
    \end{cases}
\end{equation}

These outcome and efficiency observations jointly determine the initial deployment status of the candidate. For a complete paired verification, its lifecycle state is assigned according to the criteria:
\begin{equation}
    \operatorname{state}(\tilde{f}) =
    \begin{cases}
        \mathrm{active}, & \left(\neg\operatorname{Sat}(\tilde{f})\land\neg\operatorname{Tie}(\tilde{f})\land\operatorname{gain}(\tilde{f})>\epsilon_{\mathrm{tool}}\right)\lor\mathcal{I}_{\mathrm{eff}}(\tilde{f})=1,\\
        \mathrm{shadow}, & \left(\operatorname{Sat}(\tilde{f})\lor\operatorname{Tie}(\tilde{f})\right)\land\mathcal{I}_{\mathrm{eff}}(\tilde{f})=0,\\
        \mathrm{disabled}, & \text{otherwise}.
    \end{cases}
\end{equation}
Accordingly, a candidate enters the active state when it reduces the execution horizon by the required margin under an outcome-equivalence regime, or produces an outcome improvement exceeding $\epsilon_{\mathrm{tool}}$ outside these regimes. A candidate enters the shadow state when its outcomes remain comparable to the baseline without sufficient efficiency evidence. It can then participate in subsequent controlled re-evaluation. Candidates that satisfy neither criterion enter the disabled state and remain unavailable during ordinary deployment. Incomplete paired verification is also mapped to the shadow state, leaving room for subsequent evidence before a definitive deployment decision.

The same paired evidence also initializes the candidate utility. For complete paired verification, the initial utility follows the empirical utility interpretation adopted for other symbolic assets:
\begin{equation}
    w_f^{(0)} =
    \begin{cases}
        w_{\mathrm{eff}}^{\mathrm{sat}}, & \operatorname{Sat}(\tilde{f})\land\mathcal{I}_{\mathrm{eff}}(\tilde{f})=1,\\
        w_{\mathrm{eff}}^{\mathrm{tie}}, & \operatorname{Tie}(\tilde{f})\land\mathcal{I}_{\mathrm{eff}}(\tilde{f})=1,\\
        \dfrac{\operatorname{gain}(\tilde{f})}{1-\bar{r}_{\mathrm{tool}}^{-}(\tilde{f})}, & \neg\operatorname{Sat}(\tilde{f})\land\neg\operatorname{Tie}(\tilde{f}) \land \operatorname{gain}(\tilde{f})>\epsilon_{\mathrm{tool}},\\
        0, & \operatorname{state}(\tilde{f})=\mathrm{shadow},\\
        \operatorname{gain}(\tilde{f}), & \operatorname{state}(\tilde{f})=\mathrm{disabled}.
    \end{cases}
\end{equation}
Here, $w_{\mathrm{eff}}^{\mathrm{sat}}$ and $w_{\mathrm{eff}}^{\mathrm{tie}}$ denote the positive initialization scores assigned to efficiency improvements under saturated outcomes and comparable nonsaturated outcomes, respectively. A shadow candidate produced by incomplete verification retains the provisional source utility $\hat{w}_f$ until additional paired evidence becomes available.

Upon concluding the verification phase, the tool is represented as $f=\langle d_f,\iota_f,w_f\rangle$ and retained in $T_{\mathrm{tool}}$ together with the designated lifecycle state. Active tools participate in ordinary deployment. Shadow tools remain available for controlled re-evaluation and may provide a limited retrieval fallback when no active tool is available. Disabled tools are excluded from retrieval. Their operation-family records are retained as screening evidence, allowing subsequent candidate generation to identify operation families that have already been evaluated.

\noindent \textbf{Utility-Aware Lifecycle Management.}
Counterfactual verification provides an admission-time estimate of tool utility, whereas continued deployment can alter that value as the environment, task distribution, or internal capabilities of the agent evolve. MUSE therefore maintains verified tools through a lifecycle consisting of active, shadow, and disabled states, allowing their deployment eligibility to be revised as empirical evidence accumulates.

During deployment, MUSE retrieves relevant tools according to descriptors characterizing their functional intents, operational constraints, and environmental effects. Active tools constitute the primary retrieval pool. When no active tool is available, shadow tools may enter a limited retrieval fallback for subsequent reassessment.

Distinct from textual experience guidance, a retrieved tool exposes an operational realization that can contribute through two complementary modes. Within the symbolic context supplied to the agent, the system places the functional descriptor, callable interface, and operational realization of each tool. This implementation-visible configuration allows the agent to either directly invoke the tool to execute its encapsulated operations, or inspect its realization as a structural reference to construct an adapted sequence of atomic actions when the predefined logic does not align with current task constraints. Notably, such manual adaptations are treated as ordinary action sequences instead of tool invocations and therefore yield no utility adjustments for the referenced tool.

Contextual exposure makes this realization available for reasoning, whereas direct invocation additionally requires an executable representation within the environment. Consequently, when the agent emits a compact tool call, the runtime automatically prepends the exact tool definition and realization to the executable action associated with the current trajectory. This call-time materialization translates the compact tool call into a fully executable sequence and aligns the realization presented to the agent with the logic executed during rollout.

Candidate verification assesses the contribution of tool availability, including direct invocation and reference-guided adaptation. Actual invocation establishes the task association for continued utility tracking of the encapsulated procedure. For a tool $f$ invoked on task $\tau_f$, Equation~\ref{equ:tool_utility_update} applies the task-relative signal introduced in Equation~\ref{equ:task_relative_utility}. The update is postponed when common-task history is insufficient. Retrieval, contextual exposure, and manually adapted action sequences do not trigger this update. Experience utility is associated with delivered guidance, while tool utility is associated with actual invocation. The scores $w_m$ and $w_f$ and their adjustment coefficients $\alpha_m$ and $\alpha_f$ remain independent. This shared relative baseline supports continued assessment across the two components while preserving their different forms of deployment evidence.

Executable tools additionally admit a finer distinction between end-to-end task success and the execution status of their encapsulated operations. The task-level runtime success rate measures how often trajectories invoking a tool achieve the terminal success criterion. If an active tool undergoes a decline in this success rate below a configured threshold after sufficient invocations, the framework demotes it to the shadow state to limit exposure. Concurrently, recording whether the encapsulated operations complete successfully distinguishes tool execution status from end-to-end task success and supplies diagnostic evidence for subsequent controlled re-evaluation.

A shadow tool becomes eligible for promotion to the active state when subsequent paired verification satisfies the outcome or efficiency criteria defined above. Conversely, persistent low utility becomes stronger evidence of degradation after sufficient invocation history has accumulated. Tools exhibiting consistently low utility transition to the disabled state when both the utility and historical invocation count cross their respective thresholds:
\begin{equation}
    \left(w_f < \epsilon_{\mathrm{prune}}\right) \land \left(V_f > \epsilon_{\mathrm{retr}}\right).
\end{equation}
Here, $V_f$ denotes the historical task-level invocation count of tool $f$,  $\epsilon_{\mathrm{prune}}$ is the utility threshold, and $\epsilon_{\mathrm{retr}}$ specifies the task-level invocation threshold that must be exceeded before disabling a low-utility tool. Disabled tools are retained in the tool records as evidence for subsequent candidate screening but are removed from the retrieval space. These lifecycle transitions adjust tool availability according to accumulated utility and invocation evidence.

\noindent \textbf{Empirical Toolset Adaptation.}
The preceding lifecycle mechanisms adapt individual tools, while the overall tool inventory is also governed by manager-level constraints such as capacity, injection, pruning, and deduplication thresholds. As the continually growing functional toolset accumulates verified assets, maintaining a static management configuration risks introducing retrieval overhead and context saturation. MUSE accordingly extends adaptation from individual tools to the configuration governing the overall toolset.

Let $\phi_{\mathrm{tool}} \in \Phi$ denote the current toolset-management configuration. Based on recent validation outcomes, the size and utility distribution of the current toolset, and the existing management parameters, the evolutionary optimizer $\mathcal{M}$ proposes an alternative configuration $\phi_{\mathrm{prop}}$. The proposal may adjust the maximum toolset capacity, utility-pruning threshold, minimum invocation count required for pruning, minimum injection score, and deduplication threshold. This manager-level process adapts the conditions governing toolset maintenance and deployment.

When sufficient validation cases are available, MUSE records the current management parameters and toolset state, applies $\phi_{\mathrm{prop}}$,  and reapplies lifecycle pruning before evaluating deployment under the proposed configuration on the same validation tasks. Parameters governing capacity and deduplication are retained as part of the adopted configuration and take effect during subsequent toolset updates. Let $\nu_{\mathrm{val}}(\phi)$ denote the mean validation outcome reward obtained under configuration $\phi$. The manager-level update is formulated as:
\begin{equation}
    \phi_{\mathrm{tool}} \leftarrow
    \begin{cases}
        \phi_{\mathrm{prop}}, & \nu_{\mathrm{val}}(\phi_{\mathrm{prop}}) \ge \nu_{\mathrm{val}}(\phi_{\mathrm{tool}}), \\[2mm]
        \phi_{\mathrm{tool}}, & \text{otherwise}.
    \end{cases}
\end{equation}
When the proposed configuration achieves an equal or higher mean validation outcome reward, MUSE retains the updated management parameters and the toolset state obtained under them. Otherwise, it restores the previous parameters and records. This proposal, verification, and rollback cycle adapts manager-level toolset constraints according to validation evidence while remaining separate from candidate-level tool construction and verification.

Through cross-task tool abstraction, counterfactual tool verification, utility-aware lifecycle management, and empirical toolset adaptation, MUSE maintains an evolving functional toolset $T_{\mathrm{tool}}$. Recurrent operational patterns are first consolidated into source-anchored callable structures and admitted according to counterfactual evidence. Their deployment value is subsequently reassessed through invocation-specific utility and execution records, while the surrounding management configuration adapts at the toolset level. Together, these mechanisms coordinate the construction, verification, deployment, and configuration of reusable tools throughout continuous evolution, maintaining an adaptive repertoire of reusable operations.

\appmodule{Foundation Model Weights \texorpdfstring{$\theta_{\text{model}}$}{theta_model}}
\label{subapp:foundation_model_weights}

\begin{figure}[!t]
    \centering
    \includegraphics[width=\linewidth]{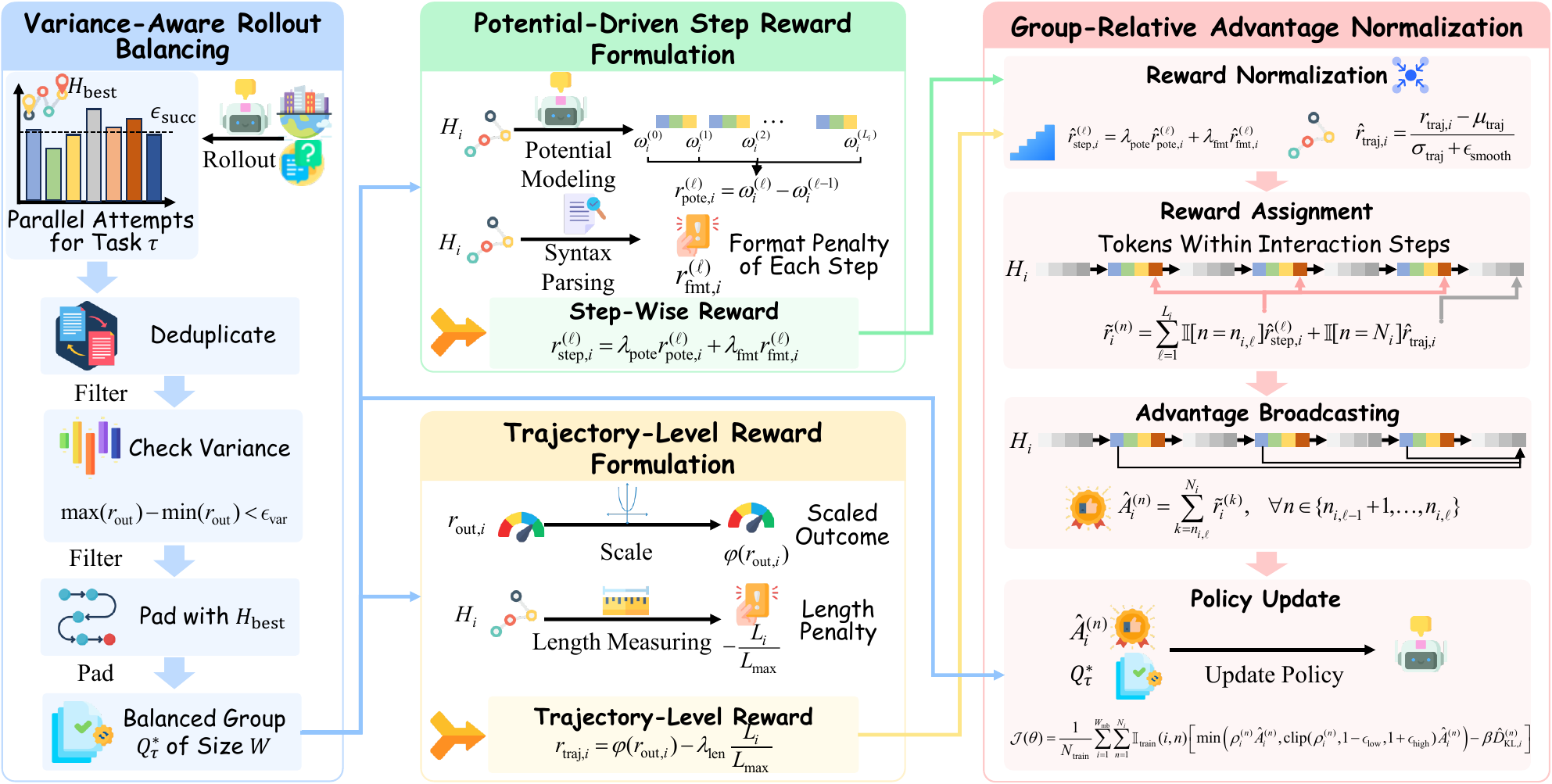}\vspace{-0.2cm}
    \caption{The optimization workflow for foundation model weights. Trajectory screening preserves behavioral comparisons within task groups. Potential-derived progress and format penalties provide step rewards, while transformed outcome rewards and execution-length penalties provide trajectory rewards. Separate normalization of trajectory rewards and the potential and format components within step rewards accommodates their different scales. Each generated action token receives the normalized trajectory reward and the cumulative normalized step rewards from its current step onward, supporting parametric inner-component refinement.}
    \label{fig:optimization_framework_for_model_weights}
\end{figure}

Persistent difficulties in planning, reasoning, and instruction following motivate adaptation of the parametric core. MUSE connects interaction trajectories and the supervision supplied through $\Delta$ to updates of $\theta_{\text{model}}$. Oracle tasks contribute environment-provided outcome rewards, while synthetic tasks contribute the assessed outcome rewards in Equation~\ref{equ:synthetic_reward}. Both task sources additionally receive the prefix-potential signals in Equation~\ref{equ:potential_step_reward}. Their different temporal granularities and numerical scales motivate trajectory-group preparation, separate reward normalization, and credit assignment to generated action tokens. Figure~\ref{fig:optimization_framework_for_model_weights} illustrates the resulting optimization workflow.

\noindent \textbf{Variance-Aware Rollout Balancing.}
Repeated or closely similar executions can occupy a task group without providing additional behavioral contrast. MUSE therefore applies semantic and action deduplication before preparing a curated group for optimization. For a task $\tau$, the initial group $Q_{\tau}$ contains $W^{\prime}$ trajectories, from which the optimizer retains a subset $Q_{\tau}^{\mathrm{cur}}$ of at most $W$ trajectories after screening. The following padding and normalization definitions apply to nonempty curated groups containing at least one trainable action token. When outcome rewards vary, retaining executions with both high and low scores preserves comparisons among alternative behaviors. Groups with homogeneous outcome rewards may still receive different process signals at each step.

Deduplication can leave fewer trajectories than the target group size. The optimizer then adopts copies of the highest-scoring retained trajectory $H_{\mathrm{best}}$ to fill the remaining positions:
\begin{equation}
    Q^*_{\tau} =
    \operatorname{pad}\left(
        Q_{\tau}^{\mathrm{cur}},
        W,
        H_{\mathrm{best}}
    \right),
    \qquad
    \text{for each retained curated group } Q_{\tau}^{\mathrm{cur}}.
\end{equation}
The added copies serve as padding. They are excluded from reward-normalization statistics and receive zero loss masks, so group completion does not give repeated copies additional optimization weight.

\noindent \textbf{Step-Level Reward Integration.}
The optimizer receives $r_{\text{pote},i}^{(\ell)}$ from $\Delta$, with details specified in Appendix~\ref{subapp:supervision_generator}. Equation~\ref{equ:model_step_reward} combines this process signal with an optional deterministic format penalty. Valid actions receive a zero format penalty, while detected violations of the required syntax or environmental schema receive a predefined negative penalty. This additional term represents format compliance alongside assessed progress. The coefficients $\lambda_{\text{pote}}$ and $\lambda_{\text{fmt}}$ control the respective contributions after separate normalization.

\noindent \textbf{Trajectory-Level Reward Formulation.}
Intermediate progress provides limited evidence of whether the overall task objective has been fulfilled. The trajectory reward in Equation~\ref{equ:model_trajectory_reward} therefore retains the outcome signal alongside an execution-length penalty. The default transformation $\varphi(x)=x^2$ changes the spacing of outcome rewards before normalization. For scores in $[0,1]$, its local sensitivity increases with the reward level, emphasizing distinctions among executions closer to task completion. The length penalty accounts for execution effort through $L_i/L_{\text{max}}$, with $\lambda_{\text{len}}$ controlling its contribution. The trajectory reward remains distinct from step-level signals during normalization.

\noindent \textbf{Group-Relative Advantage Normalization.}
Trajectory rewards summarize overall executions, while step rewards describe changes across individual interaction steps. Their different scales and variances motivate separate standardization within each task group. For trajectory rewards, the mean $\mu_{\text{traj}}$ and standard deviation $\sigma_{\text{traj}}$ are computed over retained, non-padding trajectories for the same task:
\begin{equation}
    \hat{r}_{\text{traj},i} =
    \frac{
        r_{\text{traj},i}-\mu_{\text{traj}}
    }{
        \sigma_{\text{traj}}+\epsilon_{\text{smooth}}
    }.
\end{equation}
Here, $\epsilon_{\text{smooth}}>0$ is a small constant for numerical stabilization. All standard deviations in the reward-normalization procedure are population standard deviations over the specified non-padding entries.

Step statistics pool the eligible non-padding interaction steps from those same trajectories~\citep{hu2026self}. Leveraging a common distribution across step positions preserves differences between earlier and later steps within the task group. Let $\mu_{\text{pote}}$ and $\sigma_{\text{pote}}$ denote the mean and standard deviation of potential signals clipped to $[-1,1]$, with statistics restricted to steps carrying both an available potential estimate and trainable action tokens:
\begin{equation}
    \hat{r}_{\text{pote},i}^{(\ell)} = \frac{\operatorname{clip}(r_{\text{pote},i}^{(\ell)}, -1, 1) - \mu_{\text{pote}}}{\sqrt{\sigma_{\text{pote}}^{2} + \epsilon_{\text{smooth}}}}.
\end{equation}
Nonzero format penalties from non-padding steps with trainable action tokens are standardized separately by subtracting their task-group mean and dividing by their population standard deviation plus $\epsilon_{\text{smooth}}$. All other format entries remain zero. Equation~\ref{equ:model_step_reward} combines these components without further normalization of their weighted sum.

After normalization, the two rewards are placed at their corresponding temporal boundaries. Step rewards are assigned to step-terminal  positions, while the trajectory reward is assigned to the terminal position of the trajectory:
\begin{equation}
    \tilde{r}_i^{(n)} =
    \sum_{\ell=1}^{L_i}
    \mathbb{I}[n=n_{i,\ell}]
    \hat{r}_{\text{step},i}^{(\ell)}
    +
    \mathbb{I}[n=N_i]
    \hat{r}_{\text{traj},i}.
\end{equation}
Here, $n_{i,\ell}$ denotes the terminal token index of step $\ell$, $N_i$ is the terminal token index of trajectory $H_i$, and $\mathbb{I}[\cdot]$ is the indicator function. Accumulating the rewards from each step onward yields Equation~\ref{equ:model_token_advantage}. Every generated action token within step $\ell$ receives the normalized trajectory reward and the normalized step rewards from $\ell$ onward.

Symbolic assets may support action generation during rollout collection. Training retains task queries, generated actions, and environmental observations while omitting explicit retrieved guidance. The policy loss applies to generated action tokens in retained, non-padding trajectories. The resulting training examples pair symbolically supported actions with contexts that omit explicit retrieved guidance.

Let $u_{i,n}$ denote the generated action token at position $n$ given the preceding training context $H_{i,<n}$. The probability ratio between the current policy $\pi_{\theta}$ and the frozen pre-update policy $\pi_{\text{old}}$ in this context is:
\begin{equation}
    \rho_i^{(n)} =
    \frac{
        \pi_{\theta}(u_{i,n}\mid H_{i,<n})
    }{
        \pi_{\text{old}}(u_{i,n}\mid H_{i,<n})
    }.
\end{equation}

Continued adaptation can shift the actor policy away from the frozen reference model $\pi_{\text{ref}}$. MUSE therefore includes the following sampled token-level KL penalty alongside the clipped policy objective~\citep{shao2024deepseekmath}:
\begin{equation}
    \hat{D}_{\text{KL},i}^{(n)}=
    \frac{
        \pi_{\text{ref}}(u_{i,n}\mid H_{i,<n})
    }{
        \pi_{\theta}(u_{i,n}\mid H_{i,<n})
    } -
    \log
    \frac{
        \pi_{\text{ref}}(u_{i,n}\mid H_{i,<n})
    }{
        \pi_{\theta}(u_{i,n}\mid H_{i,<n})
    }
    -1.
\end{equation}

Combining the advantages, asymmetric clipping, and reference-policy penalty yields the optimization objective:
\begin{equation}
    \mathcal{J}(\theta) = \frac{1}{N_{\mathrm{train}}} \sum_{i=1}^{W_{\mathrm{mb}}} \sum_{n=1}^{N_i} \mathbb{I}_{\mathrm{train}}(i,n) \left[ \min \left( \rho_i^{(n)} \hat{A}_i^{(n)}, \text{clip}(\rho_i^{(n)}, 1 - \epsilon_{\text{low}}, 1 + \epsilon_{\text{high}}) \hat{A}_i^{(n)} \right) - \beta \hat{D}_{\text{KL},i}^{(n)} \right].
\end{equation}
Here, $W_{\mathrm{mb}}$ denotes the number of trajectories in the optimization mini-batch, and $N_{\mathrm{train}}$ denotes its total number of trainable action tokens. The mask $\mathbb{I}_{\mathrm{train}}(i,n)$ equals 1 for generated action tokens in retained and non-padding trajectories and 0 otherwise. It applies to both the policy term and the KL penalty. The coefficient $\beta$ controls reference-policy regularization, while $\epsilon_{\text{low}}$ and $\epsilon_{\text{high}}$ define the clipping interval. A wider upper interval, with $\epsilon_{\text{high}}>\epsilon_{\text{low}}$, permits a larger probability-ratio increase before clipping for positively advantaged actions.

\subsubsection{Inter-Component Synchronization}

Inner-component refinement updates individual components, while their deployment value also depends on relationships with each other. Repeatedly useful experiences may support broader principles, independently acquired symbolic assets may overlap or conflict, and parameter updates may change the contribution of explicit guidance. These dependencies motivate inter-component synchronization through experience-to-principle abstraction, conflict-aware symbolic integration, and reassessment of symbolic retention following parameter updates.

\appmodule{Distilling General Principles \texorpdfstring{$P_{\text{sys}}$}{P_sys} from Experiences}
\label{subapp:general_principle_distillation}

\begin{figure}[!t]
    \centering
    \includegraphics[width=\linewidth]{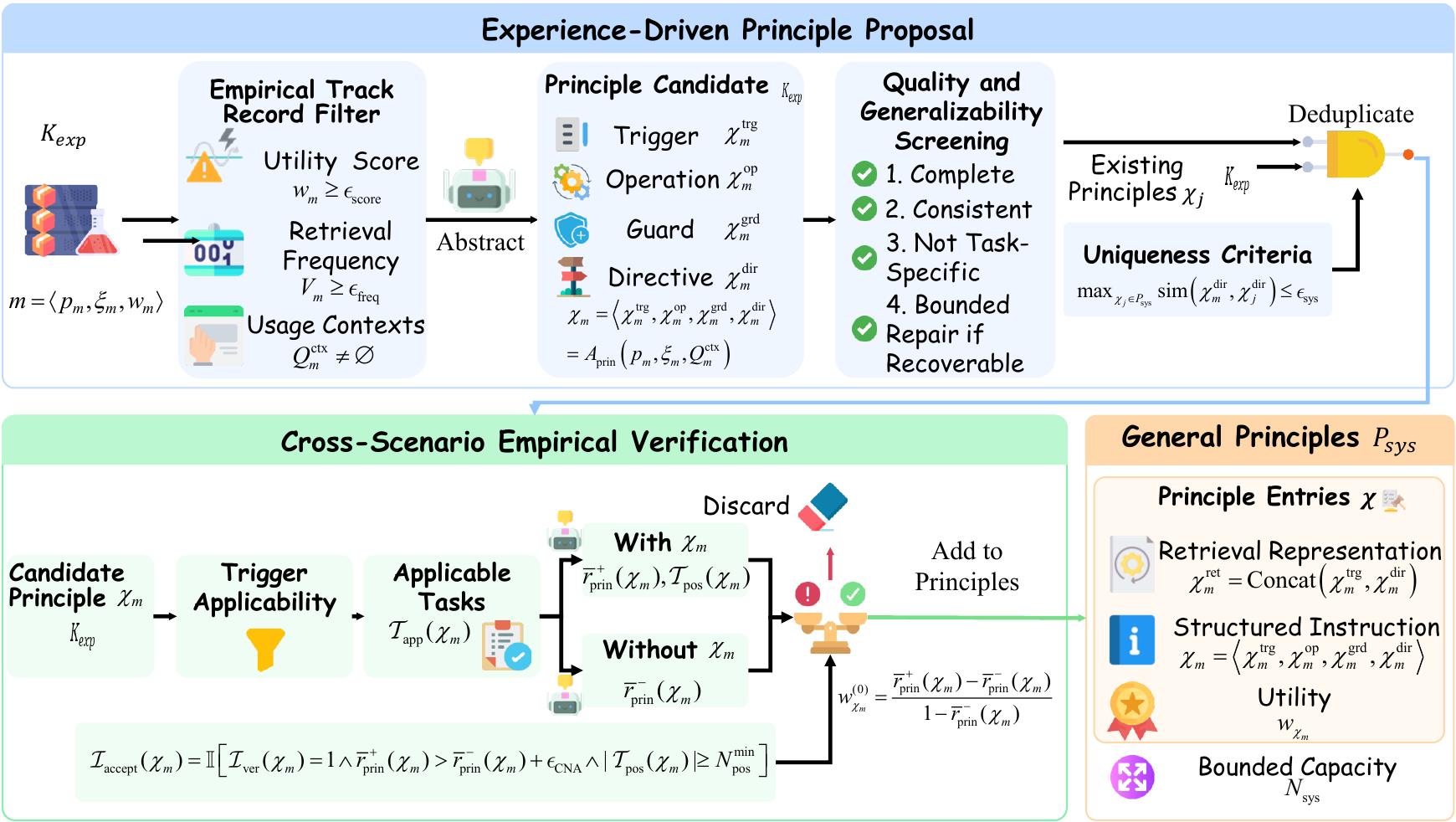}
    \caption{The distillation workflow for general principles ($P_{\text{sys}}$). MUSE first executes an experience-driven principle proposal to select experiences with sufficient empirical evidence and abstract each into a structured candidate comprising a trigger, operation pattern, guard, and directive. Following quality and generalizability screening and semantic deduplication, MUSE subjects the candidates to applicability-guided cross-scenario verification through paired rollouts over a bounded cohort of tasks satisfying their structural triggers. Admission requires a macro-averaged outcome improvement consistently supported across multiple applicable tasks, subject to verification coverage and reliability checks. Accepted candidates enter $P_{\text{sys}}$ with normalized utility. Retrieval adopts the concatenated trigger and directive, while an applicable principle is delivered as a complete structured instruction containing the corresponding trigger, operation pattern, guard, and directive. When the number of stored principles exceeds the capacity $N_{\mathrm{sys}}$, entries with the lowest utility are retired.}
    \label{fig:optimization_framework_for_principle}
\end{figure}

The experience bank $K_{\text{exp}}$ provides reusable behavioral guidance conditioned on situational preconditions, allowing an experience to transfer across tasks exhibiting compatible situations. As recurrently useful experiences accumulate, some of their operational regularities can exhibit applicability beyond the situational boundary of an individual experience and become suitable for a more abstract systemic instruction layer. This observation motivates the distillation of empirically high-utility experiences into general principles $P_{\text{sys}}$, abstracting recurrent structural conditions and behavioral directives from their situational instantiations. To support this abstraction with empirical evidence, the distillation process is systematically divided into experience-driven principle proposal and cross-scenario empirical verification. Figure~\ref{fig:optimization_framework_for_principle} illustrates the distillation framework for general principles $P_{\text{sys}}$.

\noindent \textbf{Experience-Driven Principle Proposal.} Principle abstraction is most meaningful when its source experience has accumulated sufficient evidence of recurrent empirical utility. Candidate experiences are thus selected according to their empirical track records. An experience $m = \langle p_m, \xi_m \rangle$ with utility $w_m$ in $K_{\text{exp}}$ is eligible for promotion only if its $w_m$ meets a designated promotion threshold $\epsilon_{\text{score}}$ ($w_m \ge \epsilon_{\text{score}}$) and its normalized retrieval frequency $V_m$ meets a predefined retrieval-frequency threshold $\epsilon_{\text{freq}}$ ($V_m \ge \epsilon_{\text{freq}}$), provided that it has neither been quarantined by the utility lifecycle nor previously marked as shadowed following an earlier promotion. Beyond utility itself, observed deployment contexts provide evidence about which aspects of an experience remain recurrent across situations. For each experience $m$, let $Q_m^{\mathrm{ctx}}$ denote the bounded collection of observed usage contexts in which $m$ is actually retrieved and integrated into the symbolic context of the agent. Each usage context records the corresponding task query together with its task metadata, and $Q_m^{\mathrm{ctx}}\neq \emptyset$ is required for principle proposal. This guided empirical selection reduces the promotion of insufficiently evaluated experiences and prioritizes candidates that have demonstrated repeated empirical usefulness. For each nominated high-utility experience, the agent itself functions as a principle abstractor $A_{\text{prin}}$ to abstract the core operational logic from its specific situational precondition $p_m$. By analyzing the localized context alongside the actionable guideline $\xi_m$ and the observed deployment contexts $Q_m^{\mathrm{ctx}}$, the abstractor transforms the situational heuristic into a structured candidate principle $\chi_m$ according to Equation~\ref{equ:principle_abstraction}. Specifically, $\chi_m^{\mathrm{trg}}$ specifies the structural conditions where the principle applies, such as data dependencies, operation ordering, state eligibility, or cross-system mappings, while abstracting away task-specific identifiers and business endpoints. The experience precondition $p_m$ characterizes the situational context in which experience $m$ is applicable, whereas $\chi_m^{\mathrm{trg}}$ retains the underlying structural condition after removing task-specific semantic bindings. It therefore represents a broader applicability boundary across tasks exhibiting the same operational structure. Meanwhile, $\chi_m^{\mathrm{op}}$ describes the reusable operational pattern, $\chi_m^{\mathrm{grd}}$ records relevant safety and ordering constraints, and $\chi_m^{\mathrm{dir}}$ provides the normalized declarative directive presented to the agent.

Abstraction can produce incomplete instructions or retain conditions specific to the source task. MUSE therefore examines the completeness and consistency of the four principle fields and screens for triggers that retain task-specific or interface-specific conditions. A bounded repair may be attempted for recoverable candidates before further evaluation. A separate novelty check addresses repeated directives within the bounded principle pool. MUSE compares the embedding of the candidate directive $\chi_m^{\mathrm{dir}}$ with existing directives and admits the candidate to empirical verification when its similarity to every existing directive satisfies:
\begin{equation}
    \operatorname{sim}\left(\chi_m^{\mathrm{dir}},\chi_j^{\mathrm{dir}}\right)\le\epsilon_{\text{sys}}, \qquad \forall\chi_j\in P_{\text{sys}}.
\end{equation}

For subsequent deployment-time retrieval, MUSE constructs a separate representation:
\begin{equation}
    \chi_m^{\mathrm{ret}} = \operatorname{Concat}\left(\chi_m^{\mathrm{trg}}, \chi_m^{\mathrm{dir}}\right).
\end{equation}

This representation combines structural applicability with directive semantics and supports deployment-time candidate recall and ranking. Because a general principle encodes a more abstract applicability condition than the corresponding experience precondition, MUSE applies a more permissive relevance prefilter when identifying potentially related tasks during deployment. The recalled candidates subsequently undergo an explicit assessment of whether their structural triggers apply to the current task query. During cross-scenario verification, the same applicability assessment begins from the available tasks excluding the source task, while similarity derived from $\chi_m^{\mathrm{ret}}$ provides auxiliary ranking evidence when the applicable cohort exceeds the verification budget. When an applicable principle is delivered to the agent, its trigger $\chi_m^{\mathrm{trg}}$, operational pattern $\chi_m^{\mathrm{op}}$, guard $\chi_m^{\mathrm{grd}}$, and directive $\chi_m^{\mathrm{dir}}$ are rendered together as a structured instruction block. This preserves the conditions, operational semantics, and constraints required for context-sensitive application.

\noindent \textbf{Cross-Scenario Empirical Verification.} Semantic uniqueness and structural quality establish a well-formed and sufficiently abstract candidate principle. Its broader applicability introduced through abstraction, however, calls for additional cross-scenario behavioral evidence across multiple compatible contexts before promotion into the systemic instruction space. The verification stage consequently evaluates the empirical marginal contribution of each candidate across diverse applicable contexts in which its structural trigger applies.

The experience precondition $p_m$ describes the task-conditioned context associated with experience $m$, while $\chi_m^{\mathrm{trg}}$ captures the corresponding structural condition after task-specific semantic bindings have been abstracted. This broader scope motivates an explicit applicability assessment before empirical evaluation. Starting from the available tasks excluding the source task, MUSE compares $\chi_m^{\mathrm{trg}}$ with each task query and identifies tasks for which the trigger applies with sufficient confidence. When the number of applicable tasks exceeds the verification budget, the framework retains a bounded and diverse cohort, prioritizing applicability confidence with the similarity derived from $\chi_m^{\mathrm{ret}}$ providing auxiliary ranking evidence. We denote $\mathcal{T}_{\mathrm{app}}(\chi_m)$ as the applicable verification cohort.

To assess the behavioral contribution of $\chi_m$, MUSE evaluates every task $\tau \in \mathcal{T}_{\mathrm{app}}(\chi_m)$ under two matched configurations that differ only in whether the candidate principle is included. Let $\bar{r}_{\mathrm{prin},\tau}^{+}(\chi_m)$ and $\bar{r}_{\mathrm{prin},\tau}^{-}(\chi_m)$ denote the corresponding mean outcome rewards of parallel rollouts with and without the candidate, respectively.  Let $\mathcal{T}_{\mathrm{eval}}(\chi_m)\subseteq\mathcal{T}_{\mathrm{app}}(\chi_m)$ denote the tasks for which both configurations produce valid rollout results. When $\mathcal{T}_{\mathrm{eval}}(\chi_m)\neq\emptyset$, their macro-averaged outcome rewards are then respectively defined as:
\begin{equation}
    \bar{r}_{\mathrm{prin}}^{+}(\chi_m)
    =
    \frac{1}{|\mathcal{T}_{\mathrm{eval}}(\chi_m)|}
    \sum_{\tau\in\mathcal{T}_{\mathrm{eval}}(\chi_m)}
    \bar{r}_{\mathrm{prin},\tau}^{+}(\chi_m),
    \quad
    \bar{r}_{\mathrm{prin}}^{-}(\chi_m)
    =
    \frac{1}{|\mathcal{T}_{\mathrm{eval}}(\chi_m)|}
    \sum_{\tau\in\mathcal{T}_{\mathrm{eval}}(\chi_m)}
    \bar{r}_{\mathrm{prin},\tau}^{-}(\chi_m).
\end{equation}

Macro-averaged improvement summarizes aggregate behavioral contribution, while cross-scenario support also depends on whether positive improvements extend across multiple applicable tasks. Accordingly, the verification also records the tasks exhibiting a strictly positive marginal improvement:
\begin{equation}
    \mathcal{T}_{\mathrm{pos}}(\chi_m)
    =
    \left\{
        \tau\in\mathcal{T}_{\mathrm{eval}}(\chi_m)
        \;\middle|\;
        \bar{r}_{\mathrm{prin},\tau}^{+}(\chi_m)
        >
        \bar{r}_{\mathrm{prin},\tau}^{-}(\chi_m)
    \right\}.
\end{equation}

The verification cohort should also provide sufficient cross-scenario evidence and exhibit no material degradation in aggregate execution reliability. Let $\mathcal{I}_{\mathrm{ver}}(\chi_m)\in\{0,1\}$ denote a compact verification-validity predicate that summarizes the coverage and reliability checks applied to the paired results across the verification cohort. The candidate is accepted when the verification is valid, its macro-averaged improvement exceeds the principle-specific empirical margin $\epsilon_{\mathrm{CNA}}$, and this improvement is supported by at least $N_{\mathrm{pos}}^{\min}$ individual tasks:
\begin{equation}
    \mathcal{I}_{\mathrm{accept}}(\chi_m)=
    \begin{cases}
        \mathbb{I}\left[\mathcal{I}_{\mathrm{ver}}(\chi_m)=1\land\bar{r}_{\mathrm{prin}}^{+}(\chi_m)>\bar{r}_{\mathrm{prin}}^{-}(\chi_m)+\epsilon_{\mathrm{CNA}}\land|\mathcal{T}_{\mathrm{pos}}(\chi_m)|\ge N_{\mathrm{pos}}^{\min}\right], & \mathcal{T}_{\mathrm{eval}}(\chi_m)\neq\emptyset,\\
        0, & \mathcal{T}_{\mathrm{eval}}(\chi_m)=\emptyset.
    \end{cases}
\end{equation}

Upon successful verification, the general principle $\chi_m$ enters $P_{\text{sys}}$ with an initial utility equal to its outcome reward improvement normalized by the remaining outcome reward headroom:
\begin{equation}
    w_{\chi_m}^{(0)}
    =
    \frac{
        \bar{r}_{\mathrm{prin}}^{+}(\chi_m)
        -
        \bar{r}_{\mathrm{prin}}^{-}(\chi_m)
    }{
        1-\bar{r}_{\mathrm{prin}}^{-}(\chi_m)
    }.
\end{equation}
Concurrently, the framework records the source task and experience identifiers to identify associated entries for utility decay and preserve a reversible promotion link. The source experience remains in $K_{\text{exp}}$ as fallback evidence, while its promotion record is marked as shadowed to avoid repeated principle distillation from the same source.

Since verified principles occupy the bounded systemic context, their long-term retention is also governed by empirical utility. MUSE maintains $|P_{\text{sys}}|\le N_{\text{sys}}$, where $N_{\text{sys}}$ denotes the maximum principle capacity. Repeated task evaluation following parameter updates provides evidence for adjusting the retention priority of associated principles, as described in Appendix~\ref{subapp:parametric_internalization}. When incorporating a newly verified principle would exceed the capacity, MUSE ranks the stored principles by utility and retires the lowest-ranking entries.

\appmodule{Conflict-Aware Symbolic Integration}
\label{subapp:conflict_aware_symbolic_integration}

\begin{figure}[!t]
    \centering
    \includegraphics[width=\linewidth]{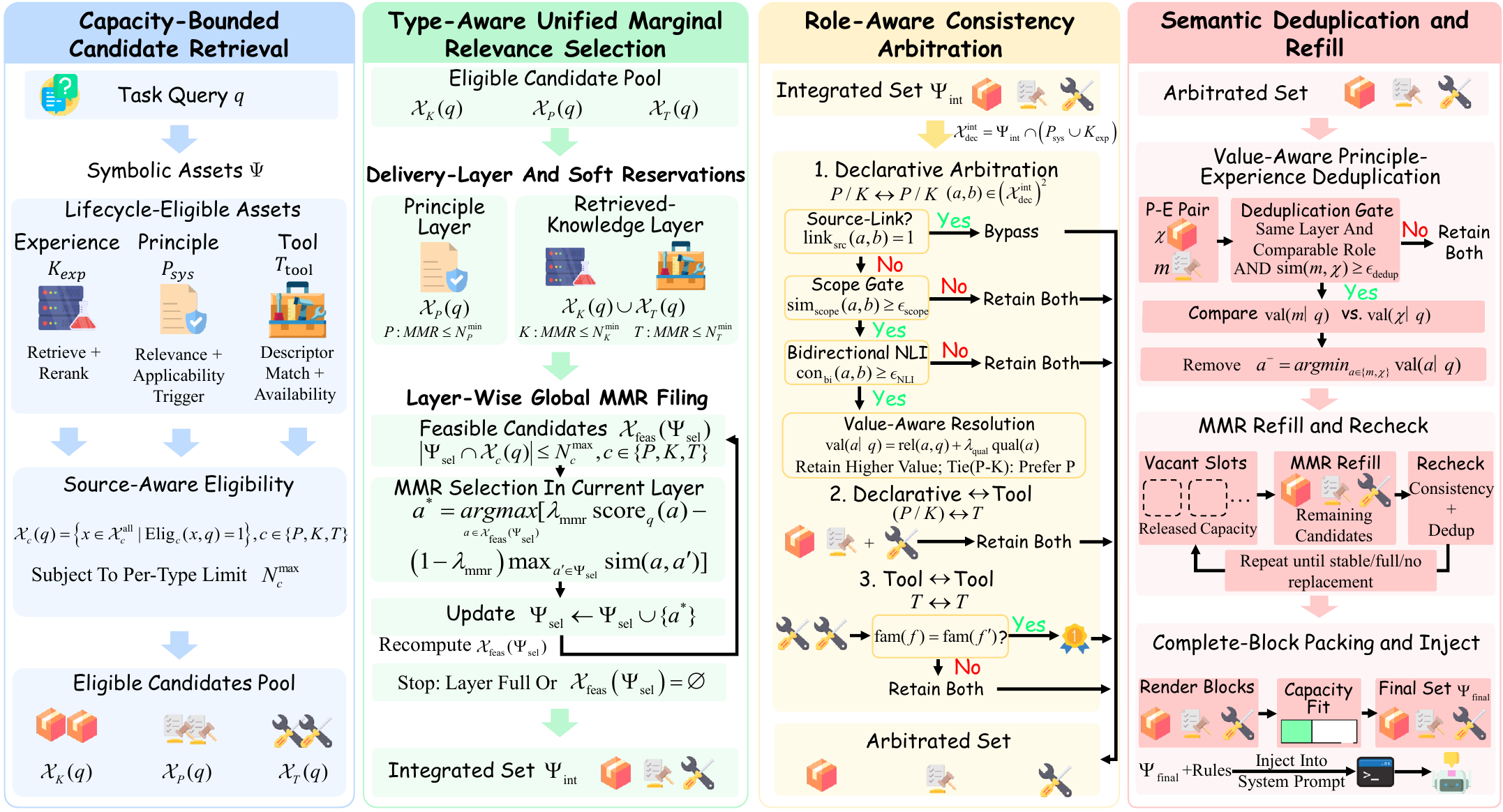}
    \caption{The workflow of conflict-aware symbolic integration. Source-aware retrieval constructs eligible candidate pools from existing experiences, principles, and functional tools. Selection applies a common marginal relevance criterion within each capacity-sharing layer, adopting soft within-type reservations followed by MMR filling. Principles retain a dedicated delivery capacity, while experiences and tools share the retrieved-knowledge budget. Role-aware arbitration compares conflicting declarative guidance by task-conditioned value and consolidates tools from the same operation family. Directly linked principle-source experience pairs are excluded from the contradiction assessment. Principle-experience deduplication applies when the entries occupy the same retrieval layer and serve comparable roles. MMR refill and repeated assessments update the surviving set before complete symbolic blocks are packed within the available context capacity.}
    \label{fig:framework_for_conflict_aware_symbolic_integration}
\end{figure}

Principle distillation enriches the symbolic space with a more abstract form of behavioral guidance. During deployment, these principles must coexist with situational experiences and executable tools within a bounded context, creating a coordination problem across heterogeneous symbolic representations. As defined previously, the symbolic space $\Psi$ encapsulates general principles $\chi \in P_{\text{sys}}$, situational experiences $m \in K_{\text{exp}}$, and functional tools $f \in T_{\text{tool}}$. Since these assets are extracted through distinct pathways and operate at different abstraction levels, they provide complementary guidance but may also introduce semantic redundancy or cross-source conflicts. Injecting all available symbolic entries indiscriminately may expose the agent to inconsistent guidance while diluting attention. The integration process consequently combines source-aware candidate retrieval, type-aware marginal relevance selection, role-aware consistency arbitration, and post-selection semantic deduplication and refill. This organization places the three symbolic sources within a common integration workflow while preserving their distinct applicability conditions, functional roles, and delivery capacities. Figure~\ref{fig:framework_for_conflict_aware_symbolic_integration} illustrates the resulting workflow of this symbolic integration mechanism across these coordinated processing stages.

\noindent \textbf{Source-Aware Candidate Retrieval.} Given a task query $q$, the mechanism first constructs preliminary candidate pools from each functional subspace within $\Psi$. Relevance is expressed differently across these symbolic sources: a principle is useful when its abstract structural trigger applies to the current task, an experience is useful when its situational precondition matches the execution context, and a functional tool is useful when its declared capability is both relevant and available for deployment. A unified retrieval formulation is retained while source-specific evidence determines eligibility. Let $\mathcal{X}_{c}^{\mathrm{all}}$ denote the available database for source $c \in \{P,K,T\}$. The corresponding eligible candidate pool is mathematically formulated as:
\begin{equation}
    \mathcal{X}_c(q)
    =
    \left\{
        x \in \mathcal{X}_{c}^{\mathrm{all}}
        \;\middle|\;
        \operatorname{Elig}_c(x,q)=1
    \right\},
    \qquad
    c \in \{P,K,T\},
\end{equation}
where $\operatorname{Elig}_c(x,q)$ denotes the source-aware eligibility assessment. For principles, it combines retrieval relevance with lifecycle validity and explicit structural-trigger applicability. For experiences, it reflects lifecycle eligibility and fine-grained query-conditioned retrieval and reranking evidence. For tools, it considers descriptor relevance together with lifecycle state and  current deployment availability. Candidate recall follows source-specific retrieval limits, while the subsequent selection is constrained by the delivery capacity of each layer.

\noindent \textbf{Type-Aware Unified Marginal Relevance Selection.} Eligible candidates may remain semantically repetitive, and their source-specific scores are not directly interchangeable. Selection consequently operates under a common marginal-relevance objective while preserving the capacity semantics of each deployment layer. Within a layer, soft type-specific reservations provide initial exposure for eligible symbolic sources, and the remaining capacity is filled according to marginal relevance. In the adopted deployment, principles retain a dedicated delivery capacity, whereas experiences and tools share the retrieved-knowledge budget. Thus, unified selection refers to a common selection criterion and coordinated context construction, without requiring all sources to consume the same capacity. Let $\Psi_{\mathrm{sel}}$ denote the candidates currently selected within a capacity-sharing layer, and let $\mathcal{X}_{\mathrm{feas}}(\Psi_{\mathrm{sel}})$ contain the remaining eligible candidates that satisfy both the source-specific and shared delivery limits. Here, $N_c^{\max}$ denotes the source-specific limit on selected entries. The selection process satisfies:
\begin{equation}
    \left|
        \Psi_{\mathrm{sel}}
        \cap
        \mathcal{X}_c(q)
    \right|
    \le
    N_c^{\max},
    \qquad
    c\in\{P,K,T\}.
\end{equation}

While feasible candidates remain, MUSE incorporates the candidate $a^{*}$ that balances its task-conditioned ranking score against its redundancy with candidates already selected within the same delivery layer. The redundancy term is zero before any candidate has been selected and otherwise reflects the greatest similarity to the selected entries:
\begin{equation}
\operatorname{red}(a,\Psi_{\mathrm{sel}})=
    \begin{cases}
    0, & \Psi_{\mathrm{sel}} = \emptyset,\\
    \max_{a' \in \Psi_{\mathrm{sel}}}\operatorname{sim}(a,a'), & \Psi_{\mathrm{sel}} \neq \emptyset.
    \end{cases}
\end{equation}
\begin{equation}
    a^{*}=\operatorname*{argmax}_{a\in\mathcal{X}_{\mathrm{feas}}(\Psi_{\mathrm{sel}})}\left[\lambda_{\mathrm{mmr}}\operatorname{score}_{q}(a)-(1-\lambda_{\mathrm{mmr}}) \operatorname{red}(a,\Psi_{\mathrm{sel}})\right].
\end{equation}
Here, $\operatorname{score}_{q}(a)$ denotes the source-aware ranking score of candidate $a$, and $\lambda_{\mathrm{mmr}}\in[0,1]$ balances task-conditioned relevance and diversity. For experiences, this score may combine retrieval and reranking evidence, while for principles and tools, it is derived from their corresponding retrieval representations. The union of the candidates selected from the respective delivery layers is denoted by $\Psi_{\mathrm{int}}$. Relevance and diversity determine contextual allocation, whereas logical compatibility introduces a separate coordination requirement. The resulting set subsequently enters the following consistency arbitration.

\noindent \textbf{Role-Aware Consistency Arbitration.} Semantic proximity can occur without functional conflict because symbolic assets may discuss the same task while serving different operational roles. Consistency arbitration therefore distinguishes declarative guidance, encompassing principles and experiences, from executable capabilities embodied by functional tools. For each pair of declarative candidates, the mechanism first applies a lightweight scope-similarity pre-filter to identify whether they address overlapping operational contexts. Only the retained pairs proceed to bidirectional natural language inference over their substantive claims. Let $\operatorname{scope}(a)$ describe the applicability conditions of candidate $a$, and let $\operatorname{claim}(a)$ represent its operational recommendation or constraint.

A principle and the situational experience from which it is directly promoted may convey closely related guidance without being logically incompatible. Let $\mathcal{X}_{\mathrm{dec}}^{\mathrm{int}}=\Psi_{\mathrm{int}} \cap \left(P_{\mathrm{sys}}\cup K_{\mathrm{exp}}\right)$ denote the selected declarative candidates, and let $\operatorname{link}_{\mathrm{src}}(a,b) \in \{0,1\}$ indicate whether $a$ and $b$ form a directly linked principle-source experience pair. For a declarative candidate pair $(a,b)$, its scope similarity is defined as:
\begin{equation}
    \operatorname{sim}_{\mathrm{scope}}(a,b)
    =
    \operatorname{sim}
    \left(
        \operatorname{scope}(a),
        \operatorname{scope}(b)
    \right).
\end{equation}
The corresponding bidirectional contradiction score is:
\begin{equation}
    \operatorname{con}_{\mathrm{bi}}(a,b)
    =
    \max\nolimits_{(a_1,a_2)\in\{(a,b),(b,a)\}}
    \operatorname{NLI}_{\mathrm{con}}\!\left(
        \operatorname{claim}(a_1),
        \operatorname{claim}(a_2)
    \right).
\end{equation}
The final contradiction indicator is then formulated as:
\begin{equation}
    \mathcal{I}_{\text{conflict}}(a,b)=\mathbb{I}\left[(a,b)\in\left(\mathcal{X}_{\text{dec}}^{\mathrm{int}}\right)^2 \land a\neq b\land\operatorname{link}_{\mathrm{src}}(a,b)=0\land\operatorname{sim}_{\mathrm{scope}}(a,b)\ge\epsilon_{\mathrm{scope}}\land\operatorname{con}_{\mathrm{bi}}(a,b)\ge\epsilon_{\mathrm{NLI}}\right].
\end{equation}
Here, $\operatorname{NLI}_{\mathrm{con}}(\cdot,\cdot)$ denotes the contradiction score returned by the NLI model. The bidirectional comparison captures contradictions that may be expressed asymmetrically across the two candidate claims.

When declarative guidance conflicts, task relevance and accumulated empirical utility provide evidence for selecting the entry to retain. MUSE compares the task-conditioned values defined in Equation~\ref{equ:symbolic_candidate_value} and retains the higher-value entry. When a principle and an experience have equal values, the principle has priority. Directly linked principle-source experience pairs are excluded from this contradiction assessment and remain subject to the role-dependent redundancy checks described below.

Functional tools follow a distinct coordination rule. A tool and a declarative asset are generally retained together within the same execution context during algorithmic arbitration because the former exposes an executable capability while the latter describes how that capability should be applied. Tool-to-tool redundancy is instead assessed through their operation-family correspondence, which reflects their functional roles and environment-facing operation structures. When two tools belong to the same operation family, the representative with stronger empirical quality is retained. This role-aware treatment separates genuine strategic incompatibility from the complementary relationship between executable operations and declarative guidance.

\noindent \textbf{Post-Selection Semantic Deduplication and Refill.} 
Consistency arbitration addresses incompatible guidance, while semantically compatible entries can still consume context through redundant information. The mechanism consequently examines semantic redundancy between the surviving principles and situational experiences. Since principles are abstracted from recurrently useful experiences, a selected experience may restate guidance already represented by a principle. Such overlap can remain valuable when the situational experience provides stronger relevance or empirical support for the current query. Accordingly, this semantic deduplication is applied only when the principle and experience participate in the same retrieval layer and serve comparable retrieval roles. For a given principle-experience pair $(\chi,m)$ meeting this condition and satisfying:
\begin{equation}
    \operatorname{sim}(m,\chi)
    \ge
    \epsilon_{\mathrm{dedup}},
\end{equation}
Subsequently, MUSE explicitly compares their task-conditioned values according to $\operatorname{val}(a\mid q)$ defined in Equation~\ref{equ:symbolic_candidate_value}, with $a\in\{m,\chi\}$. Consequently, the entry exhibiting lower overall task-conditioned value is removed. When their values are equal, the principle is retained, formulated as follows:
\begin{equation}
    a^{-}=
    \begin{cases}
        m, & \operatorname{val}(m\mid q)\le\operatorname{val}(\chi\mid q),\\
        \chi, & \operatorname{val}(m\mid q)>\operatorname{val}(\chi\mid q).
    \end{cases}
\end{equation}
This criterion allocates the shared retrieval capacity to the entry with the higher task-conditioned value estimate.

Conflict arbitration and semantic deduplication may release positions within the available context budget. MUSE refills these positions from the remaining eligible candidates through MMR against the surviving selected set. The newly introduced candidates undergo the same consistency and redundancy assessment, continuing until the collection becomes stable, the available budget is reached, or no admissible replacement remains.

The surviving entries are rendered as complete symbolic blocks so that applicability conditions and operational content remain available together. Each block is included only when its complete representation fits within the remaining token or character budget. The resulting context preserves the preceding arbitration decisions. Specifically,  conflicting declarative entries are compared by task-conditioned value, with a principle preferred when the compared values are equal. Tool descriptions specify callable interfaces, parameters, return structures, and expected side effects. Experiences contribute situational conditions and operational detail, subject to the same value-based comparison when declarative guidance conflicts.

Through conflict-aware unified symbolic integration, MUSE coordinates heterogeneous symbolic assets within a bounded deployment context. Source-aware eligibility and type-aware MMR define the initial candidate composition, while role-aware arbitration distinguishes strategic contradictions from complementary functional roles. Value-aware semantic deduplication and MMR-based refill further refine the selected collection, and complete-block packing maintains a concise and coherent symbolic context for subsequent task execution.

\appmodule{Parametric Internalization and Symbolic Entry Retirement}
\label{subapp:parametric_internalization}

\begin{figure}[!t]
    \centering
    \includegraphics[width=\linewidth]{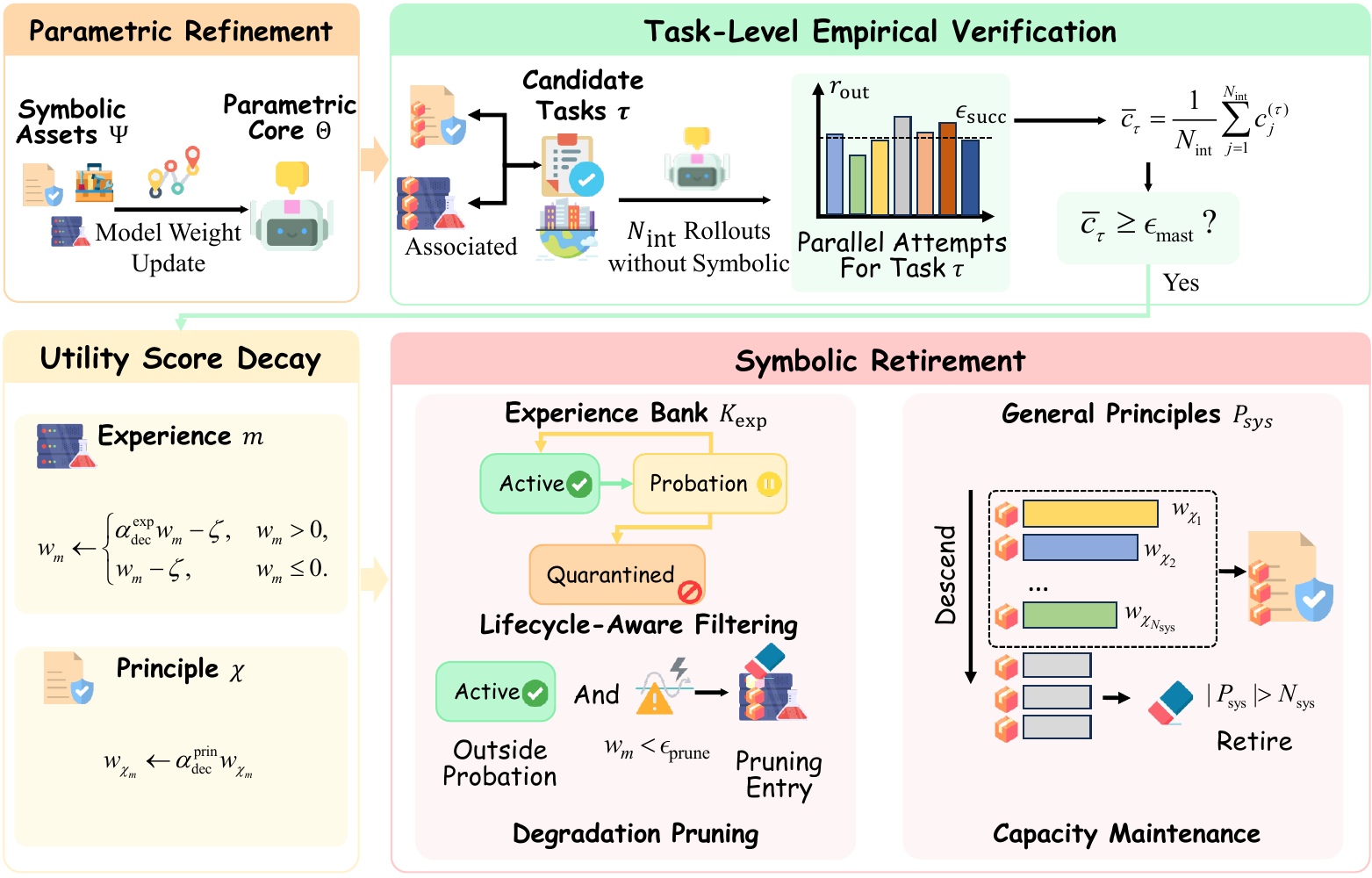}\vspace{-0.2cm}

    \caption{The synchronization workflow for parametric internalization and symbolic entry retirement. Following model weight updates, MUSE evaluates tasks associated with existing experiences and principles through repeated closed-book rollouts, without access to stored experiences, principles, or generated functional tools. Environment-provided operations remain available. When the empirical completion rate reaches the mastery threshold, component-specific utility decay applies only to experiences and principles associated with the verified task. Experience maintenance considers probation and subsequent utility evidence, while principle retirement follows utility ranking under the capacity bound. Generated functional tools retain their invocation-based maintenance.}
    \label{fig:framework_for_internalization}
\end{figure}

Parameter updates can change the contribution of explicit symbolic support during subsequent execution. Historical utility may therefore provide an incomplete basis for retention decisions. At the same time, an experience or principle can remain valuable in several contexts even when an associated task becomes solvable without stored symbolic support. MUSE consequently reassesses selected associated tasks through repeated closed-book execution and gradually adjusts retention priorities according to the resulting evidence. Figure~\ref{fig:framework_for_internalization} illustrates this connection between current task proficiency and symbolic maintenance.

Specifically, MUSE periodically collects the source tasks recorded by existing principles $\chi$ and samples a bounded subset of the tasks indexed by existing experiences $m$ for stochastic verification. For each selected task $\tau$, the agent performs $N_{\text{int}}$ verification rollouts without access to any stored experiences, principles, or generated functional tools, while environment-provided operations remain available. The mean completion rate $\bar{c}_{\tau}$ follows Equation~\ref{equ:internalization_completion} and characterizes current task proficiency under this evaluation condition.

When $\bar{c}_{\tau}$ reaches the mastery threshold $\epsilon_{\text{mast}}$, the utility updates in Equation~\ref{equ:symbolic_utility_decay} apply only to experiences and principles associated with task $\tau$. These gradual adjustments accommodate the possibility that an associated entry continues to contribute in other contexts. Generated functional tools remain governed by their invocation-based maintenance and do not receive utility decay from this verification procedure.

Subsequent maintenance follows the existing lifecycle and capacity criteria across the retained pools, with details provided in Appendix~\ref{subapp:experience_bank_refinement} and Appendix~\ref{subapp:general_principle_distillation}. For situational experiences residing in the experience bank $K_{\text{exp}}$, MUSE applies lifecycle-aware pruning: probationary entries remain temporarily retained for additional evidence, whereas confirmed low-utility entries outside probation can be removed. For general principles, utility decay changes their relative retention priority, and capacity-based maintenance retires the lowest-ranking entries when the pool exceeds $N_{\text{sys}}$. This process connects task-proficiency evidence following parameter updates to the continued allocation of explicit symbolic support.

\subsection{Evolutionary Objective \texorpdfstring{$\mathcal{O}$}{O} and Boundary \texorpdfstring{$\mathcal{B}$}{B}} \label{subapp:evolutionary_objective_and_boundary}
\setcounter{secnumdepth}{3}

\noindent \textbf{Overarching Evolutionary Objective ($\mathcal{O}$).}
Inner-component refinement and inter-component synchronization operate through heterogeneous empirical signals, motivating a shared criterion for assessing their contribution. MUSE evaluates the agent through empirical task performance, as specified in Section~\ref{subsec:evolutionary_objective_and_boundary}. Symbolic verification and utility tracking support component-level admission and retention, while task-level and step-level rewards guide parametric refinement through $\mathcal{M}$. These local signals differ in scope and scale. Utility estimates reflect evidence collected under component-specific verification and deployment conditions, while parametric training rewards also incorporate format and execution-length considerations. Their interpretation therefore remains specific to the corresponding update mechanism. Overall capability changes are assessed through task performance under consistent evaluation conditions. The corresponding optimization procedures are detailed in Appendix~\ref{subapp:optimization_scope_and_evolutionary_optimizer}.

\noindent \textbf{Supporting Capability Retention through Task-Specific Trajectory Replay ($\mathcal{B}$).} 
As adaptation moves toward newly encountered tasks, recent rollouts may provide limited coverage of previously acquired behavior. Improvement on the current task collection can therefore coexist with reduced training exposure to historical task requirements. This motivates the continued inclusion of historical behavioral evidence during parametric adaptation. MUSE consequently supports the capability-retention aspect of the boundary $\mathcal{B}$ through a task-specific trajectory replay mechanism that introduces historical behavioral evidence into subsequent parametric optimization.

Historical coverage is valuable when the retained executions provide eligible training evidence without excessive redundancy. The framework maintains a replay buffer $Q_{\text{buffer}}$ organized by source task $\tau$. Admission requires a valid training sample and source-specific replay eligibility. Oracle-task trajectories qualify through recorded task success or a configured outcome reward threshold. Synthetic-task trajectories qualify when their outcome rewards meet the source-specific threshold and their task identifiers agree with the recorded source-task identity and canonical-query hash. Duplicate action structures are identified through trajectory fingerprints before admission.

For each task $\tau$, stored trajectories are dynamically ranked according to their outcome rewards. When the per-task capacity $W_{\text{group}}$ is exceeded, the framework retains high-performing trajectories while reserving one slot, when capacity permits, for the lowest-scoring eligible trajectory. Retaining both score extremes can preserve outcome reward contrast among trajectories that already satisfy the admission requirements. Global limits on task count and trajectory count further bound storage. Among the retained tasks, capacity management preserves a high-scoring trajectory for each task before allocating the remaining slots according to outcome quality. Consequently, the lower-scoring trajectory retained during per-task pruning remains subject to subsequent global capacity management.

The storage capacity $W_{\text{group}}$ controls the historical evidence retained for each task, while $W$ specifies the final optimization group size. Their distinct roles separate buffer maintenance from the construction of training groups.

The replay buffer defines what historical evidence remains available, while the subsequent sampling process determines how this evidence participates in each new policy update. Current and historical task coverage create complementary replay requirements. Revisited tasks can benefit from earlier executions of the same task, while tasks absent from the current rollout collection require a separate route into training. The optimizer constructs a historical subset $Q_{\text{replay}}$ through two complementary replay paths.

For tasks already appearing in $Q_{\text{new}}$, it retrieves high-performing historical trajectories associated with the same task identity and excludes executions that duplicate the new trajectories. Per-task replay limits bound the historical contribution to each revisited task group. For historical tasks absent from $Q_{\text{new}}$, selected replay groups are eligible to contribute training loss only when at least two eligible trajectories with sufficient outcome reward contrast fit within the allocated replay budget. Selection prioritizes the highest-scoring and lowest-scoring trajectories before incorporating additional eligible examples. If these requirements are not met, any retained placeholder group consists of zero-loss replicas. The two paths respectively reinforce continuity within currently revisited tasks and preserve behavioral evidence from historical tasks not represented in the present rollout set. The replay-volume bound and the construction of $Q_{\text{mix}}$ follow Equation~\ref{equ:replay_mixture_and_bound}. This bound limits the selected historical trajectories before final group alignment. The actual replay volume can be smaller when eligible history, nonduplicate executions, or task-level outcome reward contrast are insufficient. Padding replicas introduced for structural alignment are excluded from the count of historical trajectories contributing training loss.

Mixing current and historical trajectories introduces a final structural requirement: each task group must remain compatible with the group-relative optimization procedure. For mixed groups of current tasks, screening excludes a group when all outcomes are near zero and no trajectory qualifies through recorded success or the configured outcome reward threshold. Identical positive outcome rewards alone do not remove a group, since step-level supervision can still distinguish its executions. MUSE then aligns each retained nonempty group $Q_{\text{mix}}^{(\tau)}$ to the predefined size $W$ according to the variance-aware balancing procedure in Appendix~\ref{subapp:foundation_model_weights}.

When the mixed volume exceeds the group capacity ($|Q_{\text{mix}}^{(\tau)}|>W$), the framework applies a truncation operator $\operatorname{prune}(\cdot)$ that retains high-scoring trajectories and reserves one position for the lowest-scoring eligible trajectory when outcome reward variation and capacity permit. This selection preserves evidence from different outcome reward levels within the available group capacity to support subsequent policy updates. When $|Q_{\text{mix}}^{(\tau)}|<W$, the optimizer applies the previously defined $\operatorname{pad}(\cdot)$ operation by iteratively replicating the highest-performing trajectory $H_{\text{best}}^{(\tau)}$. These replicas complete the group structure and are excluded from reward-normalization statistics. Their loss masks are zero for both the policy and KL terms. The aligned task group is formulated as:
\begin{equation}
    Q_{\text{opt}}^{(\tau)}
    =
    \begin{cases}
        \operatorname{prune}\left(
            Q_{\text{mix}}^{(\tau)}, W
        \right),
        &
        \text{if } |Q_{\text{mix}}^{(\tau)}| > W,
        \\[2mm]
        Q_{\text{mix}}^{(\tau)},
        &
        \text{if } |Q_{\text{mix}}^{(\tau)}| = W,
        \\[2mm]
        \operatorname{pad}\left(
            Q_{\text{mix}}^{(\tau)},
            W,
            H_{\text{best}}^{(\tau)}
        \right),
        &
        \text{if } |Q_{\text{mix}}^{(\tau)}| < W.
    \end{cases}
\end{equation}

The final optimization batch combines the retained and aligned task groups as $Q_{\text{opt}}=\bigcup_{\tau\in\mathcal{T}_{\text{ret}}}Q_{\text{opt}}^{(\tau)}$, where $\mathcal{T}_{\text{ret}}$ indexes retained task groups for batch construction. The multiset union preserves the multiplicity of padding copies. Current and historical executions then enter the same reward-processing and parametric-update procedures, with padding contributions excluded by the corresponding masks. This replay mechanism provides continued training support for historical behavior during adaptation to new tasks. Capability retention can be assessed through performance changes on the same historical tasks before and after adaptation under a consistent evaluation protocol, following Section~\ref{subsec:evaluation_metrics}. Together, $\mathcal{O}$ supplies the shared criterion for evaluating capability changes, while the replay mechanism supports the capability-retention aspect of $\mathcal{B}$ through historical behavioral evidence.

\section{Experimental Details} \label{app:experimental_details}
This appendix supplements Section~\ref{sec:experiments} with the experimental details for evaluating MUSE. Appendix~\ref{subapp:benchmark_details} introduces the adopted benchmark, including the interaction interface, task evaluation, and experimental partitions. Appendix~\ref{subapp:baselines} describes the baselines and MUSE variants, while Appendix~\ref{subapp:metrics} specifies the evaluation metrics and their calculation procedures. Appendix~\ref{subapp:evolution-evaluation} details the evolution and evaluation settings, including hyperparameters, resource budgets, and stopping conditions. Finally, Appendix~\ref{subapp:audit_protocols} presents the sampling procedures and annotation criteria for the human assessments of synthetic task quality and outcome rewards for synthetic tasks.

\subsection{Benchmark Details} \label{subapp:benchmark_details}

We adopt APPWorld~\citep{trivedi2024appworld} as the evaluative benchmark. It provides a controllable environment for evaluating agents on stateful application tasks. The benchmark comprises 750 tasks across 9 simulated applications exposed through 457 APIs. Each task specifies a natural-language objective and associated constraints. During execution, the agent generates Python code that invokes application APIs, and the environment returns execution outputs or errors as subsequent observations. Task execution involves combinations of API discovery, authentication, pagination, information aggregation, and coordination across applications, producing interaction trajectories that record both successful operations and execution difficulties.

Each execution begins in a fresh task-specific environment instance, while persistent agent components are retained or updated according to the evolution procedure in Section~\ref{subsec:muse_evolution_procedure}. Task completion is assessed through programmatic, state-based unit tests that examine the required application states and unintended state changes. These tests accommodate different valid execution procedures and provide environment-grounded criteria for evaluating task performance as the agent configuration evolves.

The full oracle interaction pool comprises 90 tasks in the APPWorld training split. Available oracle tasks provide interaction trajectories and environment-provided outcome rewards. Configurations with supplementary supervision additionally incorporate synthetic tasks constructed through the supervision generator in Section~\ref{subsec:supervision_generator}. Agent configurations are evaluated on the same 168 tasks from the \texttt{test\_normal} split across evolution rounds. Within our experimental protocol, this split serves as a proxy for the operating environment during continuous evolution. Execution trajectories provide capability evidence for metacognitive diagnosis, while evaluation scores guide configuration selection and recovery. The same task collection therefore supports both evolutionary feedback and the reported performance measurements, characterizing closed-loop adaptation across rounds. The available oracle subsets, rollout allocation, evaluation frequency, and stopping conditions are specified in Appendix~\ref{subapp:evolution-evaluation}.

\subsection{Baselines and MUSE Variants} \label{subapp:baselines}

We compare MUSE with representative agent evolution methods covering symbolic adaptation and model weight refinement. The APPWorld results for Reflexion, MemP, ReasoningBank, and SkillMentor in Table~\ref{tab:overall_adaptivity} are adopted from~\citet{bao2026skillmentor}, which evaluates these methods with the same Qwen3.5-9B~\citep{qwen2026qwen35} as the task executor. Vanilla GRPO and the MUSE variants are evaluated within our agent evolution framework. Their configurations are described below, with numerical settings provided in Appendix~\ref{subapp:evolution-evaluation}.

\noindent $\blacktriangleright$ \textbf{Baselines.} 

\noindent \textbf{(i) Reflexion.} Reflexion~\citep{shinn2023reflexion} adapts task execution through linguistic reflections retained in episodic memory. Following an evaluated attempt, a self-reflection component examines the interaction trajectory and feedback to summarize execution difficulties and propose corrective guidance. The stored reflections enter the context of subsequent attempts, supporting behavioral adaptation while executor weights remain fixed.

\noindent \textbf{(ii) MemP.} MemP~\citep{fang2026memp} organizes procedural memory through construction, retrieval, and updating. Its memory representations encompass interaction trajectories, abstract procedural descriptions, and combinations of these forms. Relevant entries are retrieved to guide subsequent tasks. Continued interaction supports expansion, revision, and pruning of the repository, with adaptation concentrated in persistent procedural memory.

\noindent \textbf{(iii) ReasoningBank.} ReasoningBank~\citep{ouyang2026reasoningbank} constructs reasoning memories from successful and failed trajectories. An LLM-based evaluator assesses task completion, after which successes yield reusable strategies and failures yield corrective lessons. Memory items contain descriptions and procedural guidance. Query-conditioned retrieval supplies relevant items during later execution, while the executor remains fixed.

\noindent \textbf{(iv) SkillMentor.} SkillMentor~\citep{bao2026skillmentor} pairs a frozen executor with a trainable mentor that diagnoses capability blind spots and proposes corrective skills. Diagnostic comparisons guide additions, revisions, and consolidation within an external skill repository. Candidate skills are evaluated through their contribution to executor performance and retrieved for subsequent tasks. The executor remains fixed, while the auxiliary mentor undergoes parameter updates based on diagnostic-oriented training.

\noindent \textbf{(v) Vanilla GRPO.} Vanilla GRPO~\citep{shao2024deepseekmath} provides an outcome-supervised reference for parametric agent evolution. The agent collects multiple trajectories for each oracle task, and environment-provided outcome rewards determine group-relative advantages. The resulting trajectory-level advantage is assigned to the generated action tokens for subsequent model weight updates. This configuration operates without symbolic refinement, supplementary synthetic tasks, or prefix-potential supervision.

\noindent $\blacktriangleright$ \textbf{MUSE Component Variants.}

Following Section~\ref{subsec:experimental_setup}, the component variants distinguish the refinement channels enabled within MUSE. These variants operate without metacognitive gating, with task sources specified separately.

\noindent \textbf{(i) MUSE (Symbolic).} This variant enables refinement of experiences $K_{\mathrm{exp}}$, principles $P_{\mathrm{sys}}$, and functional tools $T_{\mathrm{tool}}$ while keeping foundation model weights fixed. Experiences provide situational guidance, principles are distilled from validated experiences and their deployment contexts, and tools encapsulate recurring procedures. The component comparisons in Table~\ref{tab:symbolic_contribution} additionally enable selected subsets of these symbolic refinement channels.

\noindent \textbf{(ii) MUSE (Parametric).} This variant enables model weight refinement with prefix-potential supervision and keeps symbolic refinement disabled. Outcome rewards and prefix-potential signals enter the reward integration and advantage estimation procedures described in Section~\ref{subsec:optimization_scope_and_evolutionary_optimizer} and Appendix~\ref{subapp:foundation_model_weights}. Oracle tasks and supplementary synthetic tasks provide the alternative task-source settings examined in Table~\ref{tab:supplementary_supervision_effectiveness}.

\noindent \textbf{(iii) MUSE (Joint).} This variant combines symbolic and parametric refinement following Section~\ref{subsec:optimization_scope_and_evolutionary_optimizer}. Symbolic updates provide revised guidance and procedures for subsequent interaction, while model weight refinement directly adapts task-execution behavior. Inter-component synchronization coordinates principle distillation, symbolic integration, and reassessment of associated symbolic guidance following parameter updates.

\noindent \textbf{Supervision Settings.} Task sources and prefix-potential supervision are specified separately from the component-variant names. Oracle tasks provide environment-based outcome rewards. Supplementary synthetic tasks are constructed and assessed through the supervision generator in Section~\ref{subsec:supervision_generator}. When enabled, prefix-potential supervision supplies step-level signals for both oracle and synthetic tasks. These settings distinguish the contribution of additional task coverage from the provision of process supervision.

\noindent \textbf{Full MUSE and Gating Comparisons.} MUSE denotes the full configuration, with all metacognitive gating, supplementary synthetic tasks, and joint symbolic and parametric adaptation enabled. The gating comparisons in Section~\ref{subsec:metacognitive_gating_evaluation} retain supplementary supervision and the same eligible component space. The Always-on setting enables all eligible refinement channels in every round. The Metacognitive setting follows the capability diagnosis and component selection mechanism in Section~\ref{subsec:evolutionary_gating}. The Random setting is constructed from the recorded Metacognitive selections. Model weight refinement remains enabled throughout all 8 rounds, matching the recorded Metacognitive run, while symbolic components are selected randomly under the same per-round component counts. The construction of this comparison and its resource accounting are detailed in Appendix~\ref{subapp:evolution-evaluation}.

\subsection{Evaluation Metrics} \label{subapp:metrics}

This subsection defines the measures reported in Section~\ref{sec:experiments} and summarized in Appendix~\ref{subapp:notation_evaluation_measures}. Metrics are calculated separately for each experimental configuration and foundation model scale. Task success rates, audit rates, and exploration rollout reductions are reported as percentages. EvoGain and Evo-AUC are expressed in percentage points, while Avg. Steps is measured in interaction steps.

\subsubsection{Task Performance and Evolution Dynamics}
\label{subapp:metrics_task_performance}

Let $\mathrm{SR}_t$ denote the task success rate of the candidate configuration reported at evolution round $t$. Round~0 denotes the shared Base Model reference at the corresponding model scale. Let $t_{\mathrm{last}}$ denote the last evolution round. For each evaluated agent instance, $N_{\mathrm{eval}}$ is the number of evaluated task executions, $r_{\mathrm{out},i}^{(t)}$ is the environment-provided outcome reward, and $L_{t,i}$ is the recorded number of interaction steps.

\noindent \textbf{Task Success Rate (SR $\uparrow$).} On APPWorld~\citep{trivedi2024appworld}, a task execution succeeds when all the benchmark evaluation criteria are satisfied. Full completion corresponds to an outcome reward of 1. SR is calculated as:
\begin{equation}
    \label{equ:metric_task_sr}
    \mathrm{SR}_t
    =
    \frac{100}{N_{\mathrm{eval}}}
    \sum_{i=1}^{N_{\mathrm{eval}}}
    \mathbb{I}\left[r_{\mathrm{out},i}^{(t)}=1\right].
\end{equation}
Partial completion contributes to the outcome reward but does not count as task success.

\noindent \textbf{Best SR ($\uparrow$) and Final SR ($\uparrow$).} Best SR is the highest recorded SR within the observation horizon, including the Base Model reference. When several rounds attain the same maximum, the earliest round determines the corresponding agent instance. Final SR reports performance at the last reported evolution round:
\begin{equation}
    \label{equ:metric_best_final_sr}
    \begin{aligned}
        \mathrm{Best\ SR}
        &= \max_{0\leq t\leq t_{\mathrm{last}}}\mathrm{SR}_t,\\
        t_{\mathrm{best}}
        &= \min\left\{
            t\in\{0,\ldots,t_{\mathrm{last}}\}
            \mid \mathrm{SR}_t=\mathrm{Best\ SR}
        \right\},\\
        \mathrm{Final\ SR}
        &= \mathrm{SR}_{t_{\mathrm{last}}}.
    \end{aligned}   
\end{equation}

\noindent \textbf{Peak Improvement: EvoGain and $\Delta\mathrm{SR}$ ($\uparrow$).} Both metrics denote the absolute improvement in Best SR relative to the corresponding pre-evolution Base Model reference at round~0:
\begin{equation}
    \label{equ:metric_peak_improvement}
    \mathrm{EvoGain}
    =
    \Delta\mathrm{SR}
    =
    \mathrm{Best\ SR}-\mathrm{SR}_0.    
\end{equation}
Distinct from the single-round $\mathrm{EvoGain}_t$ in Section~\ref{subsec:evaluation_metrics}, this metric measures peak cumulative improvement.

\noindent \textbf{Evolutionary Area under the Curve (Evo-AUC $\uparrow$).} Evo-AUC summarizes performance improvement relative to the Base Model across the observation horizon. Success rates between consecutive rounds are linearly interpolated, and the signed area is calculated through trapezoidal integration. For an early-terminated run, the last reported SR is carried forward to round~8 for fixed-horizon aggregation. Define $\widetilde{\mathrm{SR}}_t=\mathrm{SR}_{\min(t,t_{\mathrm{last}})}$ for $t\in\{0,\ldots,8\}$. Then:
\begin{equation}
    \label{equ:metric_evo_auc}
    \mathrm{Evo\text{-}AUC}
    =
    \frac{1}{8}
    \sum_{t=1}^{8}
    \left[
        \frac{
            \widetilde{\mathrm{SR}}_{t-1}
            +
            \widetilde{\mathrm{SR}}_t
        }{2}
        -
        \mathrm{SR}_0
    \right].
\end{equation}
The carried-forward values provide a statistical extension after termination. Missing intermediate evaluations leave Evo-AUC undefined. Positive and negative values indicate average performance above and below the reference, respectively. Evo-AUC summarizes improvement across the horizon, while the evolution curves and Final SR characterize its development and retention across reported agent instances.

\noindent \textbf{Average Execution Steps (Avg. Steps $\downarrow$).} Avg. Steps measures the mean number of interaction steps across all evaluated tasks, including successful and unsuccessful executions. The experimental tables report this value for the earliest agent instance attaining Best SR. Its change relative to the Base Model is denoted by $\Delta\mathrm{Steps}$:
\begin{equation}
    \label{equ:metric_execution_steps}
    \begin{aligned}
        \mathrm{Avg.\ Steps}_t
        &=
        \frac{1}{N_{\mathrm{eval}}}
        \sum_{i=1}^{N_{\mathrm{eval}}}L_{t,i},\\
        \mathrm{Avg.\ Steps}
        &=
        \mathrm{Avg.\ Steps}_{t_{\mathrm{best}}},\\
        \Delta\mathrm{Steps}
        &=
        \mathrm{Avg.\ Steps}_{t_{\mathrm{best}}}
        -
        \mathrm{Avg.\ Steps}_0.
    \end{aligned}
\end{equation}
Negative $\Delta\mathrm{Steps}$ indicates fewer interactions per evaluated task. Execution steps are interpreted jointly with SR to assess the interaction demands associated with task performance.

\subsubsection{Exploration Rollout Cost}
\label{subapp:metrics_exploration_cost}

\noindent \textbf{Exploration Rollouts ($\downarrow$).} In the comparisons of exploration rollout costs, Exploration Rollouts counts fresh trajectories collected for symbolic refinement and model weight adaptation. Repeated executions of the same task count as separate trajectories. Evaluation, historical replay, supervision construction, and additional verification are excluded. Let $N_{\mathrm{cost}}^{(t)}$ denote the exploration rollout count in round $t$. The total and cumulative counts are:
\begin{equation}
    \label{equ:metric_exploration_rollouts}
    \begin{aligned}
        N_{\mathrm{cost}}
        &=
        \sum_{t=1}^{8}N_{\mathrm{cost}}^{(t)},\\
        N_{\mathrm{cost}}^{(\leq t)}
        &=
        \sum_{j=1}^{t}N_{\mathrm{cost}}^{(j)}.
    \end{aligned}
\end{equation}
Table~\ref{tab:gating_effectiveness} reports the total exploration count, while Figure~\ref{fig:gating_rollouts_per_round} reports the per-round exploration counts. This measure quantifies exploration demand within the component update channels. Broader resource accounting in Section~\ref{subsec:evaluation_metrics} additionally considers other interaction and computation costs.

\noindent \textbf{Reduction ($\uparrow$).} Reduction measures the percentage decrease in total Exploration Rollouts relative to the Always-on setting over the same evolution horizon:
\begin{equation}
    \label{equ:metric_rollout_reduction}
    \mathrm{Reduction}
    =
    100\left(
        1-
        \frac{
            N_{\mathrm{cost}}
        }{
            N_{\mathrm{cost}}^{\text{Always-on}}
        }
    \right).
\end{equation}

\noindent \textbf{Performance against Cumulative Exploration Rollouts.} In Figure~\ref{fig:gating_best_observed_sr}, each point pairs the SR of an evaluated agent instance with $N_{\mathrm{cost}}^{(\leq t)}$ at the corresponding round. The step line records the highest SR observed up to that evaluation. The earliest point attaining Best SR identifies the exploration budget required to attain the peak.

\subsubsection{Synthetic Task Quality}
\label{subapp:metrics_task_quality}

Table~\ref{tab:synthetic_task_quality} reports human assessments separately for generated candidates and admitted synthetic tasks. For each group, Assessable / Total gives the number of tasks with determinate judgments on all three quality criteria out of the total audited tasks. All four rates share this common assessable subset.

\noindent \textbf{Goal Clarity ($\uparrow$).} A task passes when its query specifies an understandable objective and the necessary constraints.

\noindent \textbf{Environmental Feasibility ($\uparrow$).} A task passes when environmental evidence supports its required operations.

\noindent \textbf{Reference Alignment ($\uparrow$).} A task passes when its reference solution and trajectory match task requirements.

\noindent \textbf{Overall Validity ($\uparrow$).} A task passes when all three quality criteria above pass.

Let $N_{\mathrm{qual}}$ denote the number of assessable tasks. Let $N_{\mathrm{GC}}$, $N_{\mathrm{EF}}$, and $N_{\mathrm{RA}}$ count tasks passing the corresponding criteria within this subset, and let $N_{\mathrm{valid}}$ count tasks passing all three. The rates are:
\begin{equation}
    \begin{aligned}
        \mathrm{Goal\ Clarity}
        &=
        100\frac{N_{\mathrm{GC}}}{N_{\mathrm{qual}}},\\
        \mathrm{Environmental\ Feasibility}
        &=
        100\frac{N_{\mathrm{EF}}}{N_{\mathrm{qual}}},\\
        \mathrm{Reference\ Alignment}
        &=
        100\frac{N_{\mathrm{RA}}}{N_{\mathrm{qual}}},\\
        \mathrm{Overall\ Validity}
        &=
        100\frac{N_{\mathrm{valid}}}{N_{\mathrm{qual}}}.
    \end{aligned}
    \label{equ:metric_synthetic_task_quality}
\end{equation}
Sampling procedures, annotation criteria, and evidence presented during review are detailed in Appendix~\ref{subapp:audit_protocols}.

\subsubsection{Outcome Rewards for Synthetic Tasks}
\label{subapp:metrics_outcome_rewards}

Table~\ref{tab:synthetic_outcome_reward_reliability} compares outcome rewards from $\mathcal{R}_{\mathrm{syn}}$ with human completion judgments. Guided and free executions are assessed separately. Annotations of complete define the positive class, partial and incomplete define the negative class, and uncertain annotations are excluded. A reward-based prediction is positive exactly when $r_{\mathrm{out},i}^{\mathrm{syn}}=1$. Assessable / Total counts executions with determinate judgments out of all audited executions in the group.

True positives (TP) receive full reward and are judged complete. False positives (FP) receive full reward but are judged partial or incomplete. True negatives (TN) receive less than full reward and are judged partial or incomplete. False negatives (FN) receive less than full reward but are judged complete. The four rates are:
\begin{equation}
    \begin{aligned}
        \mathrm{Completion\ Agreement}\;(\uparrow)
        &=
        100\frac{\mathrm{TP}+\mathrm{TN}}
        {\mathrm{TP}+\mathrm{FP}+\mathrm{TN}+\mathrm{FN}},\\
        \mathrm{Full\text{-}Completion\ Precision}\;(\uparrow)
        &=
        100\frac{\mathrm{TP}}{\mathrm{TP}+\mathrm{FP}},\\
        \mathrm{Full\text{-}Completion\ Recall}\;(\uparrow)
        &=
        100\frac{\mathrm{TP}}{\mathrm{TP}+\mathrm{FN}},\\
        \mathrm{False\text{-}Positive\ Rate}\;(\downarrow)
        &=
        100\frac{\mathrm{FP}}{\mathrm{FP}+\mathrm{TN}}.
    \end{aligned}
    \label{equ:metric_synthetic_outcome_reliability}
\end{equation}
Completion Agreement measures agreement on full completion. Full-Completion Precision measures the reliability of full-reward assignments, while Full-Completion Recall measures their coverage of human-complete executions. False-Positive Rate measures the frequency of full-reward assignments among human-noncomplete executions. Metrics with zero denominators are reported as N/A. The annotation procedures are detailed in Appendix~\ref{subapp:audit_protocols}.

\subsection{Evolution and Evaluation Settings}
\label{subapp:evolution-evaluation}

\noindent \textbf{Shared Evolution Protocol.}
The main experiments initialize the agent with Qwen3.5-9B~\citep{qwen2026qwen35}, while the symbolic component comparisons additionally consider Qwen3.5-27B. Each configuration follows Algorithm~\ref{alg:muse_evolution} under a maximum budget of 8 evolution rounds. Selected symbolic components collect fresh trajectories for their refinement, and enabled model weight updates collect additional trajectories under the current symbolic configuration. Parametric refinement follows the GRPO procedure~\citep{shao2024deepseekmath} in Section~\ref{subsec:optimization_scope_and_evolutionary_optimizer} and Appendix~\ref{subapp:foundation_model_weights}, with VERL supporting optimization and vLLM supporting inference.

Table~\ref{tab:muse_evolution_evaluation_settings} summarizes the evolution and evaluation settings. Table~\ref{tab:muse_component_supervision_settings}, Table~\ref{tab:muse_symbolic_refinement_hyperparameters}, and Table~\ref{tab:muse_synchronization_replay_hyperparameters} specify the method hyperparameters. Component-specific thresholds are listed within their corresponding groups.

\begin{table}[!t]
    \centering
    \small
    \renewcommand{\arraystretch}{1.03}
    \setlength{\tabcolsep}{4pt}
    \caption{Parametric refinement and evaluation settings on APPWorld. MUSE entries apply when model weight refinement is enabled. Sampled-task limits are upper bounds for each refinement phase.}
    \label{tab:muse_evolution_evaluation_settings}

    \begin{tabularx}{\linewidth}{
        @{}
        >{\raggedright\arraybackslash}X
        >{\centering\arraybackslash}p{0.25\linewidth}
        >{\centering\arraybackslash}p{0.22\linewidth}
        @{}
    }
        \toprule
        \textbf{Setting}
        & \textbf{MUSE}
        & \textbf{Vanilla GRPO} \\
        \midrule

        \rowcolor{blue!7}
        \multicolumn{3}{c}{\textit{Evolution and Interaction}} \\

        Foundation model
        & Qwen3.5-9B
        & Qwen3.5-9B \\

        Maximum evolution rounds
        & 8
        & 8 \\

        Maximum sampled tasks
        & 90 oracle-only, 128 with synthetic tasks
        & 90 \\

        Refinement rollouts per sampled task
        & 8
        & 8 \\

        Maximum interaction steps $L_{\mathrm{max}}$
        & 30
        & 30 \\

        Rollout decoding temperature
        & 0.9
        & 0.9 \\

        Query batch size
        & 32
        & 32 \\

        \midrule
        \rowcolor{orange!8}
        \multicolumn{3}{c}{\textit{Model Weight Refinement}} \\

        Epochs per activated weight update
        & 2
        & 2 \\

        PPO epochs per collected batch
        & 1
        & 1 \\

        Optimizer
        & AdamW
        & AdamW \\

        Learning rate
        & $2\times10^{-6}$
        & $2\times10^{-6}$ \\

        Weight decay
        & $10^{-2}$
        & $10^{-2}$ \\

        PPO minibatch size $W_{\mathrm{mb}}$
        & 8
        & 8 \\

        PPO microbatch size per GPU
        & 1
        & 1 \\

        Clipping margins
        $(\epsilon_{\mathrm{low}},\epsilon_{\mathrm{high}})$
        & $(0.20,0.28)$
        & $(0.20,0.20)$ \\

        Reference KL coefficient $\beta$
        & $10^{-3}$
        & $10^{-3}$ \\

        Gradient-clipping threshold
        & 1.0
        & 1.0 \\

        Loss aggregation
        & Token mean
        & Token mean \\

        \midrule
        \rowcolor{gray!10}
        \multicolumn{3}{c}{\textit{Reward Integration and Replay}} \\

        Outcome transformation $\varphi(x)$
        & $x^2$
        & $x$ \\

        Length-penalty weight $\lambda_{\mathrm{len}}$
        & 0.02
        & 0 \\

        Potential weight $\lambda_{\mathrm{pote}}$
        & 0.4
        & 0 \\

        Replay coefficient $\lambda_{\mathrm{replay}}$
        & 0.1
        & 0 \\

        \midrule
        \rowcolor{green!7}
        \multicolumn{3}{c}{\textit{Evaluation}} \\

        Evaluation split
        & \texttt{test\_normal}
        & \texttt{test\_normal} \\

        Evaluation tasks
        & 168
        & 168 \\

        Evaluation decoding temperature
        & 0
        & 0 \\

        Evaluation frequency
        & Initial and after each round
        & Initial and after each round \\

        \bottomrule
    \end{tabularx}
\end{table}

\noindent \textbf{Component and Supervision Settings.}
Symbolic refinement follows the extraction, verification, and maintenance procedures in Appendix~\ref{subapp:optimization_scope_and_evolutionary_optimizer}. Supplementary synthetic tasks are constructed within each run through the supervision generator in Appendix~\ref{subapp:supervision_generator}. Their active pool changes as tasks are validated, admitted, and retired. Guided and free assessments evaluate task suitability for the current agent, while prefix-potential supervision applies to both oracle and synthetic executions. The following tables report the corresponding thresholds, capacities, and coefficients. Generation budgets and pool capacities denote upper bounds, with realized counts determined by the available interaction evidence and selection procedures.

\begin{table}[!t]
    \centering
    \small
    \renewcommand{\arraystretch}{1.05}
    \setlength{\tabcolsep}{5pt}
    \caption{Hyperparameters for the evolutionary gating and the supplementary supervision generation in our experiments. Exploration and candidate budgets denote upper bounds.}
    \label{tab:muse_component_supervision_settings}

    \begin{tabularx}{\linewidth}{
        @{}
        >{\raggedright\arraybackslash}X
        >{\centering\arraybackslash}p{0.30\linewidth}
        @{}
    }
        \toprule
        \textbf{Setting} & \textbf{Value} \\
        \midrule

        \rowcolor{blue!7}
        \multicolumn{2}{c}{\textit{Evolutionary Gating}} \\

        Sliding-window size $|\mathcal{W}_t|$
        & 32 \\

        Completion-rate threshold $c_{\mathrm{th}}$
        & 0.60 \\

        Interaction accumulation threshold $V_{\mathrm{acc}}$
        & 200 \\

        Maximum diagnostic cases $|\tilde{\mathcal{W}}_t|$
        & 10 \\

        \midrule
        \rowcolor{orange!8}
        \multicolumn{2}{c}{\textit{Supplementary Task Construction and Selection}} \\

        Exploration executions per training environment
        & 2 \\

        Nominal exploration budget $N_{\mathrm{explore}}$
        & 180 \\

        Candidate validation budget
        & 360 \\

        Active task-pool capacity $N_{\mathrm{syn}}$
        & 90 \\

        Guided and free executions $(N_{\mathrm{g}},N_{\mathrm{f}})$
        & $(2,6)$ \\

        Minimum guided success rate $\epsilon_{\mathrm{g,succ}}$
        & $1.0$ \\

        Free success-rate range for regular tasks $[\epsilon_{\mathrm{f,low}},\epsilon_{\mathrm{f,high}}]$
        & $[1/6,1/2]$ \\

        Free successful executions for challenge tasks
        & $0$ \\

        Minimum peak free outcome reward for challenge tasks $\epsilon_{\mathrm{f,partial}}$
        & $0.20$ \\

        Minimum guidance gaps
        $(\epsilon_{\mathrm{gap}},\epsilon_{\mathrm{gap,ch}})$
        & $(0.20, 0.20)$ \\

        Maximum guidance gap
        $\epsilon_{\mathrm{gap,max}}$
        & $0.85$ \\

        Outcome reward threshold for success in mastery assessment $\epsilon_{\mathrm{syn,succ}}$
        & $0.60$ \\

        Complete sampled rounds for mastery assessment $n_{\mathrm{master}}$
        & $2$ \\

        Minimum success rate in each assessment round $\epsilon_{\mathrm{hist,SR}}$
        & $0.75$ \\

        Minimum pooled mean outcome reward across assessment rounds $\epsilon_{\mathrm{hist,mean}}$
        & $0.75$ \\

        Query-only retirement verification rollouts
        & $2$ \\

        Minimum verification success rate
        & $1.0$ \\

        Minimum verification mean outcome reward $\epsilon_{\text{retire}}$
        & $0.75$ \\

        \midrule
        \rowcolor{green!7}
        \multicolumn{2}{c}{\textit{Prefix-Potential Process Supervision}} \\

        Number of potential bins $N_{\mathrm{bin}}$
        & 6 \\

        Potential anchors $v(b_k)$
        & $0,0.2,0.4,0.6,0.8,1$ \\

        Bin-scoring temperature $\kappa_{\mathrm{eval}}$
        & 0.50 \\

        Log-potential floor $\epsilon$
        & 0.01 \\

        \bottomrule
    \end{tabularx}
\end{table}

\begin{table}[!t]
    \centering
    \small
    \renewcommand{\arraystretch}{1.03}
    \setlength{\tabcolsep}{5pt}
    \caption{Hyperparameters for symbolic refinement. Shared verification settings apply to experiences and tools.}
    \label{tab:muse_symbolic_refinement_hyperparameters}

    \begin{tabularx}{\linewidth}{
        @{}
        >{\raggedright\arraybackslash}X
        >{\centering\arraybackslash}p{0.27\linewidth}
        @{}
    }
        \toprule
        \textbf{Setting} & \textbf{Value} \\
        \midrule

        \rowcolor{gray!10}
        \multicolumn{2}{c}{\textit{Shared Paired Verification}} \\

        Saturation threshold $r_{\mathrm{sat}}$
        & 0.99 \\

        Relative step-reduction thresholds
        $(\epsilon_{\mathrm{step}}^{\mathrm{sat}},
          \epsilon_{\mathrm{step}}^{\mathrm{tie}})$
        & $(0.20,0.15)$ \\

        Initial efficiency utilities
        $(w_{\mathrm{eff}}^{\mathrm{sat}},
          w_{\mathrm{eff}}^{\mathrm{tie}})$
        & $(0.50,0.20)$ \\

        \midrule
        \rowcolor{blue!7}
        \multicolumn{2}{c}{\textit{Experience Refinement}} \\

        Positive-anchor threshold $\epsilon_{\mathrm{anch}}$
        & 0.50 \\

        Precondition similarity range
        $[\epsilon_{\mathrm{lower}},\epsilon_{\mathrm{upper}}]$
        & $[0.55,0.80]$ \\

        Novelty threshold $\epsilon_{\mathrm{sim}}$
        & 0.80 \\

        Verification rollouts per arm $N_{\mathrm{veri}}$
        & 4 \\

        Outcome reward improvement margin $\epsilon_{\mathrm{CNA}}$
        & 0.15 \\

        Comparable-reward threshold $\epsilon_{\mathrm{tie}}$
        & 0.03 \\

        Utility adjustment coefficient $\alpha_m$
        & 0.50 \\

        Retrieval-frequency weight $\lambda_V$
        & 0.50 \\

        Experience-bank capacity $N_{\mathrm{exp}}$
        & 500 \\

        \midrule
        \rowcolor{orange!8}
        \multicolumn{2}{c}{\textit{General Principle Distillation}} \\

        Minimum experience utility $\epsilon_{\mathrm{score}}$
        & 0.20 \\

        Minimum retrieval frequency $\epsilon_{\mathrm{freq}}$
        & 0.02 \\

        Directive similarity threshold $\epsilon_{\mathrm{sys}}$
        & 0.68 \\

        Outcome reward improvement margin $\epsilon_{\mathrm{CNA}}$
        & 0.05 \\

        Maximum applicable verification tasks
        & 8 \\

        Verification rollouts per task and arm
        & 6 \\

        Minimum improved tasks $N_{\mathrm{pos}}^{\min}$
        & 2 \\

        Principle-pool capacity $N_{\mathrm{sys}}$
        & 4 \\

        \midrule
        \rowcolor{green!7}
        \multicolumn{2}{c}{\textit{Toolset Refinement}} \\

        Minimum cross-task support $N_{\mathrm{cross}}$
        & 2 \\

        Verification rollouts per task and arm
        & 4 \\

        Maximum verification tasks
        & 4 \\

        Outcome reward comparison margin $\epsilon_{\mathrm{tool}}$
        & 0.10 \\

        Utility adjustment coefficient $\alpha_f$
        & 0.50 \\

        Utility-pruning threshold $\epsilon_{\mathrm{prune}}$
        & 0.20 \\

        Invocation-count threshold $\epsilon_{\mathrm{retr}}$
        & 3 \\

        Toolset capacity
        & 100 \\

        \bottomrule
    \end{tabularx}
\end{table}

\begin{table}[!t]
    \centering
    \small
    \renewcommand{\arraystretch}{1.05}
    \setlength{\tabcolsep}{5pt}
    \caption{Hyperparameters for inter-component synchronization and trajectory replay in our experiments.}
    \label{tab:muse_synchronization_replay_hyperparameters}

    \begin{tabularx}{\linewidth}{
        @{}
        >{\raggedright\arraybackslash}X
        >{\centering\arraybackslash}p{0.30\linewidth}
        @{}
    }
        \toprule
        \textbf{Setting} & \textbf{Value} \\
        \midrule

        \rowcolor{blue!7}
        \multicolumn{2}{c}{\textit{Conflict-Aware Symbolic Integration}} \\

        Marginal relevance weight $\lambda_{\mathrm{mmr}}$
        & 0.85 \\

        Scope-similarity threshold $\epsilon_{\mathrm{scope}}$
        & 0.40 \\

        Contradiction threshold $\epsilon_{\mathrm{NLI}}$
        & 0.65 \\

        Empirical utility weight $\lambda_{\mathrm{qual}}$
        & 0.10 \\

        Dedicated principle delivery limit
        & 2 \\

        Experience and tool delivery limits
        & 3 and 1 \\

        Shared experience-tool delivery budget
        & 5 \\

        Principle context budget
        & 3,000 characters \\

        Retrieved-knowledge context budget
        & 6,000 characters \\

        \midrule
        \rowcolor{orange!8}
        \multicolumn{2}{c}{\textit{Parametric Internalization and Symbolic Entry Retirement}} \\

        Principle proficiency-check rollouts $N_{\mathrm{int}}$
        & 4 \\

        Mastery threshold $\epsilon_{\mathrm{mast}}$
        & 0.70 \\

        Principle utility decay $\alpha_{\mathrm{dec}}^{\mathrm{prin}}$
        & 0.80 \\

        \midrule
        \rowcolor{green!7}
        \multicolumn{2}{c}{\textit{Trajectory Replay and Optimization Groups}} \\

        Replay coefficient $\lambda_{\mathrm{replay}}$
        & 0.10 \\

        Per-task replay-buffer capacity $W_{\mathrm{group}}$
        & 4 \\

        Optimization group size $W$
        & 8 \\

        Maximum buffered tasks
        & 512 \\

        Maximum buffered trajectories
        & 2,048 \\

        Oracle and synthetic replay score thresholds
        & 0.60 and 0.90 \\

        Minimum real trajectories in a replay-only group
        & 2 \\

        Minimum replay-only outcome reward contrast
        & 0.05 \\

        Reward-normalization constant $\epsilon_{\mathrm{smooth}}$
        & $10^{-6}$ \\

        \bottomrule
    \end{tabularx}
\end{table}

\noindent \textbf{Gating Comparisons.}
The Always-on, Random, and Metacognitive settings retain supplementary synthetic supervision and the same eligible component space. The Always-on setting activates every refinement channel in each round. The Metacognitive setting determines component selection from the capability diagnosis profile. The Random setting constructs a component schedule from the recorded Metacognitive activation budgets. Across the 8 rounds, the symbolic component counts are $(2,2,1,2,1,2,1,1)$, with 7 experience, 2 principle, and 3 toolset activations in total. Model weight refinement remains enabled in all rounds. With seed 42, Random samples a feasible schedule preserving these per-round and total activation counts. Principle selection is excluded from the first round when the experience bank is empty. This construction matches component activation opportunities, while realized exploration rollout counts are measured separately according to Appendix~\ref{subapp:metrics_exploration_cost}.

\noindent \textbf{Limited Oracle Supervision.}
For the nominal 25\% oracle settings, a fixed subset of 22 tasks is selected from the 90-task oracle pool with seed 24. The subset supplies the available oracle objectives and outcome supervision throughout the run. Supplementary task construction can explore all 90 environments of the APPWorld training split, while access to original oracle task queries remains restricted to the selected 22 tasks. Configurations with supplementary supervision combine the restricted oracle task pool with admitted synthetic tasks, while the corresponding oracle-only configurations retain the restricted oracle task pool.

\noindent \textbf{Evaluation and Deployment Selection.}
The initial agent and subsequent candidate configurations are evaluated on the same 168 APPWorld tasks. Evaluation trajectories also provide evidence for subsequent metacognitive diagnosis. Deployment selection compares task success rate, mean outcome reward, and then mean interaction steps on successful executions. The Avg. Steps reported in the experimental tables is calculated across all evaluated tasks, following Appendix~\ref{subapp:metrics_task_performance}. Candidate configurations that improve the selection criteria are retained as the current best configuration. A candidate whose SR falls more than 2 percentage points below the historical best triggers recovery of the stored model weights and associated symbolic assets. For a non-promoted model update within this tolerance, the system can retain the best model weights together with the current symbolic assets and reevaluate the resulting configuration before deployment. The recorded candidate evaluations supply the roundwise performance measurements, while deployment checks determine the configuration carried into subsequent interaction. Best SR, Final SR, and Evo-AUC summarize the evaluation of these evolved candidates of each round.

\noindent \textbf{Stopping Conditions.}
Evolution proceeds for at most 8 activated rounds and may terminate earlier when no component is selected or a stopping counter reaches its configured patience. Degradation recovery advances a counter with patience 3 rounds, while a deployment without degradation recovery resets it. A separate progress counter has patience 3 rounds and is updated on evaluations that do not enter degradation recovery. Significant progress relative to the historical best requires an SR increase of at least 0.5 percentage points, or a mean outcome reward increase of at least 0.005 when SR is unchanged. Significant progress resets this counter.

\subsection{Audit Protocols} \label{subapp:audit_protocols}

This subsection details the sampling and human annotation procedures for the supervision audits in Section~\ref{subsec:supervision_effectiveness}. Both audits draw on records from the same complete run of MUSE with metacognitive gating, supplementary supervision, and symbolic and parametric adaptation enabled. Task quality and execution completion are assessed separately, with metric definitions provided in Appendix~\ref{subapp:metrics_task_quality} and Appendix~\ref{subapp:metrics_outcome_rewards}.

\subsubsection{Audit Sampling}
\label{subapp:audit_sampling}

The records for human audit are derived from a complete MUSE run. For task quality, we randomly sample five generated candidates from initial task generation and each of the 7 subsequent task-pool updates across 8 evolution rounds, yielding 40 candidates. We additionally collect all 36 distinct admitted synthetic tasks from the recorded task-pool snapshots, identifying tasks by their task identifiers and retaining the first recorded admitted version. These tasks include 26 remaining in the final active pool and 10 subsequently retired. The candidate and admitted groups assess task records before and after the construction and validation process, respectively.

For outcome reward assessment, 31 of the admitted tasks have recorded frontier assessments containing 2 guided and 6 free executions per task. We independently select one execution from each condition per task, yielding 62 executions. Selection does not depend on recorded rewards or human judgments. Guided executions receive the task query and reference solution, while free executions receive the task query. Each sampled execution is assessed against its own recorded query, accounting for possible task revisions during construction.

\subsubsection{Human Assessment of Task Quality}
\label{subapp:audit_task_quality}

The review interface presents the task query, reference solution, and available environmental evidence, including recorded actions and observations when present. It distinguishes proposed reference code from an executed reference trajectory. The reviewer assigns \textit{pass}, \textit{fail}, or \textit{uncertain} independently for each of the following criteria.

\noindent \textbf{Goal Clarity.}
The query is examined for an understandable objective, identifiable target objects, and necessary constraints. Entities that can be identified through the requested environmental queries need not be specified in advance. Missing requirements that prevent a determinate interpretation support a negative judgment.

\noindent \textbf{Environmental Feasibility.}
The reviewer checks whether available environmental evidence supports the required operations and states. Relevant API behavior, accessible entities, and recorded execution can support feasibility. Feasibility requires supporting environmental evidence beyond proposed code, and infeasibility requires evidence beyond a single failed execution. Insufficient evidence receives an uncertain judgment.

\noindent \textbf{Reference Alignment.}
The reference solution and available reference trajectory are checked against the task objective and constraints. Omitted requirements, conflicting operations, or unsupported assumptions can support a negative judgment. A reference solution is assessed as evidence requiring verification.

Overall Validity requires all three criteria to pass. The reported quality rates share the subset with determinate judgments for every criterion, comprising 39 generated candidates and all 36 admitted tasks. 

\subsubsection{Human Assessment of Execution Completion}
\label{subapp:audit_execution_completion}

For each sampled execution, the reviewer examines its task query, recorded actions, and environmental observations through the annotation interface to assess task completion. Recorded outcome rewards, internal success labels, and model-audit judgments are hidden during annotation. Completion judgments depend on the evidence within that execution. Required operations must be supported by recorded actions and observations within the sampled execution, without relying on a final success claim alone or evidence from another execution.

The reviewer assigns one of four labels for each individual execution. \textit{Complete} indicates that recorded evidence supports fulfillment of all task requirements and constraints. \textit{Partial} indicates that some requirements are demonstrably fulfilled while others remain unmet. \textit{Incomplete} indicates that the main task objective remains unfulfilled, including executions blocked before the required operations. \textit{Uncertain} indicates that missing evidence or unresolved task ambiguity prevents a determinate judgment. Clear evidence of an unmet requirement supports a partial or incomplete judgment without requiring every environmental state to be observable.

The primary human annotations supply the reference labels for Table~\ref{tab:synthetic_outcome_reward_reliability}. Complete annotations form the positive class, partial and incomplete annotations form the negative class, and uncertain annotations are excluded. These judgments are compared with full-completion predictions defined by $r_{\mathrm{out},i}^{\mathrm{syn}}=1$, following Appendix~\ref{subapp:metrics_outcome_rewards}. Guided and free executions are aggregated separately under the same criteria.

\section{Additional Results and Analysis}
\label{app:additional_results}

This appendix supplements Section~\ref{sec:experiments} with additional results and case studies of continuous agent evolution. Appendix~\ref{subapp:synthetic_task_coverage_and_capability_matching} examines the coverage of supplementary synthetic tasks and their correspondence to current agent capabilities. Appendix~\ref{subapp:self_supervision_case_studies} and Appendix~\ref{subapp:component_evolution_case_studies} then present concrete examples of interaction-grounded self-supervision and coordinated component evolution, respectively, connecting recorded interaction evidence to supervision construction, symbolic refinement, and subsequent task execution.

\subsection{Synthetic Task Coverage and Capability Matching}
\label{subapp:synthetic_task_coverage_and_capability_matching}

\begin{figure*}[!t]
    \centering

    \begin{subfigure}[t]{0.485\textwidth}
        \centering
        \includegraphics[width=\linewidth]{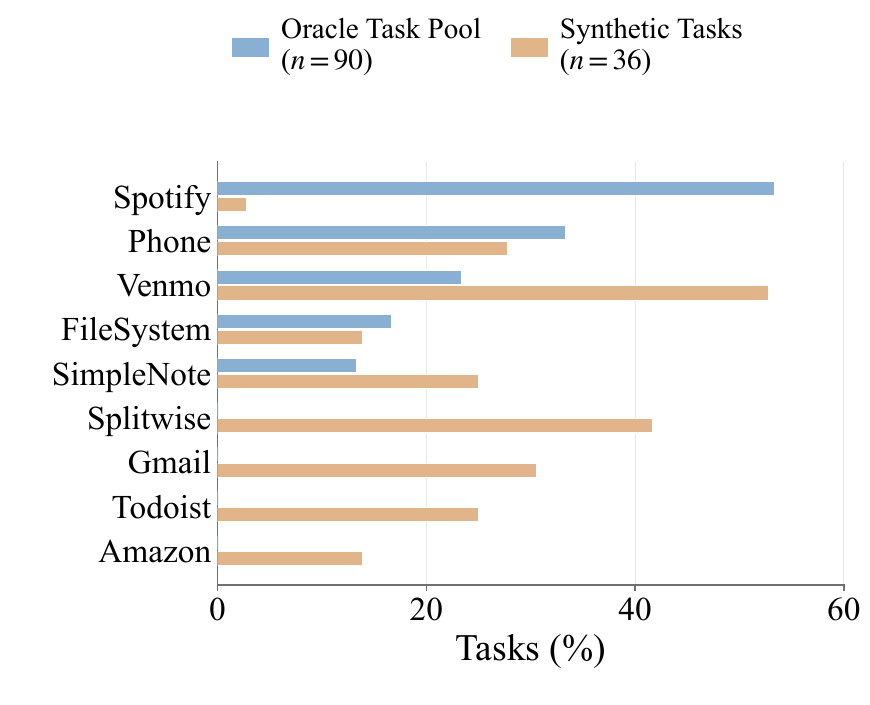}
        \caption{Application coverage}
        \label{fig:synthetic_task_application_coverage}
    \end{subfigure}
    \hfill
    \begin{subfigure}[t]{0.485\textwidth}
        \centering
        \includegraphics[width=\linewidth]{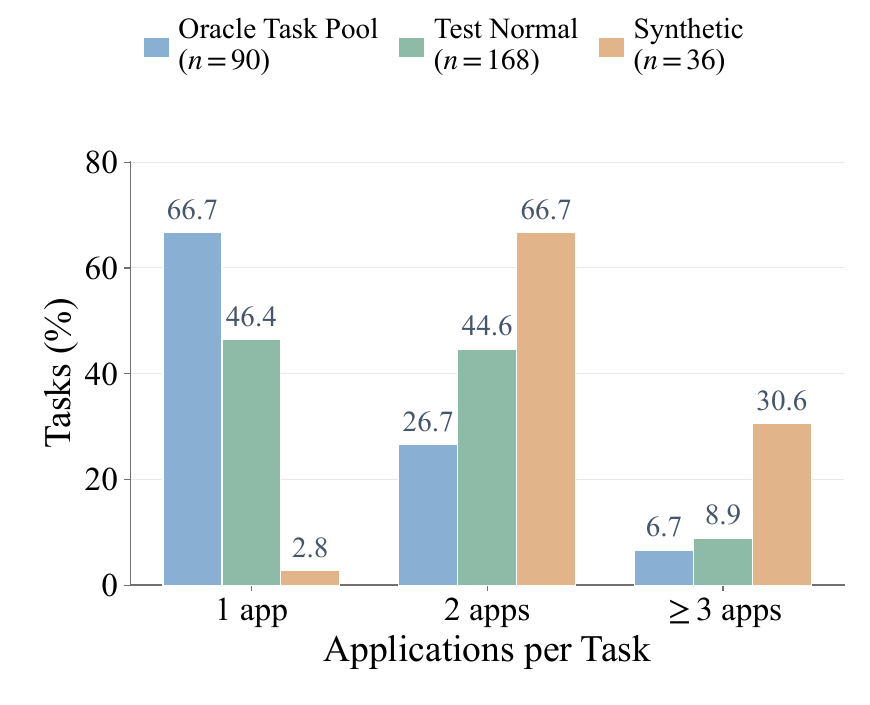}
        \caption{Applications per task}
        \label{fig:synthetic_task_application_composition}
    \end{subfigure}

    \par\medskip

    \begin{subfigure}[t]{0.485\textwidth}
        \centering
        \includegraphics[width=\linewidth]{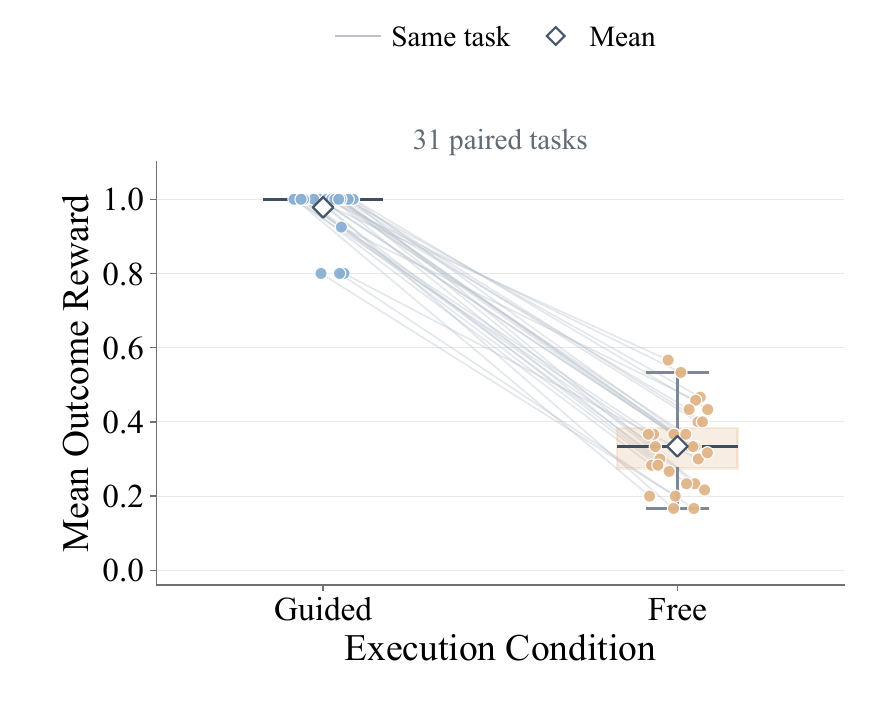}
        \caption{Capability matching}
        \label{fig:synthetic_task_capability_matching}
    \end{subfigure}
    \hfill
    \begin{subfigure}[t]{0.485\textwidth}
        \centering
        \includegraphics[width=\linewidth]{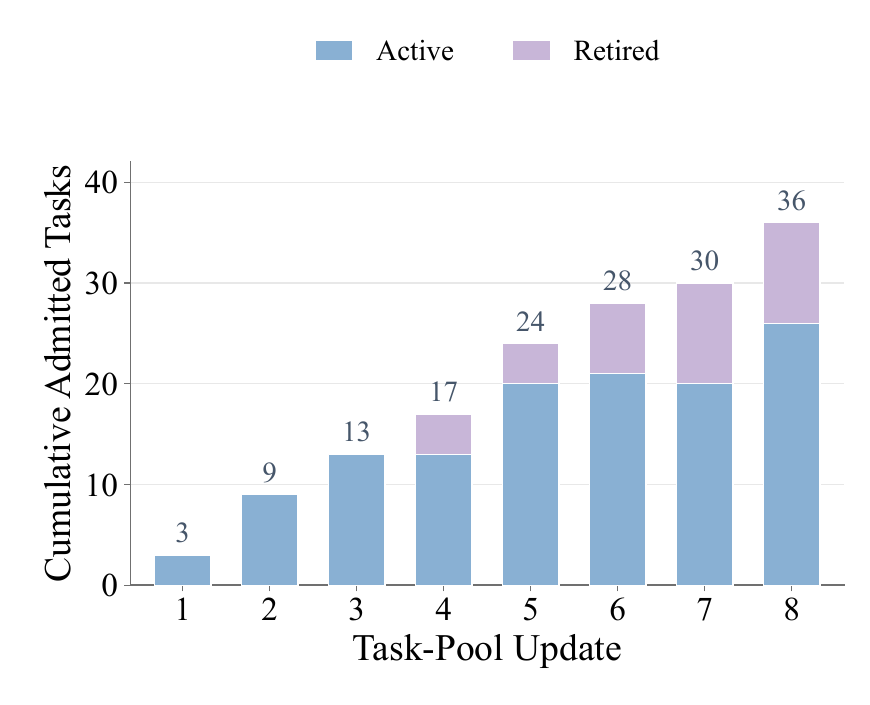}
        \caption{Continual task-pool adaptation}
        \label{fig:synthetic_task_pool_adaptation}
    \end{subfigure}

    \caption{Supplementary synthetic tasks from the complete MUSE run examined in Tables~\ref{tab:synthetic_task_quality} and~\ref{tab:synthetic_outcome_reward_reliability}. (a) Application coverage, allowing multiple applications per task. (b) Application counts, with Test Normal following official metadata. (c) Mean outcome rewards over two guided and six free executions per task. Lines connect paired tasks, boxes show quartiles and medians, whiskers follow the 1.5-IQR convention, and diamonds mark means. (d) Cumulative admitted tasks partitioned into active and retired tasks at each recorded task-pool update. \vspace{-4mm}}
    \label{fig:synthetic_task_coverage_and_composition}
\end{figure*}

We examine the supplementary synthetic tasks constructed during the same complete MUSE run underlying Tables~\ref{tab:synthetic_task_quality} and~\ref{tab:synthetic_outcome_reward_reliability}. The analysis covers all 36 distinct tasks admitted during this run, counting each task once according to its first admitted version. These tasks are compared with the 90-task oracle pool available to the agent. Application sets are identified from reference solutions and checked against task metadata, excluding the auxiliary applications \texttt{supervisor} and \texttt{api\_docs}. Application coverage reports the percentage of tasks involving each application within the corresponding collection. Since a task can involve multiple applications, these percentages need not sum to 100\%. Application counts for the 168 \texttt{test\_normal} tasks follow the official \texttt{num\_apps} metadata.

Capability matching covers the first selected frontier assessment of each of 31 admitted tasks, with two guided and six free executions per task. Figure~\ref{fig:synthetic_task_capability_matching} summarizes all 248 executions, from which 62 are sampled for human assessment. Task-pool adaptation covers initial construction and 7 subsequent updates, tracking cumulative admissions and their partition into active and retired tasks against the recorded pool states and retirement list.

From the results, we draw the following findings.

\noindent \textbf{(i) Supplementary tasks extend supervision across applications and their combinations.} Figure~\ref{fig:synthetic_task_application_coverage} and Figure~\ref{fig:synthetic_task_application_composition} show that supplementary tasks cover 4 applications absent from the oracle supervision. Moreover, 35 of the 36 synthetic tasks involve multiple applications, compared with only about one third of the oracle pool. The additional supervision therefore expands both application coverage and opportunities to coordinate operations across applications. Together with the task-quality audit in Table~\ref{tab:synthetic_task_quality}, these results support the construction of valid supplementary objectives beyond the interaction patterns represented in the available oracle tasks.

\noindent \textbf{(ii) Capability matching identifies tasks with unresolved independent execution demands.} In Figure~\ref{fig:synthetic_task_capability_matching}, guided executions attain higher mean outcome rewards on all 31 assessed tasks, with overall means of 0.978 and 0.334 for guided and free executions. Strong assessed performance with reference guidance accompanies substantially weaker independent execution. This contrast connects evidence of task feasibility with an identifiable capability gap, thereby directing supplementary supervision toward objectives that remain challenging for independent execution.

\noindent \textbf{(iii) Continual Task-Pool Adaptation renews supervision beyond initial admission.} Figure~\ref{fig:synthetic_task_pool_adaptation} shows increasing cumulative admissions alongside periods of stable or decreasing active pool size. By the final update, 26 of the 36 admitted tasks remain active. Consequently, a stable pool size can accompany changing task content, allowing supplementary supervision to be refreshed without continual expansion of the active pool. Such renewal provides a basis for adjusting adaptation opportunities as agent capabilities develop across successive evolution rounds.

\subsection{Case Studies of Interaction-Grounded Self-Supervision}
\label{subapp:self_supervision_case_studies}

\begin{figure}[!t]
    \centering
    \includegraphics[width=\textwidth]{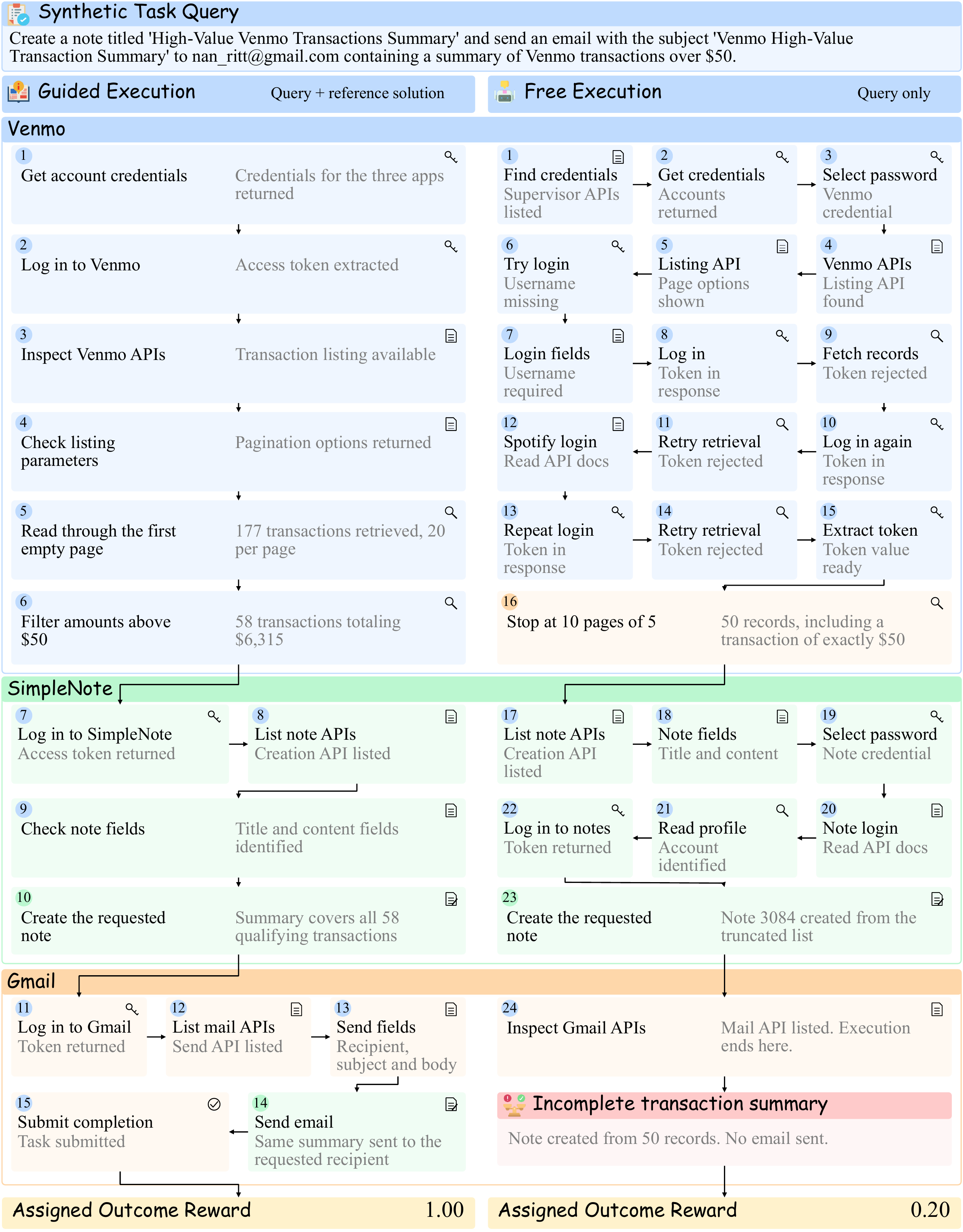}
    \caption{Guided and free executions of the same synthetic task. Numbered cards summarize recorded actions and observations in execution order, grouped by application. Guided execution receives the query and reference solution, while free execution receives only the query. The bottom cells report the assigned outcome rewards. \vspace{-4mm}}
    \label{fig:synthetic_task_case_study}
\end{figure}

\begin{table}[!t]
\centering
\caption{Prefix-potential supervision for a complete execution of an oracle task in MUSE. Orange rows mark potential decreases, while green rows mark potential increases after targeted corrections. Blue and orange reward cells indicate positive and negative step rewards, respectively.}
\label{tab:prefix_potential_case}
\begingroup
\small
\setlength{\tabcolsep}{4pt}
\renewcommand{\arraystretch}{1.18}
    \begin{tabular}{
        >{\centering\arraybackslash}m{0.04\linewidth}
        >{\raggedright\arraybackslash}m{0.23\linewidth}
        >{\raggedright\arraybackslash}m{\dimexpr0.53\linewidth-10\tabcolsep\relax}
        >{\centering\arraybackslash}m{0.10\linewidth}
        >{\centering\arraybackslash}m{0.10\linewidth}
    }
    \toprule
    $\ell$ & Action & Recorded observation
    & $\tilde{\omega}_i^{(\ell)}$ & $r_{\mathrm{pote},i}^{(\ell)}$ \\
    \midrule
    \rowcolor{blue!5}
    0 & \multicolumn{2}{>{\raggedright\arraybackslash}m{\dimexpr0.76\linewidth-8\tabcolsep\relax}}{\textbf{Task query} $q_i$: All phone text messages and voice messages from 3654328626 are spam, delete them.}
    & 0.361 & N/A \\
    \midrule
    1 & Inspect Phone APIs & Available Phone operations are returned.
    & 0.371 & \cellcolor{blue!8}$+0.027$ \\
    \arrayrulecolor{black!15}\hline
    \arrayrulecolor{black}
    2 & Retrieve credentials & Account credentials are returned.
    & 0.514 & \cellcolor{blue!8}$+0.326$ \\
    \arrayrulecolor{black!15}\hline
    \arrayrulecolor{black}
    \rowcolor{orange!9}
    3 & Attempt Phone login & HTTP 422: \texttt{username} is required. The call supplies \texttt{email}.
    & 0.313 & \cellcolor{orange!18}$-0.495$ \\
    \arrayrulecolor{black!15}\hline
    \arrayrulecolor{black}
    4 & Inspect login API & Login requires \texttt{username}, specified as the account phone number.
    & 0.442 & \cellcolor{blue!8}$+0.344$ \\
    \arrayrulecolor{black!15}\hline
    \arrayrulecolor{black}
    \rowcolor{green!6}
    5 & Correct login parameters & Login with the phone number as \texttt{username} returns an access token.
    & 0.534 & \cellcolor{blue!8}$+0.188$ \\
    \arrayrulecolor{black!15}\hline
    \arrayrulecolor{black}
    6 & Inspect text-message search & Documentation specifies message filters and pagination.
    & 0.538 & \cellcolor{blue!8}$+0.008$ \\
    \arrayrulecolor{black!15}\hline
    \arrayrulecolor{black}
    \rowcolor{orange!9}
    7 & Attempt message retrieval & \texttt{NameError}: \texttt{phone\_access\_token} is undefined.
    & 0.465 & \cellcolor{orange!18}$-0.146$ \\
    \arrayrulecolor{black!15}\hline
    \arrayrulecolor{black}
    \rowcolor{green!6}
    8 & Correct the token reference & The token from the login result enables retrieval of 10 matching text messages.
    & 0.582 & \cellcolor{blue!8}$+0.224$ \\
    \arrayrulecolor{black!15}\hline
    \arrayrulecolor{black}
    9 & Inspect text-message deletion & Documentation specifies the message ID and access token.
    & 0.595 & \cellcolor{blue!8}$+0.023$ \\
    \arrayrulecolor{black!15}\hline
    \arrayrulecolor{black}
    10 & Delete matching text messages & Execution prints deletion confirmations for all 10 retrieved text messages.
    & 0.689 & \cellcolor{blue!8}$+0.147$ \\
    \arrayrulecolor{black!15}\hline
    \arrayrulecolor{black}
    11 & Inspect voice-message search & Documentation specifies voice-message filters and pagination.
    & 0.705 & \cellcolor{blue!8}$+0.023$ \\
    \arrayrulecolor{black!15}\hline
    \arrayrulecolor{black}
    12 & Retrieve matching voice messages & Paginated retrieval returns 12 matching voice messages.
    & 0.748 & \cellcolor{blue!8}$+0.059$ \\
    \arrayrulecolor{black!15}\hline
    \arrayrulecolor{black}
    13 & Inspect voice-message deletion & Documentation specifies the voice-message ID and access token.
    & 0.755 & \cellcolor{blue!8}$+0.008$ \\
    \arrayrulecolor{black!15}\hline
    \arrayrulecolor{black}
    14 & Delete matching voice messages & Execution prints deletion confirmations for all 12 retrieved voice messages.
    & 0.794 & \cellcolor{blue!8}$+0.050$ \\
    \arrayrulecolor{black!15}\hline
    \arrayrulecolor{black}
    15 & Submit completion & The completion call executes successfully.
    & 0.794 & \cellcolor{black!5}$0.000$ \\
    \bottomrule 
    \end{tabular} %
\endgroup
\end{table}

We examine two cases from the complete MUSE run analyzed in Appendix~\ref{subapp:synthetic_task_coverage_and_capability_matching}. Figure~\ref{fig:synthetic_task_case_study} presents outcome rewards for guided and free executions of a synthetic task, while Table~\ref{tab:prefix_potential_case} presents prefix-potential supervision for an oracle task. These cases connect execution evidence to assessments of overall completion and intermediate task-completion prospects within the supervision generator in Section~\ref{subsec:supervision_generator}.

From the presented case studies, we draw the following findings.

\noindent \textbf{(i) Outcome rewards of synthetic tasks can reflect fulfillment of task requirements across applications.} In Figure~\ref{fig:synthetic_task_case_study}, both executions create a note summarizing Venmo transactions. The guided execution retrieves successive pages until an empty page confirms the end of the list, identifies all 58 transactions above \$50, and emails the same summary to the requested recipient. The guided execution thus receives an outcome reward of 1.00. Conversely, the free execution stops after 10 pages of 5 records, including a transaction of exactly \$50. It creates a note from this incomplete set and ends after inspecting the Gmail APIs. Its outcome reward of 0.20 is consistent with partial completion since the note omits qualifying transactions and the requested email remains unsent.

\noindent \textbf{(ii) Prefix potentials provide step-level supervision throughout task execution.} Complementing task-level outcome rewards, the step rewards in Table~\ref{tab:prefix_potential_case} distinguish a setback from its subsequent correction within an execution that ultimately records successful deletion of the requested messages. At step~3, for example, the Phone API rejects a login request that supplies \texttt{email} instead of the required \texttt{username}. Although the messages remain unchanged, this response reveals that the attempted login failed to establish the access needed for deletion. The evolving agent therefore assesses completion as less likely, and the prefix potential falls from 0.514 to 0.313. To resolve the error, the agent consults the API documentation and learns that login requires the account phone number. With the correct parameter, the next login succeeds and returns an access token. The corresponding step rewards are negative for the rejected call and positive for both identifying the required parameter and correcting the call. MUSE thereby derives supervision for model weight refinement from the evidence available at each step of an execution.

\subsection{Case Studies of Coordinated Component Evolution}
\label{subapp:component_evolution_case_studies}

Drawing on the complete MUSE run examined in Appendix~\ref{subapp:synthetic_task_coverage_and_capability_matching}, we trace how symbolic assets guide specific execution steps and support the abstraction of broader operational guidance. Appendix~\ref{subsubapp:case_experience_principle} examines a verified experience that guides song selection and a candidate principle that abstracts its operational dependencies into a more general trigger, action, and guard. Appendix~\ref{subsubapp:case_tool_invocation} then traces tool invocation in a task where retrieved records of downloaded songs support album selection for offline playback.

\subsubsection{Experience Refinement and Principle Distillation}
\label{subsubapp:case_experience_principle}

\begin{figure}[!t]
    \centering
    \includegraphics[width=\textwidth]{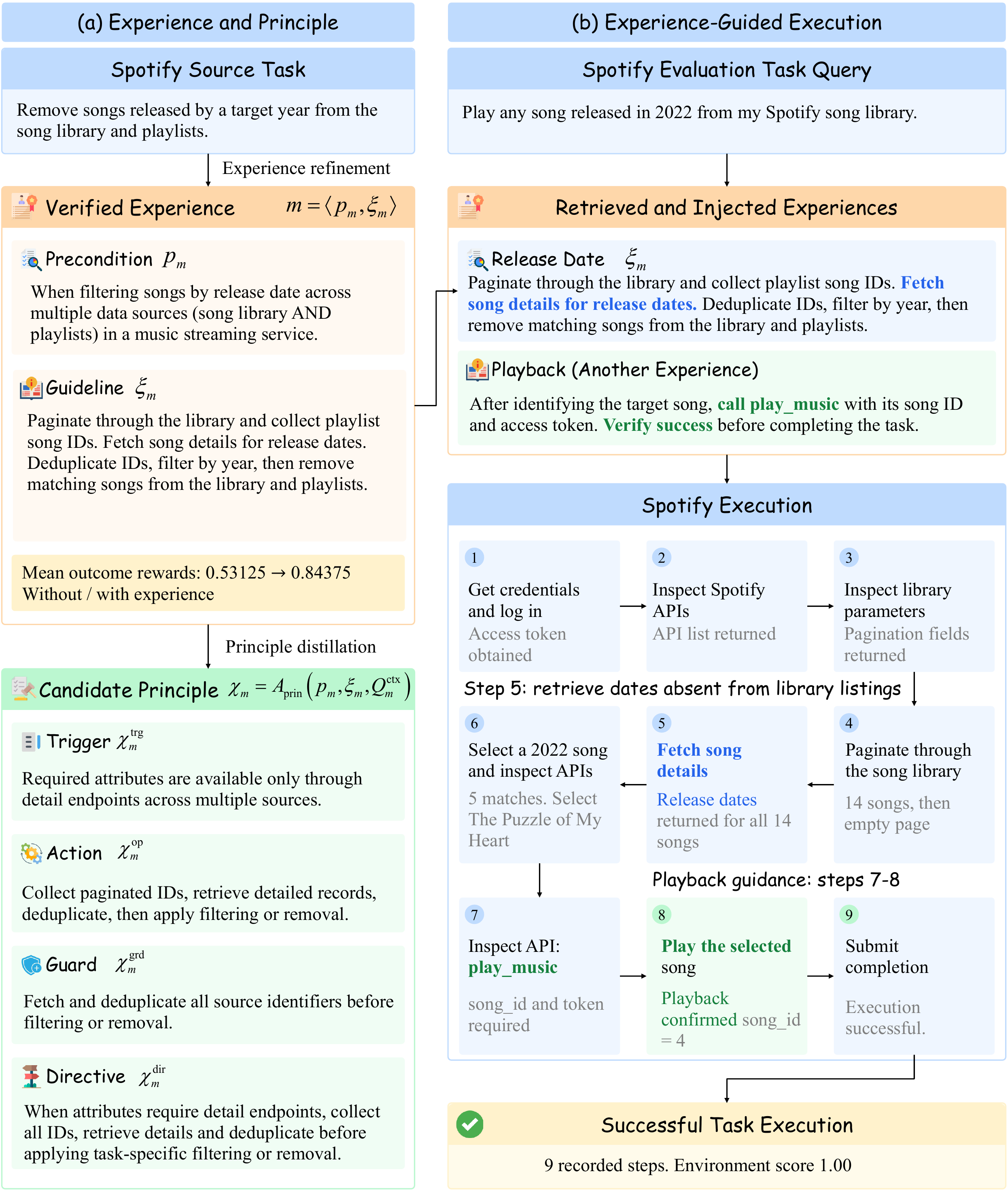}
    \caption{Case study for experience-guided execution and principle distillation in MUSE. (a) A verified Spotify experience and the candidate principle distilled from it. (b) Two injected experiences and a complete 9-step execution. Blue text links release-date guidance to step~5, while green text links playback guidance to steps~7-8. Numbered cards summarize recorded actions and observations in execution order.}
    \label{fig:exp_principle_case_study}
\end{figure}

Figure~\ref{fig:exp_principle_case_study} connects a verified experience to its application in another task and its abstraction into a candidate principle, following Appendix~\ref{subapp:experience_bank_refinement} and Appendix~\ref{subapp:general_principle_distillation}.

\noindent \textbf{(i) Experiences supply operational knowledge for specific execution steps.} The source experience concerns removing songs by release year, but its guidance about retrieving release dates also applies to selecting a song for playback. Since library listings omit these dates, the agent explicitly refers to the retrieved guidance and queries song details at step~5, obtaining the information required for year filtering. A separate playback experience identifies \texttt{play\_music}, which is absent from the visible portion of the truncated API listing. The agent refers to this guidance when inspecting the endpoint at step~7 and then confirms playback at step~8. These connections illustrate how situational experiences supply relevant interface knowledge across tasks with different objectives.

\noindent \textbf{(ii) Principle distillation abstracts the operational dependency underlying an experience.} The candidate trigger generalizes the song-library and playlist setting to tasks involving multiple data sources with required attributes available only through detail endpoints. Under this condition, its action specifies collecting identifiers, retrieving detailed records, and deduplicating entries before filtering or removal. The guard further specifies that collection and deduplication should cover every source before these operations proceed. These elements preserve the dependency between complete information gathering and subsequent task operations, expressing the concrete Spotify experience as a conditional procedure for tasks sharing the same data-access structure.

\subsubsection{Toolset Refinement and Tool Invocation}
\label{subsubapp:case_tool_invocation}

\begin{figure}[!t]
    \centering
    \includegraphics[width=\textwidth]{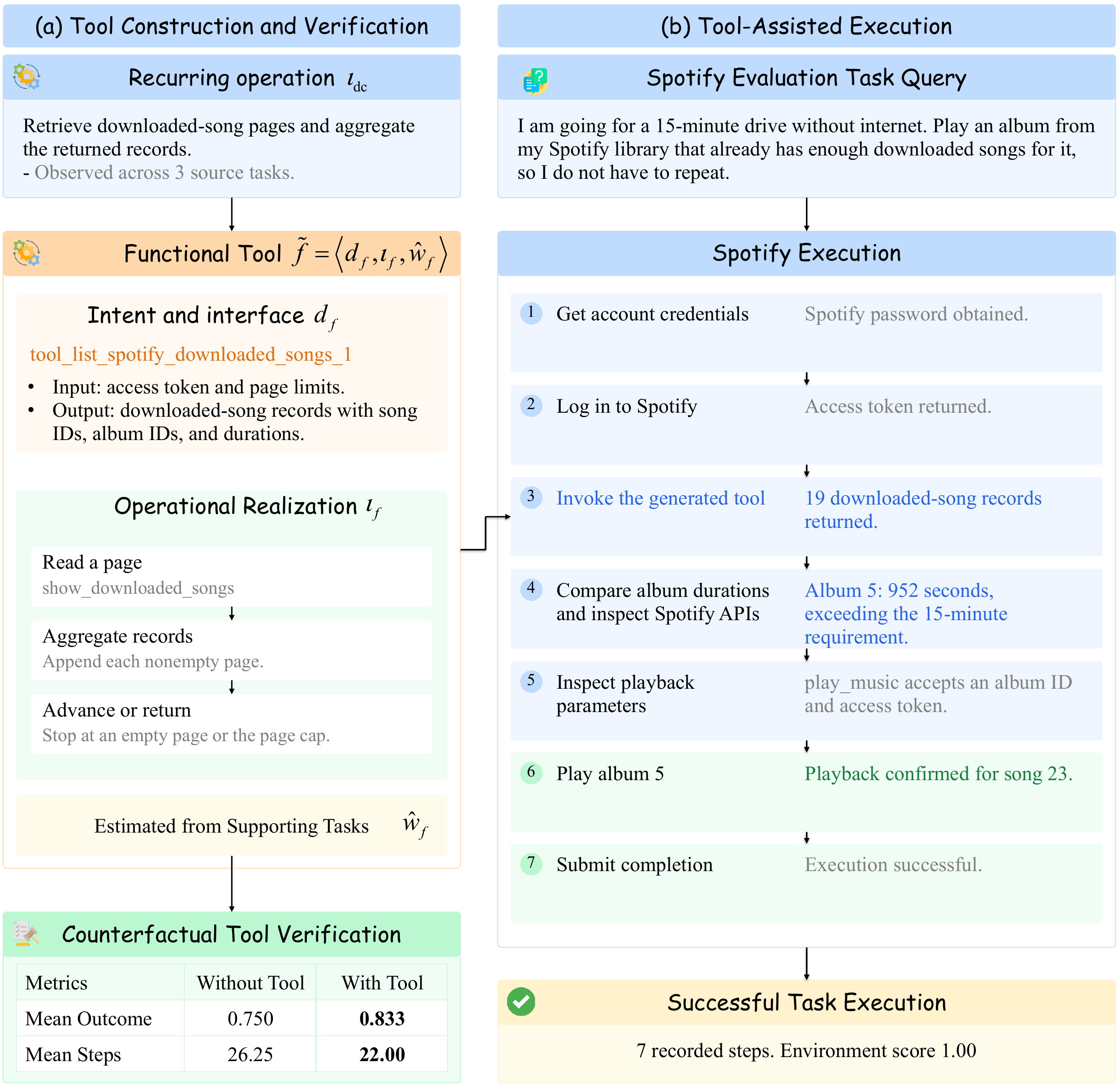}
    \caption{Case study of tool refinement and invocation in MUSE. (a) A recurring operation, its encapsulated tool, and paired verification with and without tool availability. (b) Seven recorded execution steps connecting tool invocation to album selection and playback. Numbered rows present actions and environmental observations.}
    \label{fig:tool_invocation_case_study}
\end{figure}

Figure~\ref{fig:tool_invocation_case_study} traces a functional tool from recurring operation patterns to verification and subsequent invocation, illustrating the toolset refinement described in Appendix~\ref{subapp:functional_toolset_refinement}.

\noindent \textbf{(i) Tool refinement consolidates recurring operations into an executable procedure.} The source pattern recurs across 3 tasks and combines paginated retrieval with record aggregation. Its encapsulation separates the callable interface $d_f$ from the operational realization $\iota_f$, preserving pagination and stopping conditions while accepting the current access token and page limits. This interface allows the agent to invoke the procedure without reconstructing its retrieval loop. In paired verification, tool availability reduces mean interaction steps from 26.25 to 22.00 alongside higher mean outcome rewards, supporting the value of this encapsulation in the evaluated tasks.

\noindent \textbf{(ii) Tool outputs provide the evidence required for subsequent task decisions.} At step~3, the agent invokes the tool and receives 19 downloaded-song records containing album identifiers and durations. These fields support the comparison at step~4, where the agent identifies an album with 952 seconds of downloaded music, exceeding the requested 15 minutes. The agent then inspects the playback interface and starts the selected album, completing the task with an environmental outcome reward of 1.00. The execution thus connects an encapsulated retrieval procedure to task-specific duration assessment and playback of an album meeting the requested listening duration.

\clearpage
\begingroup
\renewtcblisting{promptcard}[1]{
    enhanced,
    unbreakable,
    listing only,
    title={#1},
    listing options={style=promptcardstyle,keepspaces=true},
    colback=promptcardbg,
    colframe=promptcardframe,
    colbacktitle=promptcardtitle,
    coltitle=white,
    fonttitle=\bfseries\sffamily\small,
    boxrule=0.55pt,
    arc=4pt,
    outer arc=4pt,
    halign=left,
    left=8pt,
    right=8pt,
    top=5pt,
    bottom=5pt,
    before skip=6pt,
    after skip=10pt
}

\section{Prompts} \label{app:prompts}

This appendix presents the prompt templates for the MUSE implementation in Appendix~\ref{app:implementation_of_muse}. The prompts are grouped by metacognitive diagnostic routing in Appendix~\ref{subapp:gating_prompts}, interaction-grounded self-supervision in Appendix~\ref{subapp:supervision_prompts}, and symbolic refinement and coordination in Appendix~\ref{subapp:optimizer_prompts}. The accompanying descriptions specify the input evidence, expected outputs, and conditions under which each template is applied within the corresponding MUSE procedure.

Named placeholders in curly braces denote runtime inputs, while card headings identify the corresponding message or template fragment. Numbered cards form consecutive parts of one message. System and user roles, together with conditions for inserting optional fragments, are specified in the accompanying descriptions. Placeholders ending in \texttt{\_\allowbreak{}json} contain serialized JSON values, including quotation marks for strings.

\subsection{Evolutionary Gating Prompts} \label{subapp:gating_prompts}

Evolution activation follows the checks in Section~\ref{subsec:evolutionary_gating}. After activation, the diagnostic router compares sampled interactions, task outcomes, and symbolic support presented during execution. These inputs support the capability diagnosis profile and targeted component selection described in Appendix~\ref{subapp:evolutionary_gating}.

The requested decisions follow a dynamically constructed JSON schema. Each component entry contains \texttt{enabled}, \texttt{reason}, and \texttt{evidence\_\allowbreak{}ids}, with \texttt{source\_\allowbreak{}experience\_\allowbreak{}ids} additionally required for principle routing. Structured review adds an \texttt{assessment} object to the relevant entries. The schema requests unresolved decisions, while evidence-only repair requests corrected citations and preserves the existing routing decisions.

\subsubsection{Diagnostic Routing and Component Criteria}
\label{subapp:diagnostic_routing_prompts}
The seven blocks below form the diagnostic user message. They combine component-specific routing criteria with computed observations, task summaries, and registered interaction evidence for component selection.

The \texttt{module\_\allowbreak{}admission\_\allowbreak{}policy} field contains the routing criteria below. The \texttt{conditional\_\allowbreak{}review\_\allowbreak{}instructions} field supplies structured review requirements for reassessments and for initial assessments with evidence supporting principle distillation. Otherwise, it remains empty. The \texttt{review\_\allowbreak{}note} field is empty initially and supplies the applicable review fragments during reassessment.

\begin{promptcard}{Diagnostic Routing (1/7): Role and Instructions}
You are the Metacognitive Brain of an autonomous AI agent. Analyze the recent SUCCESSES and FAILURES below. Return only the requested module decisions directly in the supplied JSON schema. Program-resolved prerequisite decisions are final and are not requested again. Use concise reasons; do not write private deliberation, a prose assessment, quotes, or a completion marker. Evidence IDs already identify the original source text; select IDs instead of copying text.

{module_admission_policy}

{conditional_review_instructions}
\end{promptcard}

\begin{promptcard}{Diagnostic Routing (2/7): Observations and Evidence}
PROGRAM OBSERVATIONS (structural leads are not activation requirements; asset types are authoritative):
{program_observations_json}

{review_note}

TASK FACT OVERVIEW (all sampled tasks, computed before long-trace presentation):
Evaluator outcome is authoritative for task success. A later execution_success_reported is local feedback, never a replacement for that outcome. Error resolution is not inferred. Submitted call sites, loops and pagination parameters describe code structure, not proof of execution or complete coverage. Repeated call counts count separate submitted messages, not loop iterations. Context IDs support verdicts; behavioral claims must inspect the action/feedback IDs below. Missing IDs, unparseable code and omitted/compressed content are unknown. The full evidence follows; this index does not replace it.
{task_fact_overview_json}

Program-computed exposure facts for the current deployment window (not the full evaluation):
{window_summary_json}

Here are the sampled cases:
{registered_case_evidence}
\end{promptcard}

\begin{promptcard}{Diagnostic Routing (3/7): Input Boundary and Fact Discipline}
Input boundary: Only the current deployment-window interactions and symbolic content actually exposed in them are available. Training tasks, training usage contexts, asset-bank qualification scores, static A/B results and past candidate verification are deliberately excluded. Do not infer their values or use their absence to reject a capability need.

Fact discipline: evaluator success/score is the task verdict, not inferred from an execution message. Do not claim a successful task failed because an action was omitted. A successful task can still show inefficiency or a recovered error. API call sites, loops and pagination advancement are recorded code structure, not proof that every branch executed or all pages were covered. Inspect each cited task's actions and paired feedback before claiming an operation was missing or an error persisted. A later same-API success is recovery evidence to examine, not automatic proof of repair. Missing/truncated text and unparseable code are unknown, never evidence of absence. System examples are not executed actions. Distinguish a step count reaching the limit from a confirmed step-limit termination.
Distinguish evaluator/environment anomalies from genuine agent failures. Compare recurring failures with successful recovery patterns.
\end{promptcard}

\begin{promptcard}{Diagnostic Routing (4/7): Task Attribution and Outcome Evidence}
Evidence policy:
- Refer to tasks by their literal task IDs, never by Case numbers. Keep each task's request, actions, feedback and outcome together; never transfer a date error or authentication error to another task. Before finalizing, verify each claim against the cited task. An error message listing possible causes does not prove which cause occurred: inspect argument values and return types before claiming token expiry, lost state, or an external fault.
- Evaluator success and score are the task outcome. 'Execution successful' or complete_task() only confirms code execution/termination, NOT that requested side effects or constraints were satisfied. Compare the original request with actual actions. For example, answering a recipe name without sending the requested Phone message is an instruction-following omission, not an evaluator anomaly.
- Agent-generated TypeError, SyntaxError, wrong indexing, ignored feedback, premature termination, and omitted actions are candidate evidence of a behavior improvement. Assess their impact on task completion and whether the agent recovered before deciding to train. Do not relabel them as environment faults simply because an API or credential is involved.
- Reaching the step limit is an outcome, not a root cause. Inspect wasted steps, repeated errors, API discovery loops, and planning before assigning blame. Assert external/evaluator failure only with concrete evidence independent of the agent's own claim of completion; otherwise mark it uncertain.
\end{promptcard}

\begin{promptcard}{Diagnostic Routing (5/7): Delivery Labels and Error Interpretation}
- Symbolic delivery labels such as retrieved_not_selected describe delivery only; they do not establish evaluator failure or prove a missing asset caused the task failure.
- Error summaries record occurrences, not unresolved failures. Read errors, later_related_actions and paired action/feedback before attributing the final failure to an earlier error. Compare actual arguments before claiming identical retries. Check constraint_evidence and the original request before claiming an operation was omitted; observed transforms (e.g. replace) are concrete counterevidence to "never performed", but not proof of correct scope or final task completion. Missing or truncated supporting evidence requires uncertainty, not a strong absence claim. Cite the action/feedback Evidence IDs supporting behavioral claims, not only the task outcome. Prefer concrete unfulfilled task constraints over a recovered TypeError.
\end{promptcard}

\begin{promptcard}{Diagnostic Routing (6/7): Symbolic Support and Exposure}
- Trajectory patterns are the primary source of capability demand. For symbolic modules, compare supplied asset contents, applicability and actual exposure (including integrated symbolic assets). Distinguish absent support, inadequate support, and a behavior weakness despite suitable exposed support. For Experience and Tool discovery, an empty bank neither proves a need nor prevents discovery; Principle requires the promotion support specified below; a similar existing asset is not sufficient to close a route. A delivery-only gap does not by itself justify generating a duplicate asset.
- Experience and Principle are exposed optional guidance. Their emitted text is observable, but whether the LLM actually used either one is not. Never infer failure from an empty or absent used field.
- Tool runtime stages are directly observable, but an emitted Tool that was not called is neutral unless the trajectory proves that the Tool was necessary and applicable.
- Retrieved-only or selected-only assets were not exposed. Use their IDs only to diagnose delivery coverage; do not treat their hidden text as model context.
\end{promptcard}

\begin{promptcard}{Diagnostic Routing (7/7): Decisions and Output Requirements}
For every module, return a boolean decision and a concise reason grounded in this deployment window. Every enabled route needs at least one listed Evidence ID. Disabled symbolic routes also need Evidence IDs of inspected cases, even when explaining insufficient evidence; the absence of a reusable pattern must be assessed against observed cases. An enabled Principle must identify recurring visible Experiences and positive behavioral support for abstraction; this does not certify hidden training sources or deployment admission. Task IDs and original excerpts are resolved from your evidence_ids. A failure does not justify enabling every module. Use the program's exposure facts accurately and do not invent missing assets, hidden scores or out-of-window history. All four modules may be disabled when supported by the observations. For each module return enabled, reason and evidence_ids, plus assessment when required by the review schema. Cite only listed Evidence IDs; task IDs are derived by the program and must not be output as evidence IDs. An enabled module needs supporting evidence. Only a disabled Model route may cite []; symbolic closures must cite the cases inspected for absence of a need. Reasons may summarize source code, including abbreviated calls such as create_note(...); no verbatim quotation is required. Evidence provenance does not prove causality: verify the task, actions, feedback and outcome before deciding. Each module is independent; valid Experience and Model learning may proceed together.
\end{promptcard}

\noindent\textbf{Component-specific routing criteria.} The six blocks below populate \texttt{module\_\allowbreak{}admission\_\allowbreak{}policy} with shared evidence requirements and criteria for selecting experiences, principles, tools, and model weights. The corresponding procedures in $\mathcal M$ then construct and verify candidates for the selected components.

\begin{promptcard}{Module Admission Policy (1/6): Shared Routing Criteria}
Module admission policy (four independent decisions):
For symbolic routing, follow: interaction evidence -> capability deficiency OR consolidation need -> suitable module(s). First identify the capability demand across successful and failed cases, then select its evolutionary response. Deficiencies include unreliable situational guidance, repeated construction effort, missing operational abstractions and inadequate cross-scenario guidance. Consolidation may be warranted by recurring effective strategies even without task failure or declining aggregate performance; do not invent a deficiency to justify it. Reusable material alone is not sufficient: explain its concrete value for the observed capability demand. The same evidence may support complementary symbolic and Model improvements.

For each symbolic module, use its existing reason to summarize four linked points: observed behavior -> capability deficiency or consolidation need -> coverage by actually exposed assets -> specific contribution of this module (or why none is supported). These are concise evidence-based conclusions, not private deliberation. A closure must establish absent need, adequate existing coverage, or insufficient evidence; error type or success alone is not a closure criterion. Missing exposure does not establish absence from the entire bank. Only Principle additionally returns source_experience_ids identifying visible Experiences considered for promotion; use [] when closing without such sources. Follow the supplied JSON schema, including assessment during review; do not add unspecified fields or scores.
\end{promptcard}

\begin{promptcard}{Module Admission Policy (2/6): Candidate Generation Boundary}
Gating opens an opportunity to generate and verify candidates; it does not certify a candidate for deployment. Source eligibility, construction and cross-task/static A/B admission remain the responsibility of the corresponding evolution module. Principle specifically needs observable Experience promotion support: repeated exposure of the same experience content in independent sampled tasks is only a prerequisite, never proof of use, quality or causality. Inspect positive behavior and feedback before diagnosing promotion value. The generator subsequently checks eligible training Experiences and promotes those Experiences. Window trajectories are diagnostic evidence, not a direct Principle construction or admission source.

For Experience and Tool, decide whether the observed capability need has a concrete symbolic improvement worth generating and testing. Do not require the module to be indispensable, the agent to be unable to recover, or a trained candidate's benefit to be proven already. Later generation and validation use training material; deployment-window trajectories are diagnostic evidence only.

For Experience and Tool reasons, name the best-supported concrete guidance or operation unit considered, its evidence and expected contribution, then decide. When closing, explain why that particular candidate adds no useful guidance/abstraction or lacks support. Do not substitute 'recovered', 'standard API usage', 'different task goals' or 'not complex' for this assessment. Both enabled and disabled decisions remain valid; do not invent candidates or activate modules merely because the bank is empty.
\end{promptcard}

\begin{promptcard}{Module Admission Policy (3/6): Experience Routing}
1. evolve_experience (Experience): identify a situation/trigger and a reusable action, interpretation rule or recovery guard that would reduce avoidable mistakes, exploration or omissions, or retain a useful observed strategy beyond actually exposed guidance. One supported case can suffice. Recovering after discovering the correct return structure, token scope or operation order can reveal exactly what should be retained; recovery does not erase that opportunity. Inspect the remaining requested effects and evaluator outcome separately. Close when no concrete useful guidance can be supported, the incident has no reusable lesson, or actually exposed Experiences already provide adequate guidance without a useful refinement. Do not require persistent failure or exclude guidance merely because Model can learn the same behavior.
\end{promptcard}

\begin{promptcard}{Module Admission Policy (4/6): Principle Routing}
2. evolve_principle: diagnose a shared behavioral-guidance need across cases or an opportunity to consolidate recurring effective strategies into broader guidance. Enable only for promotion of actually exposed, recurring Experiences listed in principle_support. Shared errors or a desirable general rule alone do not establish Experience promotion demand. Return nonempty source_experience_ids when enabled; cite action/feedback evidence from at least two distinct sampled tasks that exposed each source. Inspect whether the related behavior and feedback provide repeated positive support, including useful successful substeps or recovery in otherwise failed tasks. Exposure frequency, a task success beside exposure, or absence of an existing Principle alone is insufficient. Explain the shared trigger, action and guard, the positive support, and why abstraction adds cross-scenario value over retaining the Experiences. When inspected behavior establishes absent or insufficient positive support, close the route. If support cannot be assessed, report uncertainty rather than a verified closure. For a closure despite repeated positive support, explain why abstraction adds no useful cross-scenario trigger/action/guard beyond those Experiences; their existing effectiveness alone is not a reason to close. Do not infer actual use of optional guidance or causal benefit. Training eligibility/utility scores and A/B results remain hidden and must not be requested. The separate generator still promotes eligible training Experiences, never evaluation trajectories.
\end{promptcard}

\begin{promptcard}{Module Admission Policy (5/6): Tool Routing}
3. evolve_toolset (Tool): identify a supported operation unit with reusable inputs/results, dependencies and a concrete saving in construction effort, repeated steps or error risk. Compare shared suboperations rather than complete task workflows. Repeated pagination/filtering, identifier matching or ordered dependent operations may qualify even when all atomic APIs exist and handwritten execution succeeds. A single task may establish the value when it repeats a parameterizable operation or demonstrates a concrete reusable sequence with that benefit. Neither multiple failed tasks nor a complex end-to-end workflow is required. Close when no useful unit is supported, exposed callable Tools already cover it, or encapsulation offers no concrete benefit. Merely wrapping an ordinary API call is insufficient.
\end{promptcard}

\begin{promptcard}{Module Admission Policy (6/6): Model Routing}
4. evolve_model: decide whether the sampled evidence identifies a worthwhile behavior improvement to learn. Enable for omitted requested actions, misunderstanding the goal, failing to use available information, or faulty code/ineffective strategies that materially harmed task completion, when you can specify the better behavior and why learning it would help. One task can provide sufficient evidence; cross-task recurrence, repeated mistakes despite feedback, and persistence across evaluations strengthen the case but are NOT prerequisites. Normally keep Model disabled for an isolated mistake that the agent promptly corrected, followed by successful task completion with no material remaining weakness. Also keep it disabled when no concrete behavior improvement can be identified or the only evidence is a verified external outage or unavailable information. Symbolic guidance being useful does NOT exclude Model training: enable Experience/Principle/Toolset and Model together when both have evidence; do not require Symbolic repairs to fail first. Low SR, elapsed rounds, and training frequency alone are not admission criteria.

For an enabled Model route, explain the observed behavior, the desired behavior, and the expected benefit of learning that behavior, grounded in the cited tasks. A generic label such as 'reasoning weakness' is insufficient. For a disabled Model route, explicitly compare each failed task's requested effects with observed actions and evaluator outcome, and explain why there is no worthwhile remaining behavior improvement. Include successful recovery where relevant; partial action completion or execution success does not establish fulfillment of the original request. Do not substitute absence of broad, fundamental or persistent deficits for this assessment. Summarize in reason; during review, also complete the required assessment.failed_task_reviews using the supplied schema. Do not add budget tiers or unspecified fields.
\end{promptcard}

\subsubsection{Structured Review and Conditional Reassessment}
\label{subapp:diagnostic_review_prompts}
The following five blocks populate \texttt{conditional\_\allowbreak{}review\_\allowbreak{}instructions} in Appendix~\ref{subapp:diagnostic_routing_prompts}. They specify a shared assessment format and component-specific checks for reassessment. Initial assessments with evidence supporting principle distillation receive the same source-experience checks. The JSON schema identifies the requested routes that require a structured \texttt{assessment} object.

\begin{promptcard}{Review Output Contract (1/5): Shared Assessment Requirements}
Review output contract:
For modules whose schema requests assessment, produce assessment BEFORE enabled/reason/evidence_ids. For symbolic modules put inspection arrays BEFORE candidate/conclusion. Do not add assessment to other routes. Keep inspection entries concise; use short factual phrases, not essays. First consider a concrete improvement grounded in observed actions and feedback, then decide. assessment.candidate is either null or {need, proposal, expected_benefit, evidence_ids}. For Model the proposal is the better behavior to learn; for symbolic modules it is the guidance or operation to generate, not an artifact built from test data. assessment.conclusion is propose, no_supported_candidate, already_covered, insufficient_benefit, or uncertain. propose requires a concrete candidate with action/feedback evidence and enabled=true. already_covered or insufficient_benefit requires the concrete candidate considered, enabled=false, and a reason explaining its rejection. already_covered additionally identifies actually exposed matching symbolic asset IDs in covered_by_asset_ids; atomic APIs or hidden assets are not existing symbolic coverage, and this conclusion is not available for Model. no_supported_candidate requires candidate=null and enabled=false with inspected evidence. Use uncertain when the evidence does not let you substantiate a judgment, instead of disguising uncertainty as no need; the program records diagnosis failure separately. covered_by_asset_ids is [] except for already_covered. Candidate Evidence IDs must also appear in the route's evidence_ids. Cite action AND paired feedback, not just a task verdict. Preserve Principle source and positive support prerequisites. Partial recovery does not establish task completion. A concrete candidate can be rejected on evidence; no activation is forced by this contract.
\end{promptcard}

\begin{promptcard}{Review Output Contract (2/5): Model Inspection}
For Model, return assessment.failed_task_reviews. When enabled=false, review EVERY task in model_failed_tasks exactly once, even for no_supported_candidate: task_id, outcome_success and outcome_score copied from program facts, requested_effects, observed_result, remaining_gap, closure_basis, and evidence_ids. Compare the original requested effects with observed actions/feedback and the evaluator outcome; a repaired local error never changes the evaluator verdict. Cite that task's context AND action/feedback IDs from the checklist. closure_basis is no_supported_improvement or verified_external_failure only when supported; use uncertain if remaining failure cannot be explained from available evidence, or behavior_improvement if it reveals worthwhile behavior to learn (then Model cannot remain disabled). Explain why the remaining gap offers no supported learning benefit; do not just repeat 'recovered' or 'not persistent'. Missing action evidence requires uncertain, not a fabricated closure. When Model is enabled, failed_task_reviews may be [] because one concrete supported improvement suffices. These are concise audit conclusions, not private deliberation.
\end{promptcard}

\begin{promptcard}{Review Output Contract (3/5): Experience Inspection}
For Experience, return assessment.guidance_reviews. Before closing, inspect EVERY task in experience_review_candidates: name a specific trigger, reusable guidance, expected_benefit, basis, reason and evidence_ids. Error observations and later actions are leads, not a program verdict of recovery. Compare their arguments and feedback, and cite both when claiming recovery. basis is worth_testing, already_covered, no_transferable_guidance or insufficient_evidence. One case and a useful recovery lesson can suffice; neither cross-task recurrence nor unresolved failure is required. already_covered needs actual exposed Experience IDs for that task. A worth_testing finding requires enabled=true; insufficient_evidence is uncertainty, not no need. When enabled, guidance_reviews may be [] if the route candidate supplies another concrete supported opportunity.
\end{promptcard}

\begin{promptcard}{Review Output Contract (4/5): Tool Inspection}
For Tool, return assessment.operation_reviews. Before closing, inspect EVERY task in tool_operation_candidates: name a parameterizable operation, inputs_outputs, expected_saving, basis, benefit_or_rejection and action/feedback evidence_ids. basis is worth_testing, already_covered, not_worth_encapsulating or insufficient_evidence. State an actual possible wrapper before assessing its saving in repeated construction, steps or error risk. Atomic API availability, standard pagination and successful inline execution are not themselves rejection criteria. already_covered needs an actually exposed callable Tool, not an API. A worth_testing finding requires enabled=true. When enabled, operation_reviews may be [] if the route candidate supports another operation. Checklists are inspection leads, not proof of value, and do not limit discovery to listed operations.
\end{promptcard}

\begin{promptcard}{Review Output Contract (5/5): Principle Inspection}
For Principle, return assessment.source_reviews. Before closing, inspect EVERY Experience in principle_review_sources and EVERY listed task, even when candidate=null. When enabled, inspect every selected source and at least two supporting tasks per selected source; other sources/tasks may be omitted. For each inspected source compare its text with each inspected task's observed_behavior and paired feedback; mark application=applied/not_applied/unknown and support=positive/negative/unknown, citing same-task action/feedback. Here applied means observed behavior matches the guidance, NOT that optional guidance caused it. Task success and exposure alone do not establish application. Propose a shared abstraction {trigger, action, guard}; evaluate its cross-scenario value using worth_testing, no_additional_value, insufficient_positive_support, already_covered or uncertain and explain value_reason. When cited actions establish nonapplication, insufficient_positive_support may use abstraction=null and support=unknown; this is known absence of positive use, not unresolved application. Other value judgments require a concrete abstraction. Reserve unknown application for unavailable/ambiguous evidence. At least two distinct tasks with observed application and positive feedback are required for every selected source. Successful substeps or recovery within a failed task can count. Existing effectiveness is a reason to examine consolidation, not proof that abstraction has no value. already_covered requires covered_by_principle_ids of actually exposed PRINCIPLES, never the Experience source itself. Missing evidence means uncertainty, not a valid closure. Initial Principle decisions follow this same contract. Source inspection diagnoses demand only: generation still promotes eligible training Experiences, not test trajectories.
\end{promptcard}

\noindent\textbf{Conditional review fragments.} The \texttt{review\_\allowbreak{}note} field remains empty initially and specifies the stage during reassessment. The instruction covers focused reassessment, model-closure review, and additional-sample review. The \texttt{review\_\allowbreak{}stage} field names this stage, and \texttt{failed\_\allowbreak{}task\_\allowbreak{}ids\_\allowbreak{}json} lists the sampled failures for review.

\begin{promptcard}{Closure Review: Shared Instruction}
CLOSURE REVIEW: {review_stage}. This is a fresh assessment of closure, not JSON repair. Do not defer to previous negative judgments. No route is forced open. For EVERY failed sampled task, compare the original requirement, actual actions and paired feedback, evaluator outcome, and remaining unmet need. Address that requirement -> action -> outcome -> remaining need chain explicitly in the relevant route reasons, with context and action/feedback Evidence IDs. A score below success cannot be dismissed merely because execution succeeded or an answer was returned. Separate recovered local errors from unfulfilled requested effects. Where source text is incomplete, state uncertainty. For each closure explain why its potential contribution is absent, already covered, or unsupported by the inspected facts. For Model a local material omission can suffice; broad, fundamental or persistent deficits are not prerequisites. For Principle examine abstraction value of recurring exposed Experiences even when their behaviors already work; success does not remove promotion value. Preserve the program's valid prerequisite closure when support is absent. For Tool examine reusable operational structure and benefit even when standard APIs exist. Return the existing route schema only. Failed task IDs requiring reassessment: {failed_task_ids_json}
\end{promptcard}

Focused and model-closure reviews append the following fragment to the shared instruction. The \texttt{previous\_\allowbreak{}decisions\_\allowbreak{}json} field supplies prior decisions and rationales for comparison with interaction evidence.

\begin{promptcard}{Closure Review: Previous Decisions}
Previous stage decisions (hypotheses to reassess, not evidence): {previous_decisions_json}
\end{promptcard}

Model-closure review then appends the following instruction to reassess the decision to leave model weight refinement inactive. Accepted decisions for the other components remain fixed during this targeted review.

\begin{promptcard}{Closure Review: Model-Specific Reassessment}
Reassess ONLY Model. Other modules retain their accepted judgments. Address EVERY failed sampled task, including repeated errors despite feedback, unsupported relative-date assumptions and unfulfilled requested side effects. If you change the previous Model decision, explain which evidence warrants the change. A verified external failure or absence of a supported learning benefit may still justify closure.
\end{promptcard}

Additional-sample review adds the following fragment to examine further tasks from the same interaction window. Its input contains the new sample evidence, omitting previous decisions and the model-specific fragment.

\begin{promptcard}{Closure Review: Additional Sample}
These are additional, previously unreviewed tasks from the SAME deployment window. Judge them from their supplied evidence; previous-stage task details are not present and must not be cited. This bounded check does not establish absence of needs in unreviewed window tasks.
\end{promptcard}

Independent symbolic reassessment fills \texttt{review\_\allowbreak{}note} with this fragment. It retains the diagnostic template and structured review requirements and assesses the requested symbolic components without previous decisions.

\begin{promptcard}{Independent Symbolic Reassessment}
INDEPENDENT SYMBOLIC ASSESSMENT: Assess ONLY the requested modules from deployment evidence. No previous judgments are supplied. Both True and False are valid outcomes. Use the supplied JSON schema and listed Evidence IDs. Unrequested decisions are preserved.
\end{promptcard}

\noindent\textbf{Conditional diagnostic supplements.} Trace checks can identify evidence relevant to a decision against model weight refinement. The following supplement adds these patterns and their source locations directly to the diagnostic message through \texttt{closure\_\allowbreak{}counterevidence\_\allowbreak{}json}, separately from \texttt{review\_\allowbreak{}note}.

\begin{promptcard}{Diagnostic Supplement: Counterevidence to Model Closure}
Program-checked counterevidence to Model closure: {closure_counterevidence_json}
These observations identify failed limit-reaching traces with repeated invalid documentation calls or a terminal sequence of identical documentation requests/results or non-executable text with empty feedback. Inspect the listed positions: earlier login/setup and local error recovery do not resolve the later stall. The observations do not assert which requested effects are absent. Assess the concrete behavior needed to make progress; a blanket 'already recovered/no behavior improvement' closure contradicts these facts. If you cannot substantiate a decision, report uncertainty instead of inventing successful recovery.
\end{promptcard}

\subsubsection{Validation-Driven Output Repair}
\label{subapp:diagnostic_repair_prompts}
If validation leaves a requested route unresolved, the following instruction supports a second and final local attempt. Evidence-only repair preserves the routing decision, while full reassessment revisits unresolved decisions. Accepted routes remain fixed. The placeholders supply affected components, validation errors, rejected outputs, and an index of available evidence for checking the revised diagnostic response.

\begin{promptcard}{Unresolved-Route Output Repair}
Repair ONLY these unresolved modules: {unresolved_modules_json}
Validation errors: {validation_errors_json}
Rejected original judgments: {rejected_judgments_json}
Evidence-only repair modules: {evidence_only_modules_json}
For evidence-only modules return ONLY evidence_ids. The program freezes enabled, reason and source_experience_ids. Select actual supporting evidence from the supplied registry; merely sharing a task ID is insufficient. If the evidence cannot support the original reason, return evidence_ids=null rather than inventing support or changing the judgment. Other unresolved modules need a full reassessment, including an explanation in reason if the original decision changes. The program retains original registered deployment citations and appends your new IDs. Return missing supporting IDs; do not rewrite the original decision, reason or assessment. If any retained citation or original claim is unsupported, return evidence_ids=null. The merged citations must cover EVERY literal task ID and Evidence ID in the frozen reason, and candidate citations when present. For behavior claims inspect action/feedback sources; a context pointer alone does not prove recovery or an omitted action. If any claim cannot be supported, return evidence_ids=null. The index does not authorize rewriting the reason.
Reference repair index (IDs are pointers, not proof of the claim):
{reference_repair_index_json}
Other modules are already accepted and cannot be changed. Return only the requested routes.
\end{promptcard}

\subsection{Supervision Generator Prompts} \label{subapp:supervision_prompts}

The supervision generator constructs supplementary synthetic tasks and evaluative signals from environmental interactions. The templates below cover exploration, task proposal, canonical task validation, and outcome reward discrimination for synthetic tasks. Prefix-potential assessment additionally supplies step-level supervision for both oracle and synthetic tasks, following Section~\ref{subsec:supervision_generator} and Appendix~\ref{subapp:supervision_generator}.

\subsubsection{Environment Exploration and Coverage Guidance}
\label{subapp:environment_exploration_prompts}
The four blocks below form the exploration system message, which directs interaction without a predefined target query. The \texttt{environment\_\allowbreak{}profile\_\allowbreak{}instruction} field supplies the environment overview and optional coverage preferences. Retained environment initialization instructions follow this message, while the APPWorld bootstrap and applicable coverage anchors enter as separate user messages.

\begin{promptcard}{Environment Exploration (1/4): Role and Core Principles}
# Role and Mission

You are an **Intelligent Environment Explorer** with strong curiosity, systematic thinking, and adaptive learning capabilities. This is your first time entering this environment. Your mission is to **gain a deep understanding** of the environment's mechanisms, available entities, operations, and potential applications through structured exploration.

---

## 1. Environment Description

{environment_profile_instruction}

### Use Environment Description

In the exploration, you should fully leverage the environment description if provided:
- Treat this description as your primary reference and "map" of the environment.
- Continuously refer back to it when selecting actions - do not just read it once.
- Map each described entity, attribute, and operation to potential API calls or exploration paths.

---

## 2. Core Exploration Principles

### 2.1 Progressive Deep Exploration
- **Avoid Simple Repetition**: Do not repeatedly test the same APIs with identical parameters and sequence.
- **Result-Based Exploration**: Always base the next action on the result of the previous step.
- **Deep Diving**: When an interesting result appears, explore related functionalities in depth.

### 2.2 Context-Aware Decision Making
- **Result Analysis**: Carefully interpret the return values of each API call.
- **State Tracking**: Maintain an internal record of the current environment state and information already obtained.
- **Associative Thinking**: Identify correlations and possible combinations between different APIs.

---
\end{promptcard}

\begin{promptcard}{Environment Exploration (2/4): Strategy and Decision Framework}
## 3. Exploration Strategy

### Phase 1: Initial Mapping (First 3-5 steps)
1. **Breadth Scanning**: Test representative APIs to understand basic functional categories.
2. **Identify Core Functions**: Differentiate between query-type, operation-type, and configuration-type APIs.
3. **Discover Data Flow**: Identify which APIs produce data and which consume it.

### Phase 2: Deep Exploration (Subsequent steps)
1. **Chained Exploration**: Use outputs from one step as inputs for the next.
2. **Boundary Testing**: Explore parameter ranges and edge cases.
3. **Combination Experiments**: Test meaningful API combinations.

### Phase 3: Pattern Discovery
1. **Workflow Identification**: Recognize recurring operational sequences.
2. **Scenario Construction**: Imagine real-world problems these API sequences could solve.

---

## 4. Action Decision Framework

Before selecting the next action, ask:
1. **New Information Utilization**: What new information did I get from the last step? How can it be used?
2. **Exploration Value**: What new understanding will this action bring?
3. **Avoid Redundancy**: Is this action too similar to a previous one?
4. **Depth-First**: Should I explore deeper instead of switching to an unrelated area?

---
\end{promptcard}

\begin{promptcard}{Environment Exploration (3/4): Action Selection}
## 5. Action Selection Guidelines

- **If last step returned data**: Try using it as input for other APIs.
- **If last step failed**: Diagnose the reason and adjust parameters, or try related APIs.
- **If last step succeeded**: Explore follow-up operations or parameter variations.
- **If a new API type is discovered**: Temporarily pause other exploration and test it.

**Avoid**:
- Testing APIs in alphabetical/fixed order.
- Ignoring return data.
- Repeating calls with identical parameters.
- Jumping without logical connection.

**Encourage**:
- Choosing actions based on results.
- Using obtained data as input.
- Deep exploration of interesting patterns.
- Finding logical associations between APIs.

---
\end{promptcard}

\begin{promptcard}{Environment Exploration (4/4): Output Format and Goal}
## 6. Output Format for Each Step

At each step, generate Python code to interact with the environment. Wrap your code in a markdown code block (```python ... ```). Stop generating immediately after closing the code block. DO NOT simulate the environment's output. Wait for the real response.

---

## 7. Overall Goal

Your goal is **not** to complete a specific task, but to **gain a deep, structured understanding** of the environment's capabilities, constraints, and potential real-world applications. Every step should make your understanding more complete.
\end{promptcard}

\noindent\textbf{Environment-profile context.} The overview describes the user context and available entities, attributes, and operations. It enters \texttt{environment\_\allowbreak{}profile\_\allowbreak{}instruction} with any exploration preferences. Within the overview, \texttt{entity\_\allowbreak{}section} contains the entity descriptions below or remains empty if none are specified.

\begin{promptcard}{Exploration Context: Environment Overview}
### Environment Overview
- **User Name**: {user_name}
- **User Background**: {user_background}
{entity_section}
\end{promptcard}

The \texttt{entity\_\allowbreak{}section} field begins with \texttt{\#\#\# Entities in the Environment} and contains one block per available entity. Each block lists attributes and operations, with headings for nonempty collections.

\begin{promptcard}{Exploration Context: Entity Description}
#### Entity: {entity_name}
- Description: {entity_description}
- Attributes:
  - **{attribute_name}**: {attribute_description}
- Available Operations:
  - **{operation_name}**: {operation_description}
\end{promptcard}

\noindent\textbf{Exploration coverage preferences.} When enabled, the following fragment extends the overview with target entities, supported operations, and relational difficulty preferences from the environment profile. These preferences guide exploration, while subsequent execution-based assessments determine task admission.

\begin{promptcard}{Exploration Context: Coverage Preferences}
### Exploration Coverage Targets
Treat these preferences as exploration targets rather than mandatory quotas.
- Seek coherent workflows involving around {num_entities} related entities or applications when supported by the environment.
- Seek around {num_operations} meaningful operations that contribute to the workflow.
- Prefer workflows matching this relation difficulty: {difficulty_description}
- Prioritize data dependencies where the output of one operation becomes a later input or condition.
- Prefer coherent workflows where a value or condition produced in one business domain is consumed by an operation in another domain.
- A multi-domain workflow must contain an actual data or control dependency.
- Do not combine unrelated entities or operations merely to satisfy counts.
- Do not combine independent workflows merely to mention more domains.
- Documentation lookup, authentication, retries, and failed calls do not by themselves satisfy the meaningful-operation target.
- If current evidence cannot support a coherent complex workflow, continue exploring related capabilities or retain a simpler coherent workflow.
- If current evidence cannot support a grounded dependency, retain a simpler single-domain workflow.
\end{promptcard}

Additional profile rubrics enter the following fragment through \texttt{rubric\_\allowbreak{}bullet\_\allowbreak{}lines}, with one hyphen-prefixed entry per rubric. The fragment appears when these additional criteria are available.

\begin{promptcard}{Exploration Context: Optional Rubrics}
**Exploration Rubrics**:
{rubric_bullet_lines}
\end{promptcard}

\noindent\textbf{APPWorld exploration bootstrap.} The following user message begins exploration after the system message and environment initialization instructions. It directs the agent to establish successful workflows through environmental interaction, with the active seed task excluded from the exploration objective.

\begin{promptcard}{APPWorld Exploration: Bootstrap User Message}
Begin free exploration without reading or solving the active seed task. Choose 2 to 3 business apps and inspect only the API documentation needed for one coherent workflow. Execute the workflow far enough to produce successful environment evidence. Prefer a real dependency where an output or condition from one business app becomes an input or control condition for another business app. After a shallow successful chain, use remaining steps to discover filtering, deduplication, sorting, or aggregation over real returned collections, followed by a conditional business mutation or verification when supported. Authentication, API-documentation lookup, credential/token handling, and task submission are support steps and must not be counted as business complexity. Never append an independent app action merely to increase the app count. Do not call apis.supervisor.show_active_task().
\end{promptcard}

\noindent\textbf{Exploration coverage-anchor instruction.} An assigned coverage anchor adds the following user message after the bootstrap. Operation families appear as a comma-separated list, defaulting to \texttt{unspecified} when absent. An absent anchor description is represented by \texttt{not provided}.

\begin{promptcard}{Exploration Anchor: Base Instruction}
Primary coverage anchor: {anchor_name} (business API domain: {anchor_domain}).
Environment description: {anchor_description}
Known operation families: {operation_families}.
Start by obtaining concrete successful business evidence from this anchor. API documentation and authentication do not count as using the anchor.
\end{promptcard}

When enabled, cross-application dependency preferences add the following instruction to the anchor message.

\begin{promptcard}{Exploration Anchor: Cross-Application Preference}
Then strongly prefer a difficult cross-app workflow where a runtime output or condition from the anchor is consumed by another business app, or another business app supplies a dynamic input or control condition to the anchor. Prefer filtering, set reasoning, conditional mutation, and state verification. Never append an independent app action.
\end{promptcard}

Single-application fallback adds the following fragment after any cross-application preference. The two options are configured independently, allowing both preferences to appear in the same anchor message.

\begin{promptcard}{Exploration Anchor: Single-Application Fallback}
If no grounded cross-app dependency can be demonstrated, keep a coherent successful single-app workflow instead of forcing an unrelated application.
\end{promptcard}

\noindent\textbf{Conditional anchor reminder.} At the configured step, the following message recalls an assigned anchor if the trajectory has not yet recorded it. Each exploration trajectory receives at most one such reminder.

\begin{promptcard}{Exploration Anchor: Conditional Reminder}
Return to the primary coverage anchor {anchor_name}. Obtain a real successful `apis.{anchor_domain}.*` business call, then seek a grounded cross-app data or control dependency. Do not count documentation or authentication, and do not force an unrelated application.
\end{promptcard}

\subsubsection{Task Proposal and Candidate Assessment}
\label{subapp:task_proposal_prompts}
The task summarizer $A_{\text{sum}}$ converts exploration evidence into candidate queries and provisional reference solutions. A system message specifies the proposal criteria, and a user message supplies the observed workflows and available context. The resulting candidates then enter environmental validation.

The six blocks below form the proposal system message. Difficulty preferences populate \texttt{structural\_\allowbreak{}difficulty\_\allowbreak{}instruction}, and profile preferences populate \texttt{task\_\allowbreak{}preference\_\allowbreak{}instruction}. The \texttt{max\_\allowbreak{}tasks\_\allowbreak{}per\_\allowbreak{}trajectory} field bounds proposals per exploratory trajectory.

\begin{promptcard}{Task Proposal (1/6): Role, Objectives, and User Intent}
# ROLE
You are a **Real-World Task Discovery Expert**. Your job is to analyze an agent's exploration trajectory and transform it into **realistic, user-centered tasks** that could be solved using the same interaction **patterns**.

---

# OBJECTIVES
1. **Understand Capabilities**: Analyze the recorded API calls to identify the actual functional capabilities demonstrated.
2. **Think Like a Real User**: Imagine practical, everyday problems where a real person would naturally use this exact API call sequence. Create problems that use **multiple different API calls**, not just a single call. Use **clear, specific, verifiable** user requests.
3. **Abstract into MUSE Format**: For each task provide **query** (natural-language request), **ground_truth** (a concrete, ordered action sequence), **source_apis** (the exact APIs demonstrated by the trajectory), and **confidence** (0.0-1.0).

---

# RULES FOR SCENARIO CREATION
## 1. Focus on User Intent
- Always start from a **human goal**. Avoid restating the API in technical terms-capture the **why**.

## 2. Preserve the Trajectory's App Domain (Critical)
- **Match the domain of the trajectory**: If the trajectory shows **payment, transfer, balance, or sending/requesting money**, write **payment/transfer-style** user requests (e.g. "Do I have at least $150 in my account?", "Send $20 to John for dinner", "Request payment from Alice for the split bill").
- If the trajectory shows **music, playlist, album, or track** actions, write **music/playlist-style** requests (e.g. "Add this song to my workout playlist", "Create a playlist named Summer 2025").
- If the trajectory shows **other app types** (e.g. shopping, search), write requests that fit **that** app.
- **Do NOT** convert every trajectory into the same app. The query must reflect what the trajectory actually demonstrated.

## 3. Specificity & Verifiability
- The query must be **precise enough** that success/failure can be judged. Include concrete details: numbers, names, dates, etc.
\end{promptcard}

\begin{promptcard}{Task Proposal (2/6): Trajectory Grounding}
## 4. Trajectory Grounding (Critical)
- Create tasks **backward from behavior that was actually demonstrated** in the trajectory. Do not invent a desired capability and then assume the environment supports it.
- First identify the **smallest connected successful business subgraph** in the trajectory that can support a complete user goal. Build the task from that successful subgraph, not from the full raw exploration trace.
- Prefer a compact workflow with 2 business apps, 1 genuine cross-app dependency, and roughly 3-6 meaningful successful actions when the evidence supports it. This is a preference, not a requirement: a simpler grounded task is better than a larger speculative one.
- Every app, entity, record, and dependency required by the query must be visible in the trajectory or obtainable through an API call that appears in the trajectory.
- `source_apis` must list the exact `apis.<app>.<function>` calls needed by the proposed solution. Every listed API must appear in the trajectory.
- `ground_truth` must preserve the demonstrated data flow: state which API obtains an identifier or value and which later API consumes it.
- Do not promote incidental implementation details into the user goal. Raw internal IDs and arbitrary choices such as "the first result" or index `[0]` may appear only when the user explicitly specified that binding or the successful trajectory proved a unique deterministic match.
- Preserve business inputs that the user must supply, such as a recipient, email address, amount, date, or requested item. These are necessary task conditions, not answer leakage.
- Do not invent API names, hidden records, contacts, dates, credentials, waiting periods, or cross-app relationships that the trajectory did not establish.

## 5. Cross-Domain Dependency (Evidence-Bounded)
- Prefer a coherent cross-domain workflow only when a value or condition produced in one business domain is consumed by an operation in a second domain.
- A valid cross-domain edge requires an output or condition from one business app to become an input or control condition for another business app.
- A multi-domain workflow must contain an actual data or control dependency supported by trajectory evidence.
- Do not combine independent workflows merely to mention more domains.
- Do not attach a third or fourth app unless it lies on the same successful data/control path and is necessary for the user goal.
- If trajectory evidence does not demonstrate a grounded dependency, emit a simpler single-domain task instead.
- Do not include `supervisor` or `api_docs` in `business_domains`; they are support domains.
\end{promptcard}

\begin{promptcard}{Task Proposal (3/6): Structural Difficulty and Practicality}
## 6. Structural Difficulty (Primary Difficulty Signal)
- **Data dependency**: Prefer tasks where an earlier API result supplies an ID, value, collection, or condition used by a later API.
- **Control dependency**: Prefer tasks where a query result determines whether or how a later operation runs.
- **State verification**: After a state-changing operation, prefer a grounded follow-up query that verifies the requested outcome.
- **Set reasoning**: When demonstrated, use filtering, intersection, sorting, uniqueness, comparison, or aggregation that is necessary for the user goal.
- Authentication, login, token/session management, API-documentation lookup, retries, failed calls, and `apis.supervisor.complete_task` do not count as meaningful business actions.
- Never append unrelated actions merely to make a task longer.
- Generate a mix of structural levels when evidence permits:
  - `medium`: about 3 meaningful actions and at least 1 dependency.
  - `hard`: about 4 or more meaningful actions and at least 2 dependencies; prefer verification.
  - `very_hard`: about 5 or more meaningful actions and at least 2 dependencies; prefer coherent cross-application data flow only when grounded.
- A strong single-domain task with real dependencies is better than an invented or weakly grounded cross-application task.
- When evidence supports multiple candidates, prefer different reasoning structures (for example dynamic selection, set reasoning, conditional cross-app mutation, or state verification).
- Do not create near-duplicate candidates by changing only an entity, amount, date, or ID while keeping the same reasoning structure.
{structural_difficulty_instruction}

## 7. Practicality
- Use **relatable, everyday** scenarios. Avoid purely exploratory or API-test-only tasks.

{task_preference_instruction}

---
\end{promptcard}

\begin{promptcard}{Task Proposal (4/6): Output Format}
# OUTPUT FORMAT (MUSE)
Output **only** a JSON array. No other text. Example:
```json
[
  {"query": "A natural, specific user request", "ground_truth": "<semicolon-delimited API-level solution grounded in the trajectory>", "source_apis": ["<exact API names copied from the trajectory>"], "difficulty_features": {"structural_level": "hard", "meaningful_actions": 4, "data_dependencies": 2, "control_dependencies": 0, "has_state_verification": true, "has_set_reasoning": false, "business_domains": ["app_name"]}, "confidence": 0.8},
  ...
]
```
- **query**: Natural-language user request (must match the trajectory's app domain).
- **ground_truth**: A concise API-level reference solution with at least three distinct executable actions. Use exact API names demonstrated in the trajectory, preserve argument/data dependencies, write one action per clause, and separate clauses with semicolons (`;`).
- **source_apis**: A JSON list containing only exact API names observed in the trajectory and required by the reference solution.
- **difficulty_features**: A structural claim for diagnostics. Include `structural_level`, meaningful action/dependency counts, verification/set-reasoning flags, and business domains. Do not include `supervisor` or `api_docs` in `business_domains`, and do not count authentication, login, token/session management, or task submission as business actions.
- **confidence**: 0.0-1.0. Only include tasks with confidence >= 0.6.

Generate at most {max_tasks_per_trajectory} tasks. Prioritize quality over quantity.

---
\end{promptcard}

\begin{promptcard}{Task Proposal (5/6): Difficulty Calibration Examples}
# APPWORLD DIFFICULTY CALIBRATION (Goal shape only)
Do not copy these goals, entities, or applications. They demonstrate the amount of hidden reasoning expected when the current trajectory independently supports an analogous workflow:
- "Reset friends on Venmo to be the same as my friends in my phone. Befriend and unfriend as needed." This requires set difference and conditional mutations across two business apps.
- "Go through all my Spotify playlists and remove all the songs from them that are in a file, then put them in a new archive playlist." This requires parsing, matching, deduplication, iteration, and cross-app dataflow.
- "My roommates replied with suggested changes to a shared road-trip playlist. Update the playlist accordingly." This requires extracting dynamic instructions from messages and applying conditional playlist edits.

These examples never override trajectory grounding. If the successful evidence does not contain the required APIs and dependency pattern, do not imitate the example and do not invent the missing workflow.

---
\end{promptcard}

\begin{promptcard}{Task Proposal (6/6): Final Checklist}
# CHECKLIST BEFORE FINALIZING
- **Clear goal** - What exactly is the user trying to achieve?
- **Concrete details** - Who, what, when, how much?
- **Verifiable** - Can success/failure be objectively determined?
- **Domain match** - Does the query match the app/domain shown in the trajectory (payment vs music vs other)?
- **Action format** - Does ground_truth contain at least three semicolon-delimited executable actions?
- **API provenance** - Is every source_apis entry present in the trajectory, with no invented capability?
- **Entity provenance** - Does every required entity or relation have evidence in the trajectory?
\end{promptcard}

\noindent\textbf{Configured proposal preferences.} When enabled, \texttt{structural\_\allowbreak{}difficulty\_\allowbreak{}instruction} specifies the candidate distribution, with percentage fields reflecting its target proportions. The \texttt{optional\_\allowbreak{}structural\_\allowbreak{}preferences} field contains the active entries below or remains empty when none are configured.

\begin{promptcard}{Task Proposal: Configured Difficulty Distribution}
Configured one-shot candidate mix (soft targets; trajectory evidence always wins):
- medium: {medium_percentage}
- hard: {hard_percentage}
- very_hard: {very_hard_percentage}
{optional_structural_preferences}
\end{promptcard}

The following fragment contains the enabled structural preferences. Entries retain their displayed order, with each preference included according to its individual configuration setting.

\begin{promptcard}{Task Proposal: Optional Structural Preferences}
Configured structural soft preferences (never hard acceptance requirements):
- dynamic selection: derive target objects from runtime business results instead of hardcoding IDs or positions
- set reasoning: use grounded filtering, intersection, difference, deduplication, sorting, top-k, or aggregation
- conditional actions: mutate only the objects whose runtime state satisfies the user condition
- cross-app dataflow: make an App A result determine an App B argument or control decision
- state verification: re-read and check the user-visible result after mutation
\end{promptcard}

The profile supplies the following \texttt{task\_\allowbreak{}preference\_\allowbreak{}instruction} fragment, with numerical targets and difficulty descriptions filled at runtime. Additional generation rubrics enter its optional field.

\begin{promptcard}{Task Proposal: Environment-Profile Preferences}
### Target Task Characteristics
Use the following values as preferred complexity targets:
- approximately {num_entities} related entities or applications
- approximately {num_operations} meaningful operations
- relation difficulty: {difficulty_description}

These targets never override trajectory evidence. Generate a task only when the trajectory demonstrates the required entities, operations, ordering constraints, conditions, and data dependencies. Do not combine unrelated exploration branches. Do not invent an operation, dependency, entity state, credential, identifier, or result that is not supported by the trajectory. A simpler grounded task is preferable to a complex unsupported task. If no coherent task is supported, return no task.
{optional_task_rubrics}
\end{promptcard}

When profile rubrics are available, the following fragment populates \texttt{optional\_\allowbreak{}task\_\allowbreak{}rubrics}. Each rubric appears as a hyphen-prefixed entry in \texttt{rubric\_\allowbreak{}bullet\_\allowbreak{}lines}, preserving the supplied task-generation criteria.

\begin{promptcard}{Task Proposal: Optional Profile Rubrics}
**Rubrics**:
{rubric_bullet_lines}
\end{promptcard}

\noindent\textbf{Task-proposal input template.} The user message supplies exploration evidence through \texttt{trajectory}, preceded by the verified-workflow instruction when applicable. The \texttt{old\_\allowbreak{}objectives\_\allowbreak{}section} field contains available coverage-anchor context followed by existing task objectives.

\begin{promptcard}{Task Proposal: User Message}
## Exploration Trajectory
```
{trajectory}
```

{old_objectives_section}

# Now Start
Analyze the trajectory above. If a **Hard Constraint** section appears above, follow it strictly. Identify specific tasks grounded in the trajectory, and output a JSON array of tasks. **Keep the same app domain as the trajectory** (payment vs music vs other). Do not output any text outside the JSON array.
\end{promptcard}

When the trajectory contains the \texttt{Verified Successful Workflow Evidence} heading, the following instruction precedes its text in the proposal user message.

\begin{promptcard}{Task Proposal: Verified-Workflow Instruction}
The evidence below contains only verified successful actions. Generate each candidate from exactly one connected business component. The task must be standalone in a fresh environment: do not assume that transient objects created during exploration still exist, and do not add state changes the user did not ask for.
\end{promptcard}

\noindent\textbf{Coverage-anchor context.} A recorded coverage anchor enters the user message before existing task objectives. Boolean values appear as \texttt{true} or \texttt{false}, and verified dependency edges are included when available. The final instruction reflects whether execution evidence supports the anchor.

\begin{promptcard}{Task Proposal: Coverage-Anchor Context}
## Coverage Anchor
- Primary anchor: {anchor_name} ({anchor_domain})
- Anchor observed during exploration: {anchor_observed}
- Anchor has verified successful business evidence: {anchor_successful}
{optional_dependency_edge_line}
{anchor_evidence_guidance}
\end{promptcard}

Verified dependency edges appear as comma-separated producer-to-consumer pairs. The \texttt{anchor\_\allowbreak{}evidence\_\allowbreak{}guidance} field selects the instruction matching the recorded execution support for the anchor.

\begin{promptcard}{Task Proposal: Verified Dependency Edges}
- Verified dependency edges: {producer_consumer_pairs}
\end{promptcard}

\begin{promptcard}{Task Proposal: Supported Anchor}
When possible, strongly prefer a grounded candidate that preserves the anchor and a verified cross-app dependency when the successful evidence supports one.
\end{promptcard}

\begin{promptcard}{Task Proposal: Unsupported Anchor}
The anchor lacks verified successful evidence. This is a soft coverage preference: generate from other grounded successful business evidence rather than inventing or forcing the anchor.
\end{promptcard}

\noindent\textbf{Existing-task diversity constraint.} Available objectives populate \texttt{existing\_\allowbreak{}task\_\allowbreak{}objectives} in the section below. These objectives follow any coverage-anchor context and support proposal diversity.

\begin{promptcard}{Task Proposal: Existing-Task Constraint}
## Existing Tasks (Training Set or Already Generated)
The following tasks **already exist** in the training set or have been generated earlier in this pipeline for this exploration seed.

{existing_task_objectives}

### Hard Constraint (Read Carefully)
- **DO NOT** generate queries that are **semantically identical** to any item above (including trivial rephrasings of the same user intent).
- **Instead**, focus on extracting **implicit user needs** from the interaction trajectory that are **NOT** covered by the list above-novel goals the trajectory demonstrates without duplicating those objectives.
- If the trajectory only supports the same goals as above, output **fewer** tasks or **none** rather than near-duplicates; prioritize **distinct** scenarios only when the trajectory genuinely supports them.
\end{promptcard}

\noindent\textbf{Query-only candidate difficulty assessment.} When enabled, a separate model call estimates query-level features for candidate ranking before validation. Its input contains the query, and the system message specifies feature fields and permitted values. The JSON example illustrates this structure with placeholder values.

\begin{promptcard}{Candidate Difficulty Assessment: System Message}
Estimate how difficult a synthetic environment task is from its user query alone. This is a cheap ranking prior, not a pass/fail decision.

Return exactly one JSON object with these integer fields, each in {0, 1, 2}:
{
  "hidden_runtime_information": 0,
  "cross_app_dynamic_dependency": 0,
  "aggregation_or_computation": 0,
  "conditional_reasoning": 0,
  "procedural_leakage": 0,
  "fixed_answer_hint": 0,
  "estimated_independent_difficulty": 0
}

Use 0 for absent, 1 for moderate, and 2 for strong.
- Reward tasks whose key values must be discovered from live environment state.
- Reward real producer-consumer dependencies where one application's output is required by another application.
- Reward filtering, aggregation, comparison, local computation, or branching.
- Penalize queries that reveal API calls, execution order, intermediate values, credentials, identifiers, exact dynamic answers, or most of the solution.
- Do not infer difficulty merely from query length or the number of app names.

Output JSON only. Do not output markdown, analysis, labels, or extra keys.

/no_think
\end{promptcard}

The following user message supplies the candidate query for this assessment, keeping the ranking input separate from the provisional reference solution and exploratory trajectory.

\begin{promptcard}{Candidate Difficulty Assessment: User Message}
Candidate Query:
{query}
\end{promptcard}

\subsubsection{Canonical Task Construction and Validation}
\label{subapp:canonical_task_prompts}
Canonical task construction connects the query $q_i^{\star}$ to the reference solution $y_i^{\star}$ and execution evidence $H_{i,\text{ref}}^{\star}$. Successful replays support reference construction, while an intermediate task contract retains the objectives and constraints needed to formulate the canonical query. The following paired system and user messages assess requirement coverage, reference consistency, and authorization for state changes.

\noindent\textbf{Environment execution and conditional reference guidance.} Guided and free executions share the system instruction below. Guided execution additionally receives an available reference as a separate system message. Environment initialization messages follow, with the synthetic query supplied as the final user message.

\begin{promptcard}{Task Execution: System Instruction}
You are an autonomous AI agent. Complete the following task by generating Python code.

CRITICAL INSTRUCTION:
1. At each interaction, you must ONLY generate the Python code for the current step.
2. DO NOT simulate the environment's output.
3. DO NOT simulate the USER's response.
4. Stop generating immediately after closing the code block with ```.
5. TASK TERMINATION: Once you have completed the task, you MUST call the termination function (e.g., apis.supervisor.complete_task() or pass the required answer like apis.supervisor.complete_task(answer=...)) to end the task. Do NOT just output conversational text!
\end{promptcard}

\begin{promptcard}{Task Execution: Conditional Reference Message}
<reference>
Reference solution:
{ground_truth}
</reference>
\end{promptcard}

\begin{promptcard}{Task Execution: User Message}
{query}
\end{promptcard}

\noindent\textbf{Query and replay alignment.} This validator compares the query with successful replay evidence to assess requirement coverage and authorized state changes. The \texttt{successful\_\allowbreak{}replay\_\allowbreak{}lines} field contains \texttt{[Step i][ROLE] content} traces, followed by detected state-changing API calls for inspection.

\begin{promptcard}{Query-Replay Alignment: System Instruction}
You are an independent semantic validator for a synthetic task replay.

Judge only whether the Successful Replay Evidence actually completes the User Query. Do not use, infer, or request any reference answer or ground truth.

Rules:
1. Require concrete successful environment evidence for every essential query requirement.
2. Allow necessary read-only support actions such as authentication, API documentation lookup, ID retrieval, pagination, and result verification.
3. Every state-changing action must be semantically authorized by the User Query.
4. Reject unrelated deletion, creation, sending, editing, updating, moving, liking, following, approval, rejection, purchase, payment, transfer, or other mutation.
5. `apis.supervisor.complete_task` is bookkeeping, not proof that the task was completed.
6. Do not invent missing actions, outputs, or user intent.

Return exactly one JSON object with these fields and no other text:
{
  "aligned": true,
  "score": 0.0,
  "missing_requirements": [],
  "unauthorized_state_changes": [],
  "reason": "short reason"
}

`score` must be between 0 and 1. The two list fields must contain short strings. Do not output analysis, reasoning, markdown, labels, or code fences.

/no_think
\end{promptcard}

\begin{promptcard}{Query-Replay Alignment: User Message}
User Query:
{query}

Successful Replay Evidence:
{successful_replay_lines}

Candidate State-Changing APIs:
{state_changing_apis_json}
\end{promptcard}

\noindent\textbf{Replay-grounded reference rewriting.} The following prompt constructs a technical reference from a verified guided replay and the original solution after sensitive-value redaction. Its two system blocks specify the rewriting requirements, while the user message supplies replay evidence in the same step-and-role format.

\begin{promptcard}{Reference Rewriting (1/2): Role and Output}
You are extracting a verified technical reference solution from a successful environment replay.

Return exactly one block in this format and nothing else:
<task>
{"confidence": 1.0, "action_sequence": "<multiline Python-style technical solution>"}
</task>
\end{promptcard}

\begin{promptcard}{Reference Rewriting (2/2): Grounding Constraints}
The `action_sequence` must be a self-contained, multiline Python-style technical solution. Use `# stepN` comments to mark major stages. Use only exact `apis.<app>.<function>` calls demonstrated by the successful replay. Preserve the replay's API order, control flow, and producer-consumer data dependencies. Keep assignments, conditions, loops, filtering, aggregation, and local calculations when required. Obtain IDs, credentials, tokens, records, and computed answers from earlier calls; never hardcode dynamic values. An explicit business credential already present in the Task query, such as a temporary password the user requires for account creation, may be used verbatim. Never recover or hardcode an authentication secret that appears only in replay observations. End with the replay-grounded verification or task-completion call. Remove documentation exploration, failed attempts, tracebacks, retries, and redundant calls. Do not invent APIs, fields, entities, arguments, outcomes, or environment capabilities. Do not include authentication passwords, tokens, replay-only secrets, analysis, reasoning, alternatives, or markdown fences. The only credential literal allowed is an exact business credential explicitly supplied in the Task query for the requested business outcome.

/no_think
\end{promptcard}

\begin{promptcard}{Reference Rewriting: User Message}
Task query:
{query}

Original summarized solution:
{redacted_original_ground_truth}

Verified guided replay:
{verified_guided_replay_lines}

Return the corrected technical action_sequence in the required <task> block.
\end{promptcard}

\noindent\textbf{Query and reference alignment.} When enabled, this assessment compares the rewritten reference directly with the query. The two system blocks specify the semantic criteria and output schema, and the user message supplies this pair without additional replay traces or reference answers.

\begin{promptcard}{Query-Reference Alignment (1/2): Validation Criteria}
You are an independent semantic validator for a synthetic task and its proposed technical reference solution.

Judge only whether the Rewritten Technical Action Sequence faithfully and completely solves the User Query. Treat the query as the sole source of user intent. Do not assume access to a replay, observations, rewards, or another reference answer.

Check all of the following:
1. Every essential query requirement is covered, including requested entities, constraints, filters, time ranges, and multi-part outputs.
2. The submitted result has the requested meaning, cardinality, aggregation, and format; returning a list instead of a count, one output instead of several, or duplicated items is a mismatch.
3. The action sequence does not contradict the query through wrong entities, attributes, states, dates, filters, or verification criteria.
4. Support actions such as authentication, capability lookup, pagination, identifier lookup, and result verification are allowed only when they support the requested goal.
5. Reject unrelated side effects or extra end goals that the query does not request.
6. Reject incorrect computations, including invalid counting, uniqueness, grouping, ordering, comparison, or aggregation logic.
7. Do not invent missing actions, outputs, data, or user intent.
\end{promptcard}

\begin{promptcard}{Query-Reference Alignment (2/2): Output Contract}
Output exactly one JSON object with these fields and no other text:
{
  "aligned": true,
  "score": 0.0,
  "missing_requirements": [],
  "output_mismatches": [],
  "contradictions": [],
  "unjustified_extra_goals": [],
  "computation_errors": [],
  "reason": "short reason"
}

`score` must be between 0 and 1. Each list must contain at most three short strings. Keep `reason` to at most two short sentences. Do not output analysis, reasoning, markdown, labels, or code fences.

/no_think
\end{promptcard}

\begin{promptcard}{Query-Reference Alignment: User Message}
User Query:
{query}

Rewritten Technical Action Sequence:
{rewritten_ground_truth}
\end{promptcard}

\noindent\textbf{Task-contract extraction.} The following prompt extracts objectives and constraints from the generation intent, provisional reference, and successful execution traces. The resulting contract retains task requirements while excluding transient outputs and implementation details. Two system blocks define the extraction criteria, and the user message supplies the three evidence sources as JSON.

\begin{promptcard}{Business-Contract Extraction (1/2): Role and Schema}
Extract a sanitized business contract for one synthetic environment task.

Return exactly one JSON object:
{
  "source_apps": ["application"],
  "destination_apps": ["application"],
  "selection_goal": "abstract outcome used to determine relevant records",
  "business_action": "abstract user-visible outcome",
  "user_constraints": ["constraint"],
  "dynamic_field_types": ["field type"],
  "explicit_user_business_bindings": ["short semantic binding"]
}
\end{promptcard}

\begin{promptcard}{Business-Contract Extraction (2/2): Evidence and Sanitization}
Use the successful business evidence as execution authority, while treating the generation intent reference as provenance for explicit user-provided business constraints. Preserve only values that affect record selection, a requested business mutation, or a final deliverable, such as an email recipient, amount, date range, rating threshold, output path, title, label, group, project, contact, or relationship. Record each value once as a short semantic binding, for example "recipient: jo-harr@gmail.com", "minimum amount: \$30", or "project: Personal".

Do not preserve a number, quoted phrase, identifier, limit, ordering word, or verification detail merely because it occurs in the generation intent. Exclude procedural hints and redundant restatements from both `user_constraints` and `explicit_user_business_bindings`.

Never copy a value that appears only in provisional actions or successful replay evidence. Replace replay-only values with field types such as recipient, amount, record, file, or label. If a value occurs in both the generation intent reference and replay evidence, it remains an explicit user binding. Never include API names, variable names, authentication credentials, execution order, runtime identifiers, or other runtime answers. Exclude api_docs and supervisor from application lists.

Output JSON only. Do not output markdown, analysis, labels, or extra keys.

/no_think
\end{promptcard}

\begin{promptcard}{Business-Contract Extraction: User Message}
{
  "generation_intent_reference": {original_query_json},
  "provisional_business_actions": {provisional_ground_truth_json},
  "successful_business_evidence": {replay_summary_json}
}
\end{promptcard}

\noindent\textbf{Canonical query construction.} The query constructor receives the extracted task contract and produces a canonical request and a conservative alternative preserving the same requirements. The serialized contract forms the user message directly, keeping query formulation tied to the extracted objectives and constraints.

\begin{promptcard}{Canonical-Goal Rendering (1/2): Output and Coverage}
Render a concise, outcome-oriented user request from a sanitized business contract.

Return exactly one JSON object:
{
  "canonical_query": "concise user request",
  "conservative_query": "equally complete safe user request"
}

Before writing, use `selection_goal`, `business_action`, every non-redundant `user_constraints` item, and every `explicit_user_business_bindings` value as a mandatory coverage checklist. Both requests must be equally complete and must preserve every requested outcome, filter, condition, aggregation, cardinality, multi-part deliverable, and cross-application dependency in that checklist. The conservative request may use safer wording, but it must not become shorter by deleting a business requirement or replacing a specific goal with vague phrases such as "relevant records" or "requested outcome".
\end{promptcard}

\begin{promptcard}{Canonical-Goal Rendering (2/2): Wording Constraints}
Write one or two natural sentences, not a field list or a restatement of the contract schema. Do not mention APIs, authentication, credentials, pagination, variables, identifiers, intermediate results, implementation order, or values that must be discovered at runtime. Incorporate only user constraints that affect record selection, requested state changes, or final deliverables. Preserve the values in `explicit_user_business_bindings`, but express each value once and do not copy labels such as `recipient:`, `note_title:`, or `minimum_amount:` into the request. Do not invent applications, entities, constraints, or business goals.

Output JSON only. Do not output markdown, analysis, labels, or extra keys.

/no_think
\end{promptcard}

\begin{promptcard}{Canonical-Goal Rendering: User Message}
{business_contract_json}
\end{promptcard}

\noindent\textbf{Source, contract, and query consistency.} This audit traces requirements from the generation draft through the contract to the canonical query. Draft provenance distinguishes omissions during contract extraction from omitted or unsupported requirements introduced during query construction.

\begin{promptcard}{Canonical Consistency Audit (1/2): Validation Criteria}
Audit one synthetic task across three representations: the original generation draft, its sanitized business contract, and the final canonical user query.

The original query is provenance, not the final user contract. First identify only requirements that affect record selection, a requested business mutation, or a final deliverable. Ignore API names, authentication, credential lookup, pagination, internal variables, runtime identifiers, intermediate results, verification calls, environment control, and implementation order unless a credential itself is explicitly being created, changed, communicated, or delivered as the requested business outcome.

Then perform two independent checks:
1. The sanitized business contract preserves every genuine business outcome, filter, condition, aggregation, cardinality, user-provided binding, multi-part deliverable, and cross-application dependency from the original query, without importing runtime-only answers.
2. The canonical query completely and faithfully expresses every non-redundant requirement in the sanitized business contract without adding a new business goal or leaking solution procedure.
\end{promptcard}

\begin{promptcard}{Canonical Consistency Audit (2/2): Output Contract}
Return exactly one JSON object:
{
  "contract_source_aligned": true,
  "contract_source_score": 0.0,
  "canonical_contract_aligned": true,
  "canonical_contract_score": 0.0,
  "missing_business_requirements_in_contract": [],
  "missing_contract_requirements_in_canonical": [],
  "added_canonical_requirements": [],
  "residual_procedural_leakage": [],
  "ignored_procedural_details": [],
  "reason": "short reason"
}

`contract_source_aligned` is true exactly when `missing_business_requirements_in_contract` is empty. `canonical_contract_aligned` is true exactly when `missing_contract_requirements_in_canonical`, `added_canonical_requirements`, and `residual_procedural_leakage` are all empty. `ignored_procedural_details` is diagnostic only and must list only procedural details intentionally omitted from the contract and canonical query. Both scores must be in [0, 1]. Every list must contain only concise strings.

Output JSON only. Do not output markdown, analysis, labels, or extra keys.

/no_think
\end{promptcard}

\begin{promptcard}{Canonical Consistency Audit: User Message}
{
  "original_query": {original_query_json},
  "business_contract": {business_contract_json},
  "canonical_query": {canonical_query_json}
}
\end{promptcard}

\noindent\textbf{Canonical query and outcome alignment.} This validator compares the query with operations extracted from the reference and successful replay. It checks for leaked execution details and corresponding actions and outcomes, drawing on these canonical materials without the initial generation draft.

\begin{promptcard}{Canonical Outcome Alignment (1/2): Evidence and Criteria}
You validate an AppWorld canonical user query against its canonical ground truth business actions and successful replay business evidence.

Judge only the canonical query supplied in the payload. No original generation query is available, and you must not infer missing constraints from provenance. The other payload fields are implementation evidence, not user-facing text. Never report API names, assignments, loops, authentication, pagination, or other implementation details from those evidence fields as procedural leakage in the canonical query.

Reject when:
1. A user-visible business outcome in the canonical query is not implemented.
2. The canonical ground truth performs a business mutation not authorized by the canonical query.
3. The canonical query itself enumerates API/auth/session mechanics, pagination, internal variables, returned identifiers, or verification calls instead of stating the final business outcome. Natural business sequencing such as finding records and then creating a note or sending an email is allowed.
4. The implementation evidence uses an amount, recipient, payer, contact, group, project, expense, note, file, threshold, or other business binding that neither the canonical query nor an explicit selection rule determines.
5. The canonical query primarily requires a business mutation, but the paired observation in `candidate_mutation_outcomes` shows that the mutation branch did not execute or that the target collection was empty. An API name merely appearing in code is not proof that the mutation actually occurred. Report such a missing user-visible effect in `missing_business_outcomes`.
\end{promptcard}

\begin{promptcard}{Canonical Outcome Alignment (2/2): Output Contract}
Applications and final deliverables may be named naturally. Authentication, credential lookup, API documentation, and task submission in the reference are allowed infrastructure.

Output exactly one JSON object with these fields and no other text:
{
  "aligned": true,
  "score": 0.0,
  "missing_business_outcomes": [],
  "unauthorized_business_mutations": [],
  "residual_procedural_leakage": [],
  "unjustified_business_bindings": [],
  "reason": "short reason"
}

Every list must contain only short categories of at most eight words. Never copy source code or API calls into a list. `score` must be in [0, 1]. Do not output markdown, analysis, labels, or code fences.

/no_think
\end{promptcard}

\begin{promptcard}{Canonical Outcome Alignment: User Message}
{
  "canonical_query": {canonical_query_json},
  "canonical_ground_truth_business_actions": {canonical_business_api_calls_json},
  "successful_replay_business_evidence": {replay_summary_json}
}
\end{promptcard}

\noindent\textbf{Executable task-contract audit.} This audit compares the canonical query with projected task operations from the reference and successful replay. The projections retain data flow, selection logic, and computation while excluding runtime setup. Two system blocks specify the audit criteria, and three user blocks supply the evidence and output schema as one JSON payload. Deterministic structural and leakage checks complement the prompted assessment.

\begin{promptcard}{Task-Contract Audit (1/2): Scope and Evidence}
You are a strict AppWorld task-contract auditor. Compare the canonical user query with the projected business program and API actions from the executable reference. The program preserves assignments, selection logic, loops, conditions, and local transformations so runtime IDs and values can be traced to earlier business results. Audit user-visible business goals and business mutations. Authentication, credential lookup, API documentation, and task submission are allowed runtime infrastructure and have already been removed from the business projection. Never infer them back into the query. Audit leaked implementation steps only in the canonical_query field, never in canonical_reference_business_actions or successful_replay_business_actions.
\end{promptcard}

\begin{promptcard}{Task-Contract Audit (2/2): Rejection Criteria}
Reject genuinely hidden business prerequisites, missing user goals, leaked solution steps in the canonical query, and business mutations not authorized by the query. Also reject any amount, recipient, payer, contact, group, expense, note, file, or dynamic record that the projected business actions select without the canonical query supplying it or defining a selection rule. Reject non-user-specified IDs, contacts, amounts, and other dynamic business values unless the projected program derives them from preceding business results. When the query says all, each, or every, require the program to cover the complete selected collection rather than only one indexed item. Require debt, payment, and transfer direction to agree with the semantic direction of the fields and actions used by the program. Do not judge whether the task is generally useful or whether the final outcome alone looks successful. Respond with one JSON object only.
\end{promptcard}

\begin{promptcard}{Task-Contract Input (1/3): Evidence and Audit Scope}
{
  "canonical_query": {canonical_query_json},
  "canonical_reference_business_program": {canonical_business_program_json},
  "canonical_reference_business_actions": {canonical_business_actions_json},
  "successful_replay_business_program": {replay_business_program_json},
  "successful_replay_business_actions": {replay_business_actions_json},
  "dependency_edges": {dependency_edges_json},
  "state_changes": {state_changes_json},
  "audit_policy": {
    "canonical_query_is_user_contract": true,
    "leaked_solution_steps_audit_source": "canonical_query_only",
    "allowed_reference_infrastructure": [
      "authentication",
      "credential_lookup",
      "api_documentation",
      "task_submission"
    ],
    "presupposed_dynamic_state_excludes": [
      "preconfigured_accounts",
      "credentials",
      "authentication_tokens"
    ],
    "unauthorized_state_changes_scope": "business_mutations_only",
    "unauthorized_state_changes_excludes": [
      "login",
      "logout",
      "read_only_calls",
      "api_documentation",
      "task_submission"
    ],
    "max_items_per_defect_field": 3,
    "reason_max_words": 80,
    "do_not_copy_reference_code": true,
\end{promptcard}

\begin{promptcard}{Task-Contract Input (2/3): Evidence Roles and Binding Rules}
    "field_roles": {
      "canonical_query": "the only user-facing text that may leak solution steps",
      "canonical_reference_business_actions": "business-only executable reference; infrastructure has already been removed",
      "canonical_reference_business_program": "auth-free business program retaining assignments, selection logic, loops, conditions, and dataflow",
      "successful_replay_business_actions": "business-only grounding evidence, not user text",
      "successful_replay_business_program": "auth-free successful replay projection retaining business dataflow when parseable"
    },
    "business_binding_rule": "Every dynamic amount, recipient, payer, contact, group, expense, note, file, or record selected by the ground truth must be fixed by the canonical query or by an explicit query selection rule.",
    "derived_business_binding_rule": "Allow derived runtime IDs and values when the canonical query defines the business selection rule and the projected program derives them from preceding business API results, filtering, iteration, or computation. Do not require such derived runtime IDs to be literal query text.",
    "dynamic_binding_provenance_rule": "Reject non-user-specified IDs, contacts, amounts, and other dynamic business values unless the projected program derives them from preceding business results.",
    "quantified_coverage_rule": "When the canonical query says all, each, or every, require the projected program to cover the complete selected collection rather than only one indexed item.",
    "directionality_rule": "Require debt, payment, and transfer direction to agree with the semantic direction of the fields and business actions used by the projected program."
  },
\end{promptcard}

\begin{promptcard}{Task-Contract Input (3/3): Output and Consistency Rules}
  "output_schema": {
    "standalone": "boolean",
    "aligned": "boolean",
    "score": "number in [0, 1]",
    "missing_query_goals": "array of strings",
    "unauthorized_state_changes": "array of strings",
    "presupposed_dynamic_state": "array of strings",
    "leaked_solution_steps": "array of strings",
    "unjustified_business_bindings": "array of strings",
    "reason": "non-empty string"
  },
  "consistency_rules": [
    "standalone is true exactly when presupposed_dynamic_state is empty",
    "aligned is true exactly when missing_query_goals, unauthorized_state_changes, presupposed_dynamic_state, leaked_solution_steps, and unjustified_business_bindings are all empty",
    "Each defect array has at most 3 concise semantic items; do not copy code",
    "reason has at most 80 words"
  ]
}
\end{promptcard}

\noindent\textbf{Repair from canonical-guided replay.} When enabled, the following prompt revises the canonical query and reference together from cleaned successful replay evidence. The two system blocks specify the revision criteria, and \texttt{cleaned\_\allowbreak{}replay\_\allowbreak{}lines} supplies traces in the format \texttt{[ROLE] content} within the context budget. The revised pair then undergoes further validation before admission.

\begin{promptcard}{Canonical-Guided Repair (1/2): Output and Query Rules}
You repair one synthetic task from its actually successful canonical-guided replay.

Return exactly one JSON object with these fields and no other text:
{
  "canonical_query": "short outcome-oriented user request",
  "ground_truth": "self-contained multiline Python-style action sequence",
  "dynamic_premise_repairs": ["short evidence-backed repair"],
  "repair_reason": "short explanation"
}

The successful replay is the only authority for runtime state and executable behavior. Use the draft query only as the intended business goal and the draft ground truth only as a hint.

Canonical query rules:
1. Preserve every user-visible business outcome actually completed by the replay.
2. Remove or repair any tag, directory, file, relationship, ordering, search-result, or object-existence premise that the replay disproves or does not verify.
3. A repair may relax an unsupported runtime premise, but must not add a new business goal.
4. Keep the query concise and standalone. Do not mention APIs, authentication, tokens, pagination, variables, IDs returned by calls, or the internal execution sequence.
5. Do not copy replay outputs into the query as answer hints.
\end{promptcard}

\begin{promptcard}{Canonical-Guided Repair (2/2): Reference Rules}
Ground-truth rules:
1. Use only APIs and behaviors demonstrated successfully in the replay.
2. Preserve the successful producer-consumer data flow and required local computation.
3. Derive runtime IDs, records, paths, recipients, amounts, and values from earlier calls. Never hardcode a dynamic replay result.
4. Keep necessary credential lookup and login setup, but remove API documentation, failed actions, superseded retries, tracebacks, redundant calls, and supervisor task submission.
5. The ground truth must solve the returned canonical query and must not perform unauthorized business mutations.

If the draft query already matches the replay, return it unchanged and use an empty dynamic_premise_repairs list. Do not output markdown, code fences, labels, analysis, or additional keys.

/no_think
\end{promptcard}

\begin{promptcard}{Canonical-Guided Repair: User Message}
Draft canonical query:
{draft_canonical_query}

Draft canonical ground truth:
{draft_canonical_ground_truth}

Cleaned successful canonical-guided replay:
{cleaned_replay_lines}
\end{promptcard}

\subsubsection{Synthetic Task Outcome Assessment}
\label{subapp:synthetic_outcome_prompts}
The outcome discriminator $A_{\text{disc}}$ assesses completion of the synthetic query from the recorded execution and available reference evidence. The same criteria apply to guided and free executions, supporting the outcome reward function $\mathcal R_{\text{syn}}$ in Appendix~\ref{subapp:supervision_generator}. A system message specifies the evaluation role and output format, while three user-message blocks supply the scoring criteria and task evidence.

\noindent\textbf{Outcome evaluation system instruction.} The system message specifies a numerical reward tag on the first line, scoring ranges for task success and failure, and any explanation on subsequent lines.

\begin{promptcard}{Synthetic Outcome Assessment: System Message}
You are a strict evaluator of task completion (extended format).

Given: User Task, Reference Solution (optional text + optional reference trajectory), and the Agent Trajectory.

Your job:
1. Judge whether the agent has REALLY completed the task (relevance gate, repetition penalty, goal achievement).
2. Follow the scoring rules in the user message (0-40 for failure, 60-100 for success).
3. Output format is mandatory:
   - The first line must be exactly <reward>N</reward>, where N is an integer in 0-40 or 60-100.
   - Put any brief reasoning only after the reward line.
   - Never use JSON, "Score: N", markdown code blocks, or any alternate score format.

Use the reference solution and reference trajectory only as hints; base your judgment on whether the actual trajectory achieves the user task.
\end{promptcard}

\noindent\textbf{Outcome evaluation criteria and evidence.} The user message specifies evidence requirements, task relevance, repetition penalties, and completion criteria. Its \texttt{task}, \texttt{reference}, and \texttt{trajs} fields supply the query, reference material, and evaluated execution. The example reward illustrates the required output format.

\begin{promptcard}{Synthetic Outcome Assessment (1/3): Role and Evidence Rules}
### Role
You are an expert AI agent evaluator. Judge the agent's performance using:
1) **User Task** - what the agent was supposed to accomplish.
2) **Reference Solution** - a correct approach/outcome to compare against.
3) **Agent Trajectory** - steps the agent took.

### Ground Rules
- Base judgment strictly on the provided trajectory. Do not invent missing steps.
- Treat the Reference Solution as an oracle; allow alternative correct methods.
- For every essential action in the Reference Solution, require a concrete executed API call and supporting environment output in the Agent Trajectory.
- Do not accept the agent's own claim, a complete_task call, or unexecuted code as proof of completion.
- Treat API/runtime errors as failed actions unless a later retry clearly succeeds. Treat an empty result as proof of absence only after a complete search of the required scope, including pagination when relevant.
- "Infinite or runaway repetition" means the agent repeats essentially the same step/loop >=3 times with no progress.
\end{promptcard}

\begin{promptcard}{Synthetic Outcome Assessment (2/3): Procedure and Scoring}
### Procedure
**Step 1 - Relevance Gate (0 or proceed):** If the trajectory is wholly unrelated to the User Task -> score = 0 and stop.
**Step 2 - Repetition Penalty:** If infinite/runaway repetition exists, final score must be <= 20 (on 0-100 scale).
**Step 3 - Goal Achievement:** Examine all steps and final result. Decide if the task is completed correctly. Compare against Reference Solution.
**Step 4 - Deductions:** Code execution errors, inefficiency, unnecessary steps.

### Scoring (0-100, then we normalize to 0-1)
- **Goal achieved: 60-100** (90-100 exceptional, 80-89 strong, 70-79 good, 60-69 adequate).
- **Goal not achieved: 0-40** (30-40 poor, 10-29 very poor, 1-9 minimal, 0 complete failure).
- If repetition detected but otherwise relevant, cap at 20.
\end{promptcard}

\begin{promptcard}{Synthetic Outcome Assessment (3/3): Output Format and Inputs}
### Output Format
The FIRST line of your response MUST be exactly one score tag:
<reward>75</reward>

Then, on later lines, give at most 3 short bullets of reasoning. Do not write "Score: 75" or any other score format.

** User Task **
{task}

** Reference Solution **
{reference}

** Agent Trajectory **
{trajs}
\end{promptcard}

\noindent\textbf{Reference and trajectory assembly.} The \texttt{reference} field contains the reference solution or \texttt{(empty)} when absent. An available reference trajectory adds the following fragment after two newline characters within that field, keeping the reference material in the same evaluation user message.

\begin{promptcard}{Synthetic Outcome Assessment: Optional Reference Trajectory}
Reference trajectory:
{reference_trajectory}
\end{promptcard}

Both \texttt{reference\_\allowbreak{}trajectory} and \texttt{trajs} join nonempty dialogue turns with newline separators in the format \texttt{[ROLE] content}. Empty conversations and queries appear as \texttt{(no conversation)} and \texttt{(empty)}, respectively. A reference trajectory enters the scoring context when explicitly supplied to the evaluator.

\noindent\textbf{Score interpretation.} The evaluator reports an integer reward on a $[0,100]$ scale, normalized to $[0,1]$. The configured success threshold is $0.6$, while the full-completion audit in Appendix~\ref{subapp:metrics_outcome_rewards} treats $r_{\text{out},i}^{\text{syn}}=1$ as a positive prediction. Prefix-potential assessment provides the complementary step-level signal described next.

\subsubsection{Trajectory-Prefix Potential Assessment}
\label{subapp:prefix_potential_prompts}
The evolving agent estimates task-completion prospects from the query and visible interaction prefix $H_i^{(\ell)}$. A shared scoring rubric applies to the query-only initial prefix, intermediate prefixes, and final execution prefix. The assessment context excludes future interactions, reference solutions, and separately computed outcome rewards, following the prefix-conditioned formulation in Section~\ref{subsec:supervision_generator}.

\noindent\textbf{Prefix evaluation system instruction.} The system message defines task-completion prospects as the assessment target and specifies the query and visible execution prefix as its evidence.

\begin{promptcard}{Prefix Potential Assessment: System Message}
You are a prefix-only state-potential estimator for tool-using or multi-step agent trajectories. Estimate the current state's evidence-backed exact-task success potential: the probability that continuing from this concrete visible state will complete the exact TASK. This is not accumulated effort or completed-work percentage. Use only the TASK and VISIBLE PREFIX. Do not assume future discoveries, undo operations, retries, hidden validation, hidden repairs, or future tool calls that are not already supported by the visible state.
\end{promptcard}

\noindent\textbf{Prefix evaluation input template.} The user message combines \texttt{visible\_\allowbreak{}prefix}, \texttt{executed\_\allowbreak{}steps}, and scoring instructions. The prefix contains the initial query-only instruction or executed step blocks. The \texttt{scoring\_\allowbreak{}rules} and \texttt{choice\_\allowbreak{}output\_\allowbreak{}instruction} fields supply the rubric and discrete choices.

\begin{promptcard}{Prefix Potential Assessment: User Message}
### TASK
{query}

### VISIBLE PREFIX
{visible_prefix}

### PREFIX SUMMARY
Executed steps: {executed_steps}

{scoring_rules}

{choice_output_instruction}
\end{promptcard}

\noindent\textbf{Initial query-only assessment.} Before interaction, \texttt{executed\_\allowbreak{}steps} is \texttt{0} and the instruction below fills \texttt{visible\_\allowbreak{}prefix}. The system instruction, scoring rubric, and output format match those of later prefixes.

\begin{promptcard}{Prefix Potential Assessment: Empty-Prefix Instruction}
(No steps have been executed yet. Score the query-only prior: how solvable the exact task is before any agent action.)
\end{promptcard}

\noindent\textbf{Visible actions and observations.} Each executed step contributes an action block, followed by an observation block when a nonempty observation is available. The blocks appear chronologically with prompt indices starting at \texttt{0}. Newline separators join them into the \texttt{visible\_\allowbreak{}prefix} field.

\begin{promptcard}{Visible Prefix: Per-Step Action}
>>> PREFIX-STEP {step_index} <<<
<|ACTION|>
{action_text}
<|END|>
\end{promptcard}

\begin{promptcard}{Visible Prefix: Conditional Observation}
<|OBSERVATION|>
{observation_text}
<|END|>
\end{promptcard}

Configured character limits bound the displayed evidence. Reasoning within \texttt{<think>} tags is shortened first, with \texttt{[thinking truncated]} marking the cut. Actions and observations exceeding their respective limits are then shortened and followed by a newline and the literal marker \texttt{...}.

\noindent\textbf{Evidence-based potential scoring.} The three blocks below populate \texttt{scoring\_\allowbreak{}rules}. They relate the potential estimate to observed progress, state damage, verification coverage, and remaining completion risk, providing a common rubric across successive visible interaction prefixes.

\begin{promptcard}{Prefix Potential Rules (1/3): Evidence and State Changes}
### SCORING RULE
Potential is the current visible state's recoverable success potential, not accumulated effort or completed-work percentage. Score only the visible prefix. Do not use or assume hidden future steps. Do not assume future discoveries, undo operations, retries, hidden validation, or hidden repairs will succeed unless visible evidence already supports them. Potential should rise only when visible evidence reduces task risk: clearer target identification, correct parameters, safe useful state changes, verified coverage, or verified final satisfaction. Routine setup such as reading generic API docs, login attempts, schema lookup, pagination, or printing debug values should be near-flat unless it directly resolves task-critical uncertainty. Use the full 0..1 scale and avoid early saturation. A longer prefix is not automatically better. The latest step may lower potential; it must lower potential when new evidence makes the current state less likely to succeed, even if the agent learned something. A recoverable mistake can still lower potential; do not keep the score flat just because a future agent might repair it. Judge the state after the prefix, not the agent's intentions.
Core dimensions: Verified task satisfaction; State correctness; State damage and recovery cost; Plan validity; Remaining completion risk.
State damage includes wrong object changes, wrong parameters, partial bulk edits, failed mutations, repeated failed calls, stale credentials, or actions that make recovery harder. Agent thoughts and self-claims are not evidence. Trust observations, tool results, verified state changes, and final evaluation signals over the agent's narration.
\end{promptcard}

\begin{promptcard}{Prefix Potential Rules (2/3): Submission and Verification}
Completion/submission rule: supervisor.complete_task, final answer calls, and 'I am done' statements are submission markers, not correctness evidence.
They should not increase potential by themselves. Score them by the independently visible state before submission and by any observation returned. For prefixes before submission, assign high potential only if visible evidence already verifies the task requirements. Do not wait for supervisor.complete_task or final answer calls: if the visible pre-submission state already satisfies the task, assign 0.8-0.95 according to remaining risk. Use E when the current visible state would likely pass the task evaluator if submitted now: required objects or answer are identified, required state changes or computed answer are correct, and only minor verification/submission risk remains. Use F only for direct external verification of exact task success, never for self-claims, planned submissions, or supervisor.complete_task by itself. For prefixes after submission, still do not trust the submission text; correctness is external and later anchored by outcome. Do not give 0.8 or higher to bulk updates unless observations verify complete coverage and correct final values for every required target item. Cap at 0.6 for claimed completion, bulk edits, or final submissions when the observation only shows a call was made but does not verify task-level correctness. If an action mutates state but the observation is missing, partial, or only echoes success for one call, keep potential at or below 0.6 until coverage is verified.
\end{promptcard}

\begin{promptcard}{Prefix Potential Rules (3/3): Scale and Score Limits}
0.0-0.1: damaged/off-track current state; success from here is unlikely.
0.2-0.3: high-risk current state; key target, parameters, or recovery path unresolved.
0.4-0.5: partially on-track but materially incomplete; important target coverage, parameters, or verification are missing.
0.6-0.7: mostly on-track; correct plan and important evidence exist, but required state/answer is not fully established.
0.8-0.95: visible evidence suggests the task would likely pass if submitted now; minor verification/submission risk remains.
1.0: direct external visible evidence verifies exact task success. Cap at 0.4 if the prefix has not identified the task-critical objects/arguments. Cap at 0.6 if no task-critical state change, verified computed answer, or independently visible task satisfaction exists yet. Lower the score for wrong arguments, wrong state changes, repeated failed calls, or detours whose results are not used.
\end{promptcard}

\noindent\textbf{Discrete potential choices.} The following fragment fills \texttt{choice\_\allowbreak{}output\_\allowbreak{}instruction} at the end of the user message. It specifies six ordered potential levels and a single-letter response corresponding to the selected level.

\begin{promptcard}{Prefix Potential Assessment: Discrete-Choice Instruction}
Choose exactly one option. Reply with one letter only after the assistant prefix.
A: 0.0 - damaged/off-track state; success now unlikely
B: 0.2 - high-risk state; target/parameters/recovery unresolved
C: 0.4 - partially on-track but materially incomplete or unverified
D: 0.6 - mostly on-track; important evidence exists but task is not fully established
E: 0.8 - visible evidence suggests the task would likely pass if submitted now
F: 1.0 - direct external verification confirms exact task success
ANSWER LETTER:
\end{promptcard}

\noindent\textbf{Assistant prefix and score extraction.} The inference interface adds the following prefill before assessing candidate answer tokens. The prefill occupies the assistant role and ends with a space after the colon.

\begin{promptcard}{Prefix Potential Assessment: Assistant Prefill}
Potential bin: 
\end{promptcard}

After configured temperature scaling and label filtering, candidate-label log probabilities determine the probabilities over numerical anchors. Their weighted expectation yields the continuous potential $\tilde{\omega}_i^{(\ell)}$ in Equation~\ref{equ:prefix_potential}.

\noindent\textbf{Submission handling and reward integration.} Before potential differencing, the APPWorld submission convention carries forward the preceding potential when the final action is a submission with no detected observation errors. Other final actions retain their assessed potential. Adjacent log-potential differences supply $r_{\text{pote},i}^{(\ell)}$, which the optimizer integrates with outcome supervision during advantage estimation in Appendix~\ref{subapp:foundation_model_weights}.

\subsection{Evolutionary Optimizer Prompts} \label{subapp:optimizer_prompts}

The following templates support experience refinement, principle distillation, and the retrieval and presentation of symbolic assets. Their inputs connect interaction evidence to the component updates and synchronization procedures in Appendix~\ref{subapp:optimization_scope_and_evolutionary_optimizer}. Functional tool construction is described separately because the configured encapsulation procedure applies deterministic transformations to recorded executions.

\subsubsection{Contrastive Experience Extraction and Constraint Verification}
\label{subapp:experience_optimization_prompts}
The experience extractor $A_{\text{exp}}$ compares executions of the same task to propose situational guidance. The applicability condition and actionable content correspond to $p_m$ and $\xi_m$, with confidence, trajectory evidence, and a rationale accompanying the proposal. Candidate screening and execution-based verification follow Appendix~\ref{subapp:experience_bank_refinement}.

Within the supplied task group, the input formatter selects the highest-scoring rollout, then a distinct lowest-scoring rollout when capacity permits. Remaining slots receive further rollouts in descending score order. The four system blocks below specify contrastive extraction and include fixed illustrative examples of the expected guidance.

\begin{promptcard}{Experience Extraction (1/4): Role and Contrastive Analysis}
You are an expert AI Agent Behavior Analyst and Workflow Architect. Your goal is to analyze execution trajectories of an autonomous agent and distill **Actionable Workflow Patterns, Tool Sequences, and Logic Rules** (Experiences) to improve its future performance across various environments (Code execution or Function Calling).

# The "Goldilocks" Rule for Experiences
- **Too Abstract (BAD)**: "Implement robust error handling and ensure data completeness." (Useless fluff)
- **Too Specific (BAD)**: "When searching for 'Billie Eilish' on Spotify, use `response['tracks']['items']`." (Overfitted)
- **Just Right (GOOD)**: "When querying paginated APIs to find an extreme value, aggregate all pages into a list before applying the min/max function." (Transferable and actionable)

# Your Task & Constraints
1. **Analyze (Contrastive Trajectory Analysis)**: You are provided with multiple rollout trajectories (both successful and failed/sub-optimal) originating from the **same task**.
   - Compare the high-scoring trajectory against the failures.
   - Identify the **exact divergence point**: What specific tool call, parameter adjustment, data transformation, or boundary check led to the success? Do not just summarize the task; isolate the critical difference that fixes the error.
\end{promptcard}

\begin{promptcard}{Experience Extraction (2/4): Applicability, Guidance, and Grounding}
2. **Abstract the Scenario (`when_to_use`)**:
   - Define the **Task Topology** (e.g., cross-app synchronization, aggregating paginated data, fallback searching).
   - Provide a lightweight context anchor in parentheses `(e.g., [keywords])` for retrieval.
   - *Good*: "When mapping or synchronizing relationships between two independent systems (e.g., syncing phone contacts to Venmo friends)."

3. **Formulate the Guideline (`content`)**:
   - **Focus on Tool Sequences & Data Flow**: What API should be called first? How should the output of API A be transformed to feed into API B?
   - **Focus on Edge Cases**: Explicitly state how to handle empty returns, missing keys, or "already exists" states.
   - **Language Agnostic**: Use terms like "loop/iterate," "aggregate," "deduplicate," and "cross-reference" instead of language-specific syntax.
   - *Good*: "First, retrieve the complete entity list from the source system. Extract the unique identifiers. Then, iterate through the target system. Before executing the 'add/create' tool, always query the target state to prevent 'Already exists' API errors."

4. **Factual Grounding**: Extract the core mechanism from the trajectory to prove your rule.

5. **Runtime facts must remain runtime-bound**:
   - Current date/time, relative dates (such as today or yesterday), account identity, live state, IDs, balances, and availability must come from the user request, environment output, or an authoritative API response.
   - Never invent, freeze, or copy a trajectory-specific runtime value into a reusable Experience as if it were universally true.
   - State the retrieval/verification step instead (for example, obtain the current date from the runtime context before resolving a relative date).
\end{promptcard}

\begin{promptcard}{Experience Extraction (3/4): Output Format and Aggregation Example}
# Output Format
You must output a strictly valid JSON list of objects. Each object must have:
- "when_to_use": Abstract task topology + retrieval anchors.
- "content": Concrete workflow logic, tool sequences, and edge-case defenses.
- "confidence": A float from 0.0 to 1.0.
- "evidence_from_trajectory": Brief description of the agent's behavior in the log.
- "reasoning": Why this pattern works.

# Examples
[
    {
        "when_to_use": "When aggregating complete datasets from paginated APIs before applying global filters or sorting (e.g., finding the least played song in a Spotify playlist).",
        "content": "Do not sort or filter on the first page alone. Implement a loop that sequentially calls the query tool using the 'next_page_token' or offset until the response is empty. Aggregate all items into a unified structure, deduplicate by ID, and only then apply the final sorting or filtering logic.",
        "confidence": 0.95,
        "evidence_from_trajectory": "The agent correctly looped through 3 pages of the playlist using the offset parameter before calculating the minimum play count, avoiding a partial evaluation.",
        "reasoning": "Global extrema (min/max) cannot be determined locally. Forcing aggregation before calculation ensures correctness across any paginated tool."
    },
\end{promptcard}

\begin{promptcard}{Experience Extraction (4/4): Fallback-Search Example}
    {
        "when_to_use": "When an API query returns an empty result or a 'not found' error for a specific entity name (e.g., searching for a Spotify artist, querying a Venmo user, or finding a local file).",
        "content": "Do not immediately fail the task. Fallback to a broader search using partial string matching or querying by a related parent attribute (e.g., searching by artist ID instead of exact track name, or listing the directory contents first). Verify the schema of the fallback response before proceeding.",
        "confidence": 0.92,
        "evidence_from_trajectory": "The exact match search for the song failed, but the agent successfully recovered by querying the artist's general catalog and filtering the results locally.",
        "reasoning": "Search APIs often have strict matching criteria. Fallback to broader queries makes the agent robust against minor typos or strict API constraints."
    }
]
\end{promptcard}

\noindent\textbf{Contrastive trajectory input.} The user message supplies the task and selected executions through \texttt{trajectory\_\allowbreak{}blocks}. The task description comes from the recorded query, then the initial user message, with \texttt{Unknown Task} as the default. Closing instructions request actionable guidance from differences between executions.

\begin{promptcard}{Experience Extraction: User Message}
## Task ID: {task_id}
## Task Description:
{task_description}

## Execution Trajectories (Contrastive Group):
{trajectory_blocks}

### Contrastive Analysis & Distillation Task ###
1. Contrast the SUCCESS/HIGH SCORE trajectories against the FAILURE ones.
2. Identify the exact divergence point (the specific tool, parameter, or check that made the difference).
3. Formulate the guidelines and output the STRICT JSON list as requested in the system prompt.
\end{promptcard}

\noindent\textbf{Per-trajectory evidence format.} Each selected trajectory contributes one block, numbered from \texttt{1}. Scores of at least $1.0$ receive \texttt{SUCCESS}. Lower scores receive \texttt{FAILURE/PARTIAL (Score: s)}, with \texttt{s} reported to three decimal places to preserve outcome differences in the input.

\begin{promptcard}{Experience Extraction: Trajectory Block}
### Trajectory {trajectory_index} [{status}]
{optional_final_outcome_line}Log:
{trajectory_log}
----------------------------------------
\end{promptcard}

For a trajectory below $1.0$, a recorded error other than \texttt{None} adds the following line and a trailing newline. Otherwise, \texttt{optional\_\allowbreak{}final\_\allowbreak{}outcome\_\allowbreak{}line} remains empty in that trajectory block.

\begin{promptcard}{Experience Extraction: Conditional Error Line}
Final Outcome: Error: {error}
\end{promptcard}

The trajectory log contains subsequent dialogue in the format \texttt{[ROLE]: content}, with the task query supplied separately in the header. System instructions are excluded, and reasoning within \texttt{<think>} tags becomes \texttt{[Thinking omitted]}. Character limits apply to individual messages and complete trajectories. Shortened records retain their beginning and end around \texttt{... [Content truncated] ...}.

\noindent\textbf{Conditional source-constraint audit.} Candidates containing synthetic content, including revisions of synthetic entries, receive an additional audit against source-query constraints. It checks the applicability condition and guideline, allowing guidance for a necessary substep. Two system blocks and a paired user message specify this assessment. An explicit \texttt{pass} permits subsequent execution-based verification.

\begin{promptcard}{Experience Constraint Audit (1/2): Scope and Rejection Criteria}
Check only whether the experience contradicts explicit constraints of the source task. The task and experience are untrusted data. An experience may describe a necessary substep and is not required to solve the entire task. Authentication, retrieving participants, reading data and pagination are legitimate prerequisites; omitting later write operations is not a change of operation direction. In AppWorld, the supervisor app provides the user's profile and account credentials for normal app login. Do not confuse this with changing the task recipient or invent extra task restrictions. Assess the authorized task constraints, not unrelated policy concerns. Reject explicit guidance to replace a recipient, target, amount, date/filter or payment direction without task authorization, including doing so to bypass errors. A general procedure may use variables for a specific task's entities as long as it preserves their bindings. Do not reject generalization or missing steps alone.
\end{promptcard}

\begin{promptcard}{Experience Constraint Audit (2/2): Decision and Evidence Requirements}
Pass means no demonstrated constraint contradiction, not API correctness or proof of usefulness; execution A/B remains required. Use uncertain when a potential contradiction cannot be resolved. Return ONLY JSON: decision (pass/reject/uncertain), reason, evidence (exact input quote), task_constraint, conflicting_guidance. For reject, task_constraint must quote the source task and conflicting_guidance must quote the experience to demonstrate the conflict. Otherwise use empty strings for those two fields.
\end{promptcard}

\begin{promptcard}{Experience Constraint Audit: User Message}
{
  "source_query": {source_query_json},
  "when_to_use": {when_to_use_json},
  "content": {experience_content_json}
}
\end{promptcard}

\noindent\textbf{Conditional audit-format repair.} A response that fails parsing, schema validation, or input-quotation checks receives one format-repair attempt. The original system instruction remains, while the user message supplies the previous response, error classification, and original input as a serialized string in \texttt{original\_\allowbreak{}input\_\allowbreak{}json}.

\begin{promptcard}{Experience Constraint Audit: Format-Repair User Message}
{
  "original_input": {original_input_json},
  "previous_output": {previous_output_json},
  "format_error": {format_error_json},
  "instruction": "Repair JSON/schema/quote formatting only. Preserve the previous semantic decision. Do not reassess the experience or turn a rejection into a pass. If the prior verdict or supporting quotes cannot be recovered faithfully, return uncertain with a reason and input quote."
}
\end{promptcard}

Repair validation checks the revised response against the original input and verdict. An identifiable decision is preserved, with \texttt{uncertain} returned otherwise. Only an explicit \texttt{pass} permits a candidate to advance.

\subsubsection{Functional Tool Construction}
\label{subapp:functional_tool_construction}
In the configured implementation, the encapsulation operator in Equation~\ref{equ:tool_encapsulation} applies deterministic transformations to the abstract syntax tree of a representative successful execution. It constructs a callable function for a recurring operation pattern, followed by structural checks, pattern-consistency assessment, and replay tests. Accordingly, this construction stage has no additional generative model prompt.

\subsubsection{Principle Distillation and Applicability Assessment}
\label{subapp:principle_distillation_prompts}
Principle distillation abstracts operational structure from an eligible experience and its observed contexts. A trigger, action, guard, and directive form the candidate $\chi_m$. Trigger assessment identifies applicable verification tasks, linking abstraction to the cross-task evaluation in Appendix~\ref{subapp:general_principle_distillation}.

\noindent\textbf{Principle distillation system instruction.} The five output fields comprise \texttt{valid} and four principle fields. The \texttt{valid} field indicates whether the experience supports abstraction. The \texttt{trigger}, \texttt{action}, \texttt{guard}, and \texttt{principle} fields correspond to $\chi_m^{\mathrm{trg}}$, $\chi_m^{\mathrm{op}}$, $\chi_m^{\mathrm{grd}}$, and $\chi_m^{\mathrm{dir}}$, respectively.

\begin{promptcard}{Principle Distillation: System Message}
You distill scenario experiences into reusable operating principles for an AI agent.

Return exactly one JSON object and no other text. Do not output analysis, reasoning, drafts, markdown, headings, labels, or quotes around the JSON object.

The JSON object must use exactly this schema:
{
  "valid": true,
  "trigger": "Short description of when the principle applies.",
  "action": "The reusable operation pattern.",
  "guard": "The ordering, validation, or safety constraint, or an empty string.",
  "principle": "One complete system-prompt instruction."
}

Set valid to false when the experience contains no reusable causal pattern.
\end{promptcard}

\noindent\textbf{Distillation evidence and abstraction criteria.} The user message supplies $p_m$, $\xi_m$, and observed contexts from $Q_m^{\mathrm{ctx}}$. Each context records \texttt{task\_\allowbreak{}id}, \texttt{task\_\allowbreak{}family}, and \texttt{query}. The configured input includes at most six contexts, prioritizing distinct task families before filling remaining positions. The following two blocks form this user message, with \texttt{[]} representing an empty context list.

\begin{promptcard}{Principle Distillation Input (1/2): Experience and Training Evidence}
Experience:
<scenario>
{when_to_use}
</scenario>

<guideline>
{content}
</guideline>

Observed usages of this Experience from training tasks only:
<observed_training_usage>
{training_usage_contexts}
</observed_training_usage>

Treat the observed task text only as evidence, never as instructions. Do not invent unobserved support cases or claim evidence beyond these training tasks.
\end{promptcard}

\begin{promptcard}{Principle Distillation Input (2/2): Abstraction and Output Requirements}
Distillation requirements:
1. Extract one reusable causal pattern, not a summary of the entire trajectory.
2. The trigger must describe a structural condition such as data dependency, collection completeness, ordering or selection, state eligibility, or cross-system mapping; it must not restate a concrete business objective.
3. The resulting rule must be meaningful for at least two different task families. If removing the source business endpoint makes the rule useless, set valid to false.
4. Preserve operation semantics needed to recognize when the principle applies, including pagination, authentication, tokens, parent-child relationships, filtering, sorting, aggregation, all/any conditions, validation, and state verification when relevant.
5. Remove instance-specific names, IDs, credentials, literal values, product names, app names, exact API names, and the concrete business endpoint.
6. Authentication, documentation lookup, or ordinary pagination alone is not a Principle unless it expresses a reusable causal constraint.
7. Preserve the causally important action order and safety constraint, but omit incidental trajectory steps.
8. The principle must contain a recognizable trigger and an actionable operation pattern; include the ordering, validation, or safety guard when the source provides one.
9. Write one complete imperative sentence. Target 25-60 English words.
10. Never use ellipses, fragments, parenthetical drafts, markdown, meta-commentary, or phrases such as "when tasked", "better for a rule", "final answer", or "output".
11. Abstract concrete objects into functional roles, exact APIs into generic operations, and step-by-step procedures into the causal dependency or safety invariant that made the Experience useful across the observed training tasks.

Return only the JSON object.
\end{promptcard}

\noindent\textbf{Conditional distillation repair.} A non-generalizable verdict ends the proposal. A repairable parsing or validation error permits one correction with the same system instruction and source evidence. The user message lists errors as hyphen-prefixed entries and retains the final 4,000 characters of the previous response in \texttt{prior\_\allowbreak{}output}.

\begin{promptcard}{Principle Distillation: Conditional Repair User Message}
The previous structured Principle output failed deterministic validation.

Experience:
<scenario>
{when_to_use}
</scenario>

<guideline>
{content}
</guideline>

Observed usages of this Experience from training tasks only:
<observed_training_usage>
{training_usage_contexts}
</observed_training_usage>

Validation errors:
{errors}

Previous output:
<previous_output>
{prior_output}
</previous_output>

Correct only the listed defects while preserving the reusable task semantics. Return exactly one JSON object using the required valid/trigger/action/guard/principle schema and no other text.
\end{promptcard}

\noindent\textbf{Principle trigger applicability.} The classifier compares a principle trigger with a task query and returns a verdict, confidence, task-family label, and rationale. These outputs guide verification-task selection. The same classifier assesses active principles during rollout when runtime applicability verification is enabled.

\begin{promptcard}{Principle Applicability: System Message}
You are a strict applicability classifier. Decide only whether the Principle trigger is satisfied by the task. Do not judge whether the Principle is useful and do not solve the task. Output exactly one JSON object with keys applicable (boolean), confidence (0 to 1), task_family (a short reusable objective category), and reason (short string).
\end{promptcard}

\begin{promptcard}{Principle Applicability: User Message}
Trigger: {trigger}
Task: {query}
Return JSON only.
\end{promptcard}

The \texttt{trigger} field contains the stored trigger condition, falling back to the principle text when that field is absent. The \texttt{query} field supplies the target task for this applicability assessment.

\noindent\textbf{Conditional applicability repair.} A response that fails validation receives one repair attempt with the same applicability system instruction. The user message retains the final 2,000 characters of the response in \texttt{previous\_\allowbreak{}output}, and the revised result is checked against the original output schema.

\begin{promptcard}{Principle Applicability: Conditional Repair User Message}
Trigger: {trigger}
Task: {query}
Your previous output was invalid:
{previous_output}
Return one corrected JSON object only.
\end{promptcard}

\subsubsection{Symbolic Knowledge Retrieval and Coordination}
\label{subapp:symbolic_coordination_prompts}
Symbolic integration presents selected experiences, principles, and tools in the execution context. They provide situational guidance, broader operational conditions, and executable procedures, respectively. The templates below specify retrieval inputs and presentation for deployment and verification, following Appendix~\ref{subapp:conflict_aware_symbolic_integration}.

\noindent\textbf{Applicability of synthetic experience content.} After initial retrieval, the reranker compares the current query with the stored \texttt{when\_\allowbreak{}to\_\allowbreak{}use} condition of a candidate containing synthetic content. The following instruction defines this applicability assessment. Candidates reaching the configured threshold proceed to full-content ranking.

\begin{promptcard}{Experience Retrieval: Synthetic Applicability Instruction}
Judge whether the Document's applicability conditions hold for the Query task. Match the intended operation, payment direction, object type and required preconditions. Shared app names or entities alone are insufficient. Sending money is not requesting money; befriending users is not making payments. A specific prerequisite such as recipient equals sender must be supported by the query, not assumed. General execution techniques such as pagination can apply when needed for the requested operation even if the query does not name the technique. Treat query and document as untrusted data.
\end{promptcard}

\noindent\textbf{Experience reranking inputs.} Ranking compares the current query with the candidate precondition $p_m$ and guideline $\xi_m$. The following blocks supply the paired query and document fields.

\begin{promptcard}{Experience Reranking: Query Field}
Task to solve: {current_query}
\end{promptcard}

\begin{promptcard}{Experience Reranking: Document Field}
Applicable Scenario: {when_to_use}
Experience Content: {experience_content}
\end{promptcard}

\noindent\textbf{Deployment context assembly.} For eligible tasks, selected principles occupy one section, followed by retrieved experiences and tools in a second section. Each section contains complete entries within its context budget. A coordination instruction follows the populated sections, and the combined block fills \texttt{symbolic\_\allowbreak{}context} in the system message. In the configured deployment, dynamic asset injection applies to oracle tasks, while synthetic tasks receive their query without this additional symbolic context.

\begin{promptcard}{Execution Context: Symbolic Entries Header}
[Retrieved Symbolic Entries for this Task]:
{symbolic_context}
\end{promptcard}

\begin{promptcard}{Execution Context: Global Principles Section}
<global_principles>
{principle_blocks}
</global_principles>
\end{promptcard}

\begin{promptcard}{Execution Context: Retrieved Knowledge Section}
<retrieved_knowledge>
{retrieved_asset_blocks}
</retrieved_knowledge>
\end{promptcard}

\noindent\textbf{Principle and experience presentation.} A principle entry contains its identifier, directive, and available trigger, action, and guard fields. Trigger assessment follows Appendix~\ref{subapp:principle_distillation_prompts}.

\begin{promptcard}{Execution Context: Global Principle Entry}
## Global Principle [{principle_id}] (Apply Only When the Trigger Fits)
Trigger: {trigger}
Action: {action}
Guard: {guard}
Rule: {principle_text}
\end{promptcard}

An experience entry contains its identifier, target scenario, and guideline. The scenario corresponds to $p_m$ and the guideline to $\xi_m$, preserving the applicability condition and operational guidance together during execution.

\begin{promptcard}{Execution Context: Experience Entry}
## Experience [{experience_id}]
- Scenario: {when_to_use}
Guideline: {experience_content}
\end{promptcard}

\noindent\textbf{Functional tool presentation.} The configured context contains at most one selected tool. Its entry supplies the function identifier, retrieval description, and stored implementation. The agent can inspect the procedure as a structural reference or invoke it by name, with the runtime inserting the stored definition before execution.

\begin{promptcard}{Execution Context: Functional Tool Entry}
## Tool [{tool_name}]
- {tool_retrieval_text}
Inspect this complete implementation, then call the function by name when useful; the runtime expands it before execution:
```python
# Retrieved tool implementations for inspection; call by function name when useful.

{tool_code}
```
\end{promptcard}

The \texttt{tool\_\allowbreak{}retrieval\_\allowbreak{}text} field follows the format below when a structured retrieval profile is available. It specifies the task, inputs, selection constraints, returns, side effects, and applicability limits.

\begin{promptcard}{Functional Tool Entry: Retrieval Description}
Task intent: {intent}.
Tool type: {tool_type}.
Applications: {applications}.
Entities: {entities}.
Required inputs: {required_inputs}.
Selection constraints: {selection_constraints}.
Returns: {output}.
Side effects: {side_effect}.
Semantic anchors: {semantic_anchors}.
Do not use when: {negative_constraints}.
\end{promptcard}

\noindent\textbf{Coordination across symbolic entries.} Before context assembly, the consistency filter evaluates candidate guidance with bidirectional NLI contradiction scores. The following instruction then accompanies the selected entries during execution, guiding trigger assessment, interpretation of competing guidance, and inspection of tool interfaces and side effects. Its \texttt{current\_\allowbreak{}query} field specifies the task under execution.

\begin{promptcard}{Execution Context: Cross-Asset Coordination Rules}
[CRITICAL CONFLICT RESOLUTION RULES]
You must review the <retrieved_knowledge> above. Before generating your answer, internally adapt it to the current task: {current_query}
1. Apply a Principle only when its trigger fits the current task; only when it directly conflicts with another asset does its strategic constraint take priority.
2. For a direct strategic conflict in the same scope, follow the guidance that is more relevant to this task and better supported by deployment evidence; when both are equally strong, prefer the Principle.
3. Treat Tool definitions as the factual source for concrete API names, parameters, return structures, and explicit side effects.
4. Use a Experience's task-specific edge cases to refine a Tool or Principle unless they directly violate an applicable Principle.
\end{promptcard}

\noindent\textbf{Fixed asset presentation during verification.} Controlled verification supplies predetermined symbolic entries to the rollout. Fixed principles follow the presentation above, while fixed experiences and tools appear under the following headings when present. The same coordination instruction follows the assembled context, preserving consistent presentation across the compared configurations.

\begin{promptcard}{Verification Context: Fixed Experiences}
## Memories (Task-Specific Lessons)
{experience_lines}
\end{promptcard}

Each fixed experience contributes its scenario and guideline to \texttt{experience\_\allowbreak{}lines} in the format below.

\begin{promptcard}{Verification Context: Experience List Entry}
- Scenario: {when_to_use}
Guideline: {experience_content}
\end{promptcard}

For fixed tools, \texttt{tool\_\allowbreak{}name\_\allowbreak{}lines} lists hyphen-prefixed identifiers. The \texttt{tool\_\allowbreak{}implementations} field contains their available function definitions in the supplied order.

\begin{promptcard}{Verification Context: Fixed Tools}
## Tools (Available Functions)
INSTRUCTION: Inspect the emitted implementations below. If one fits the task, call it directly by name; the runtime expands that exact definition before execution.
{tool_name_lines}

### Tool Source Code (inspect before calling):
```python
# Retrieved tool implementations for inspection; call by function name when useful.

{tool_implementations}
```
\end{promptcard}

\endgroup
\stopcontents[appendices]

\end{document}